\documentclass[acmtog,authorversion]{acmart}

\AtBeginDocument{%
}

\usepackage{subcaption}
\usepackage{tikz}

\acmJournal{TOG}

\begin{document}
\title{A Divide-and-Conquer Approach for Global Orientation of Non-Watertight Scene-Level Point Clouds Using 0-1 Integer Optimization}

\subtitle{}
\author{Zhuodong Li}
\orcid{0009-0007-1865-0561}

\email{lizd@ios.ac.cn}

\author{Fei Hou}
\authornote{Corresponding author}
\orcid{0000-0001-8226-6635}
\email{houfei@ios.ac.cn}

\author{Wencheng Wang}
\orcid{0000-0001-5094-4606}
\email{whn@ios.ac.cn}
\affiliation{%
 \institution{Key Laboratory of System Software (CAS), State Key Laboratory of Computer Science, Institute of Software, Chinese Academy of Sciences}
 \country{China}
}
\affiliation{%
 \institution{University of Chinese Academy of Sciences}
 \country{China}
}

\author{Xuequan Lu}
\orcid{0000-0003-0959-408X}
\affiliation{%
 \institution{Department of Computer Science and Software Engineering, The University of Western Australia}
 \country{Australia}
}
\email{Bruce.lu@uwa.edu.au}
\author{Ying He}
\orcid{0000-0002-6749-4485}
\affiliation{%
 \institution{College of Computing and Data Science, Nanyang Technological University}
 \country{Singapore}
 }
\email{yhe@ntu.edu.sg}

\renewcommand\shortauthors{Z. Li et al.}

\begin{abstract}
Orienting point clouds is a fundamental problem in computer graphics and 3D vision, with applications in reconstruction, segmentation, and analysis. While significant progress has been made, existing approaches mainly focus on watertight, object-level 3D models. The orientation of large-scale, non-watertight 3D scenes remains an underexplored challenge. To address this gap, we propose \textit{DACPO} (Divide-And-Conquer Point Orientation), a novel framework that leverages a divide-and-conquer strategy for scalable and robust point cloud orientation. Rather than attempting to orient an unbounded scene at once, DACPO segments the input point cloud into smaller, manageable blocks, processes each block independently, and integrates the results through a global optimization stage. For each block, we introduce a two-step process: estimating initial normal orientations by a randomized greedy method and refining them by an adapted iterative Poisson surface reconstruction. To achieve consistency across blocks, we model inter-block relationships using an an undirected graph, where nodes represent blocks and edges connect spatially adjacent blocks. To reliably evaluate orientation consistency between adjacent blocks, we introduce the concept of the \textit{visible connected region}, which defines the region over which visibility-based assessments are performed. The global integration is then formulated as a 0-1 integer-constrained optimization problem, with block flip states as binary variables. Despite the combinatorial nature of the problem, DACPO remains scalable by limiting the number of blocks (typically a few hundred for 3D scenes) involved in the optimization. Experiments on benchmark datasets demonstrate DACPO's strong performance, particularly in challenging large-scale, non-watertight scenarios where existing methods often fail. The source code is available at https://github.com/zd-lee/DACPO.
\end{abstract}

%
%


%
%


\begin{teaserfigure}
\centering
    \includegraphics[width=0.325\textwidth]{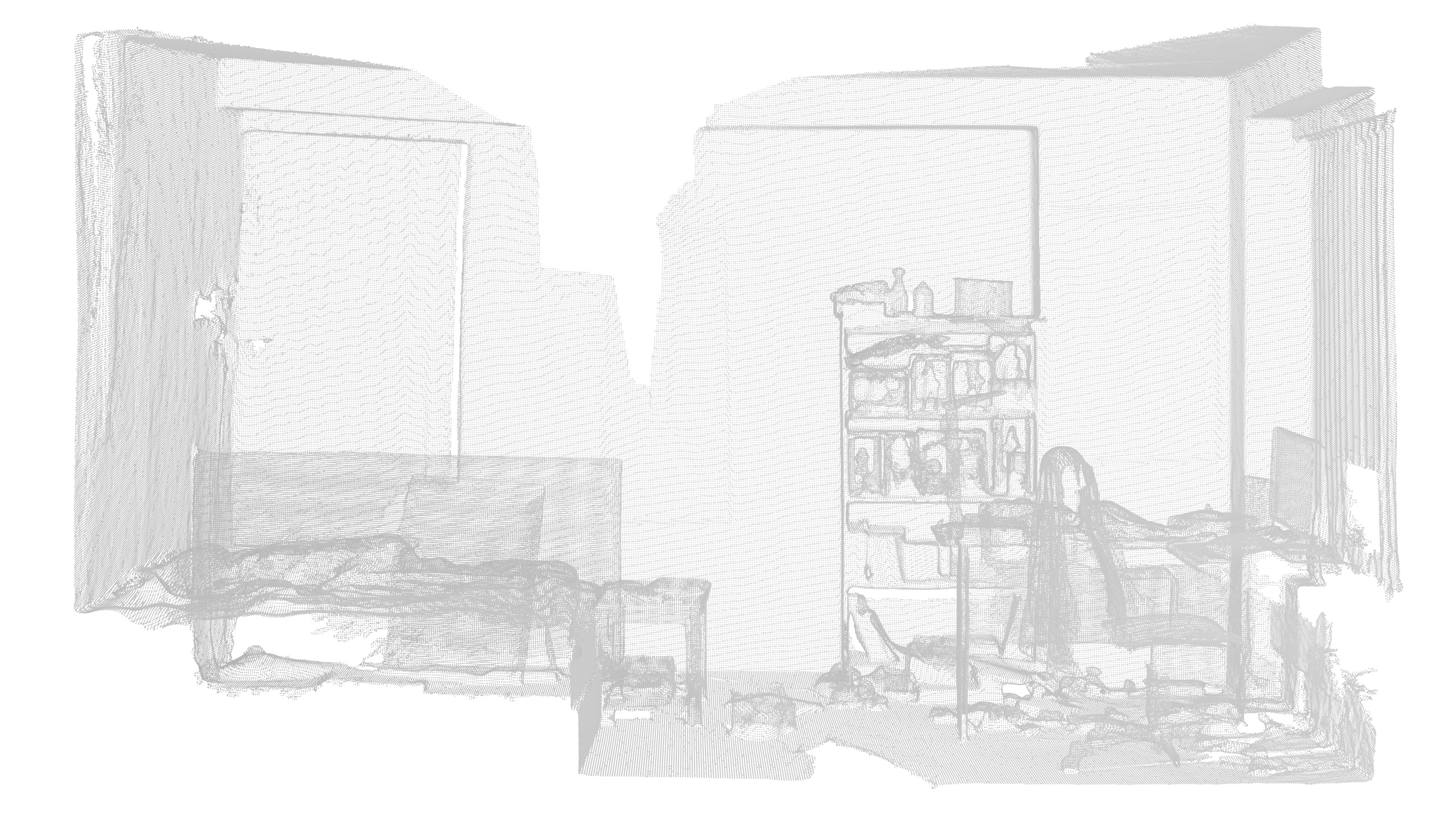}
    \includegraphics[width=0.325\textwidth]{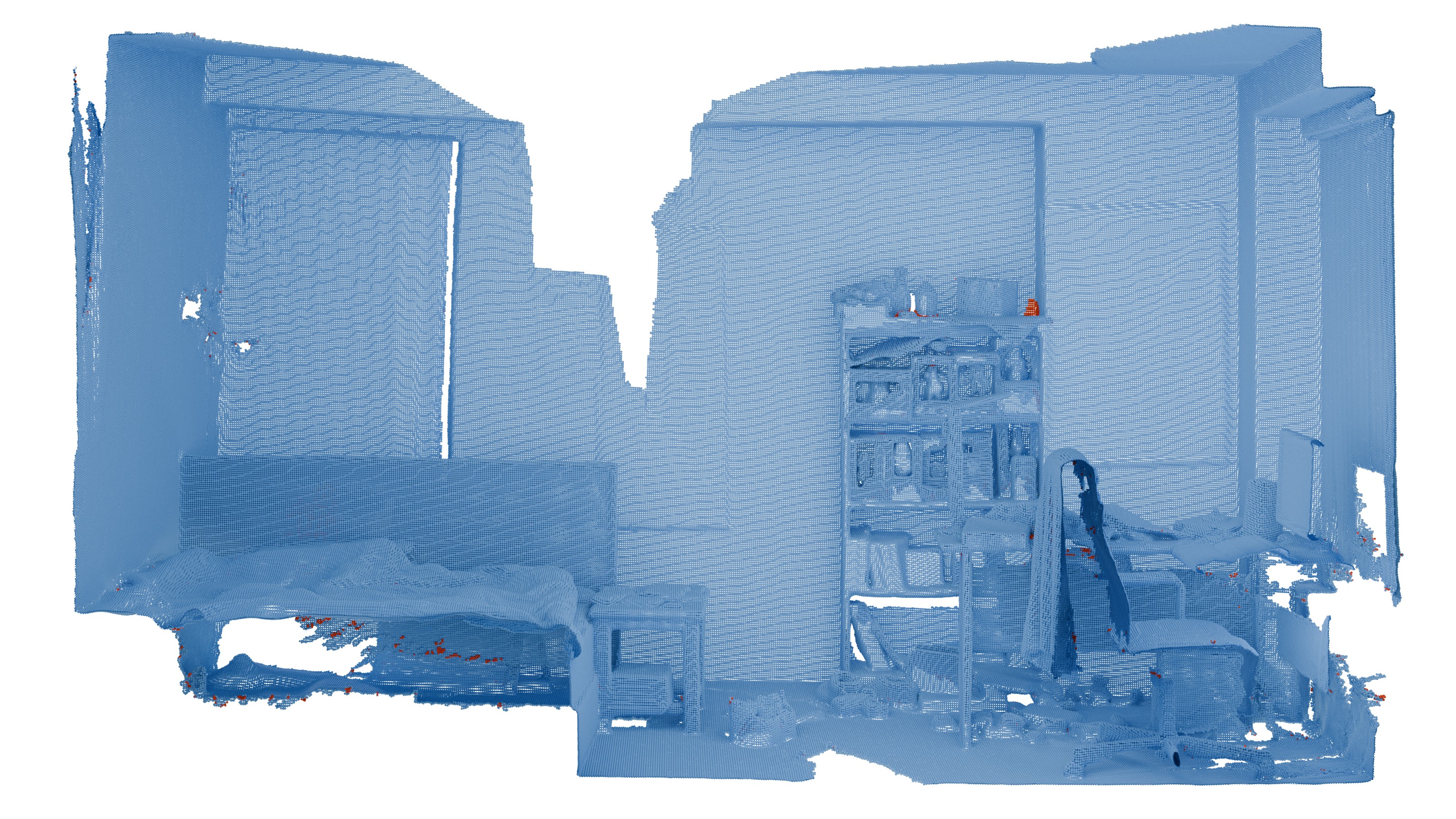}
    \includegraphics[width=0.325\textwidth]{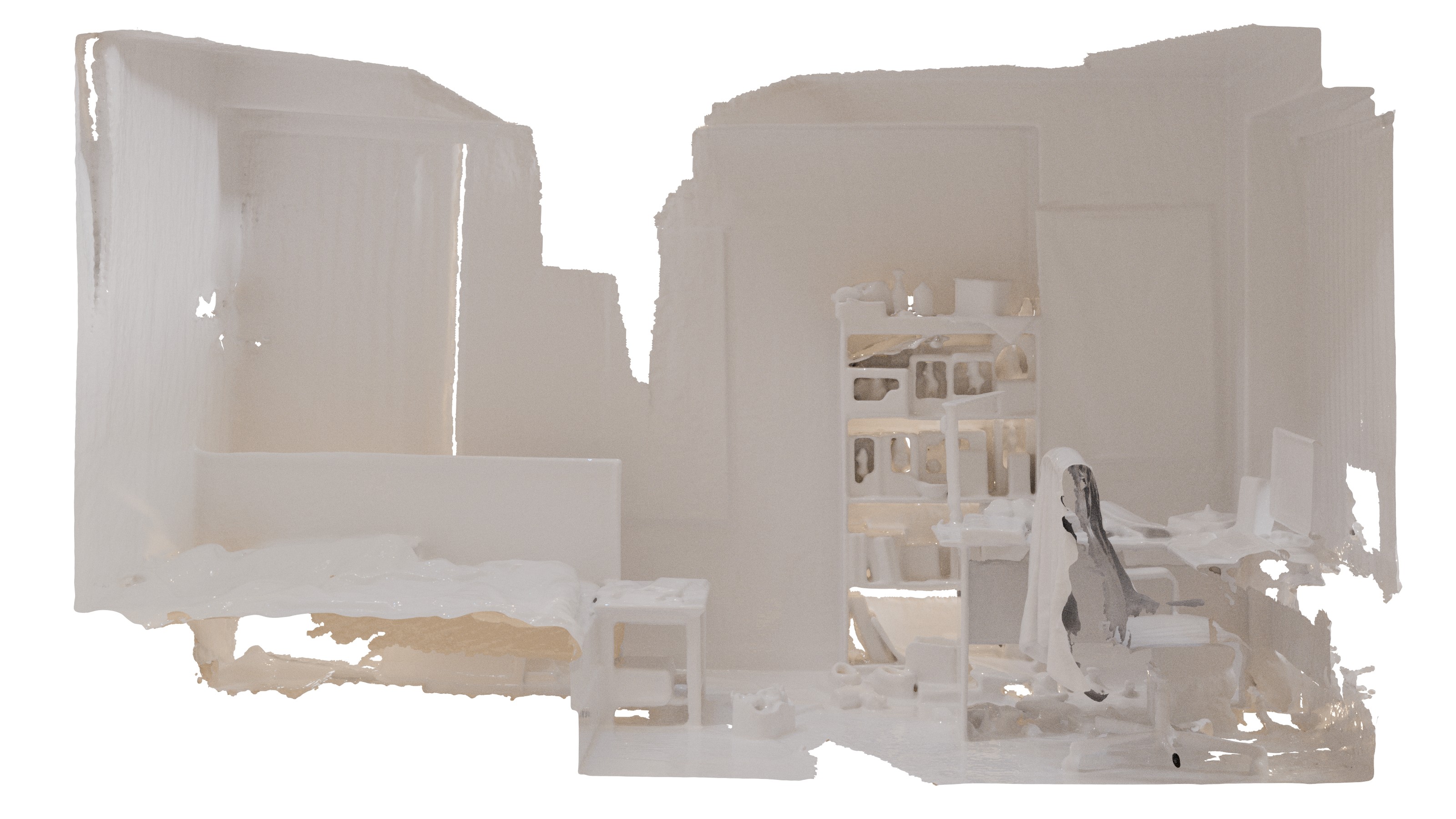}
  \caption{Our method addresses the global orientation of non-watertight, scene-level point clouds using a divide-and-conquer approach. We first segment the point cloud into smaller blocks and orient each block independently. The orientations are then globally aligned through 0-1 integer-constrained optimization. Blue points represent correctly oriented normals, while red points indicate incorrectly oriented normals. Finally, the oriented point cloud is used to reconstruct the surface via screened Poisson surface reconstruction with Neumann boundary condition~\cite{Kazhdan2013}}
  \label{fig:teaser}
\end{teaserfigure}

\maketitle

\section{Introduction} \label{sec:introduction}

 Classical propagation-based methods typically adopt a two-step strategy. First, they estimate point normals using local shape analysis techniques, such as principal component analysis (PCA) or local tangent-plane fitting. Second, they propagate normal directions across the point cloud by enforcing local consistency, under the assumption that neighboring points should have similar orientations~\cite{Hoppe1992,Xie2003,Huang2009,Yang2024NRSC}. While these methods are efficient, they tend to fail in ambiguous cases such as sparse regions, noisy inputs, outliers, and thin structures.

More recently, optimization-based methods have gained popularity by solving both point normals and their orientations jointly~\cite{Lin2023PGR,Xu2023GCNO,Liu2024BIM,Lin2024WNNC,Huang2024Stochastic,Gotsman2024linear}. These methods often rely on implicit functions, such as generalized winding number (GWN)~\cite{Jacobson2013GWN}, to infer globally consistent normal orientation. However, their applicability is limited to watertight surfaces, where a clear distinction between interior and exterior regions can be established. As such, they are not directly applicable to non-watertight, scene-level data.

To address this challenge, we introduce \textit{DACPO} (\textbf{D}ivide-\textbf{A}nd-\textbf{C}onquer \textbf{P}oint \textbf{O}rientation), a novel framework for globally orienting non-watertight, scene-level point clouds. DACPO follows a divide-and-conquer strategy: it partitions the input into smaller, connected blocks and orients each block independently. Orienting individual blocks is significantly easier than resolving the orientation of the entire scene at once. For per-block orientation, we adapt the iterative Poisson surface reconstruction (iPSR)~\cite{Hou2022iPSR}, replacing its Dirichlet boundary condition with a Neumann condition to better accommodate open surfaces.

After per block orientation, the block orientations are not globally consistent. To resolve this, we model the relationships between blocks as an undirected, connected graph, where nodes represent blocks and edges link spatially adjacent ones. We define edge weights using a novel visibility-based consistency criterion that measures the alignment cost between adjacent blocks under different flip states. The global orientation is then determined by solving a 0-1 integer optimization problem, where each binary variable indicates whether a block’s orientation should be flipped.
Although such optimization is inherently combinatorial, DACPO remains efficient and scalable due to its divide-and-conquer design and the limited number of blocks (typically a few hundred in standard 3D scenes). In our experiments, most of the computation time is spent on per-block orientation. We validate our method on large-scale, non-watertight 3D scene benchmarks, demonstrating its strong performance.

\section{Related Works}
\label{sec:related-works}

Point cloud orientation and surface reconstruction from unoriented points are closely intertwined. On the one hand, oriented normals are essential for surface reconstruction algorithms, such as Poisson Surface Reconstruction~\cite{Kazhdan2006,Kazhdan2013}; on the other hand, surfaces reconstructed from unoriented points can be leveraged to infer globally consistent orientations. This \textit{mutual dependency} has motivated a variety of methods over the years.

\paragraph{Classic Approaches} Early methods for point cloud orientation and surface reconstruction rely on computational geometry techniques~\cite{Bernardini1999,Amenta2001,Alliez2007,Dey2003}. Although efficient and elegant, these approaches are sensitive to parameter settings and lack robustness in the presence of noise. Implicit function-based methods reconstruct surfaces via extracting iso-surfaces from  computed implicit function, typically signed or unsigned distance fields~\cite{Hornung2006,Mullen2010}. These methods offer improved robustness and have been widely used in real-world applications. However, they depend on distinguishing between interior and exterior regions and are thus limited to watertight models.

\paragraph{Propagation Methods}
Propagation-based methods solve the orientation problem by iteratively propagating normal orientations across neighboring points using a priority scheme.
Minimum spanning tree-based strategies~\cite{Hoppe1992} are highly efficient but prone to errors, often flipping entire subtrees, especially in sparse data. To address this, improved propagation priorities were proposed~\cite{Xie2003,Huang2009}. Metzer et al.~\shortcite{Metzer2021Dipole} proposed using dipole electric fields to guide orientation, while Yang et al.~\shortcite{Yang2024NRSC} introduced a hierarchical propagation framework for indoor scenes. These methods can handle both watertight and non-watertight models but are greedy in nature and often lack robustness under noise.

\begin{figure}[!htbp]
    \centering
    \subcaptionbox{GT}{\includegraphics[width=.24\linewidth]{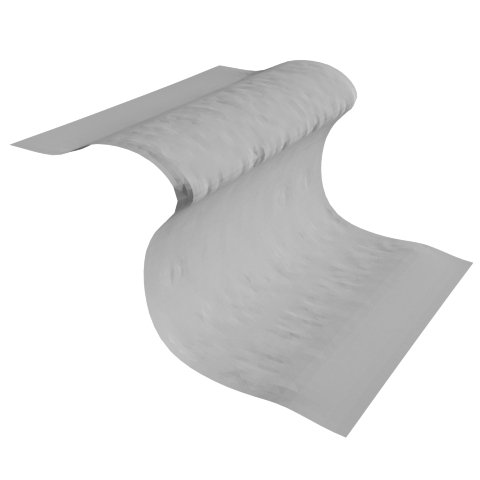}}\hspace{2pt plus 4pt minus 2pt}%
    \subcaptionbox{WNNC}{\includegraphics[width=.24\linewidth]{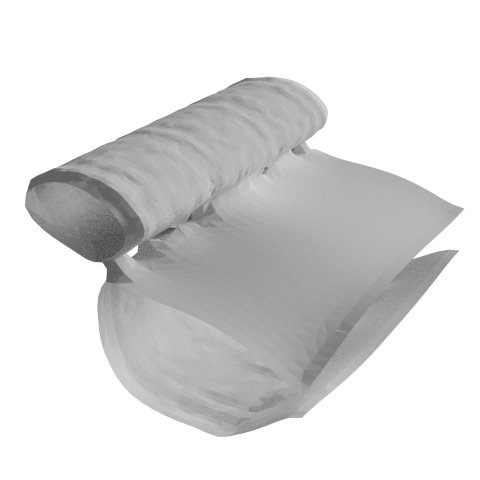}}\hspace{2pt plus 4pt minus 2pt}%
    \subcaptionbox{GCNO}{\includegraphics[width=.24\linewidth]{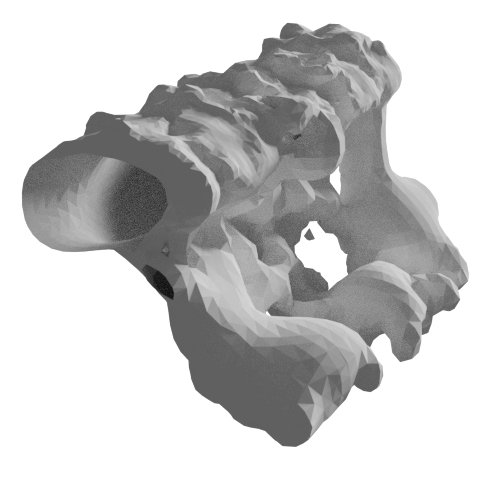}}\hspace{2pt plus 4pt minus 2pt}%
    \subcaptionbox{iPSR}{\includegraphics[width=.24\linewidth]{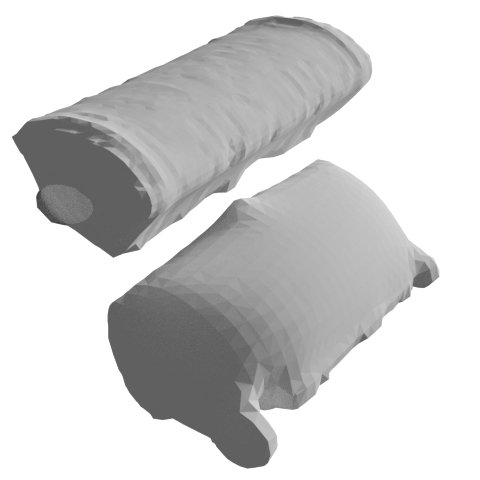}}
    \caption{Recent implicit function-based methods, such as iPSR~\cite{Hou2022iPSR}, GCNO~\cite{Xu2023GCNO}, and WNNC~\cite{Lin2024WNNC}, rely on interior-exterior separation, making them well-suited for watertight models. When applied to open surfaces, they often produce significant orientation errors. Surfaces shown here are reconstructed via sPSR~\cite{Kazhdan2013} using the predicted point normals.}
    \label{fig:non-watertight}
\end{figure}

\paragraph{Global Methods}
 In contrast to propagation-based methods, recent advances have demonstrated a growing preference to orient normals within a global framework. iPSR~\cite{Hou2022iPSR} applies the screened Poisson surface reconstruction (sPSR)~\cite{Kazhdan2013} iteratively to refine surface geometry and orientation. In each iteration, point normals are updated based on the reconstructed surface from the previous step. This method converges quickly for clean point clouds but may require more iterations for complex or noisy inputs. Other recent works~\cite{Lin2023PGR,Xu2023GCNO,Liu2024BIM,Liu2024DWG,Lin2024WNNC} utilize the GWN field~\cite{Jacobson2013GWN} for orientation. These approaches treat normals as optimization variables and regularize the GWN field to achieve consistency. Key properties such as harmonicity for non-surface points~\cite{Liu2024BIM} and a target value of $0.5$ on surfaces~\cite{Lin2023PGR} are exploited to drive optimization. These methods are robust to noise, outliers, and thin structures. However, they are designed for watertight models and struggle with open surfaces due to their reliance on interior-exterior separation. Figure~\ref{fig:non-watertight} highlights these limitations of existing global methods when applied to open surfaces.

\paragraph{Graph-based Methods}
Schertler et al.~\shortcite{Schertler2017SNO} proposed a graph-based method to orient large-scale point clouds, enabling support for non-watertight models. Their method constructs a normal orientation graph and optimizes flip states globally. Although promising, it remains sensitive to noise and sparsity, which limits its robustness.

\paragraph{Deep Learning Methods}
Deep learning methods have shown promise in learning signed distance fields (SDFs)~\cite{Erler2020Poins2Surf, Ma2021NeuralPull, Wang2022IDF, Ma2023,Li2023NGLO,Li2024} and unsigned distance fields (UDFs)~\cite{Chibane2020NDF,Zhou2022CAPUDF,Ren2023,Zhou2023levelset, Xu2024DEUDF} from point clouds. 

While UDFs are conceptually beneficial for non-watertight surface reconstruction, they cannot determine globally consistent surface orientations. SDF learning, on the other hand, is capable of determining surface orientations. IDF~\cite{Wang2022IDF} requires oriented normals as input. Point2Surf~\cite{Erler2020Poins2Surf} and SHS-Net~\cite{Li2024} are supervised learning methods, and their performance may depend on the training dataset. Neural-Pull~\cite{Ma2021NeuralPull} learns SDFs from raw point cloud input alone. Building on this, NGLO~\cite{Li2023NGLO} introduces a novel two-stage normal estimation pipeline, comprising orientation initialization and direction refinement phases, where its Normal Gradient Learning (NGL) module enhances Neural-Pull’s loss function for orientation initialization. We compare with NGL in our experiments.

\section{Overview}
\label{sec:overview}
Our work addresses the challenge of point cloud orientation in 3D scenes, which are characterized by the following difficulties:

\begin{itemize} \item \textit{Open surfaces}: The point clouds originate from unbounded 3D scenes, making many existing object-level approaches inapplicable.
\item \textit{Sparsity}: The sparse distribution of points reduces the effectiveness of commonly used methods such as propagation-based techniques.
\item \textit{Noise}: Measurement noise further complicates the estimation of consistent normal orientations.
\end{itemize}

\begin{figure*}[!htbp]
    \centering
    \includegraphics[width=1\linewidth]{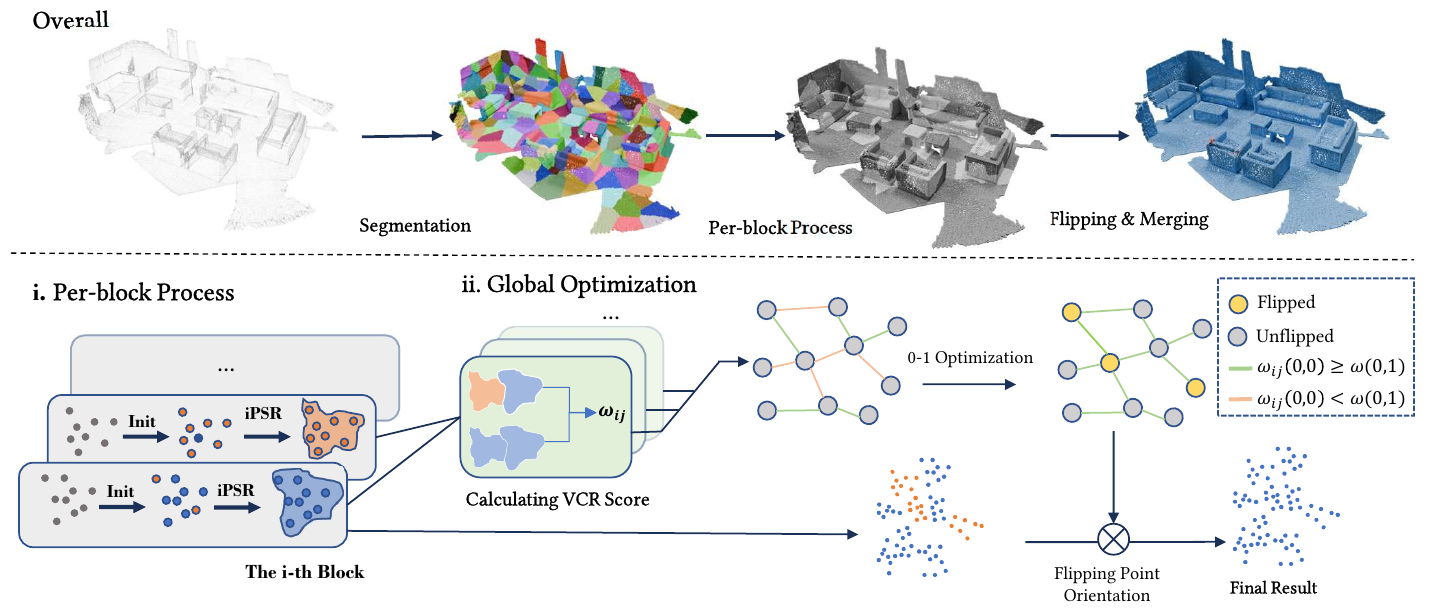}
    \caption{Algorithmic pipeline. First, our method segments the input scene into multiple blocks and orients each block individually.  Block normal orientations are initialized using a randomized propagation method and then refined using iPSR with Neumann boundary conditions. Second, for global orientation, the blocks are represented as nodes in an undirected graph, which is optimized via 0-1 integer-constrained optimization. Edge weights $\omega_{ij}$ are calculated based on a visibility measure derived from visible connected regions.}
    \label{fig:pipeline}
\end{figure*}

Given these challenges, achieving globally consistent orientations in large-scale 3D scenes is particularly difficult. To overcome this, we propose a divide-and-conquer strategy that segments the input point cloud into smaller, geometrically coherent blocks. Each block is oriented individually, and a global 0-1 integer optimization ensures consistency across blocks. The overall pipeline, illustrated in Figure~\ref{fig:pipeline}, consists of the following steps: 

\begin{enumerate}
\item \textit{ Block Segmentation} (Section~\ref{sec:segmentation}):
The input point cloud is partitioned into connected, spatially coherent blocks to facilitate local orientation.

\item \textit{Per-Block Orientation}: Each block is initialized using a randomized propagation method (Section~\ref{subsec:initialization}). To refine orientations, we adapt iPSR~\cite{Hou2022iPSR} with Neumann boundary conditions, making it suitable for open surfaces (Section~\ref{subsec:ipsr}).

\item \textit{Visible Connected Region} (Section~\ref{subsec:vcr}):
We introduce the concept of a visible connected region to assess orientation consistency between adjacent blocks. Orientations within this region must be view-aligned if the block orientations are considered consistent.

\item \textit{Global Consistency Optimization} (Section~\ref{subsec:graph-optimization}):
Using visibility-based consistency as a criterion, we solve a 0-1 integer-constrained optimization problem over the block graph to ensure globally consistent orientation across the entire scene.
\end{enumerate}

Throughout this paper, we assume that the input point cloud represents a scene with a single connected component\footnote{For point clouds with highly non-uniform distributions, regions with sparse sampling may be treated as disconnected from the rest.}. If this assumption does not hold, the definition of consistent visible regions becomes ambiguous and may no longer be well defined.

\section{Per-Block Orientation}
\label{sec:block_orientation}

\subsection{Block Segmentation}
\label{sec:segmentation}

Real-world scenes often exhibit diverse geometries and topologies, making it computationally expensive to process the entire scene as a whole. A natural solution is to adopt a divide-and-conquer strategy by segmenting the scene into smaller, manageable blocks. Dipole~\cite{Metzer2021Dipole} performs this by dividing the scene into regular grids, effectively reducing the problem size. However, this grid-based approach often leads to disconnected components, making consistent normal orientation ambiguous due to undefined relationships between disconnected regions.

Let the input point cloud be denoted by $\mathcal{P} = \{\mathbf{p}_i\}_{i=1}^{n}$.
Our goal is to segment $\mathcal{P}$ into $N$ blocks such that $\mathcal{P} = \bigcup_{i=1}^{N} \mathcal{B}_i$, where each block $\mathcal{B}_i$ forms a spatially connected subset of points and contains a relatively uniform number of points.

To achieve this, we begin by applying a kd-tree-like partitioning to divide the point cloud into small subsets, each containing approximately the same number of points. However, points within a subset may not be spatially connected. To enforce connectivity, we construct a $k$-NN graph (with $k = 10$) by connecting mutually nearest neighboring points. A seed point is then selected from each subset. Starting from these seed points, we grow spatially connected subsets using breadth-first search (BFS) on the $k$-NN graph. Since the input point cloud is assumed to represent a single connected component, this procedure ensures that each resulting subset is spatially connected. After this refinement, each connected subset is referred to as a \textit{block}.

After segmentation, each block is significantly smaller and geometrically simpler than the original model. However, orienting these open surfaces remains challenging. While propagation-based methods can be applied to open surfaces, they often lack robustness due to their greedy nature. Recently, GWN-based methods have gained popularity for their effectiveness in orienting normals, but they are limited to watertight models, as GWN distinguishes between ``inside'' and ``outside'' of a surface using binary values (1 and 0). Therefore, these approaches are not directly applicable to open, non-watertight blocks. 

To address this issue, we adopt a two-step approach for orienting points within each block. First, we initialize point orientations using a dipole-like vector field (Section~\ref{subsec:initialization}). Then we apply an adapted iPSR to further refine orientation accuracy (Section~\ref{subsec:ipsr}).

\subsection{Orientation Initialization}
\label{subsec:initialization}

\begin{figure}[!htbp]
    \centering
    \includegraphics[width=0.9\linewidth]{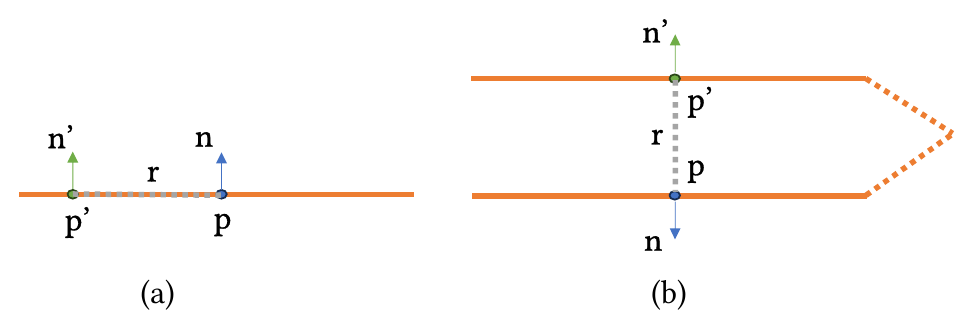}
\caption{Relative orientation between two input points, $\mathbf{p}$ and $\mathbf{p}'$, that are close to each other in terms of Euclidean distance. There are two typical situations: one where the two points are on the same side of a planar face, and the other where the points are on different sides of two parallel faces (often observed in thin structures). In both situations, the normal vectors are parallel, i.e., $\mathbf{n}\parallel \mathbf{n}'$. To determine the relative orientation between $\mathbf{p}$ and $\mathbf{p}'$, we consider the vector $\mathbf{r}=\mathbf{p}'-\mathbf{p}$. (a) If $\mathbf{r}\perp \mathbf{n}$, the orientation of point $\mathbf{p}'$ should match that of point $\mathbf{p}$. (b) If $\mathbf{r}\parallel \mathbf{n}$, the orientation of point $\mathbf{p}'$ should be opposite to that of point $\mathbf{p}$. }
    \label{fig:ppr}
\end{figure}

We begin by estimating the normals of all points using (PCA)~\cite{Hoppe1992}. To determine the relative orientation between two nearby input points, $\mathbf{p}$ and $\mathbf{p}'$, with estimated normals $\mathbf{n}$ and $\mathbf{n}'$, we examine the direction vector $\mathbf{r}=\mathbf{p}'-\mathbf{p}$. As illustrated in Figure~\ref{fig:ppr}, if $\mathbf{r}\parallel \mathbf{n}$, then $\mathbf{n}'$ should be opposite to $\mathbf{n}$. Conversely, if $\mathbf{r}\perp \mathbf{n}$, the orientation of $\mathbf{n}'$ should match that of $\mathbf{n}$. Note that this observation can be extended from planar faces to curved surfaces with small curvatures by relaxing the exact orthogonality and parallelism criteria.

\begin{figure*}[!htbp]
    \centering
    \includegraphics[width=7.0in]{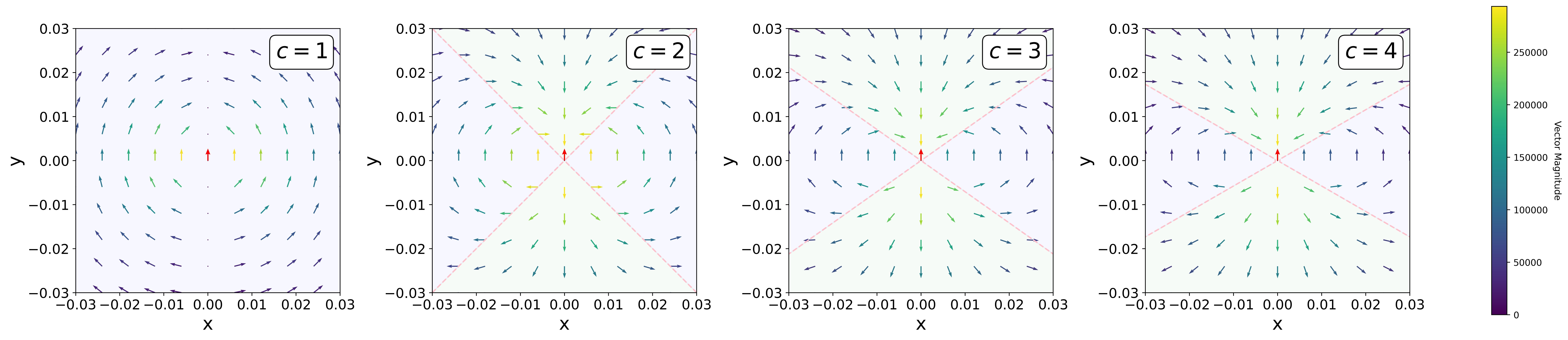}
 \caption{Visualizations of vector fields $\mathbf{F}_{\mathbf{o},\mathbf{n}}$ for the origin $\mathbf{o}=(0,0)^\intercal$ with an upward-pointing normal (red), shown for different values of $c$. Arrows indicate field directions at various spatial locations, while color encodes the magnitude, which decays with increasing distance from $\mathbf{o}$. The red dashed lines separate regions where the vector directions tend to point upward or downward. Note that with higher values of $c$, the wing-like region (light purple) becomes narrower, while with lower values of $c$, the region becomes wider. Therefore, higher values of $c$ are beneficial for orienting thin structures, and lower values of $c$ are beneficial for noise resistance.}
\label{fig:dipole_like_fields}
\end{figure*}

Inspired by this observation, we model point interactions using a dipole-like vector field $\mathbf{F}$. Specifically, the vector field generated by point $\mathbf{p}$ with normal $\mathbf{n}$ is defined as:
\begin{equation}
\label{eqn:electric_field}
\mathbf{F}_{\mathbf{p},\mathbf{n}}(\mathbf{p}') = - \frac{\left(c \,\mathbf{\hat r}\, {\mathbf{\hat r}}^\intercal - \mathbf{I}\right)\mathbf{n}}{\left\|{\mathbf{r}}\right\|^{3}},
\end{equation}
where $\mathbf{r}=\mathbf{p}'-\mathbf{p}$, $\mathbf{\hat r} = \mathbf{r} / \|\mathbf{r}\|$, and $c>1$ is a tunable constant. Vector fields with different values of $c$ are illustrated in Figure~\ref{fig:dipole_like_fields}. When $c = 3$, Equation~(\ref{eqn:electric_field}) reproduces the standard dipole field. The red dashed lines in the figure separate regions where the vector directions tend to point upward or downward. As $c$ increases, the downward regions gradually expand. Increasing $c$ improves the ability to orient extremely thin structures, while smaller values of $c$ enhance noise resistance. In our implementation, we set $c = 2$ for noisy models and $c=4$ for others.

We use the dot product to compute the influence of the vector field $\mathbf{F}_{\mathbf{p},\mathbf{n}}$ on the normal $\mathbf{n}'$ of point $\mathbf{p}'$:
\begin{equation}
e\left((\mathbf{p},\mathbf{n}),(\mathbf{p}',\mathbf{n}')\right) = \mathbf{F}_{\mathbf{p},\mathbf{n}}(\mathbf{p}')\cdot\mathbf{n}'
\end{equation}
Note that $e\left((\mathbf{p},\mathbf{n}),(\mathbf{p}',\mathbf{n}')\right)$ is not symmetric, i.e., $e\left((\mathbf{p},\mathbf{n}),(\mathbf{p}',\mathbf{n}')\right) \neq e\left((\mathbf{p}',\mathbf{n}'),(\mathbf{p},\mathbf{n})\right)$. 
To determine the orientation of each point, let $\mathbf{o}=\{o_i | o_i\in\{0,1\}\}$ represent binary variables indicating whether the normals need to be flipped. The optimal solution is then obtained by maximizing:
\begin{equation}
\label{eqn:init_opt1}
\arg\max_{\mathbf{o}}\sum_{\forall\mathbf{p}_i,\forall\mathbf{p}_j,i \neq j} e \left((\mathbf{p}_i,(-1)^{o_i}\mathbf{n}_i),(\mathbf{p}_j,(-1)^{o_j}\mathbf{n}_j)\right).
\end{equation}
However, directly maximizing Equation~(\ref{eqn:init_opt1}) is computationally prohibitive due to its exponential time complexity. To address this, we adopt a randomized greedy algorithm. Specifically, for each block, we construct an unweighted $k$-NN graph and perform BFS on the graph to generate a spanning tree and derive an ordered point sequence. We then choose the orientation of each point $ \mathbf{p}_j $ in sequence to maximize:
\begin{equation}
\label{eqn:init_opt2}
\arg\max_{o_j}\sum_{\forall \mathbf{p}_i \prec \mathbf{p}_j} e \left((\mathbf{p}_i,\mathbf{n}_i),(\mathbf{p}_j,(-1)^{o_j}\mathbf{n}_j)\right), 
\end{equation}
where $\forall \mathbf{p}_i \prec \mathbf{p}_j $ denotes all points $ \mathbf{p}_i $ that precede $ \mathbf{p}_j $ in the sequence. When the orientations of $\{\mathbf{n}_1,\cdots,\mathbf{n}_i\}$ are fixed, we choose the orientation of $\mathbf{n}_{j}$ to maximize Equation~(\ref{eqn:init_opt2}). This iterative procedure is repeated until all points have been processed.

This process is repeated $M$ times ($M = 5$ in our experiments) to generate a set of orientations for the block point set $\mathcal{P}$. The final solution is obtained by voting over these orientations. However, before voting, the orientations must be aligned, which may require flipping some orientations to ensure consistency across all of them. We use an exhaustive search over $2^M$ possible alignments, as $M$ is small. Alternatively, a 0-1 integer-constrained optimization can be applied for larger values of $M$.

\textbf{Remark.} Xie et al.~\shortcite{Xie2003} and Dipole~\cite{Metzer2021Dipole} are special cases of Equation~(\ref{eqn:electric_field}) with $c=2$ and $c=3$, respectively, and both methods propagate normals greedily. We propose a more general equation that unifies these cases, along with a randomized greedy algorithm to improve robustness in per-block orientation initialization. Subsequently, the adapted iPSR is used to further enhance the orientation accuracy within each block.

\subsection{iPSR Adaption}
\label{subsec:ipsr}

In iPSR~\cite{Hou2022iPSR}, initial normals are either randomly assigned or estimated using a visibility-based heuristic~\cite{Katz2007}. Although visibility initialization typically reduces the number of iterations required for convergence, it is not suitable for non-watertight models. To enhance both convergence speed and robustness, we propose a new strategy specifically designed for non-watertight surfaces. In this subsection, we introduce an adaption of iPSR specifically designed to handle open surfaces.

\begin{figure}[!htbp]
    \centering
    \includegraphics[width=\linewidth]{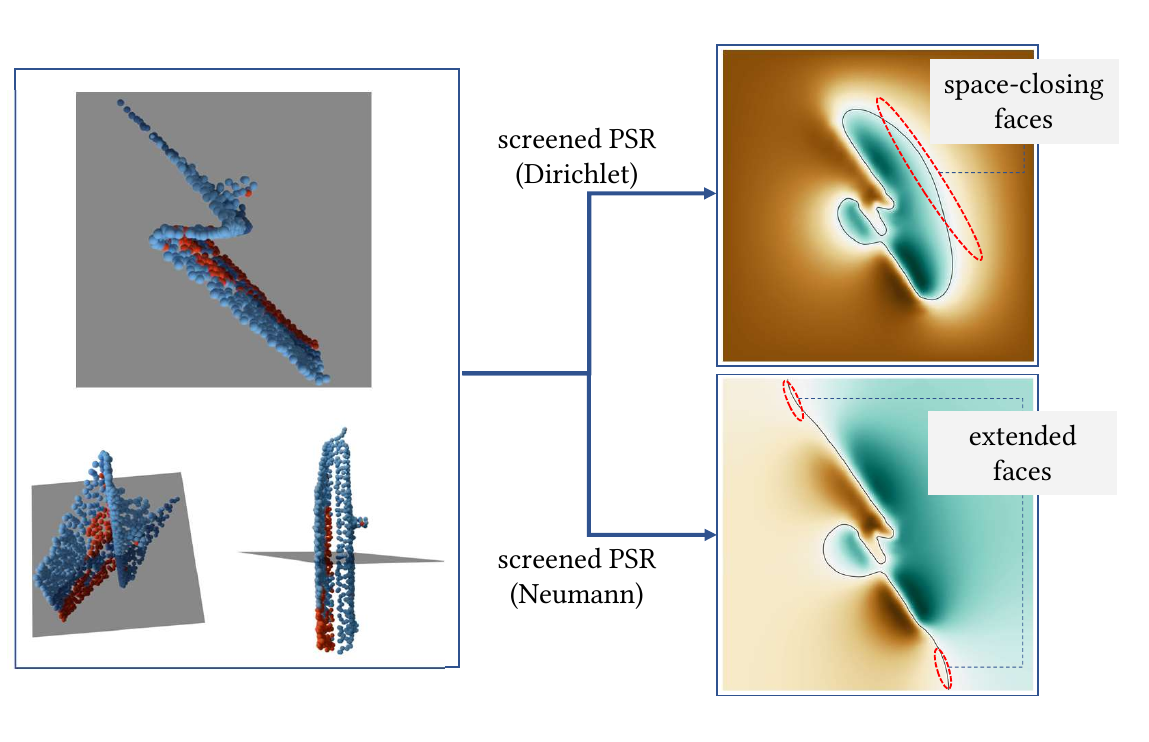}
    \caption{Illustration of the effects of different boundary conditions in sPSR~\cite{Kazhdan2013} on an open block surface with initial normal errors (marked in red). A cross-section of the reconstructed field shown on the right. Using the Dirichlet boundary condition results in a watertight model (top-right), while the Neumann boundary condition preserves the open boundary (bottom-right). The field visualizations indicate that both boundary conditions partition the space into two regions: Dirichlet separates it into ``inside'' and ``outside,'' whereas Neumann divides it into ``left'' and ``right.'' Dirichlet consistently introduces space-closing faces to seal the surface, whereas Neumann avoids this by appending extended faces from the surface boundaries to the bounding box.}
    \label{fig:space_closing_faces}
\end{figure}

\paragraph{iPSR Review} iPSR~\cite{Hou2022iPSR} was originally proposed for orienting normals on closed surfaces using the screened PSR framework~\cite{Kazhdan2013}. Starting from randomly initialized normals, iPSR runs sPSR with Dirichlet boundary conditions to reconstruct an immediate surface mesh $\mathcal{M}$. Correspondences between mesh triangles in $\mathcal{M}$ and the input points $\mathcal{P}$ are established by finding the $k$ nearest neighbors from each face of $\mathcal{M}$ to the point set $\mathcal{P}$. Each point normal in $\mathcal{P}$ is then updated using an area-weighted average of its corresponding faces normals. This iterative process continues until the normals converge.

\paragraph{iPSR Adaptation for Open Surface Orientation} The original iPSR framework is not directly applicable to non-watertight models, which complicates the process of updating normal orientations. Specifically, screened PSR with Dirichlet boundary condition generates a watertight model by partitioning the space into ``inside'' and ``outside'' regions. As illustrated in Figure~\ref{fig:space_closing_faces}, this leads to the creation of a large number of redundant faces, which are referred as space-closing faces, that artificially seal the surface. These faces are inadvertently mapped to the input points,  disrupting the normal update process. While it would be ideal to remove these space-closing faces, distinguishing them from valid reconstructed geometry is non-trivial. 

As demonstrated in \cite{Kazhdan2013}, screened PSR with Neumann boundary condition offers a practical alternative for handling open surfaces. Unlike Dirichlet conditions, which divide the space into ``inside'' and ``outside,'' Neumann conditions partition space into ``top/bottom'' or ``left/right.'' As a result, space-closing faces do not form. Although some extended faces may appear near the boundaries, their impact is minimal and they are easily identifiable. This makes Neumann conditions well-suited for adapting iPSR to open surface reconstruction.

\begin{figure}[!htbp]
    \centering
    \includegraphics[width=\linewidth]{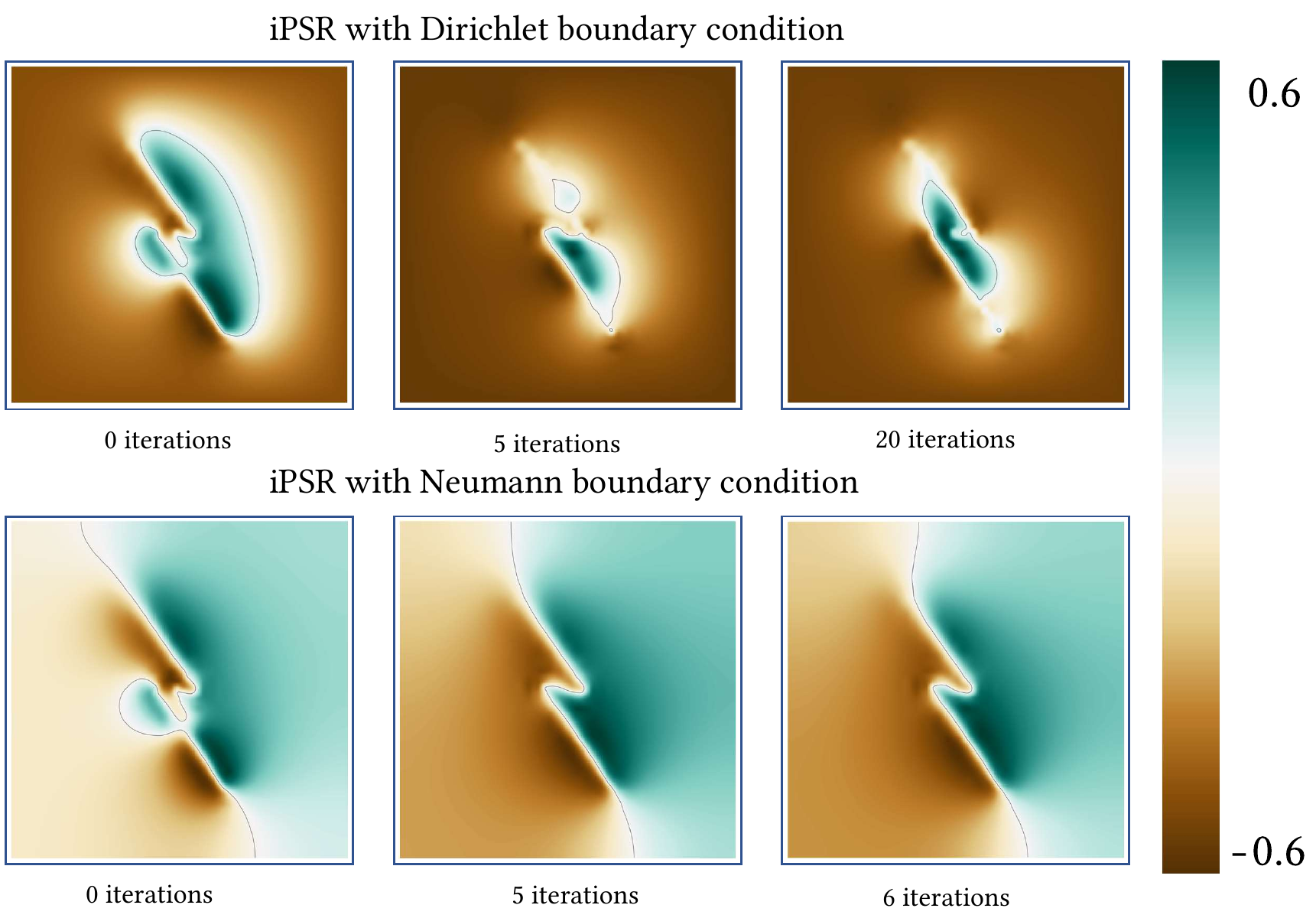}
    \caption{We iteratively apply iPSR with both Dirichlet and Neumann boundary conditions to the point block in Figure~\ref{fig:space_closing_faces}, using the same initial normals obtained from our initialization method. The top row visualizes the indicator function under Dirichlet boundary conditions, which fails to converge even after 20 iterations. In contrast, the bottom row shows the indicator function under Neumann boundary conditions, which converges successfully within 6 iterations.}
    \label{fig:ipsr}
\end{figure}

In our adaptation, we apply the screened PSR with  Neumann boundary conditions in every iteration of iPSR, replacing the Dirichlet variant. Since the extended boundary faces mainly affect only boundary points, we retain them during early iterations and remove them only in the final stages, which specifically when refining the normals near boundaries. These faces are incrementally trimmed from the mesh boundary until they are sufficiently close to the input points. The overall convergence behavior is illustrated in Figure~\ref{fig:ipsr}.

\section{Global Orientation}
\label{sec:global}

After completing per-block orientations, each block has reliable normals. The next step is to align the normal orientations across all blocks. To achieve this, we propose a visibility-based method to evaluate the consistency of adjacent pairs of blocks, as detailed in Section~\ref{subsec:vcr}. The global orientations are then aligned using a 0-1 optimization approach, described in Section~\ref{subsec:graph-optimization}.

\begin{figure}[!htbp]
    \centering
    \subcaptionbox{}{\includegraphics[width=1.1in]{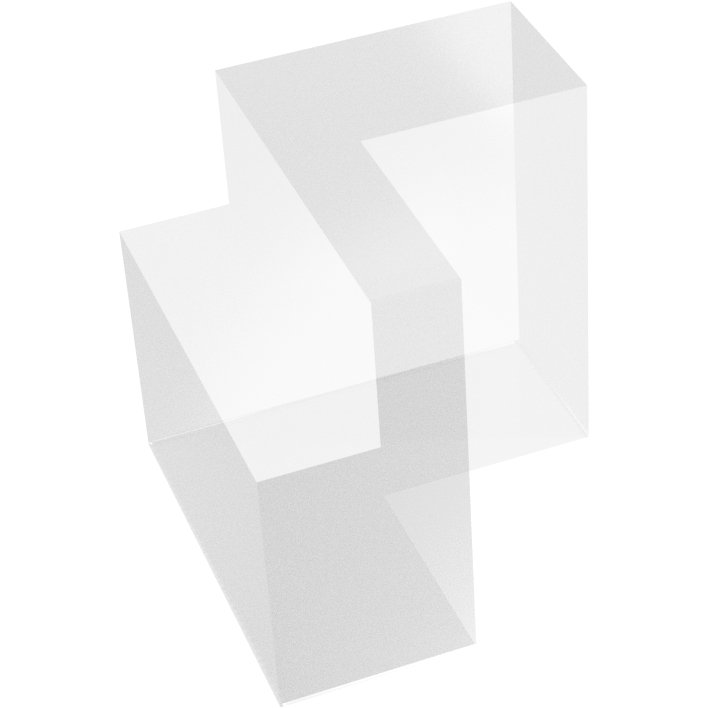}}~%
    \subcaptionbox{}{\includegraphics[width=1.1in]{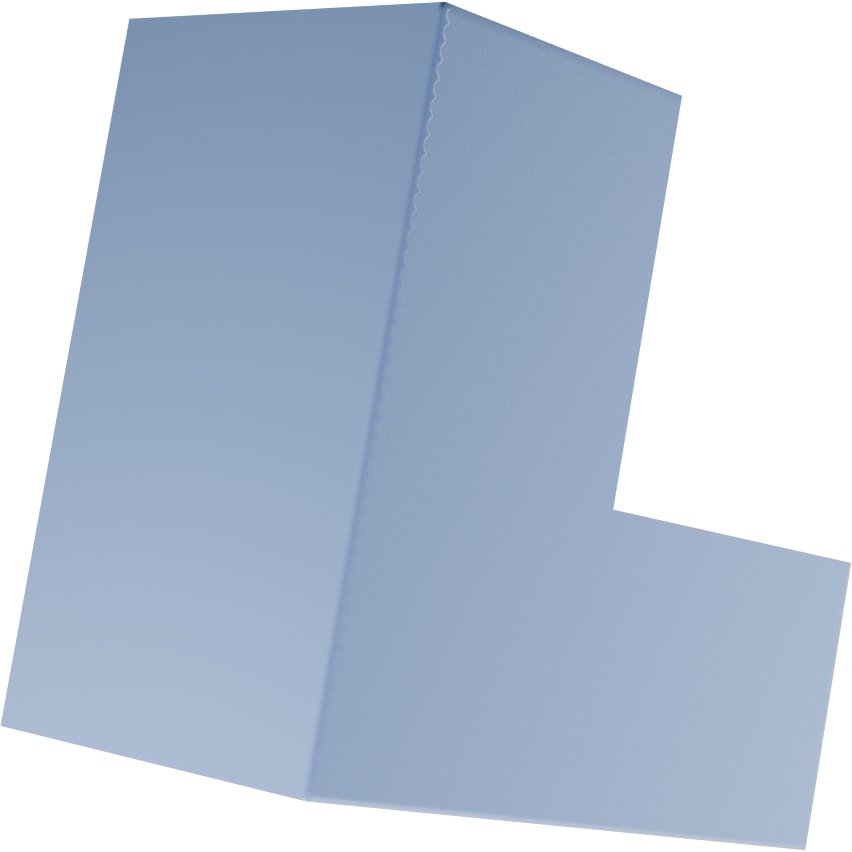}}~%
    \subcaptionbox{}{\includegraphics[width=1.1in]{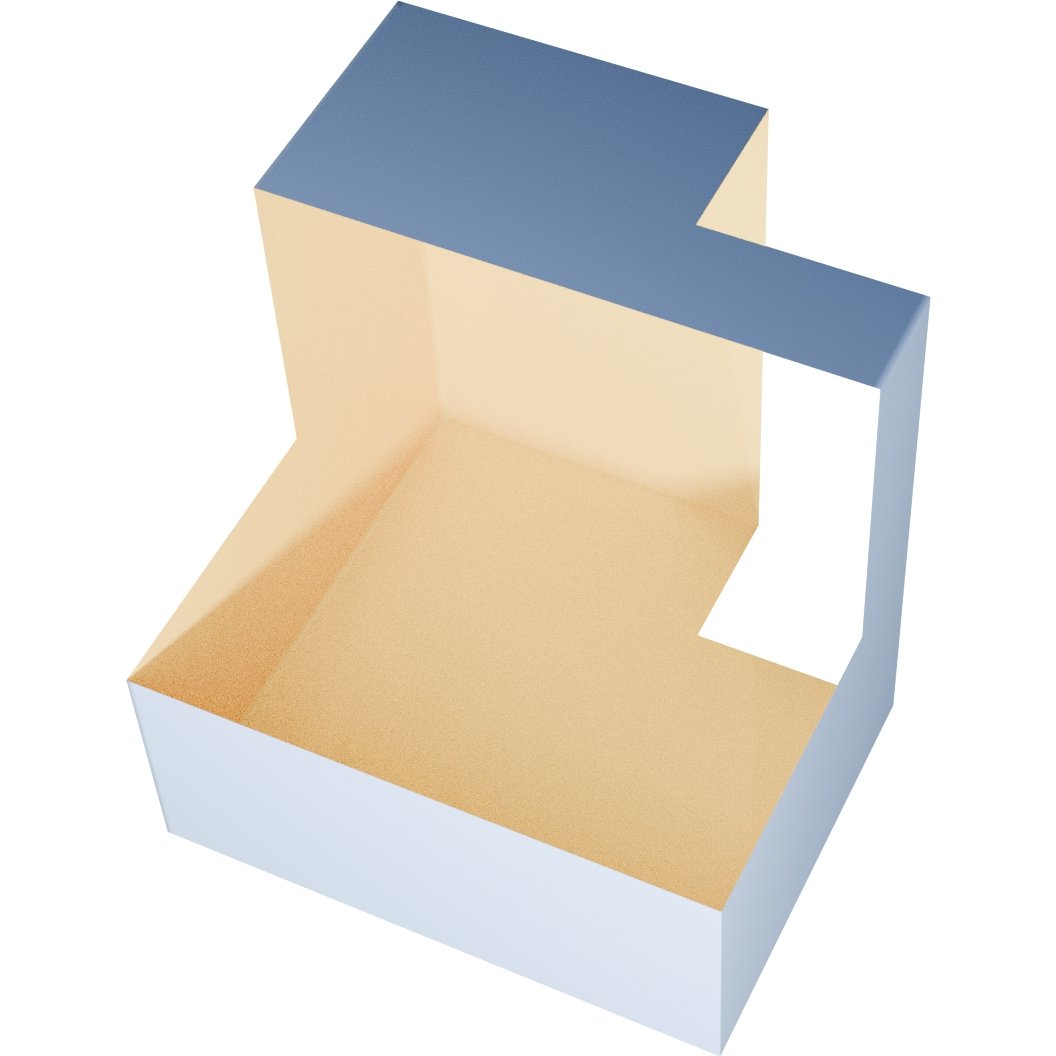}}%
    \caption{View alignment. Consider an orientable open surface rendered in transparent mode in (a). The rendering in (b) is  view-aligned, since only the front side is visible with respect to the viewing direction. In (c), the rendering is not view-aligned, since both the front and back sides of the model are visible.}

    \label{fig:visibility}
\end{figure}

If the orientations of two blocks are correctly aligned, we say that their orientations are \textit{consistent}. For a connected region $R$ (which may be within a single block or span two adjacent blocks), we say that it is  \textit{view-aligned} with respect to a viewing direction $\bf v$ if the signs of $\mathbf{v}\cdot\mathbf{n}_i$ are consistent for all points within the region $R$. 

Note that view alignment relies on the viewing direction, whereas normal consistency does not. Thus, for two blocks to have consistent normal orientations, they do not need to be always view-aligned.  Conversely, while an orientable, open surface is view-aligned under a specific viewing direction, it may not be view-aligned under other viewing directions. Figure~\ref{fig:visibility} illustrates the concept of view alignment, along with an example in which a globally oriented surface that is view aligned under one viewing direction is not necessarily view aligned under other viewing directions. Therefore, given a viewing direction, we need to identify the regions where the orientations must be view-aligned if two adjacent blocks are consistent.

\begin{figure*}[!htbp]
    \centering
    \includegraphics[width=\linewidth]{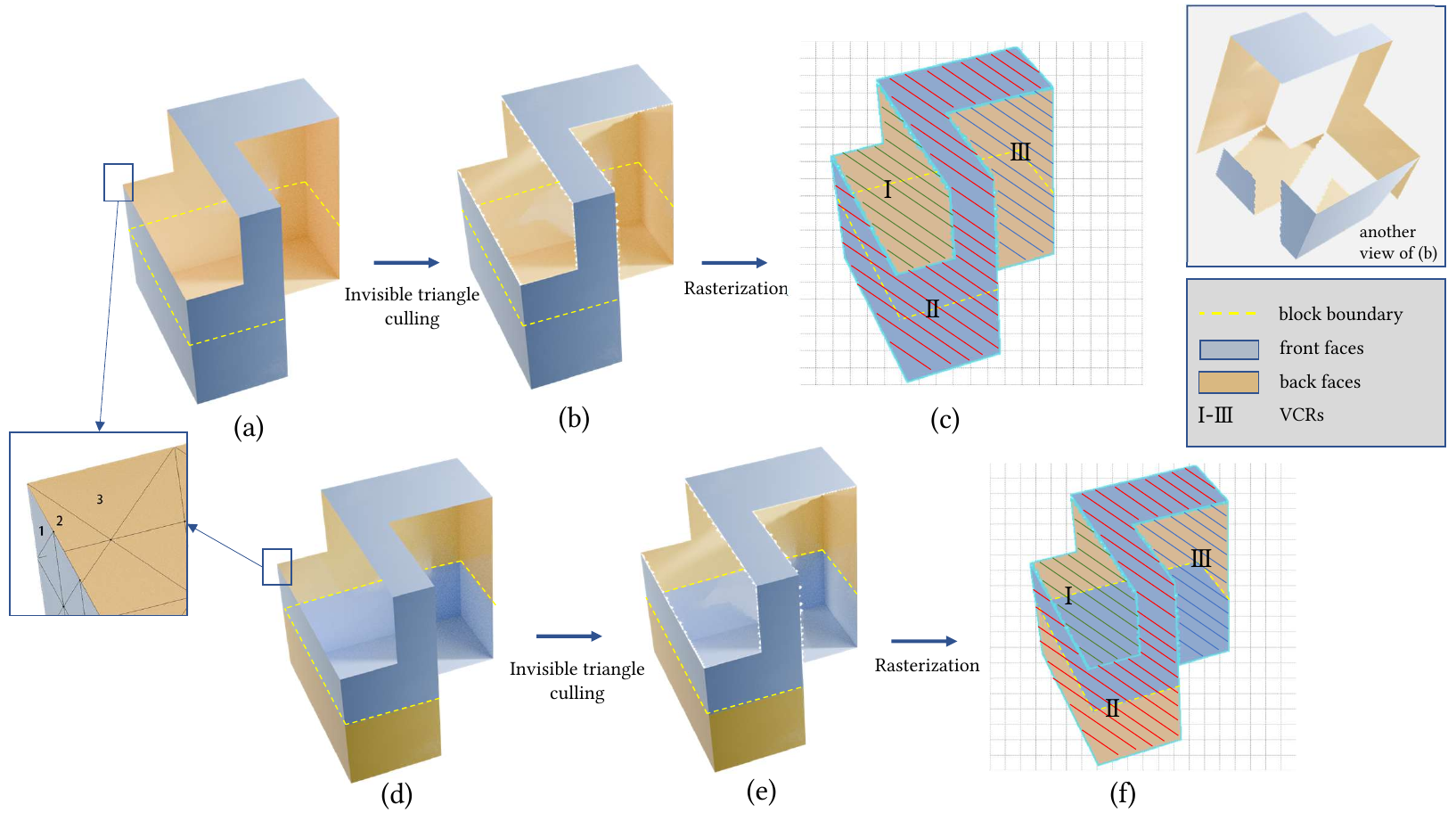}
    \caption{Illustrations of visible connected regions. Two adjacent blocks are separated by yellow dashed lines, with one side colored in yellow and the other in blue. We show two cases: when the two blocks are consistently oriented (a,b,c), and when they are not (d,e,f). To generate the VCRs, we remove invisible triangular faces (b,e). For example, in the zoomed-in inset, triangular face 2 is removed since it is only partially visible under the viewing direction, while faces 1 and 3 are retained, as they are fully visible. The remaining faces are then projected onto the viewing plane, resulting in three VCRs: I, II, and III (c,f). In (c), the two blocks are consistently oriented, so the orientations within each VCR are view-aligned, though the orientations of the three VCRs differ. In (f), where the two blocks are not consistently oriented, the orientation within each VCR may not be view-aligned.}
    \label{fig:vcr}
\end{figure*}

\subsection{Visible Connected Regions}
\label{subsec:vcr}

A \textit{Visible Connected Region} (VCR) on a manifold surface is defined as a geometrically connected region where all points are visible from a particular viewpoint and viewing direction, and the projection of the region onto the viewing plane is also connected. A key observation is that within any VCR, only one side of the surface---either the front or the back---can be seen from the viewpoint, but never both simultaneously. For more details, refer to Appendix~\ref{sec:app_vcr}.

In general, the combination of two adjacent blocks forms a manifold. Based on the visibility observation, if the orientations of two adjacent blocks are consistent, the orientations of all the points in a single VCR within the blocks must be view-aligned. However, the orientations of different VCRs do not necessarily have to be view-aligned. For example, as illustrated in Figure~\ref{fig:vcr}, if the normals of the two adjacent blocks are consistently oriented, the orientation within each VCR is view-aligned, although the orientations of different VCRs may not match. Conversely, if the normals of the two blocks are not consistently oriented, the orientation within each VCR may not be view-aligned

\subsubsection{VCR Generation}
VCRs are view-dependent. Given two adjacent blocks, $\mathcal{B}_1$ and $\mathcal{B}_2$, along with a specific viewpoint and viewing direction, we project $\mathcal{B}_1$ and $\mathcal{B}_2$ perspectively onto a viewing plane, $\mathcal{S}$, which is perpendicular to the viewing direction. This projection is used to generate the VCRs.

As illustrated in Figure~\ref{fig:vcr}, the plane $\mathcal{S}$ is rasterized into pixels with a resolution of $400 \times 400$, equipped with a z-buffer. We project all triangular faces of $\mathcal{B}_1\cup\mathcal{B}_2$ onto the plane $\mathcal{S}$. Using the z-buffer test, for each pixel $q_i\in\mathcal{S}$, we determine the corresponding visible triangular face and its depth. The corresponding faces of two adjacent pixels, $q_i$ and $q_j$, are considered geometrically connected if their depth difference is less than a threshold $\eta$, which is determined by the maximum depth variation within any triangle. Formally, $\eta$ is defined as $\eta=\max_{F\in\mathcal{B}_1\cup\mathcal{B}_2}\left(\max_{\mathbf{v}\in F}D(\mathbf{v})-\min_{\mathbf{v}\in F}D(\mathbf{v})\right)$, where $F$ is a triangular face, $\mathbf{v}$ is a vertex of $F$, and $D(\mathbf{v})$ is the projection depth of vertex $\bf v$. We then derive the projections of the VCRs of the manifold patch $\mathcal{B}_1\cup\mathcal{B}_2$ onto the plane $\mathcal{S}$.

\begin{figure*}[!htbp]
    \centering
    \begin{tabular}{cccccc}  
\textbf{Hoppe} & \textbf{Dipole} & \textbf{Konig} & \textbf{SNO} & \textbf{NGL} & \textbf{DACPO (Ours)} \\
\includegraphics[width=0.15\linewidth]{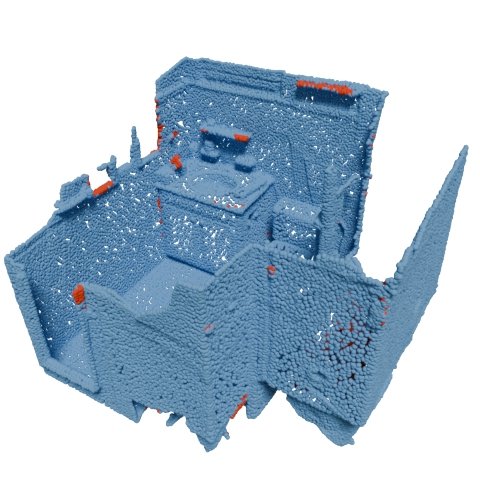} & \includegraphics[width=0.15\linewidth] {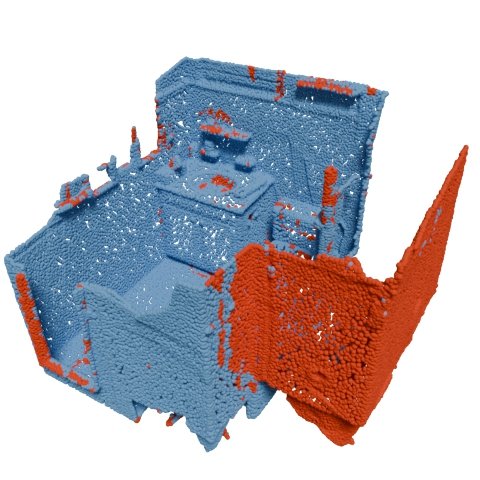} 
& \includegraphics[width=0.15\linewidth]{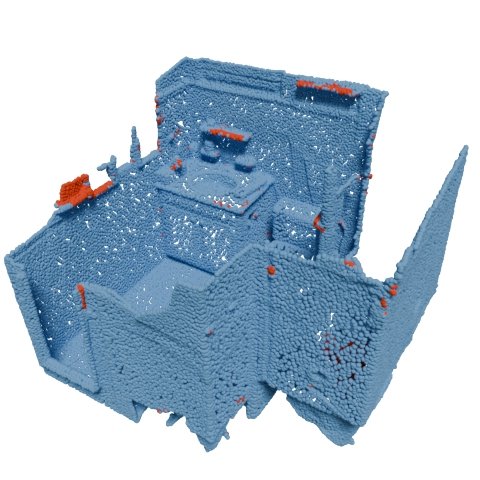} & \includegraphics[width=0.15\linewidth]{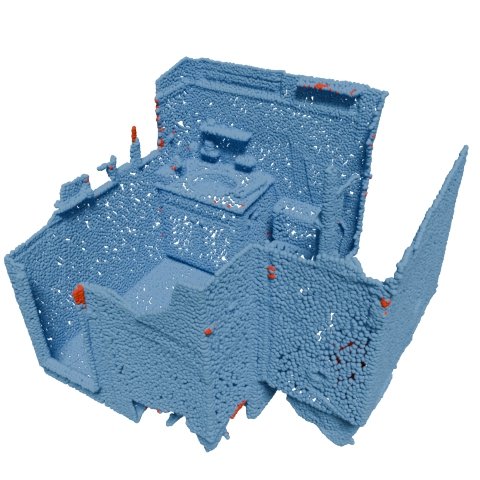} & \includegraphics[width=0.15\linewidth]{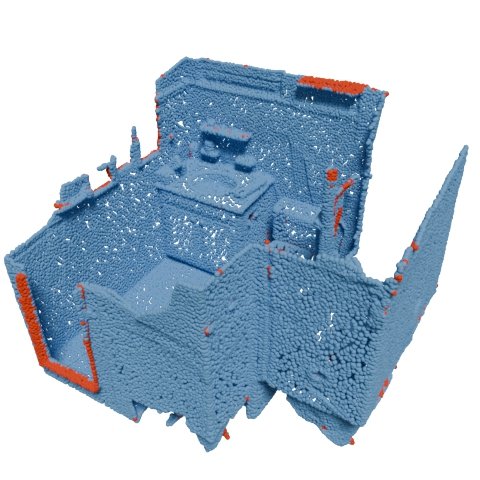} & \includegraphics[width=0.15\linewidth]{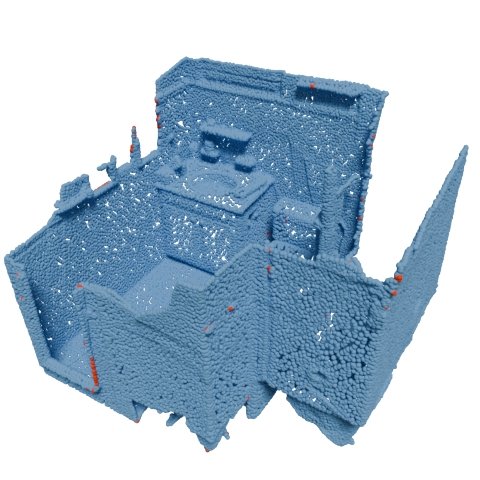} \\ 
\includegraphics[width=0.15\linewidth]{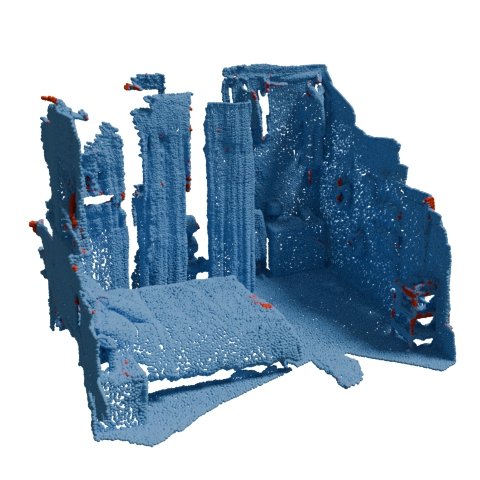} & \includegraphics[width=0.15\linewidth]{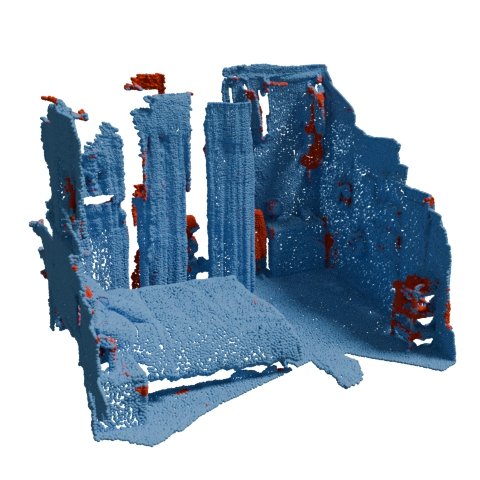} & \includegraphics[width=0.15\linewidth]{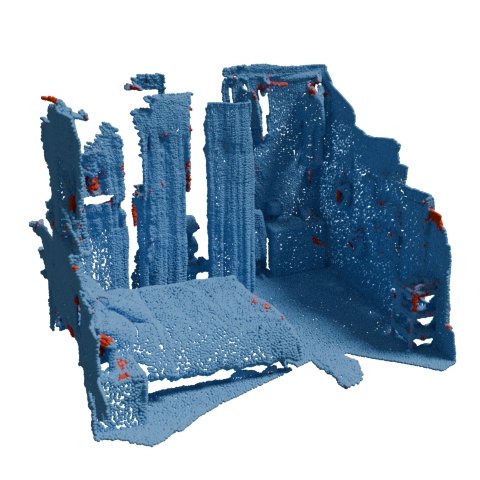} & \includegraphics[width=0.15\linewidth]{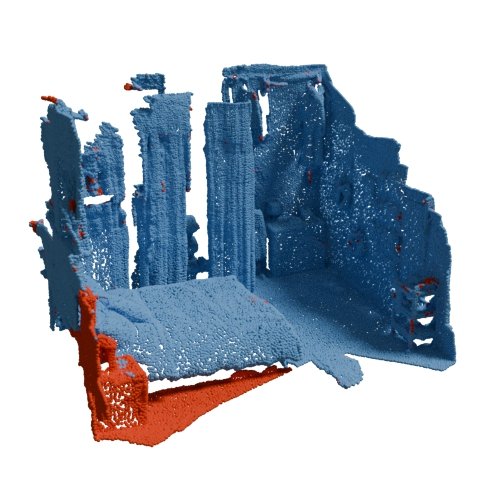} & \includegraphics[width=0.15\linewidth]{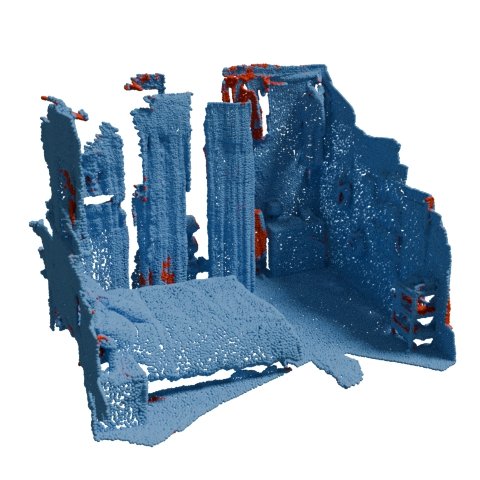} & \includegraphics[width=0.15\linewidth]{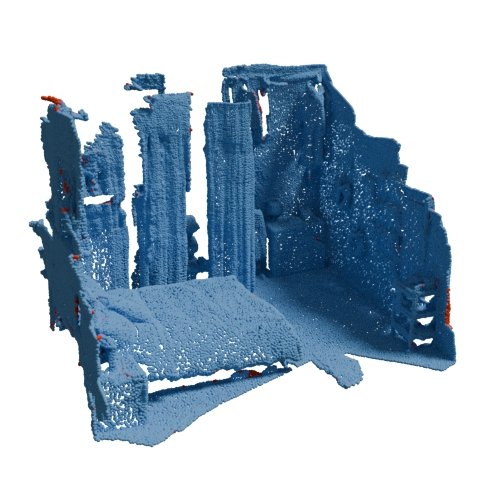} \\ 
\includegraphics[width=0.15\linewidth]{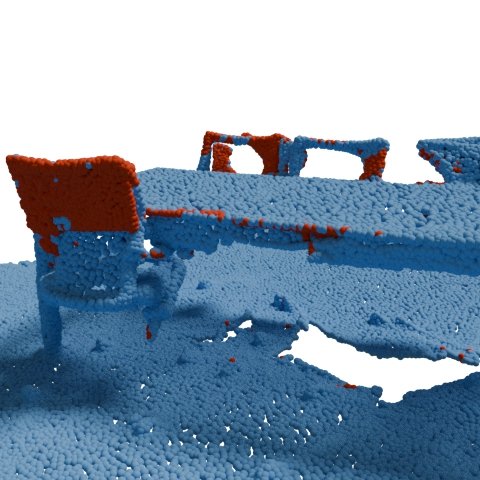} & \includegraphics[width=0.15\linewidth]{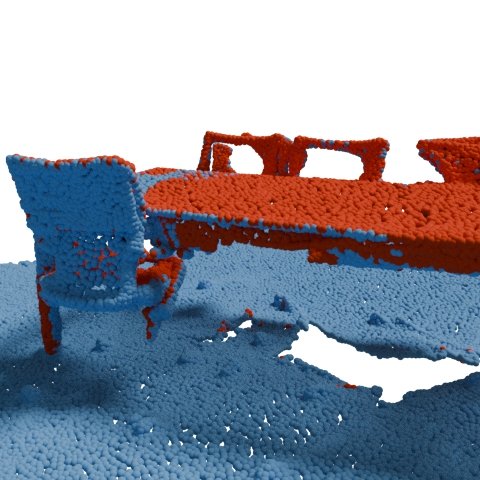} & \includegraphics[width=0.15\linewidth]{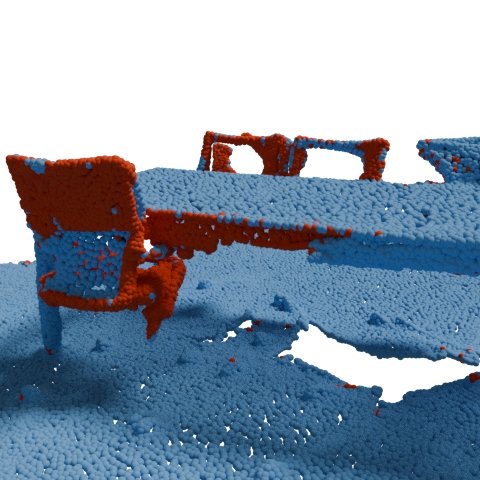} & \includegraphics[width=0.15\linewidth]{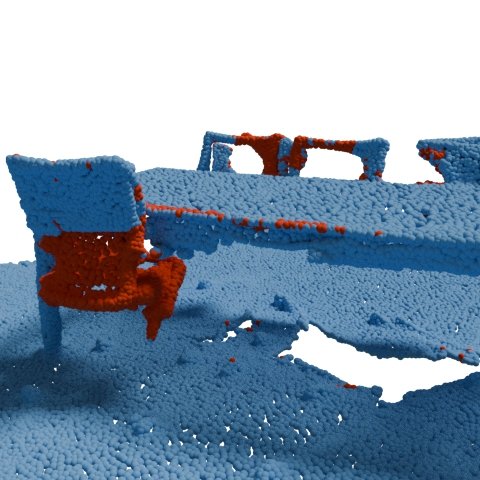} & \includegraphics[width=0.15\linewidth]{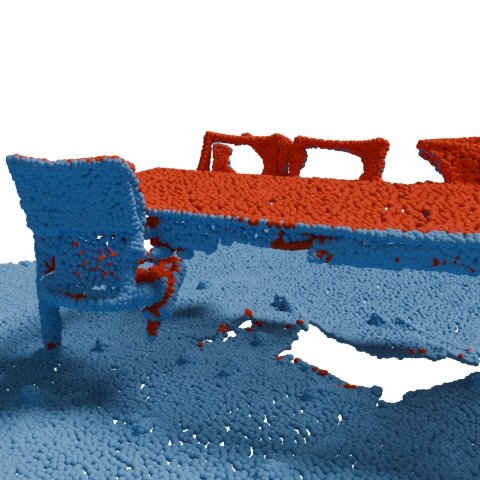} & \includegraphics[width=0.15\linewidth]{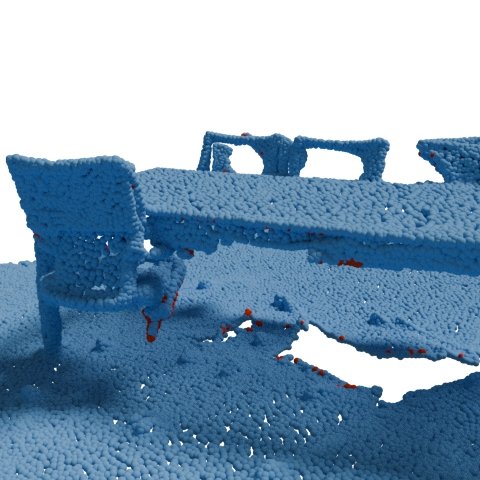} \\ 
\includegraphics[width=0.15\textwidth]{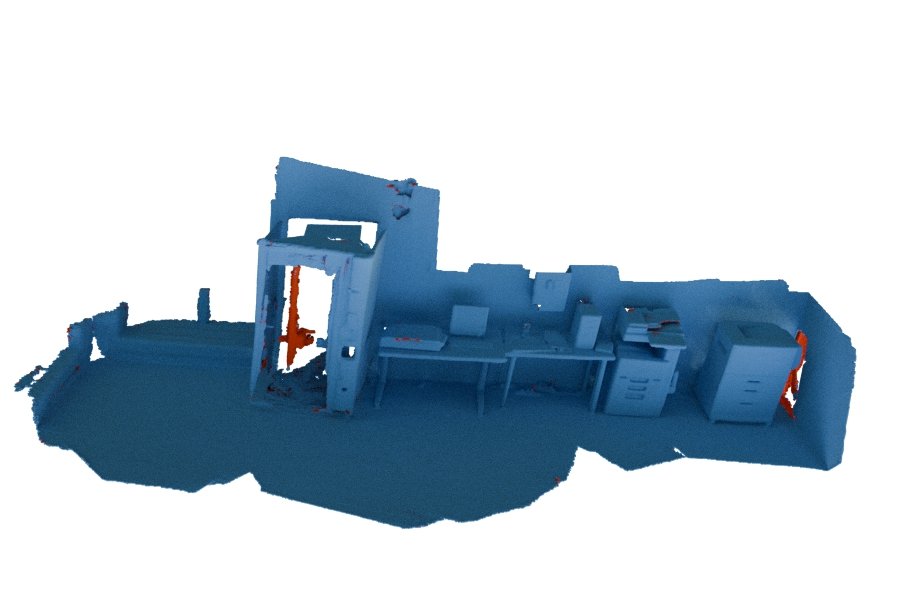} & \includegraphics[width=0.15\textwidth]{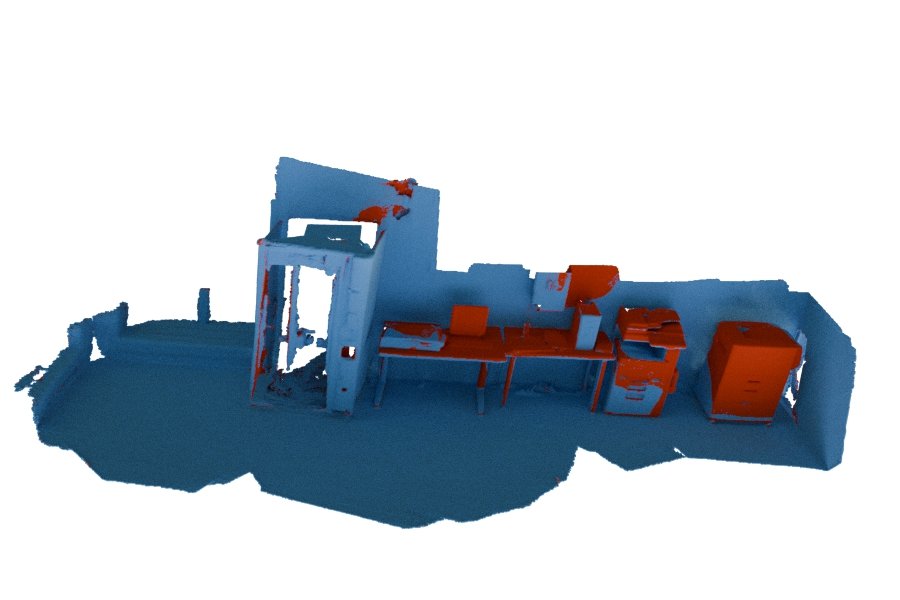} & \includegraphics[width=0.15\textwidth]{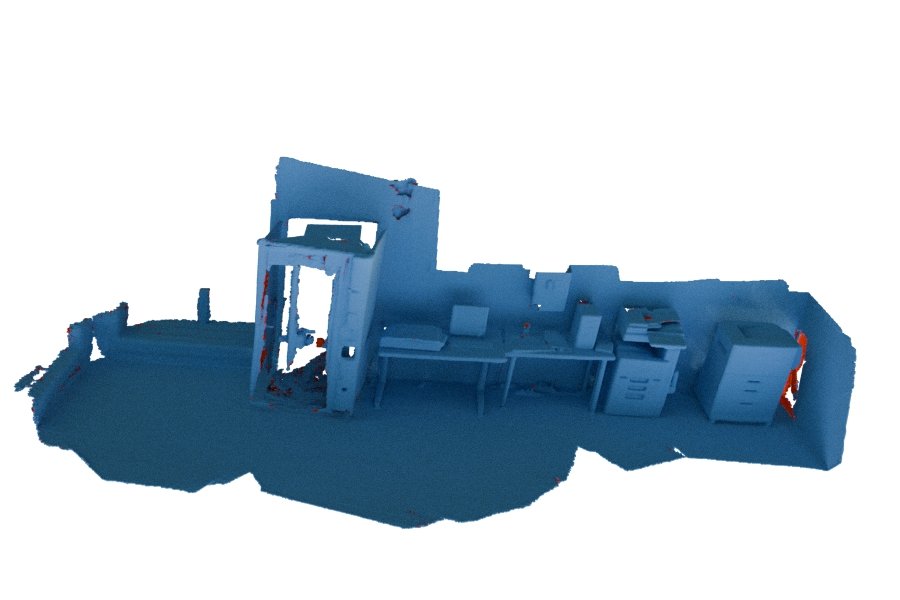} & \includegraphics[width=0.15\textwidth]{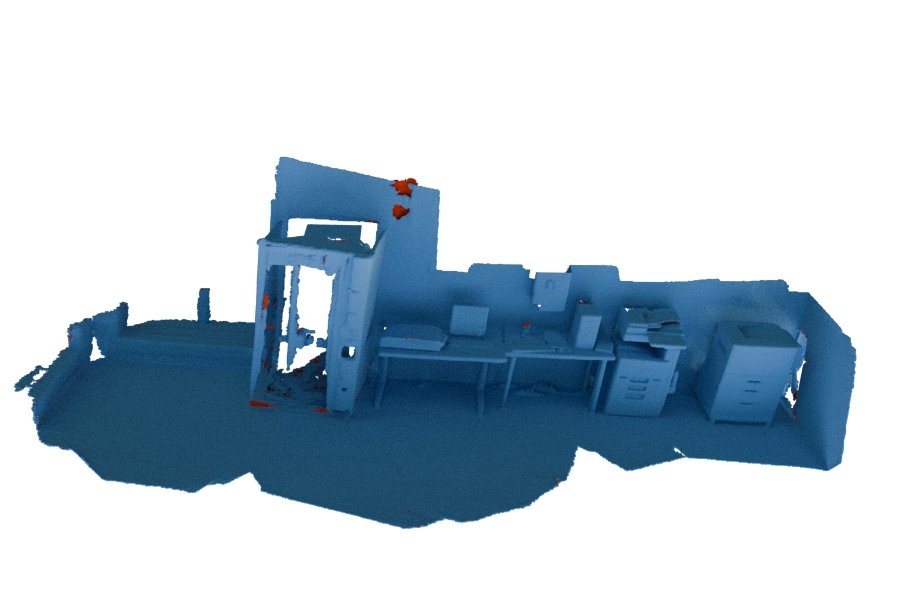} & \includegraphics[width=0.15\textwidth]{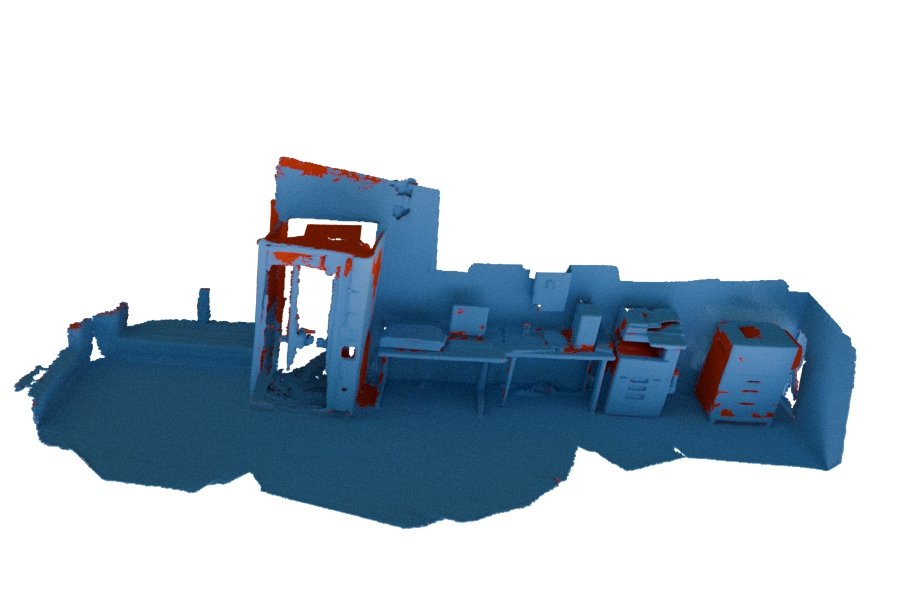} & \includegraphics[width=0.15\textwidth]{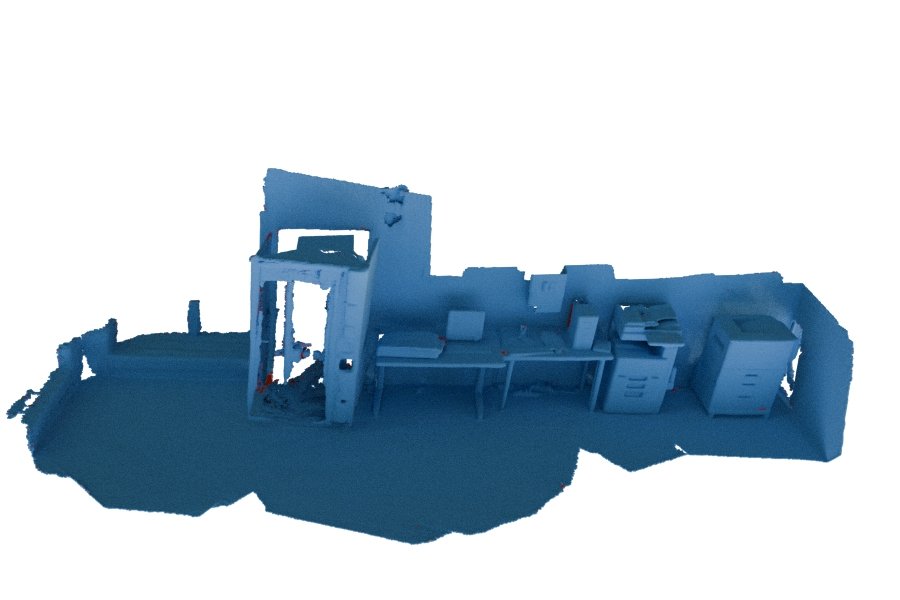} \\ 
\includegraphics[width=0.15\textwidth]{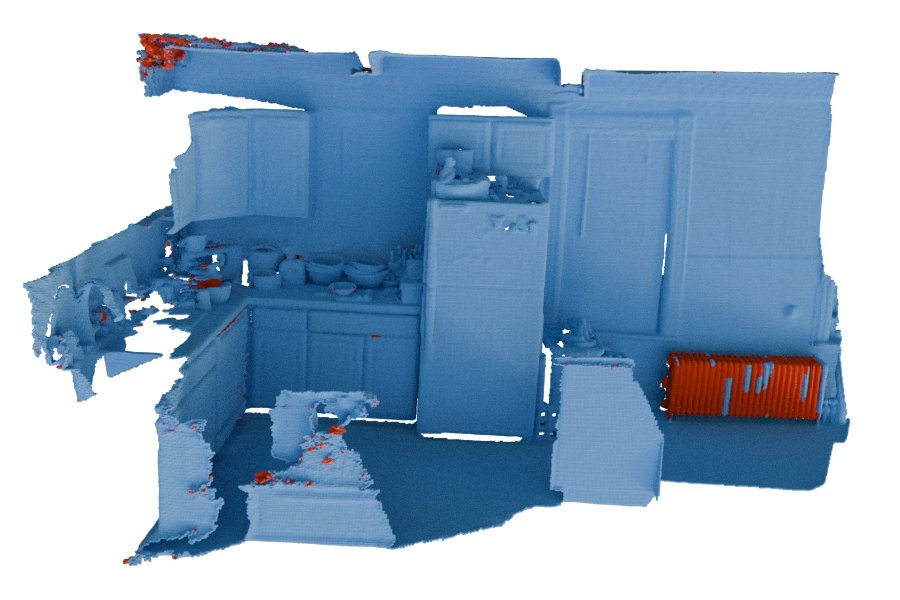} & \includegraphics[width=0.15\textwidth]{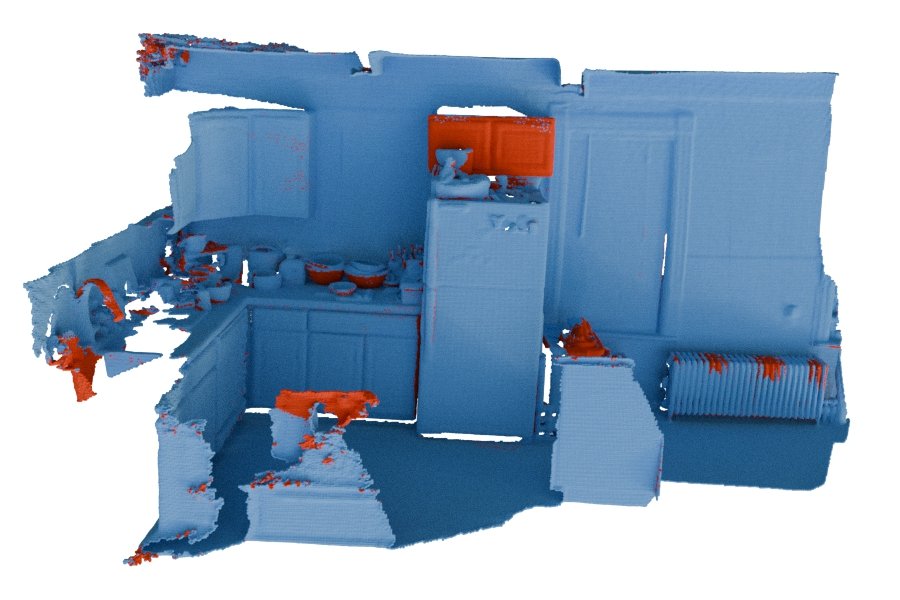} & \includegraphics[width=0.15\textwidth]{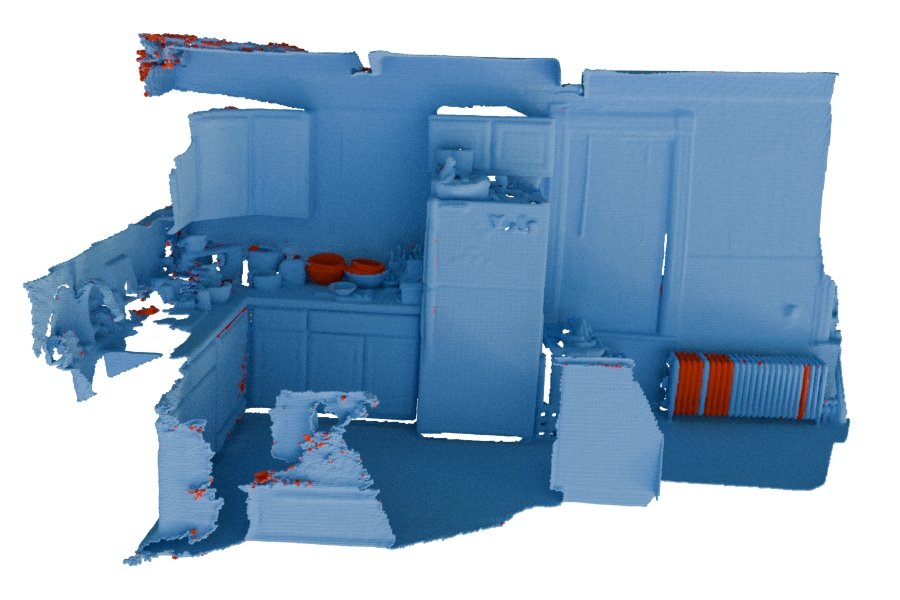} & \includegraphics[width=0.15\textwidth]{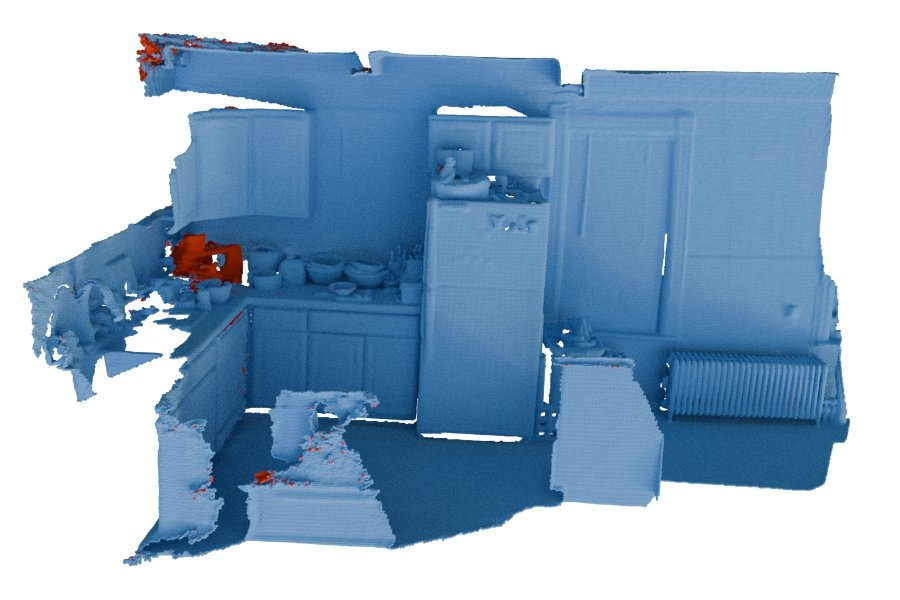} & \includegraphics[width=0.15\textwidth]{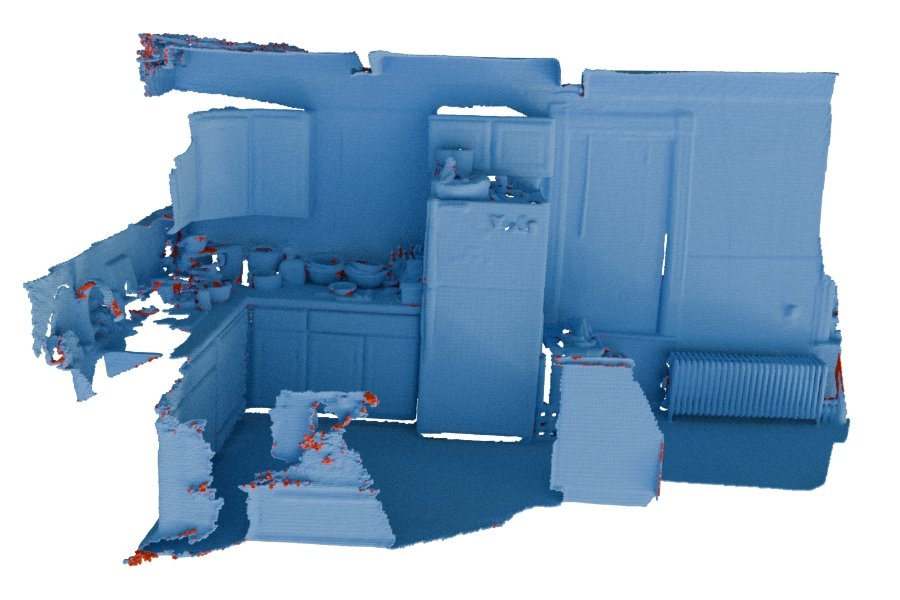} & \includegraphics[width=0.15\textwidth]{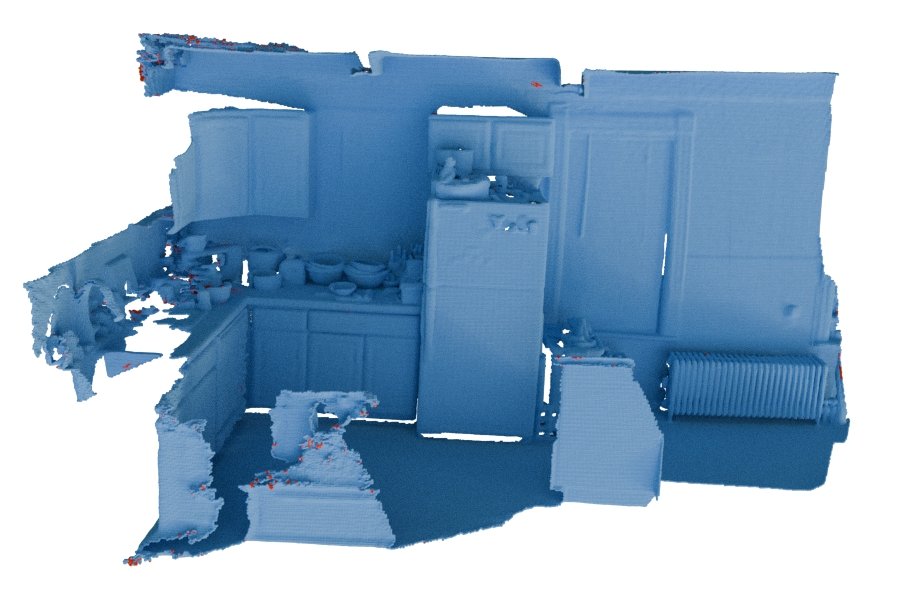} \\ 

\end{tabular}
    \caption{Normal orientation results. Blue points indicate correctly orientated normals, while red points represent incorrect ones. Rows 1--3 show models from ScanNet v2~\cite{Dai2017ScanNet}, with 41K, 10K, and 90K points, respectively. Rows 4--5 show models from SceneNN~\cite{Hua2018}, containing 1,066K and 735K points, respectively.}
    \label{fig:comparison_original}
\end{figure*}

\subsubsection{Orientation Consistency Evaluation} To evaluate the consistency of normal orientations between adjacent blocks, we focus on VCRs that span the boundary between $\mathcal{B}_1$ and $\mathcal{B}_2$ in the image plane. If the orientations of two adjacent blocks are consistent, then any VCR that spans both blocks should be view-aligned. This view-alignment within a VCR can serve as a metric for orientation consistency between the blocks.

After obtaining the VCRs, we divide each VCR into sub-regions, ensuring that each sub-region is view-aligned. Let $\mathcal{V}^\mathcal{S}_k(\mathcal{B}_1\cup\mathcal{B}_2)$ denote the $k$-th sub-region of the VCRs between blocks $\mathcal{B}_1$ and $\mathcal{B}_2$ in the projection plane $\mathcal{S}$. We define $C_1\left(\mathcal{V}^\mathcal{S}_k(\mathcal{B}_1\cup\mathcal{B}_2)\right)$ and $C_2\left(\mathcal{V}^\mathcal{S}_k(\mathcal{B}_1\cup\mathcal{B}_2)\right)$ as the number of pixels corresponding to blocks $\mathcal{B}_1$ and $\mathcal{B}_2$ within the sub-region $\mathcal{V}^\mathcal{S}_k(\mathcal{B}_1\cup\mathcal{B}_2)$. 

Take Figure~\ref{fig:vcr} as an example. In (c), each VCR is fully view-aligned, meaning both $C_1$ and $C_2$ are non-zero. In contrast, in (f), the unaligned VCRs are divided into six sub-regions, where either $C_1=0$ or $C_2=0$.
In addition to pixel counts, we consider the strength of the boundary between adjacent blocks. If the neighbors of a pixel are projections from both $\mathcal{B}_1$ and $\mathcal{B}_2$, we label this as a boundary pixel. The number of such boundary pixels 
 within the sub-region $\mathcal{V}^\mathcal{S}_k(\mathcal{B}_1\cup\mathcal{B}_2)$ is denoted by $C_B\left(\mathcal{V}^\mathcal{S}_k(\mathcal{B}_1\cup\mathcal{B}_2)\right)$, which reflects the strength of the adjacency between the blocks within the sub-region. 

To quantify the view-alignment strength of $\mathcal{V}^\mathcal{S}_k$, we compute:
\begin{displaymath}
\mathcal{I}\left(\mathcal{V}^\mathcal{S}_k(\mathcal{B}_{12})\right) = C_B\left(\mathcal{V}^\mathcal{S}_k(\mathcal{B}_{12})\right) C_1\left(\mathcal{V}^\mathcal{S}_k(\mathcal{B}_{12})\right) C_2\left(\mathcal{V}^\mathcal{S}_k(\mathcal{B}_{12})\right),
\end{displaymath}
where $\mathcal{B}_{12}$ denotes the union of $\mathcal{B}_1$ and $\mathcal{B}_2$. If all pixels in $\mathcal{V}^\mathcal{S}_k(\mathcal{B}_{12})$  correspond to only one block (either $\mathcal{B}_1$ or $\mathcal{B}_2$), then $\mathcal{I}\left(\mathcal{V}^\mathcal{S}_k(\mathcal{B}_{12})\right)=0$, meaning that the sub-region $\mathcal{V}^\mathcal{S}_k(\mathcal{B}_{12})$ does not cross the boundary between $\mathcal{B}_1$ and $\mathcal{B}_2$, and thus contributes nothing to the consistency metric. 

To evaluate the consistency across different viewpoints, we construct a regular dodecahedron and use the centroids of its faces as viewpoints. The view directions are defined as vectors pointing from each viewpoint to the geometric center of the target block. For each viewpoint-direction pair, we generate a corresponding projection plane $\mathcal{S}_j$. The orientation consistency metric $\alpha_{12}$ between blocks $\mathcal{B}_1$ and $\mathcal{B}_2$ is then computed by aggregating the contributions from all projection planes:
\begin{equation}
\label{eq:confidence}
\alpha_{12} = \sum_{j}\sum_{k} \mathcal{I}\left(\mathcal{V}^{\mathcal{S}_j}_k(\mathcal{B}_{12})\right).
\end{equation}

\subsection{Graph Optimization}
\label{subsec:graph-optimization}
To ensure global orientation consistency, we need to determine whether the orientation of each block requires flipping after per-block orientation. Each block has two possible states: flipped or not. Our objective is to find an assignment that maximizes consistency between all pairs of adjacent blocks. This can be modeled as a graph optimization problem.

The graph consists of one node for each block, and edges $ (i, j) \in E $ indicate that the corresponding blocks $ \mathcal{B}_i $ and $ \mathcal{B}_j $ are adjacent. Let $ \mathbf{o} = \{o_i | 1 \leq i \leq N, o_i \in \{0, 1\} \} $ represent the flipping states of each block, where $ o_i = 0 $ means no flip and $ o_i = 1 $ means the block is flipped. The weight of edge $ (i, j) $, which measures the consistency between the two adjacent blocks, is evaluated by $ \alpha_{ij} $. However, $ \alpha_{ij} $ is not normalized to account for the number of block points. To normalize it, we define:
\begin{equation}
\omega_{ij}(o_i,o_j)=\left\{
\begin{aligned}
\frac{\alpha_{ij}}{\alpha_{ij} + \overline{\alpha}_{ij} + \epsilon}, && o_i=o_j\\
\frac{\overline\alpha_{ij}}{\alpha_{ij} + \overline\alpha_{ij} + \epsilon}, && o_i \neq o_j
\end{aligned}
\right.
\end{equation}
where $ \overline{\alpha}_{ij} $ represents the consistency when one of the two blocks is flipped (i.e., flipping either $ \mathcal{B}_i $ or $ \mathcal{B}_j $ yields the same result). The term $\epsilon$ is a small scalar used to avoid division by zero. 

To maximize the consistency between all adjacent blocks, we aim to solve the following optimization problem:
\begin{displaymath}
    \arg\max_{\mathbf{o}}\sum_{(i,j)\in E}\omega_{ij}(o_i,o_j),
\end{displaymath}
which can be reformulated as a 0-1 integer-constrained optimization problem: 
\begin{equation}
\label{eqn:01opt}
\arg\max_{\mathbf{o}}\sum_{(i,j) \in E} \left( (o_i - o_j)^2 \omega_{ij}(0,1) + \left(1 - (o_i - o_j)^2\right) \omega_{ij}(0,0)\right).
\end{equation}

\begin{table*}[!htbp]
\centering
\caption{Ratio of incorrectly oriented normals on the ScanNet v2~\cite{Dai2017ScanNet} and SceneNN~\cite{Hua2018} datasets, expressed as a percentage. ``Original'' refers to the original point clouds in ScanNet V2 and SceneNN. ``Noise1'' and ``Noise2'' denote point clouds in ScaneNet V2 with Gaussian noise applied, having standard deviations of $0.004$ and $0.008$, respectively. ``10K'' and ``100K'' denote downsampled point clouds from SceneNN, containing 10K and 100K points, respectively.}
\begin{tabular}{lcccccccc}
\toprule
 & \multicolumn{4}{c}{ScanNet v2} & \multicolumn{4}{c}{SceneNN} \\
\cmidrule(lr){2-5} \cmidrule(lr){6-9}
Methods & Original & Noise1 & Noise2 & Average & Original & 100K & 10K & Average \\
\midrule
SNO~\shortcite{Schertler2017SNO} & 7.333 & 17.376 & 42.242 & 22.317 & 3.536 & 18.005 & 25.728 & 15.756 \\  
NGL~\shortcite{Li2023NGLO} & 11.846 & 11.886 & 12.329 & 12.020 & 11.683 & 17.939 & 30.972 & 20.198 \\ 
Hoppe et al.~\shortcite{Hoppe1992} & 10.039 & 12.230 & 20.680 & 14.316 & 6.101 & 20.279 & 26.556 & 17.645 \\ 
Dipole~\shortcite{Metzer2021Dipole} & 18.121 & 18.777 & 21.556 & 19.485 & 15.295 & 29.023 & 30.594 & 24.971 \\ 
K\"{o}nig et al.~\shortcite{Konig2009} & 10.553 & 16.598 & 30.139 & 19.097 & 5.615 & 20.912 & 27.830 & 18.119 \\ 
WNNC~\shortcite{Lin2024WNNC} & 15.384 & 19.305 & 29.089 & 21.259 & 10.930 & 25.812 & 27.439 & 21.394 \\ 
\textbf{DACPO (Ours)} & \textbf{5.270} & \textbf{7.689} & \textbf{9.800} & \textbf{7.586} & \textbf{2.430} & \textbf{12.434} & \textbf{15.603} & \textbf{10.156} \\
\bottomrule
\end{tabular}
\label{tab:normal_ratio}
\end{table*}

\section{Experiments}
\label{sec:experiments}

\subsection{Experimental Setup}
\label{subsec:setup}

\paragraph{Implementation }
We use the mixed-integer programming solver from Gurobi 12.0~\footnote{\url{https://www.gurobi.com}} to solve the 0-1 integer optimization problem. The experiments were conducted on a workstation with an Intel Core i9-11900K CPU and 64GB RAM.

\paragraph{Datasets}
We evaluated our method on two indoor scene datasets: ScanNet v2~\cite{Dai2017ScanNet} and SceneNN~\cite{Hua2016SceneNN,Hua2018}. For ScanNet v2, we used the last 106 scenes from a total of 706, based on their sequence numbers. The number of points in these models ranges from 8K to 400K. We introduced two noisy variants, ScanNetV2-Noise1 and ScanNetV2-Noise2, by adding Gaussian noise with standard deviations 0.004 and 0.008, respectively. For SceneNN, we used all 76 models as in~\cite{Hua2018}, with point counts ranging from 400K to 4000K. Furthermore, we used two official downsampled SceneNN datasets\footnote{{\url{https://hkust-vgd.ust.hk/scenenn/downsample/}}} with 100K and 10K points, respectively.
Scene models may contain multiple disconnected components due to the inherent limitations (e.g., occlusion) in point cloud scanning. Since orienting unconnected models is problematic, we extracted the largest connected component from the triangular meshes provided in the original datasets for each model. All subsequent evaluations and comparisons are based on these connected components, rather than the entire scene models. We used the mesh vertices as input and the meshes themselves as ground truths for evaluations.

\paragraph{Baselines} 
We conducted comprehensive comparisons with both classical approaches and state-of-the-art methods, including the propagation-based methods such as Hoppe~\cite{Hoppe1992}, K\"{o}nig~\cite{Konig2009}, and Dipole~\cite{Metzer2021Dipole}; the graph-based method SNO~\cite{Schertler2017SNO}; the winding number-based method WNNC~\cite{Lin2024WNNC}; and the deep learning-based method NGLO~\cite{Li2023NGLO}. NGLO consists of two steps: NGL for normal orientation computation, followed by GVO to refine the orientations. For most datasets, the normal orientations of NGL+GVO are close to those of NGL alone, except for the original SceneNN dataset, where performance deteriorates significantly. The reason for this anomaly remains unclear. Considering this as an outlier, we compare with NGL for fairness. These methods represent a wide range of strategies for normal estimation and orientation of point cloud data, providing a comprehensive evaluation of our method's performance across different paradigms.

\paragraph{Parameters}
We used consistent parameter settings across all experiments, with the exception of the value of $c$ in Equation~(\ref{eqn:electric_field}). Specifically, we set $c=2$ for the ScanNetv2-Noise1 and ScanNetV2-Noise2 datasets, and $c=4$ for all other datasets. All other parameters remain the same across experiments. Each model was partitioned into 200 blocks. For iPSR, we set the octree depth to 10 and the maximum number of iterations to 20.

\paragraph{Metrics}
We evaluated the quality of both normal orientations and reconstructed surfaces. For normal orientations, we calculated the ratio of incorrectly oriented normals, following the approach in~\cite{Xu2023GCNO} and \cite{Yang2024NRSC}. A normal is considered correctly oriented if the angle between the the predicted normal and the ground truth normal is less than $90^\circ$; otherwise, it is deemed incorrect. Since the scenes are non-watertight, ground truth normals are still considered valid even if flipped. If more than $50\%$ of the normals are incorrectly oriented, we flipped the ground truth normals. Thus, the maximum ratio of incorrectly oriented normals is capped at $50\%$. For surface reconstruction, we applied screened PSR~\cite{Kazhdan2013}. Since screened PSR typically extends surface boundaries to the bounding box, we trimmed redundant faces that are far from the input points. Finally, we computed the Chamfer distance (CD) to the ground truth mesh to assess the quality of  reconstructed surfaces.

\begin{table}
    \caption{Chamfer distances of the reconstructed meshes on the ScanNet v2~\cite{Dai2017ScanNet} dataset. ``Noise1'' and ``Noise2'' correspond to the two noisy variants, with Gaussian noises applied at standard deviations of 0.004 and 0.008, respectively.}
    \label{tab:chamfer}
    \centering
    \begin{tabular}{cccc}
        \toprule
        Mehtods & Original & Noise1 & Noise2 \\
        \midrule
        SNO~\shortcite{Schertler2017SNO} & 0.011910 & 0.012284  & 0.014963\\
        NGL~\shortcite{Li2023NGLO} & 0.012319 &0.012623  & 0.013370 \\
        Hoppe et al.~\shortcite{Hoppe1992} & 0.012008  &0.012426  & 0.013722 \\
        Dipole~\shortcite{Metzer2021Dipole} & 0.012304 & 0.012647 &0.013548 \\
        K\"{o}nig et al.~\shortcite{Konig2009} & 0.011997 &  0.012550 &0.014263 \\
        WNNC~\shortcite{Lin2024WNNC} &0.011969  &0.012438  &0.014029 \\
        \textbf{DACPO (Ours)} & \textbf{0.011669} &\textbf{0.011957}  &\textbf{0.012882} \\
        \bottomrule
    \end{tabular}
\end{table}

\subsection{Results}
\label{sec:results}

Real-world 3D scene data often comes with various imperfections, including noise, outliers, misalignment from multiple views, and occluded or missing regions. Additionally, these indoor scenes feature numerous complex local structures, such as tables, chairs, and cabinets, where weak connections and thin structures frequently arise. Moreover, due to the limitations of scanning devices, objects in these datasets are often incomplete or fragmented. All of these artifacts make the task of orientation particularly challenging.

In this subsection, we first present experimental results on the original point clouds from ScanNet V2~\cite{Dai2017ScanNet} and SceneNN~\cite{Hua2018}. We then report results on two variants: one with Gaussian noises applied and the other with downsampling to induce sparsity, representing more challenging scenarios. Finally, we discuss the issues of weak connections.

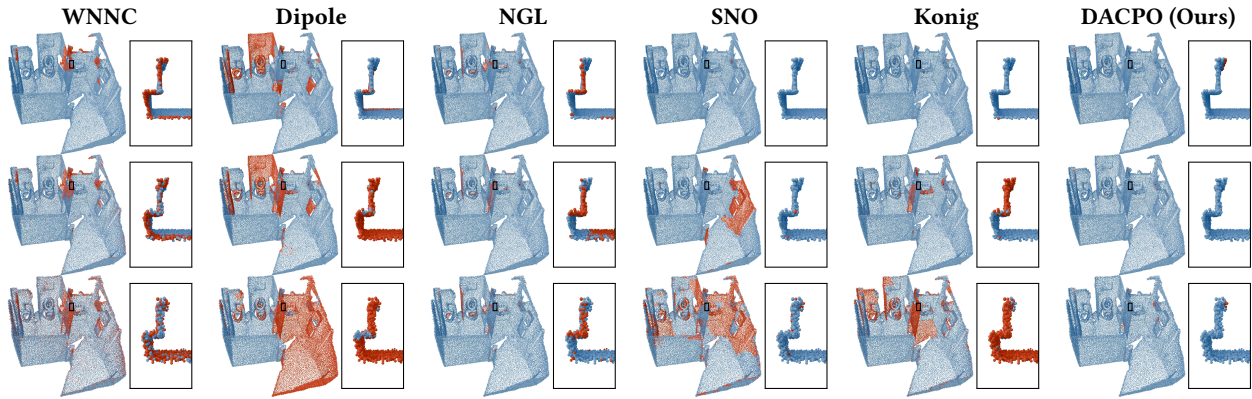
\begin{figure*}[!htbp]
    \centering
    \begin{tabular}{cccccc}
\textbf{WNNC} & \textbf{Dipole} & \textbf{NGL} & \textbf{SNO} & \textbf{Konig} & \textbf{DACPO (Ours)} \\

\begin{tikzpicture}[baseline=(current bounding box.center)]
    \node[anchor=south west,inner sep=0] (far) at (0,0) {\includegraphics[width=0.09\linewidth]{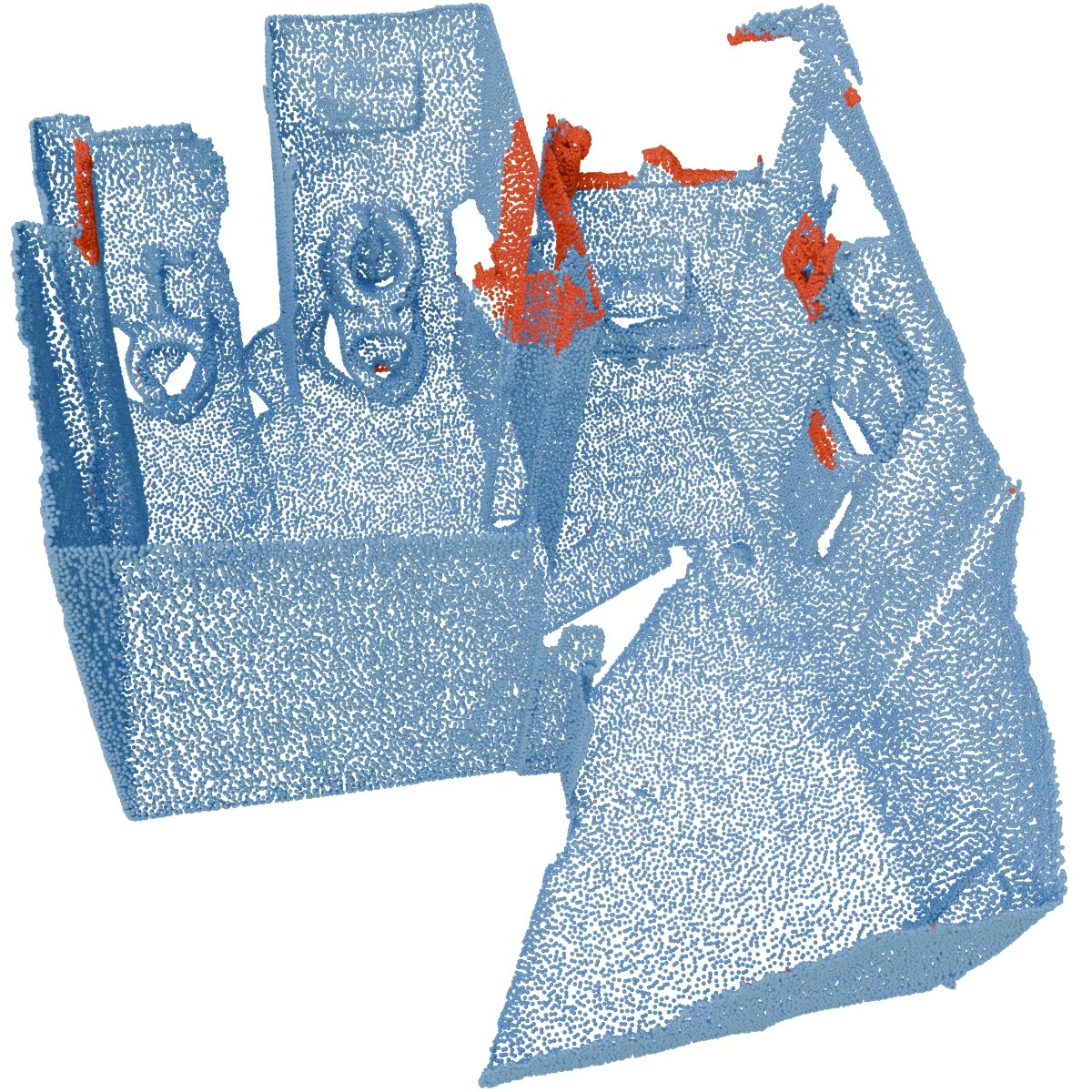}};
    \draw[black,thin]  (0.84,1.15) rectangle (0.89,1.25);
    \node[anchor=west,inner sep=0] (near) at (far.east) {\fcolorbox{black}{white}{\includegraphics[width=0.045\linewidth]{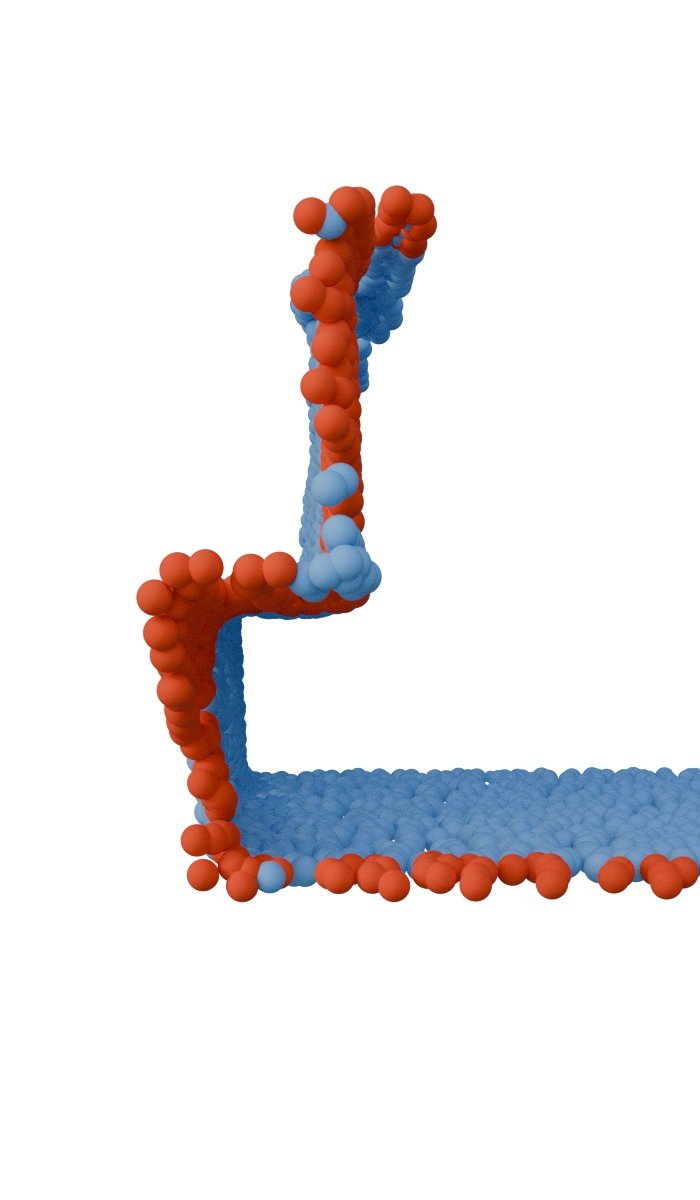}}};
\end{tikzpicture} &
\begin{tikzpicture}[baseline=(current bounding box.center)]
    \node[anchor=south west,inner sep=0] (far) at (0,0) {\includegraphics[width=0.09\linewidth]{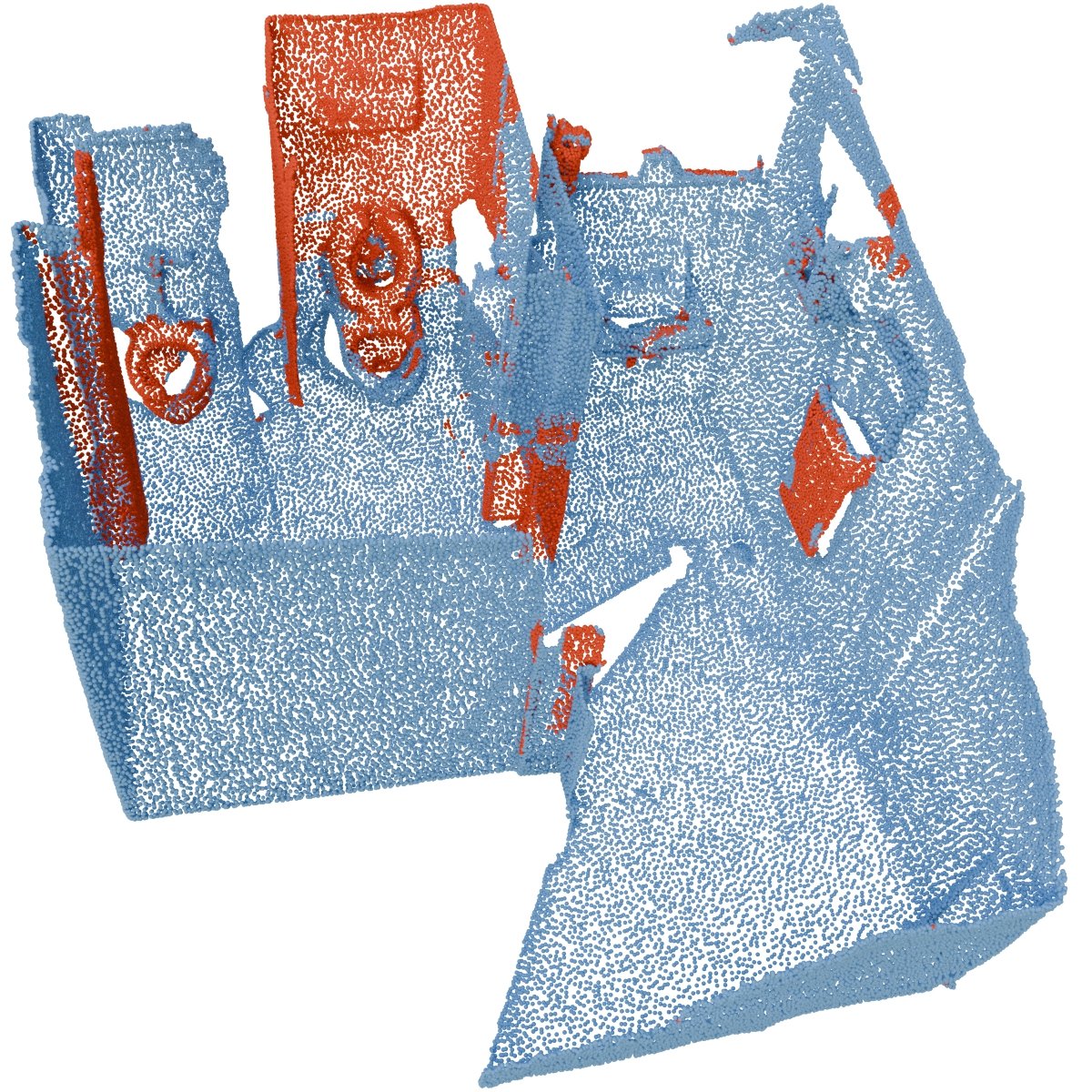}};
    \draw[black,thin]  (0.84,1.15) rectangle (0.89,1.25);
    \node[anchor=west,inner sep=0] (near) at (far.east) {\fcolorbox{black}{white}{\includegraphics[width=0.045\linewidth]{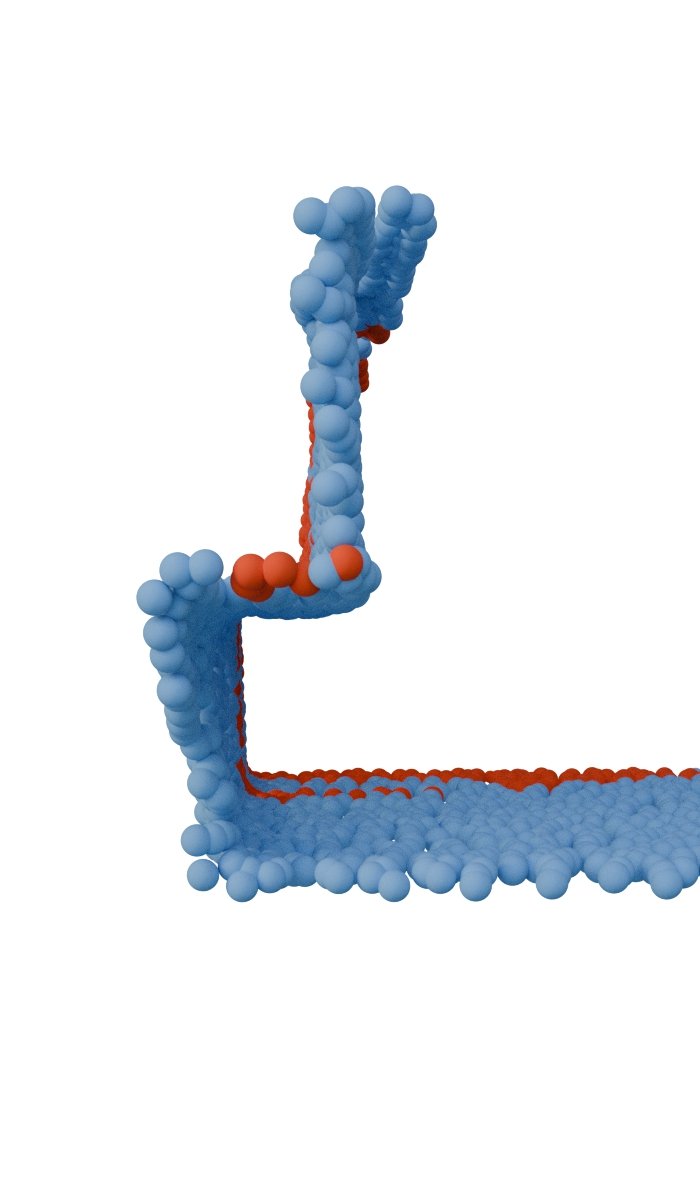}}};
\end{tikzpicture} &
\begin{tikzpicture}[baseline=(current bounding box.center)]
    \node[anchor=south west,inner sep=0] (far) at (0,0) {\includegraphics[width=0.09\linewidth]{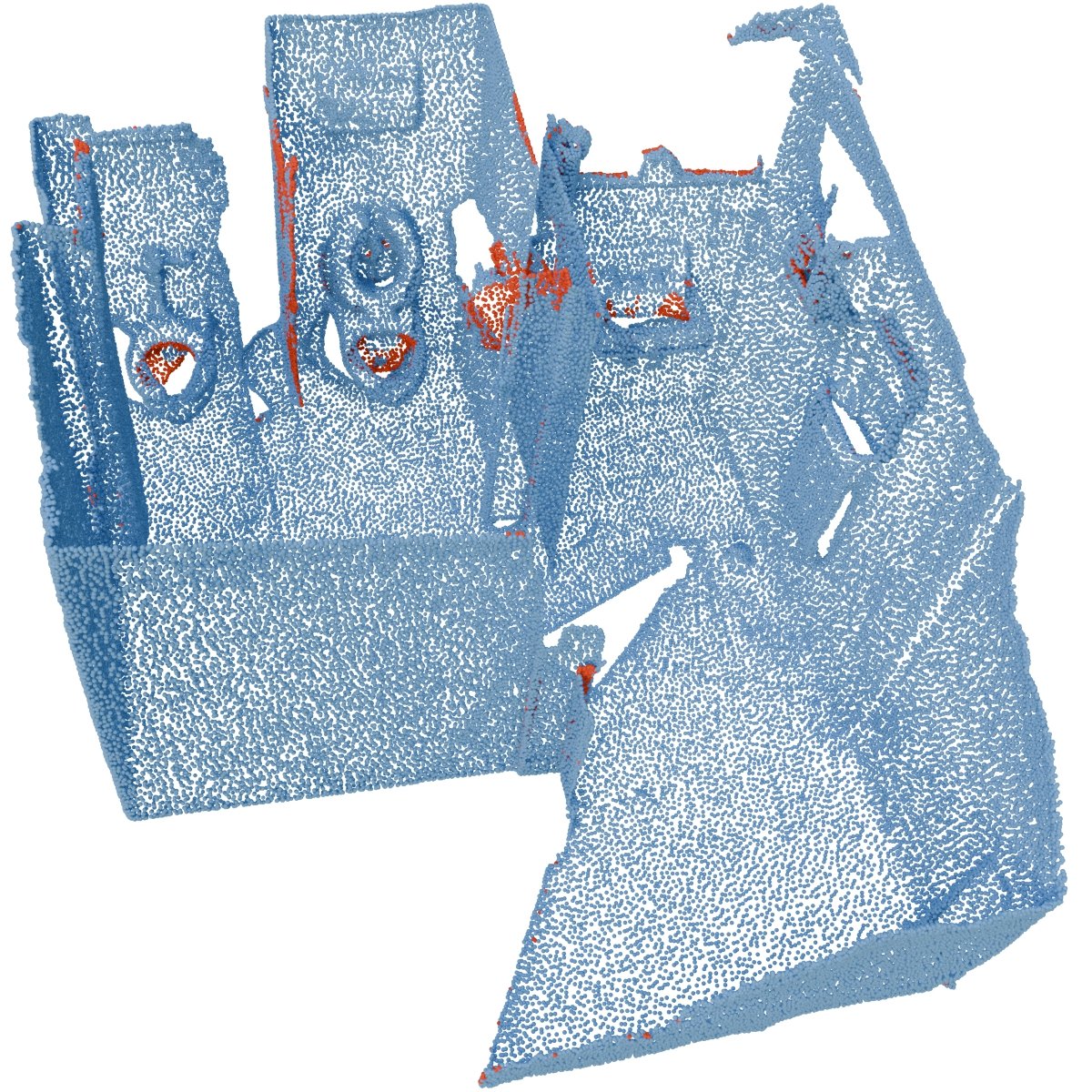}};
    \draw[black,thin]  (0.84,1.15) rectangle (0.89,1.25);
    \node[anchor=west,inner sep=0] (near) at (far.east) {\fcolorbox{black}{white}{\includegraphics[width=0.045\linewidth]{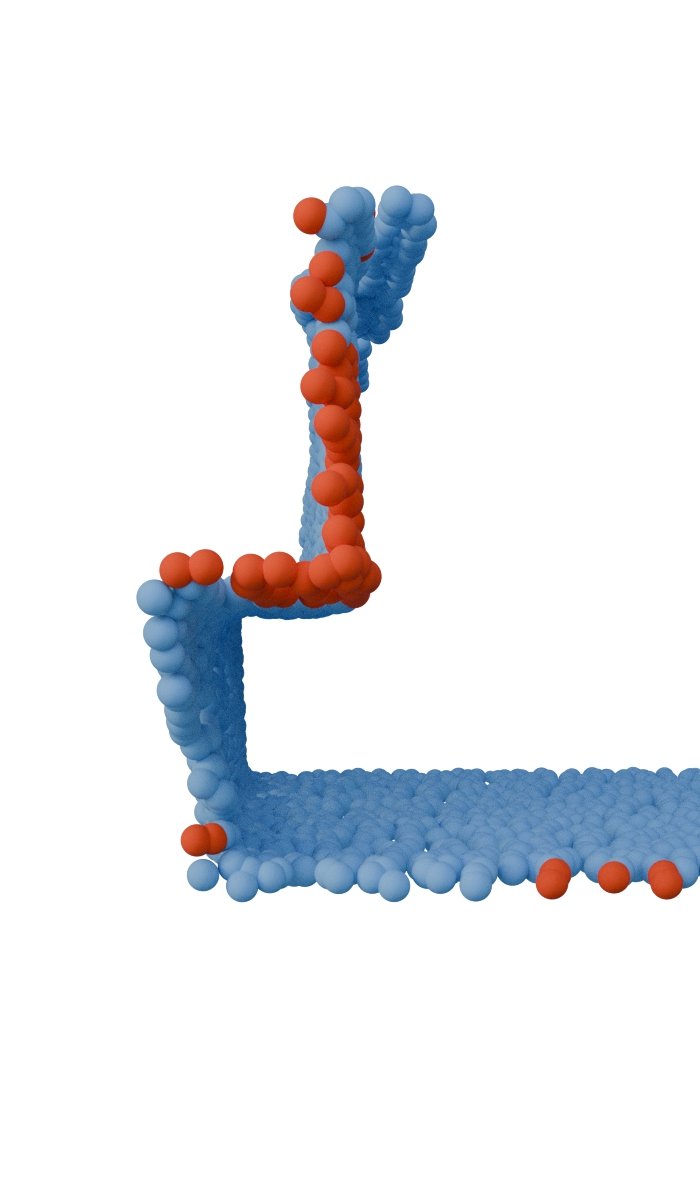}}};
\end{tikzpicture} &
\begin{tikzpicture}[baseline=(current bounding box.center)]
    \node[anchor=south west,inner sep=0] (far) at (0,0) {\includegraphics[width=0.09\linewidth]{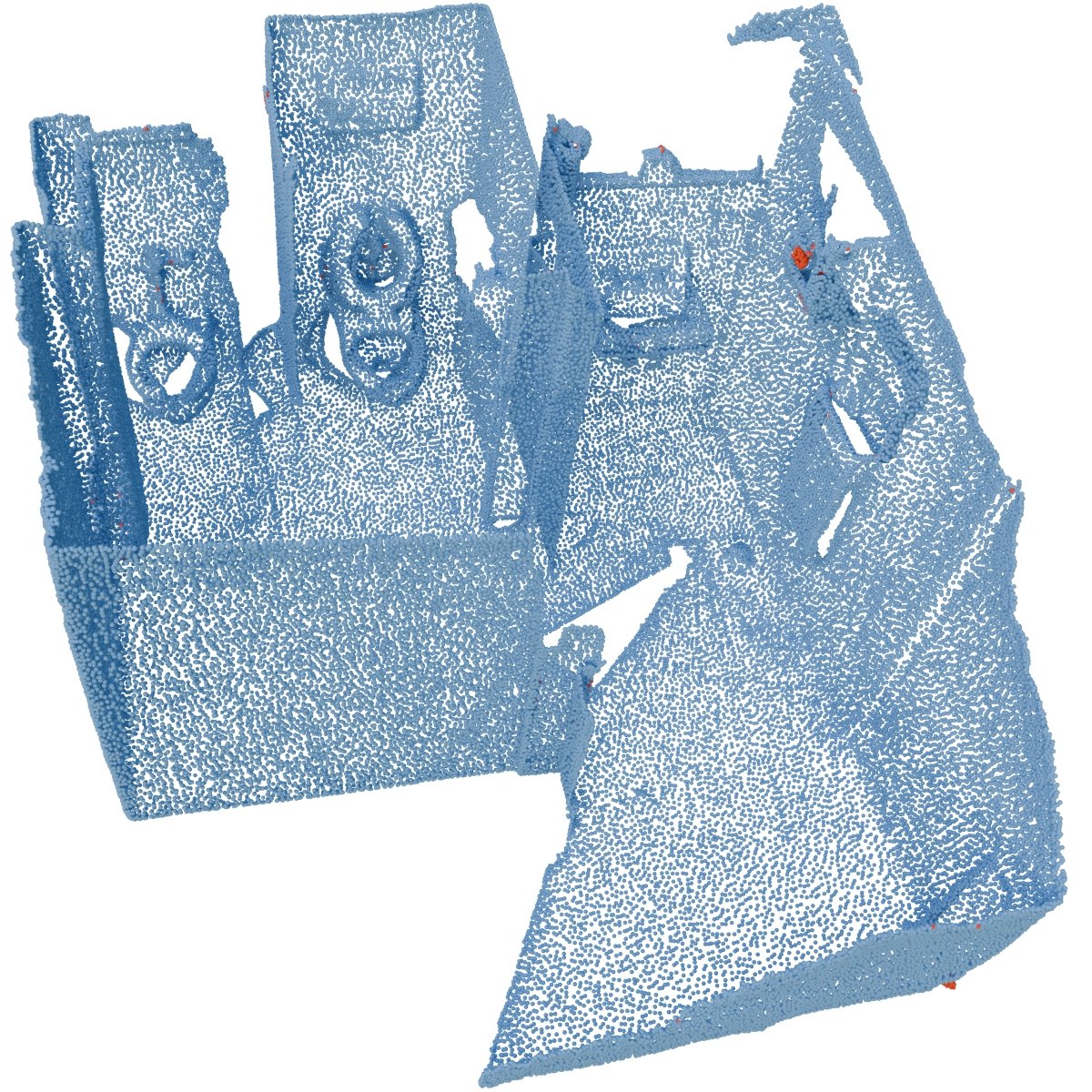}};
    \draw[black,thin]  (0.84,1.15) rectangle (0.89,1.25);
    \node[anchor=west,inner sep=0] (near) at (far.east) {\fcolorbox{black}{white}{\includegraphics[width=0.045\linewidth]{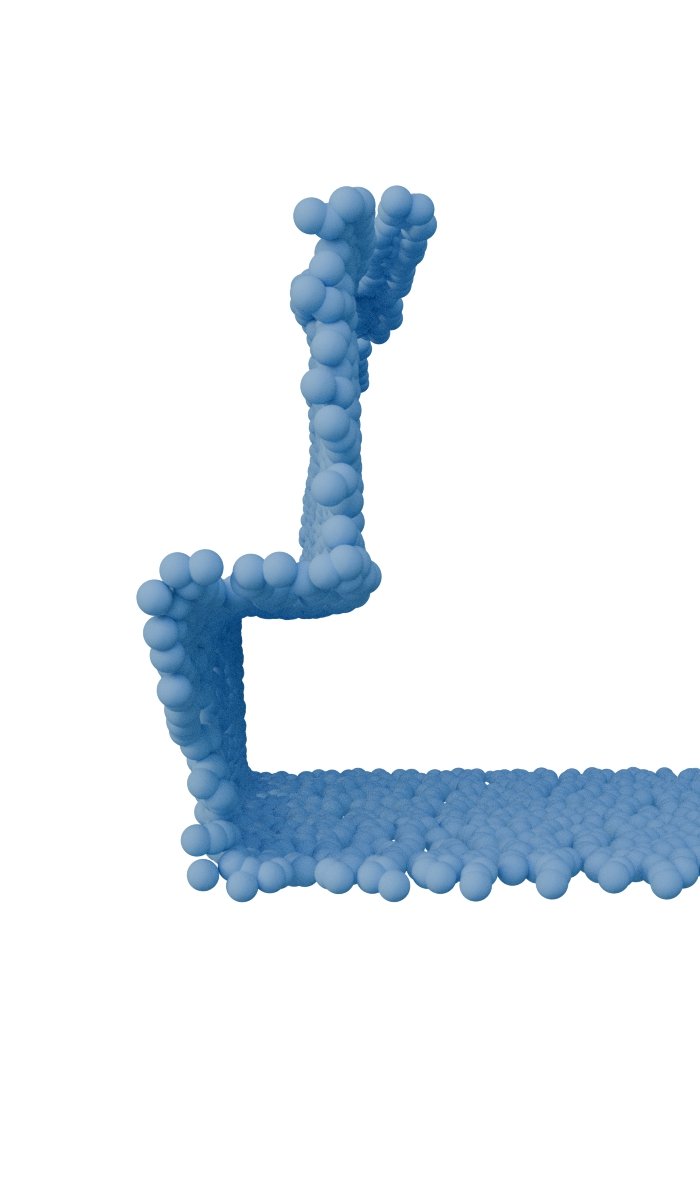}}};
\end{tikzpicture} &
\begin{tikzpicture}[baseline=(current bounding box.center)]
    \node[anchor=south west,inner sep=0] (far) at (0,0) {\includegraphics[width=0.09\linewidth]{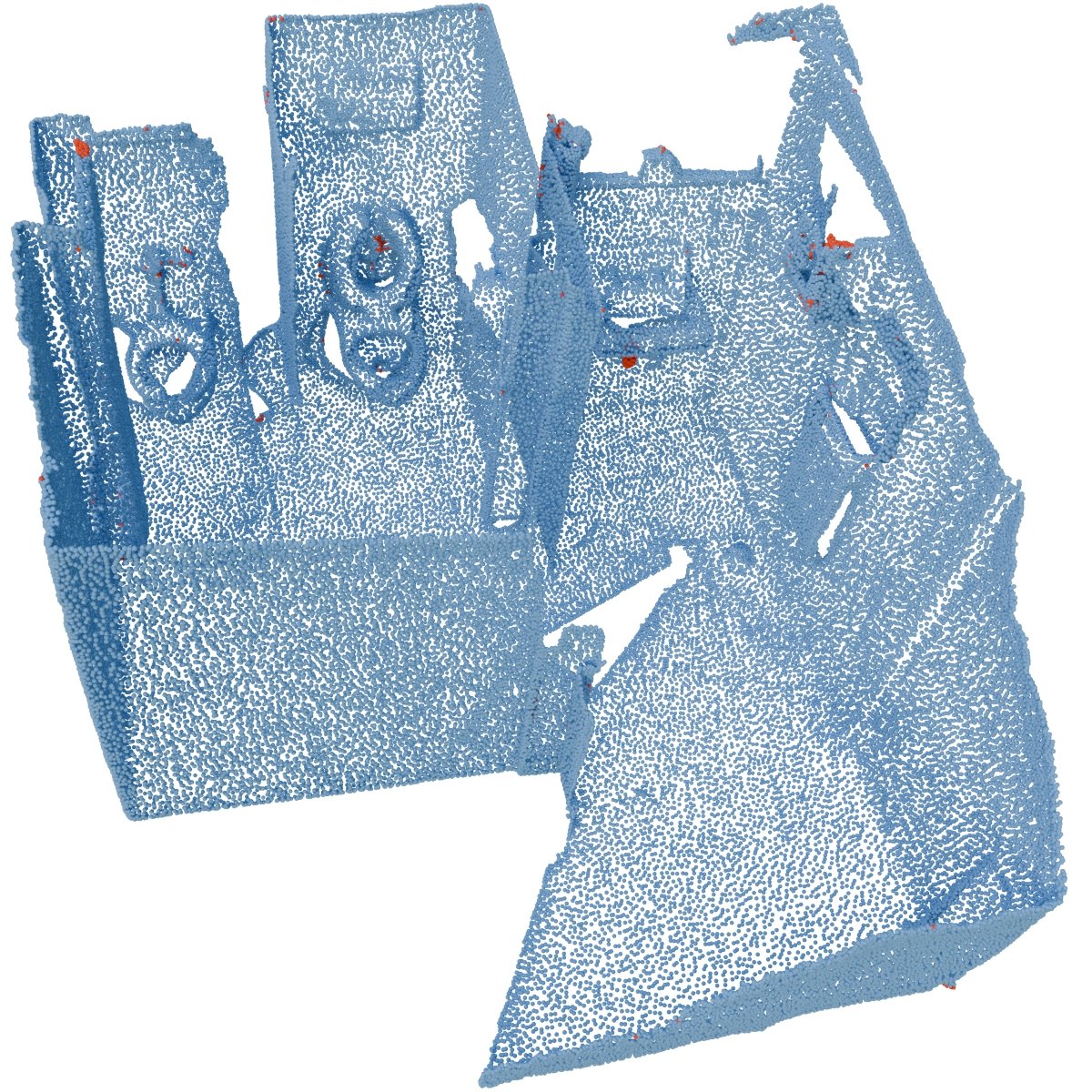}};
    \draw[black,thin]  (0.84,1.15) rectangle (0.89,1.25);
    \node[anchor=west,inner sep=0] (near) at (far.east) {\fcolorbox{black}{white}{\includegraphics[width=0.045\linewidth]{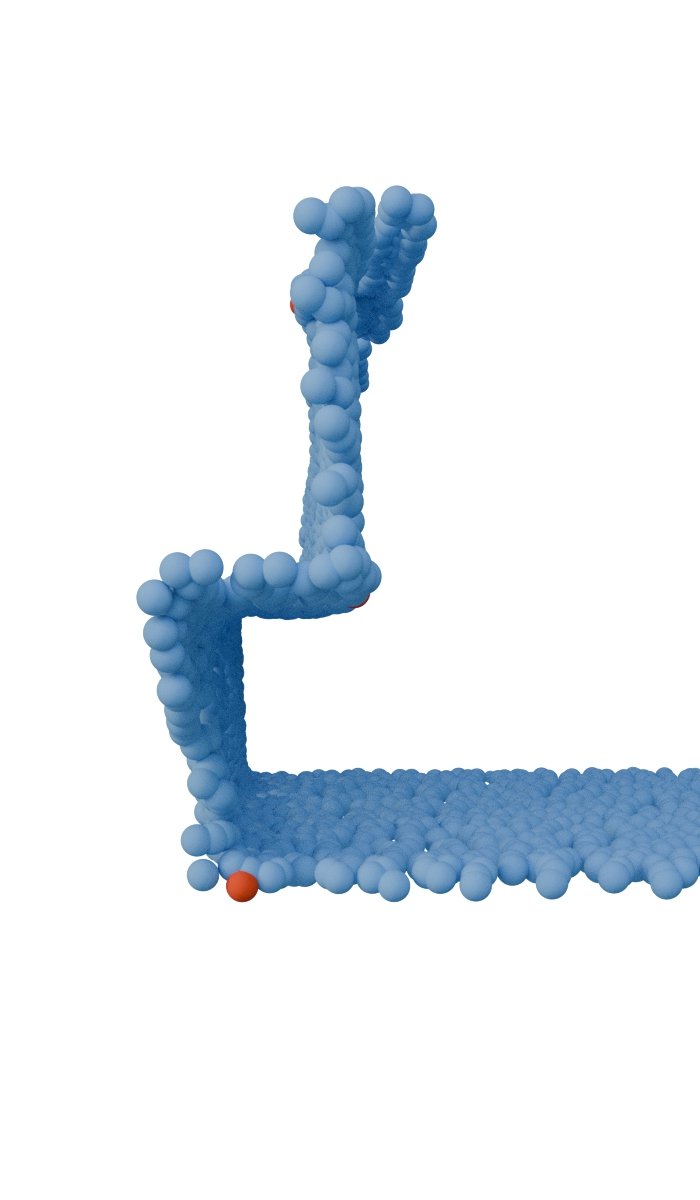}}};
\end{tikzpicture} &
\begin{tikzpicture}[baseline=(current bounding box.center)]
    \node[anchor=south west,inner sep=0] (far) at (0,0) {\includegraphics[width=0.09\linewidth]{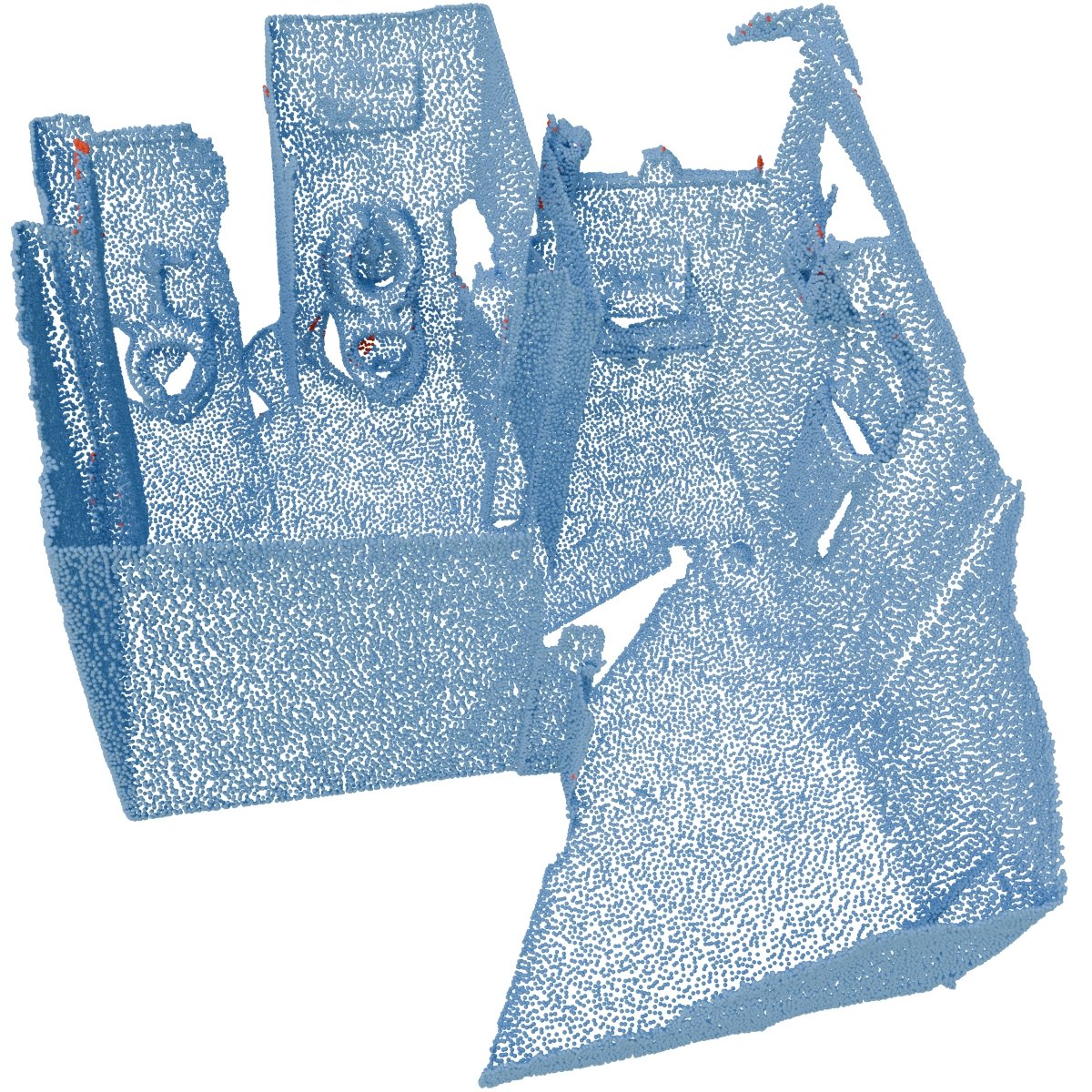}};
    \draw[black,thin]  (0.84,1.15) rectangle (0.89,1.25);
    \node[anchor=west,inner sep=0] (near) at (far.east) {\fcolorbox{black}{white}{\includegraphics[width=0.045\linewidth]{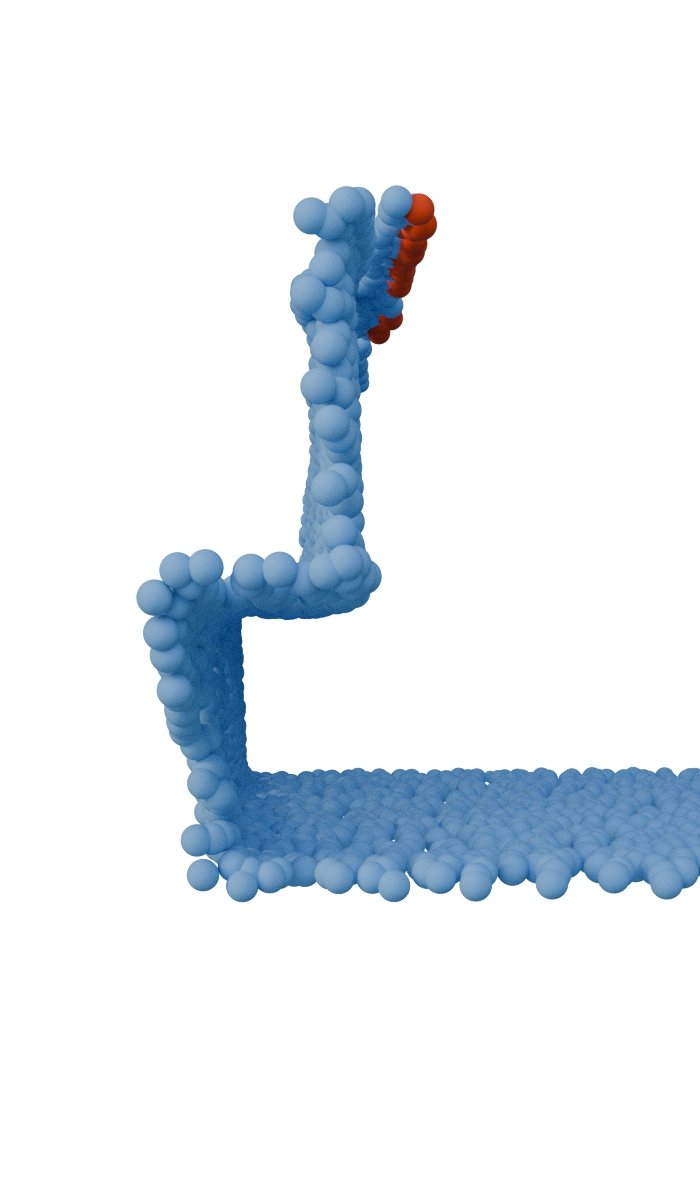}}};
\end{tikzpicture} \\

\begin{tikzpicture}[baseline=(current bounding box.center)]
    \node[anchor=south west,inner sep=0] (far) at (0,0) {\includegraphics[width=0.09\linewidth]{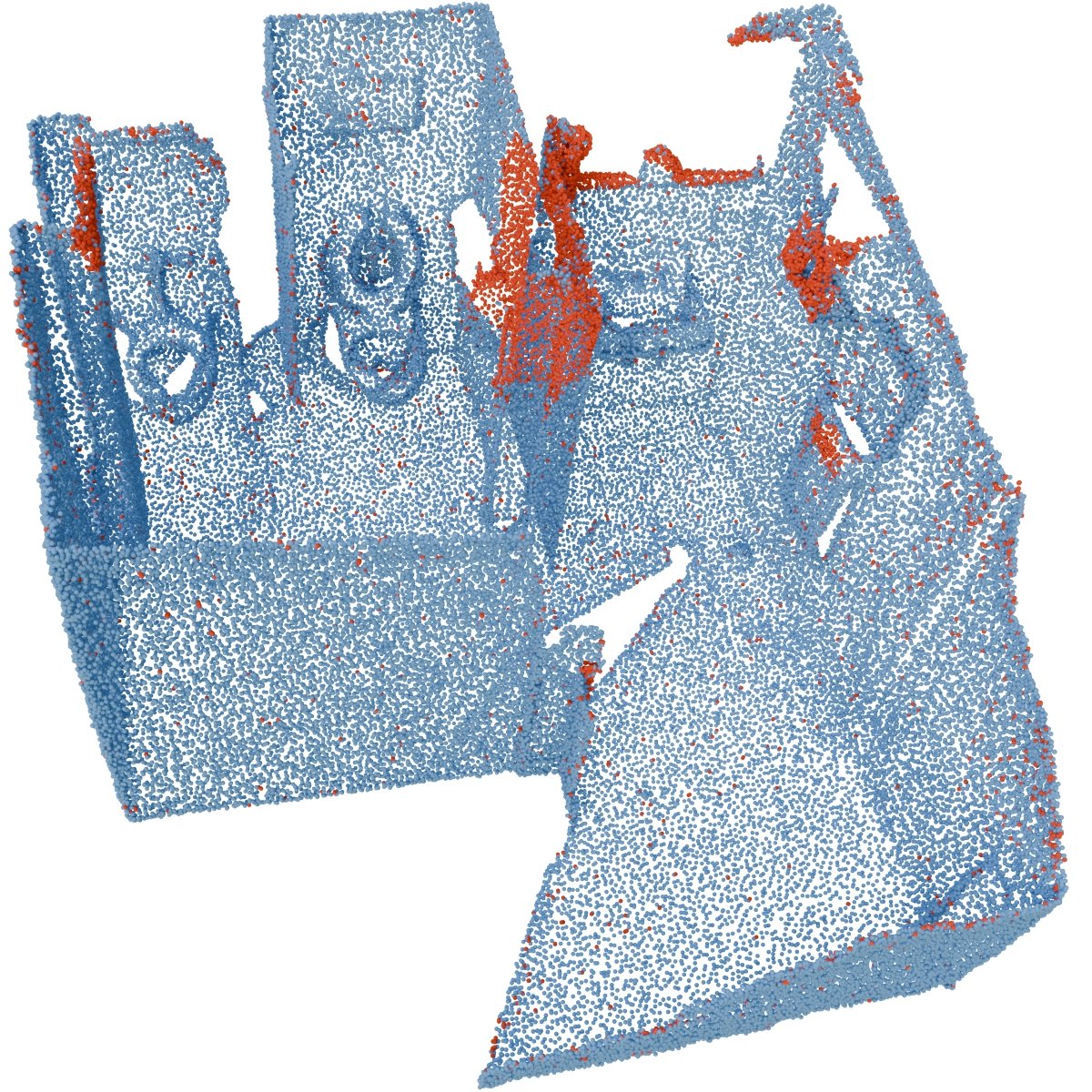}};
    \draw[black,thin]  (0.84,1.15) rectangle (0.89,1.25);
    \node[anchor=west,inner sep=0] (near) at (far.east) {\fcolorbox{black}{white}{\includegraphics[width=0.045\linewidth]{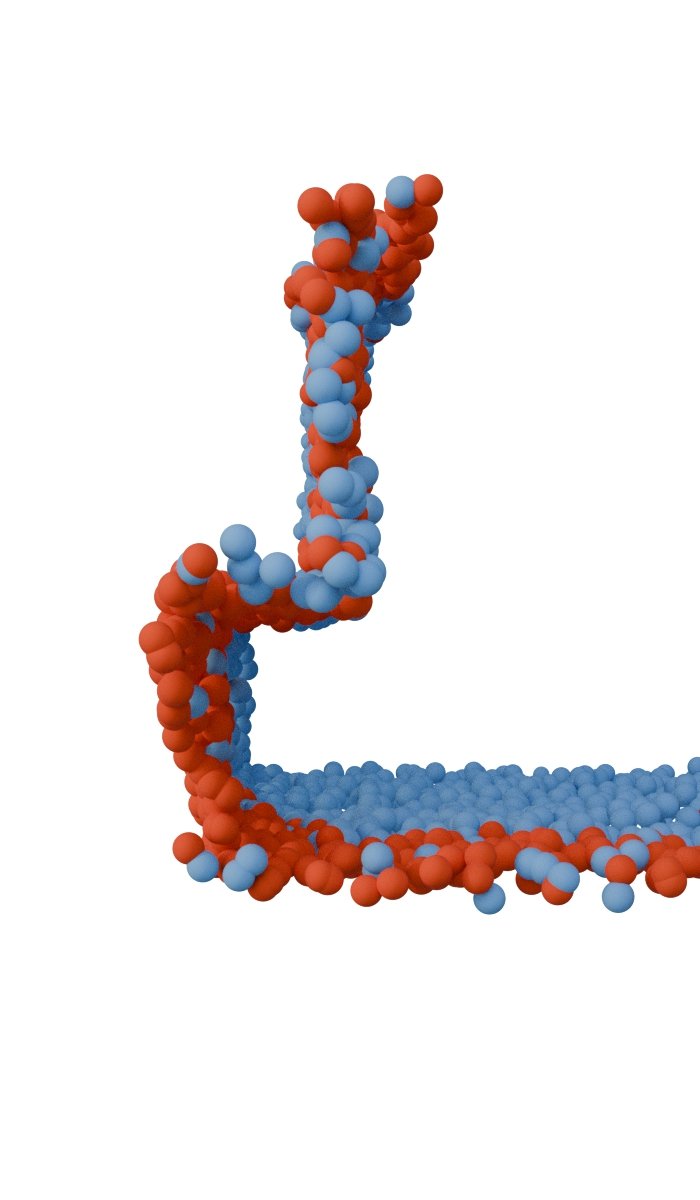}}};
\end{tikzpicture} &
\begin{tikzpicture}[baseline=(current bounding box.center)]
    \node[anchor=south west,inner sep=0] (far) at (0,0) {\includegraphics[width=0.09\linewidth]{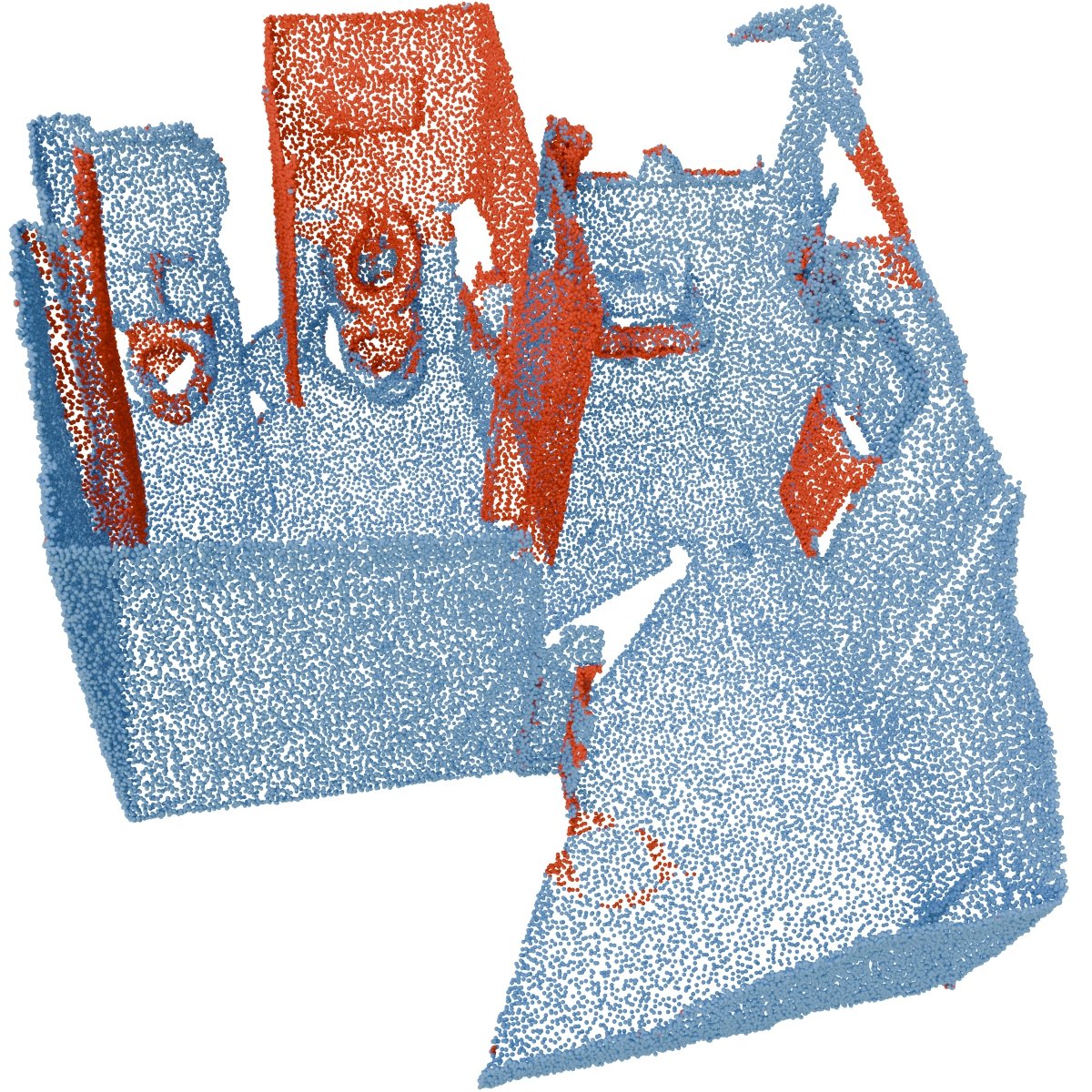}};
    \draw[black,thin]  (0.84,1.15) rectangle (0.89,1.25);
    \node[anchor=west,inner sep=0] (near) at (far.east) {\fcolorbox{black}{white}{\includegraphics[width=0.045\linewidth]{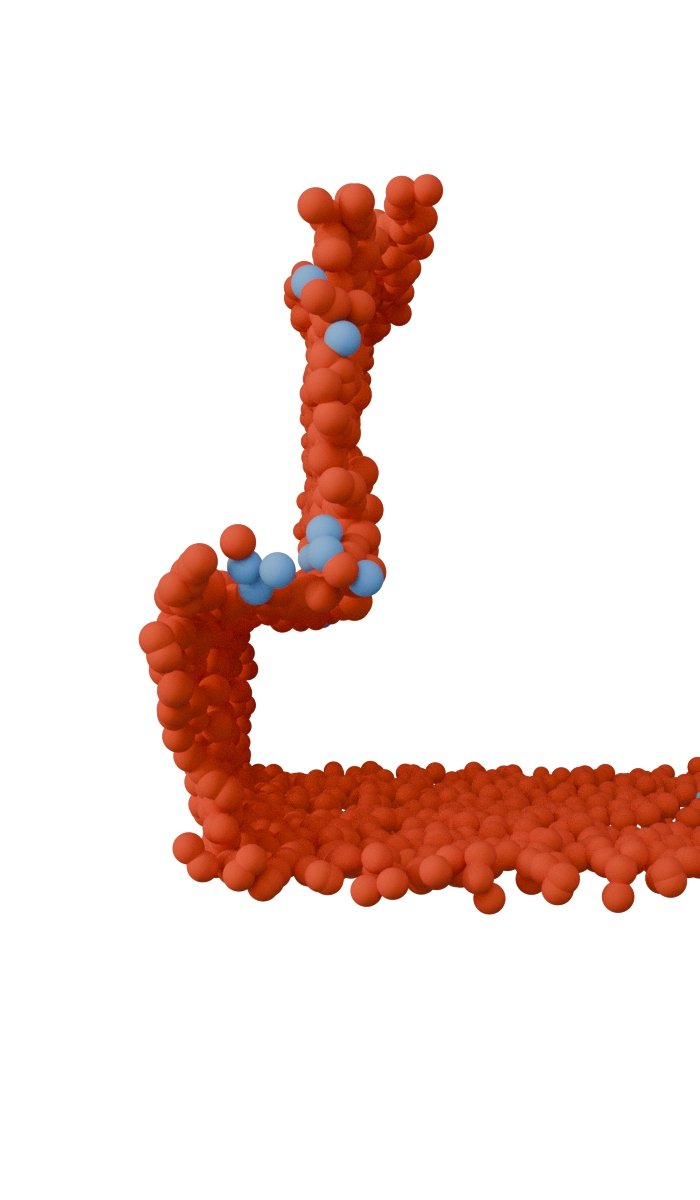}}};
\end{tikzpicture} &
\begin{tikzpicture}[baseline=(current bounding box.center)]
    \node[anchor=south west,inner sep=0] (far) at (0,0) {\includegraphics[width=0.09\linewidth]{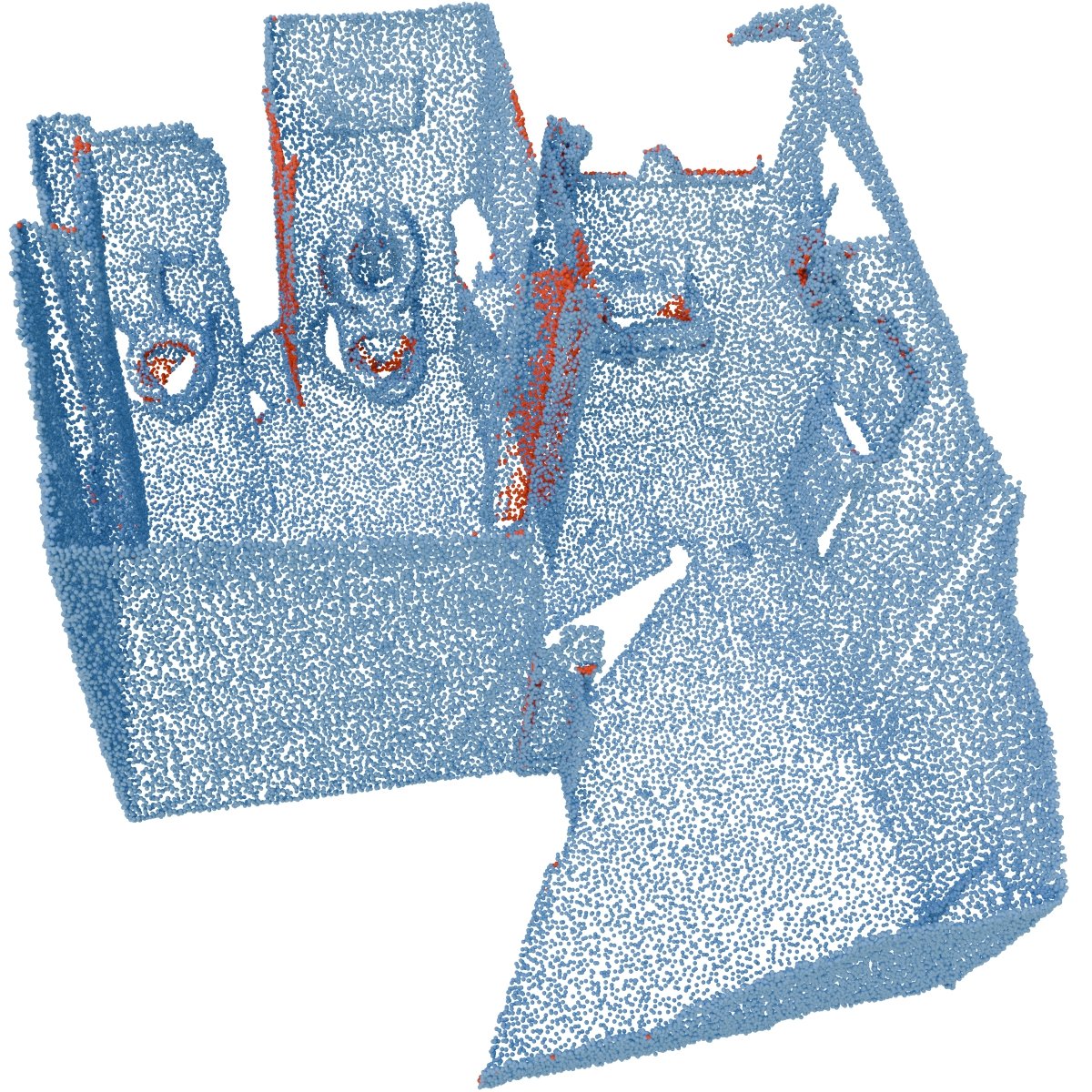}};
    \draw[black,thin]  (0.84,1.15) rectangle (0.89,1.25);
    \node[anchor=west,inner sep=0] (near) at (far.east) {\fcolorbox{black}{white}{\includegraphics[width=0.045\linewidth]{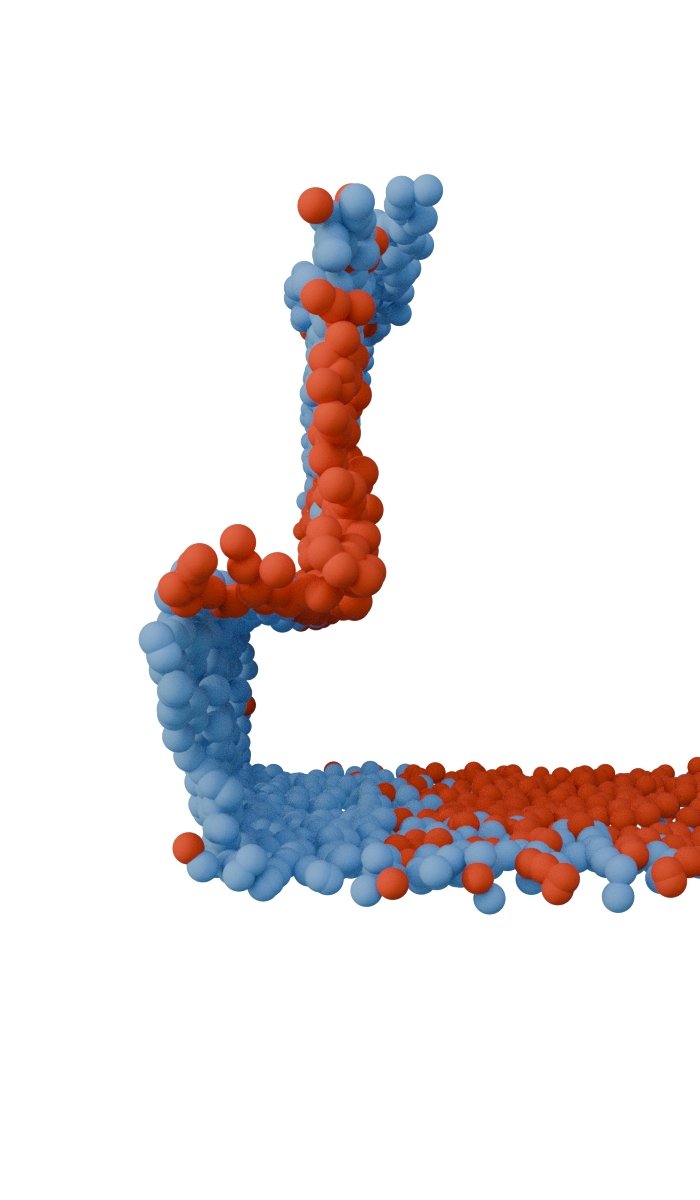}}};
\end{tikzpicture} &
\begin{tikzpicture}[baseline=(current bounding box.center)]
    \node[anchor=south west,inner sep=0] (far) at (0,0) {\includegraphics[width=0.09\linewidth]{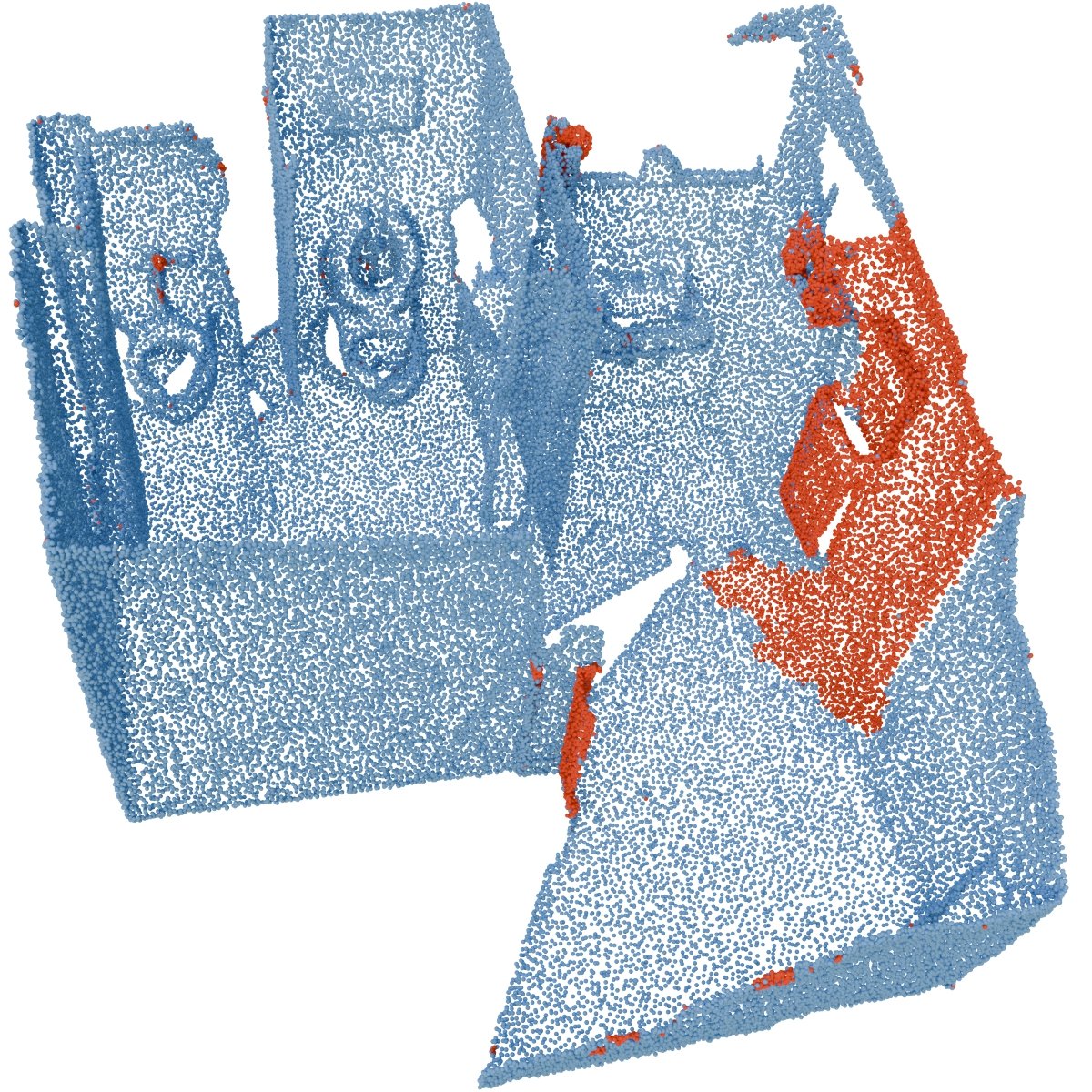}};
    \draw[black,thin]  (0.84,1.15) rectangle (0.89,1.25);
    \node[anchor=west,inner sep=0] (near) at (far.east) {\fcolorbox{black}{white}{\includegraphics[width=0.045\linewidth]{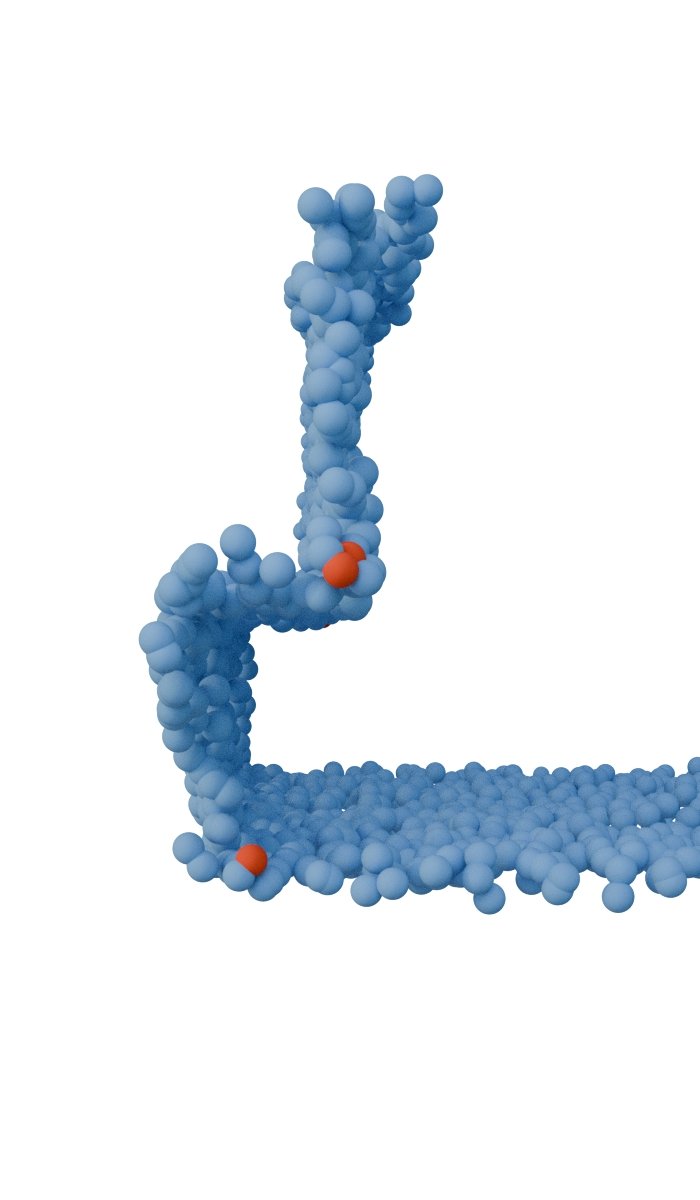}}};
\end{tikzpicture} &
\begin{tikzpicture}[baseline=(current bounding box.center)]
    \node[anchor=south west,inner sep=0] (far) at (0,0) {\includegraphics[width=0.09\linewidth]{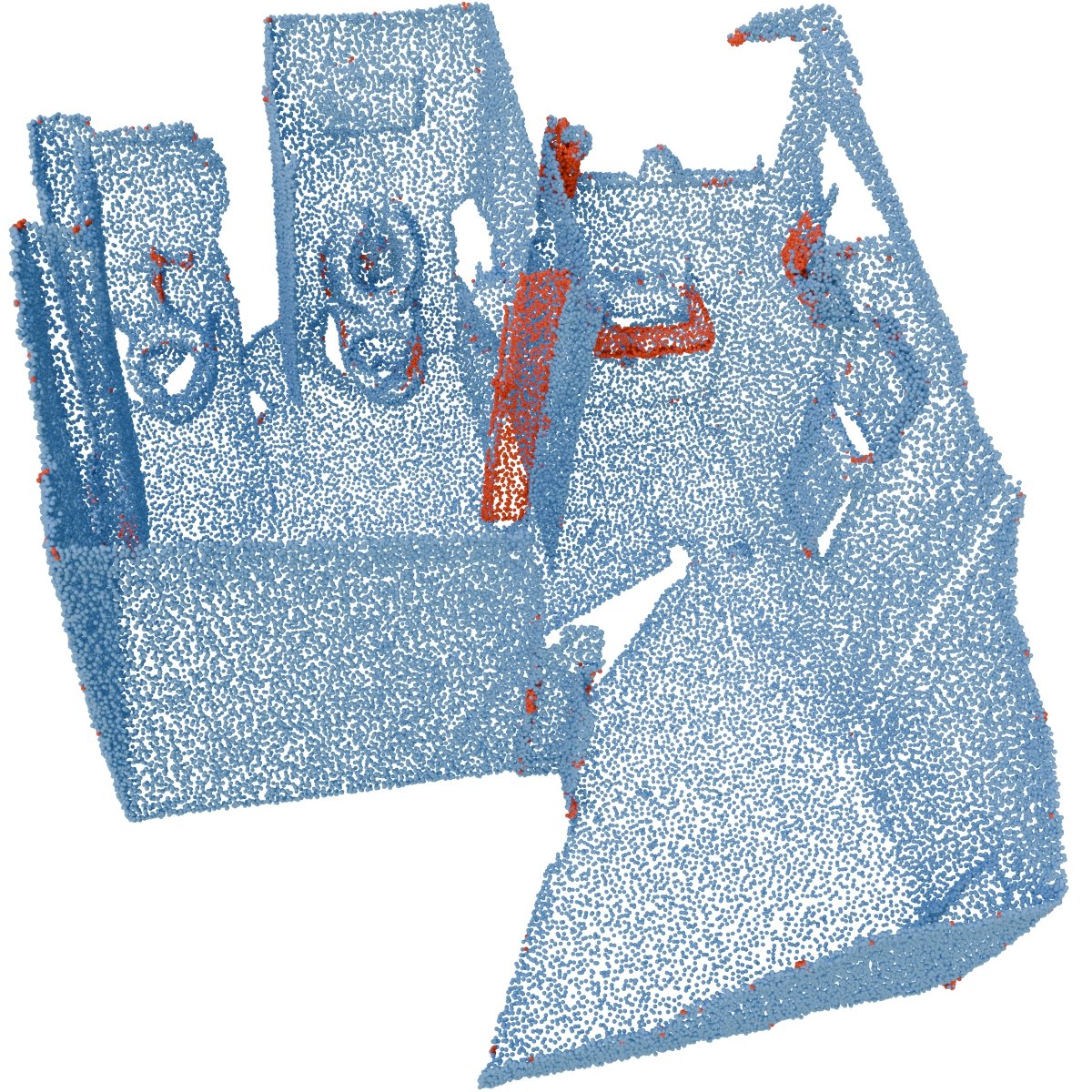}};
    \draw[black,thin]  (0.84,1.15) rectangle (0.89,1.25);
    \node[anchor=west,inner sep=0] (near) at (far.east) {\fcolorbox{black}{white}{\includegraphics[width=0.045\linewidth]{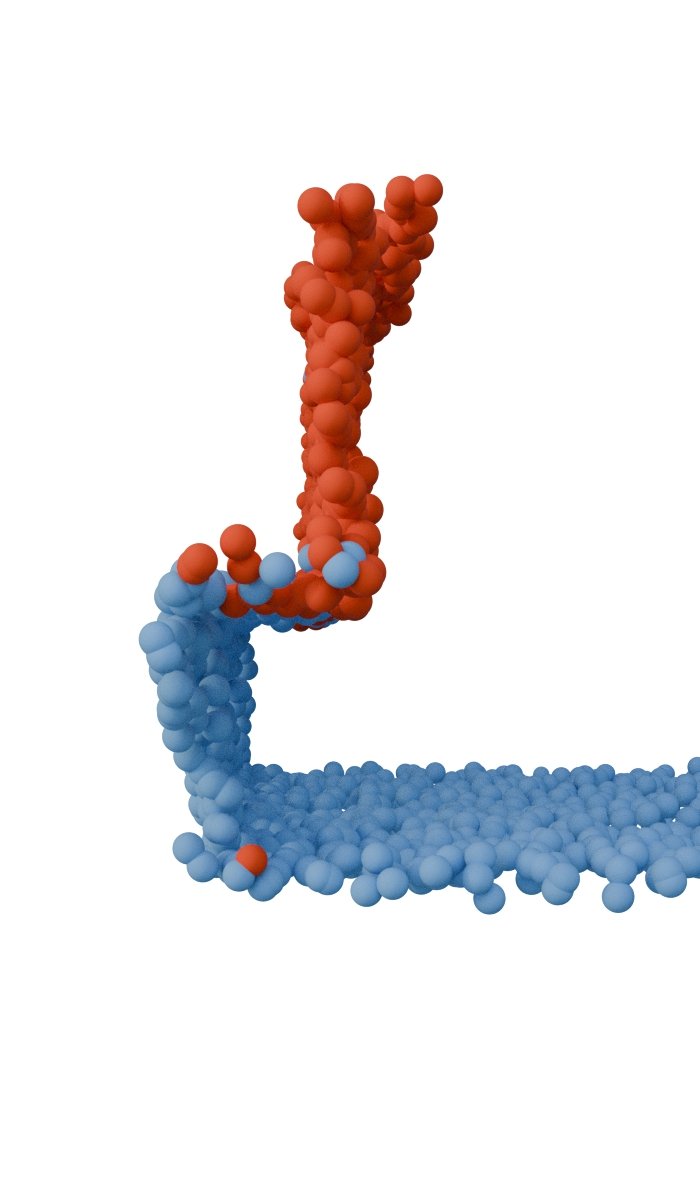}}};
\end{tikzpicture} &
\begin{tikzpicture}[baseline=(current bounding box.center)]
    \node[anchor=south west,inner sep=0] (far) at (0,0) {\includegraphics[width=0.09\linewidth]{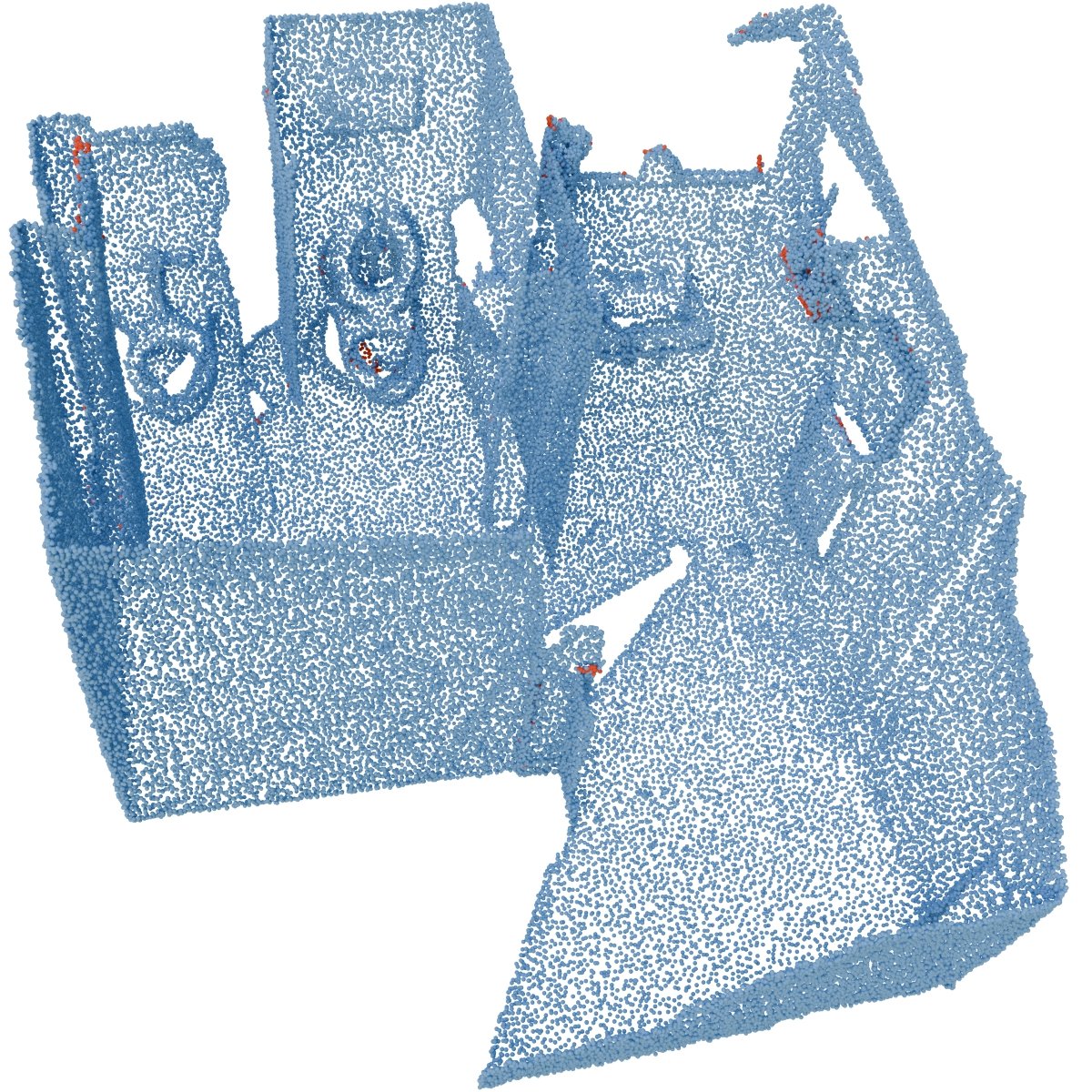}};
    \draw[black,thin]  (0.84,1.15) rectangle (0.89,1.25);
    \node[anchor=west,inner sep=0] (near) at (far.east) {\fcolorbox{black}{white}{\includegraphics[width=0.045\linewidth]{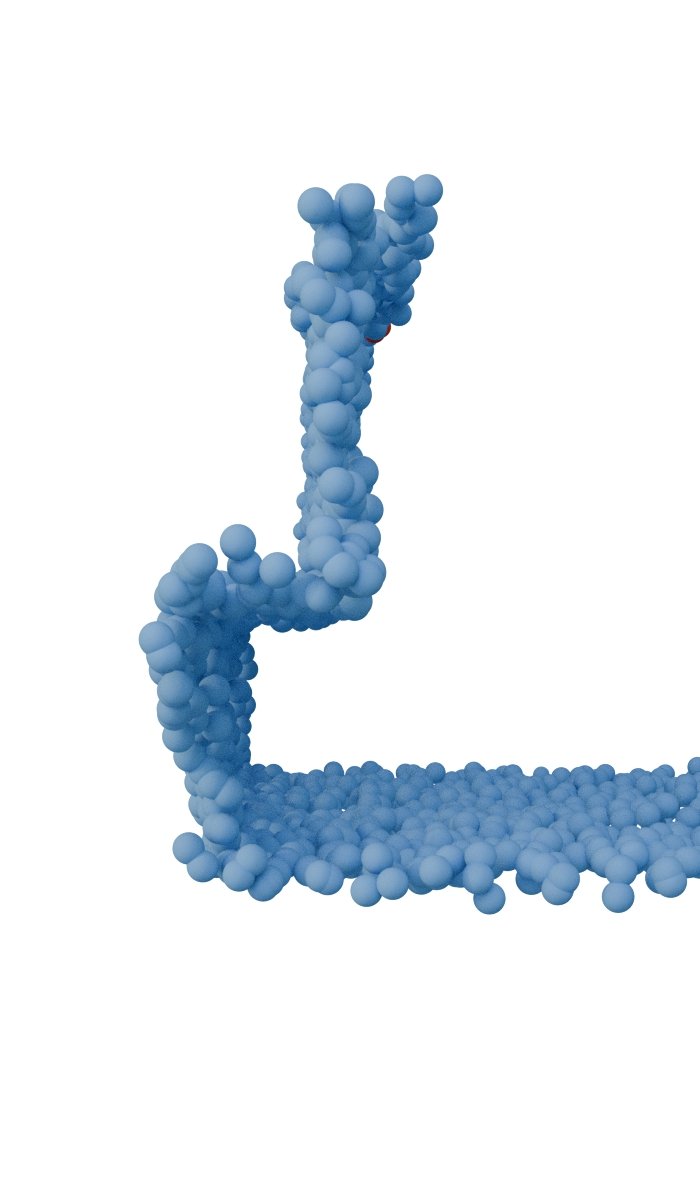}}};
\end{tikzpicture} \\

\begin{tikzpicture}[baseline=(current bounding box.center)]
    \node[anchor=south west,inner sep=0] (far) at (0,0) {\includegraphics[width=0.09\linewidth]{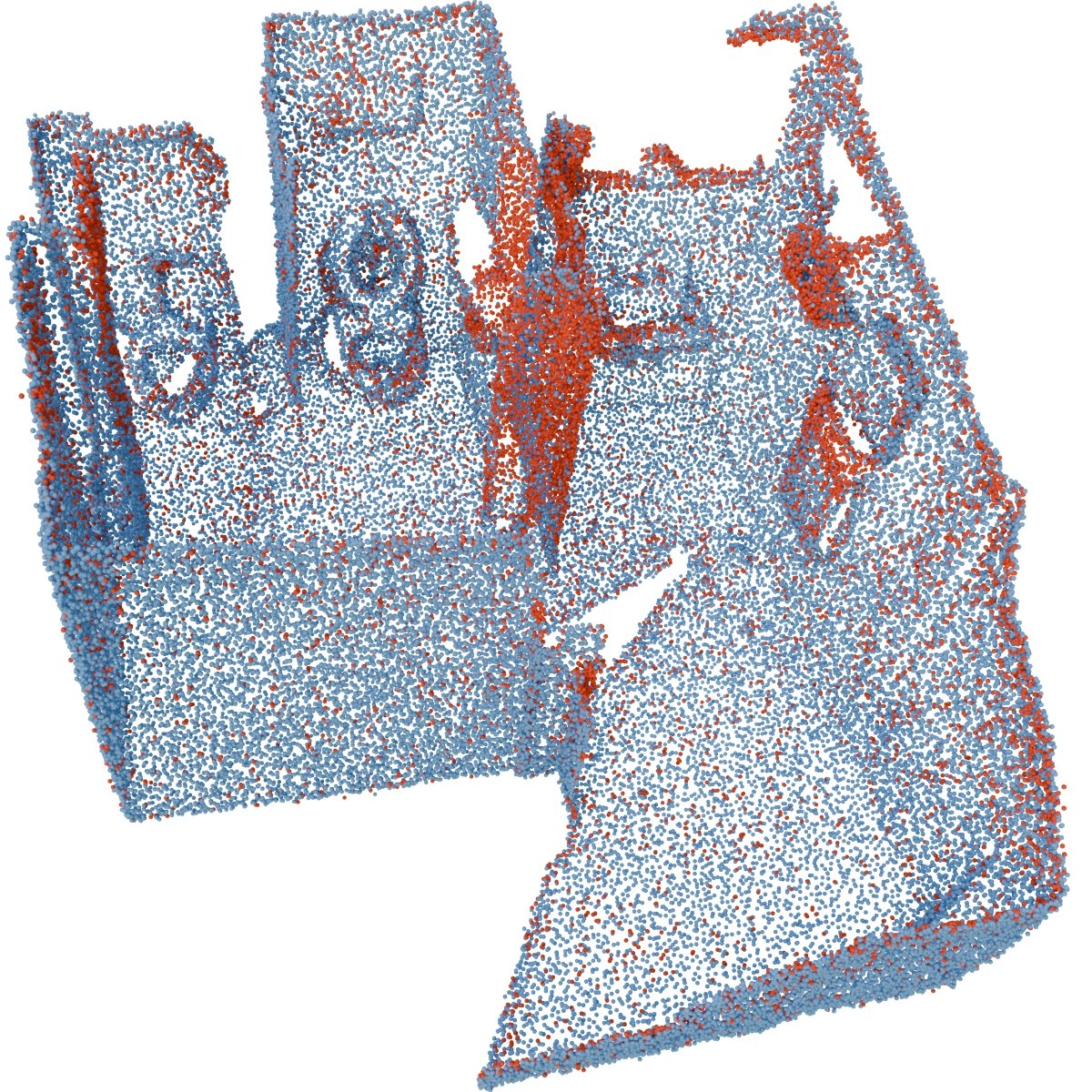}};
    \draw[black,thin]  (0.84,1.15) rectangle (0.89,1.25);
    \node[anchor=west,inner sep=0] (near) at (far.east) {\fcolorbox{black}{white}{\includegraphics[width=0.045\linewidth]{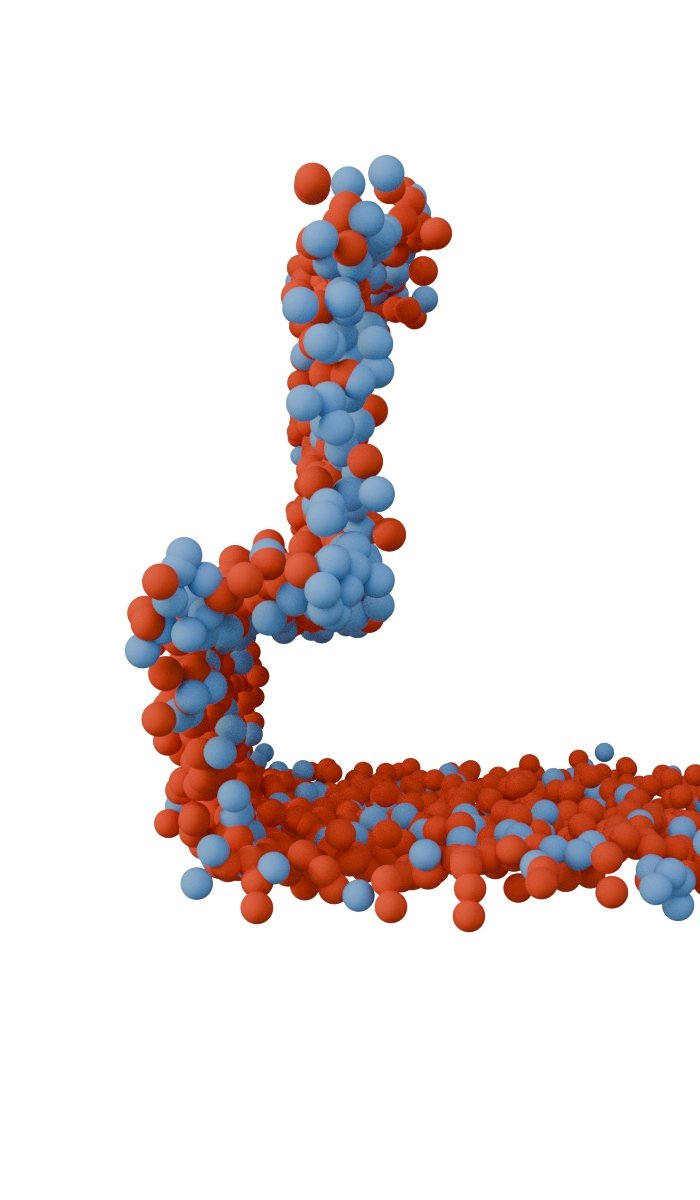}}};
\end{tikzpicture} &
\begin{tikzpicture}[baseline=(current bounding box.center)]
    \node[anchor=south west,inner sep=0] (far) at (0,0) {\includegraphics[width=0.09\linewidth]{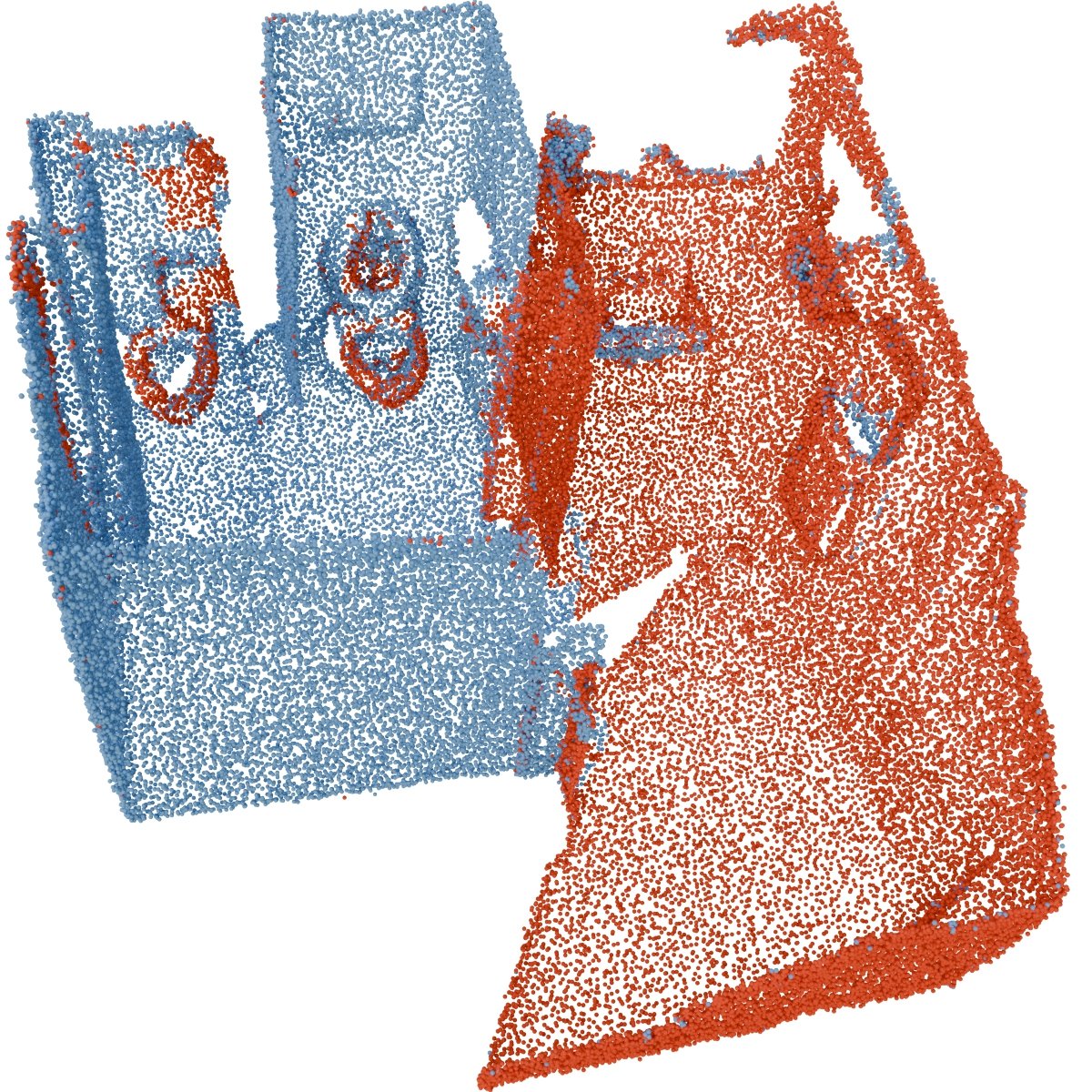}};
    \draw[black,thin]  (0.84,1.15) rectangle (0.89,1.25);
    \node[anchor=west,inner sep=0] (near) at (far.east) {\fcolorbox{black}{white}{\includegraphics[width=0.045\linewidth]{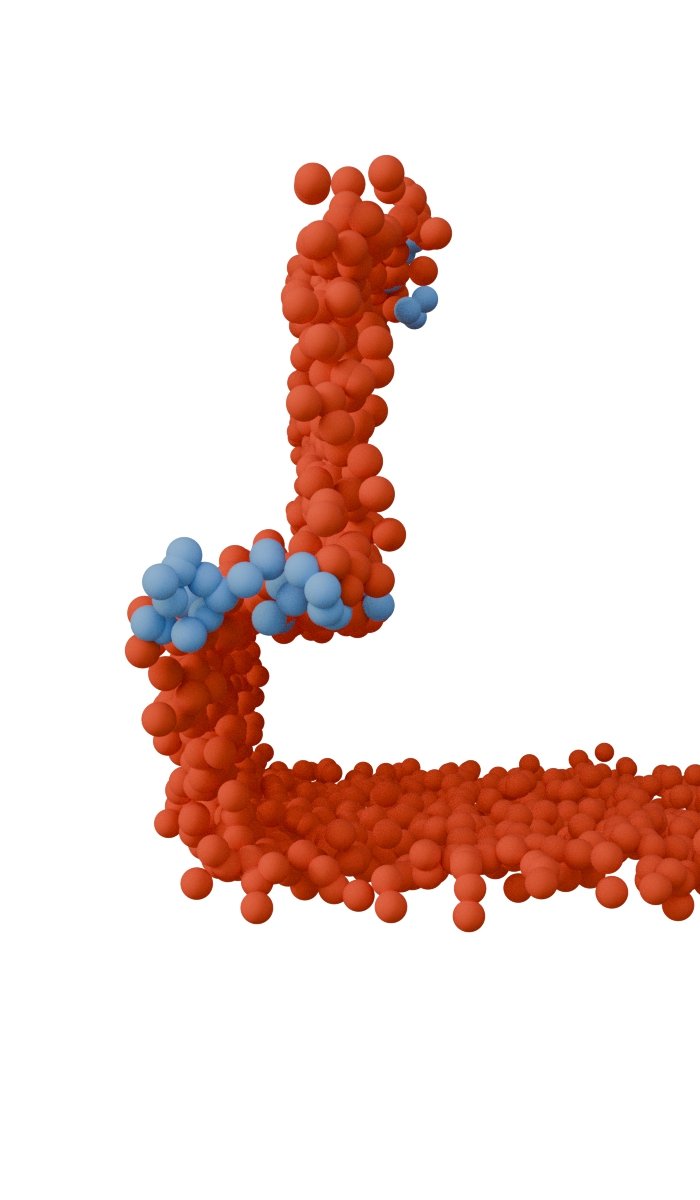}}};
\end{tikzpicture} &
\begin{tikzpicture}[baseline=(current bounding box.center)]
    \node[anchor=south west,inner sep=0] (far) at (0,0) {\includegraphics[width=0.09\linewidth]{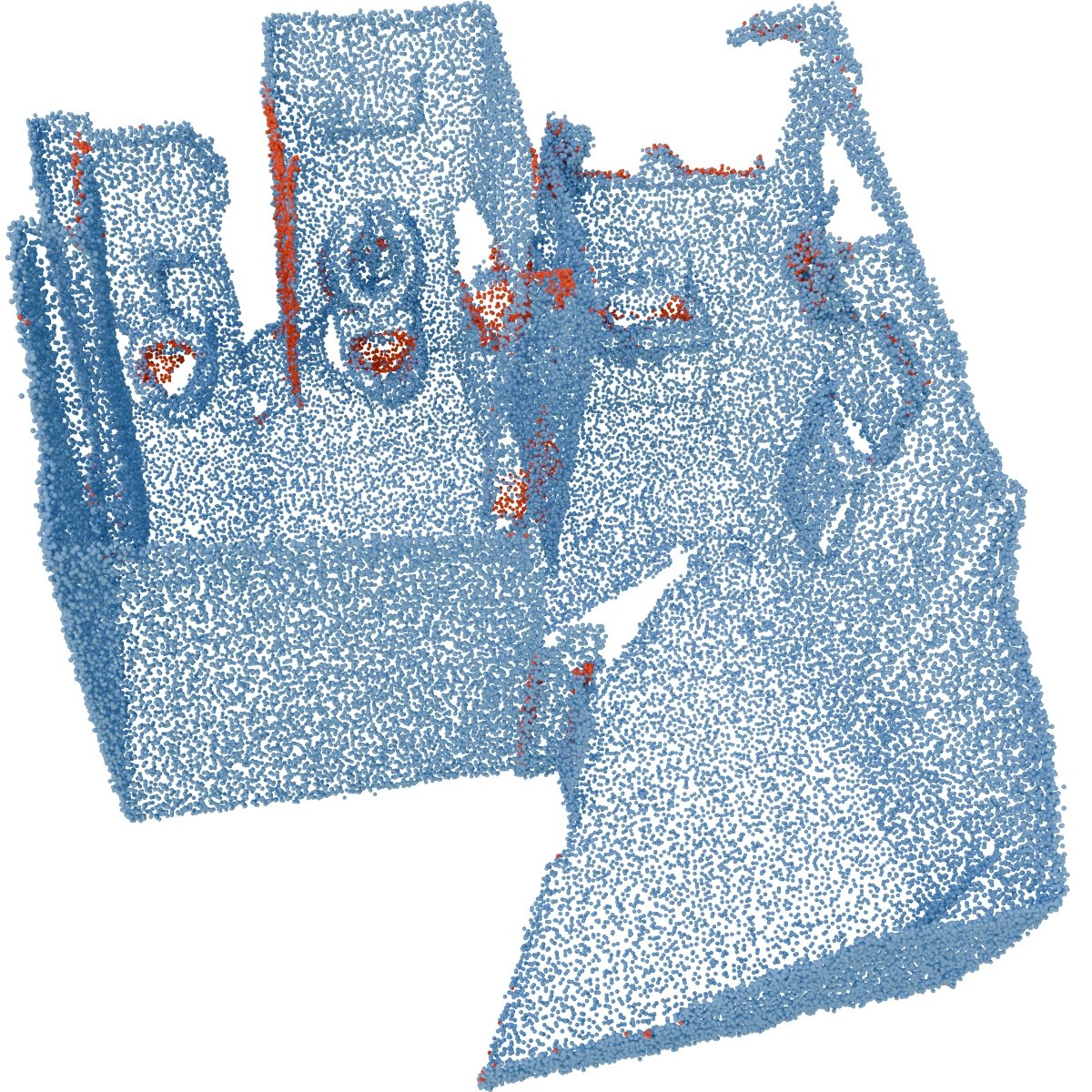}};
    \draw[black,thin]  (0.84,1.15) rectangle (0.89,1.25);
    \node[anchor=west,inner sep=0] (near) at (far.east) {\fcolorbox{black}{white}{\includegraphics[width=0.045\linewidth]{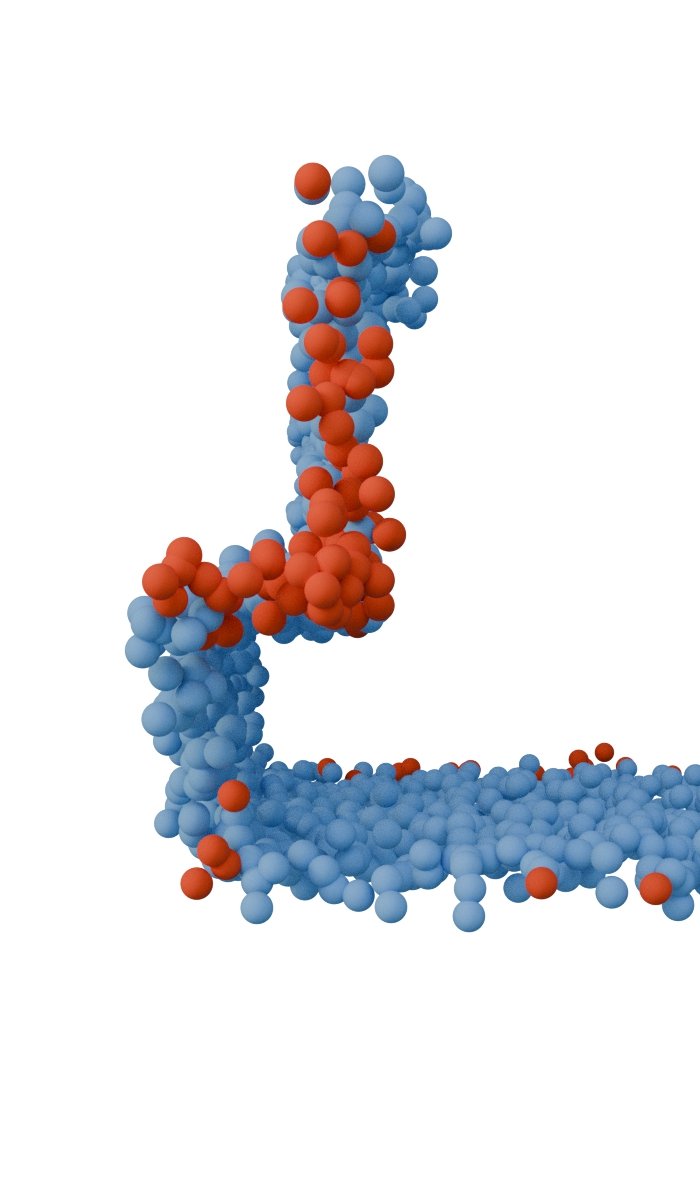}}};
\end{tikzpicture} &
\begin{tikzpicture}[baseline=(current bounding box.center)]
    \node[anchor=south west,inner sep=0] (far) at (0,0) {\includegraphics[width=0.09\linewidth]{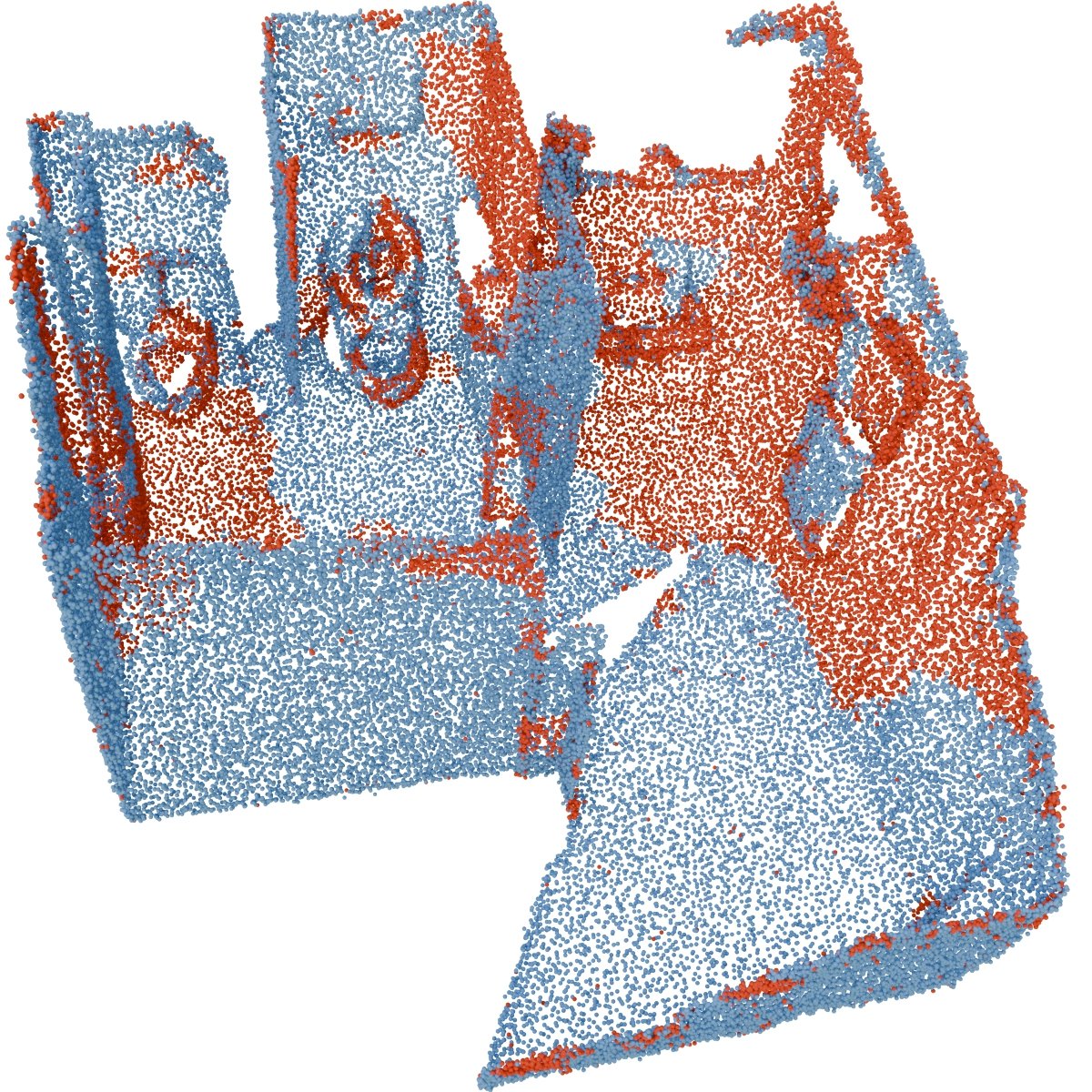}};
    \draw[black,thin]  (0.84,1.15) rectangle (0.89,1.25);
    \node[anchor=west,inner sep=0] (near) at (far.east) {\fcolorbox{black}{white}{\includegraphics[width=0.045\linewidth]{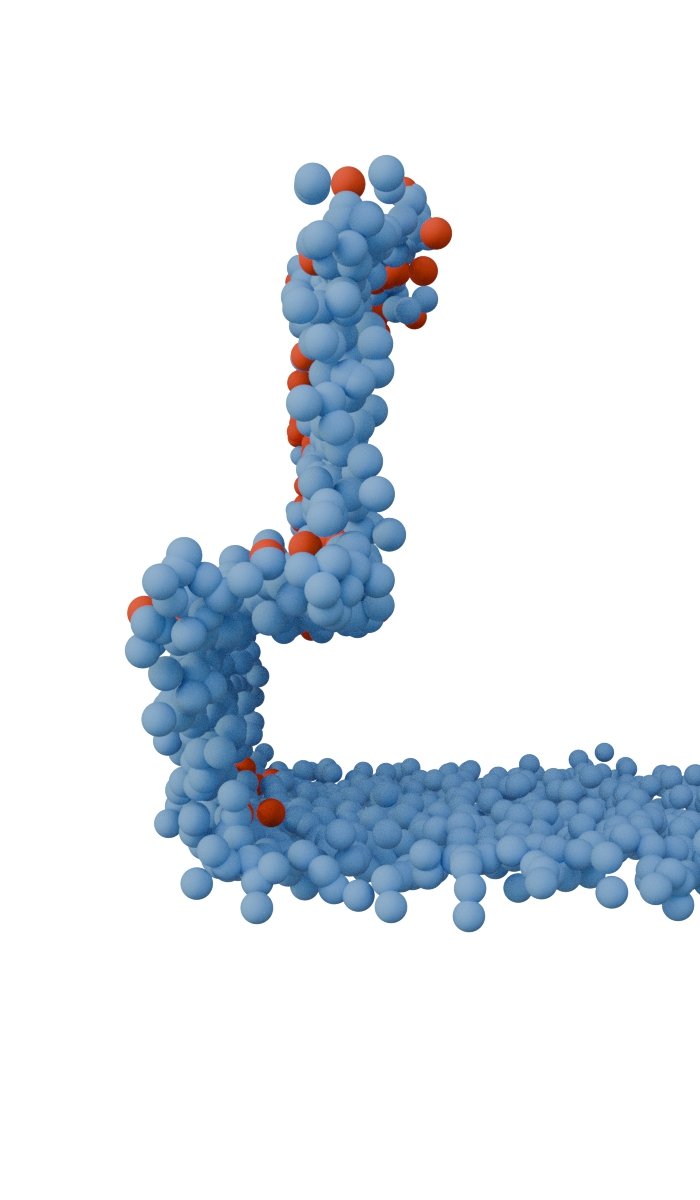}}};
\end{tikzpicture} &
\begin{tikzpicture}[baseline=(current bounding box.center)]
    \node[anchor=south west,inner sep=0] (far) at (0,0) {\includegraphics[width=0.09\linewidth]{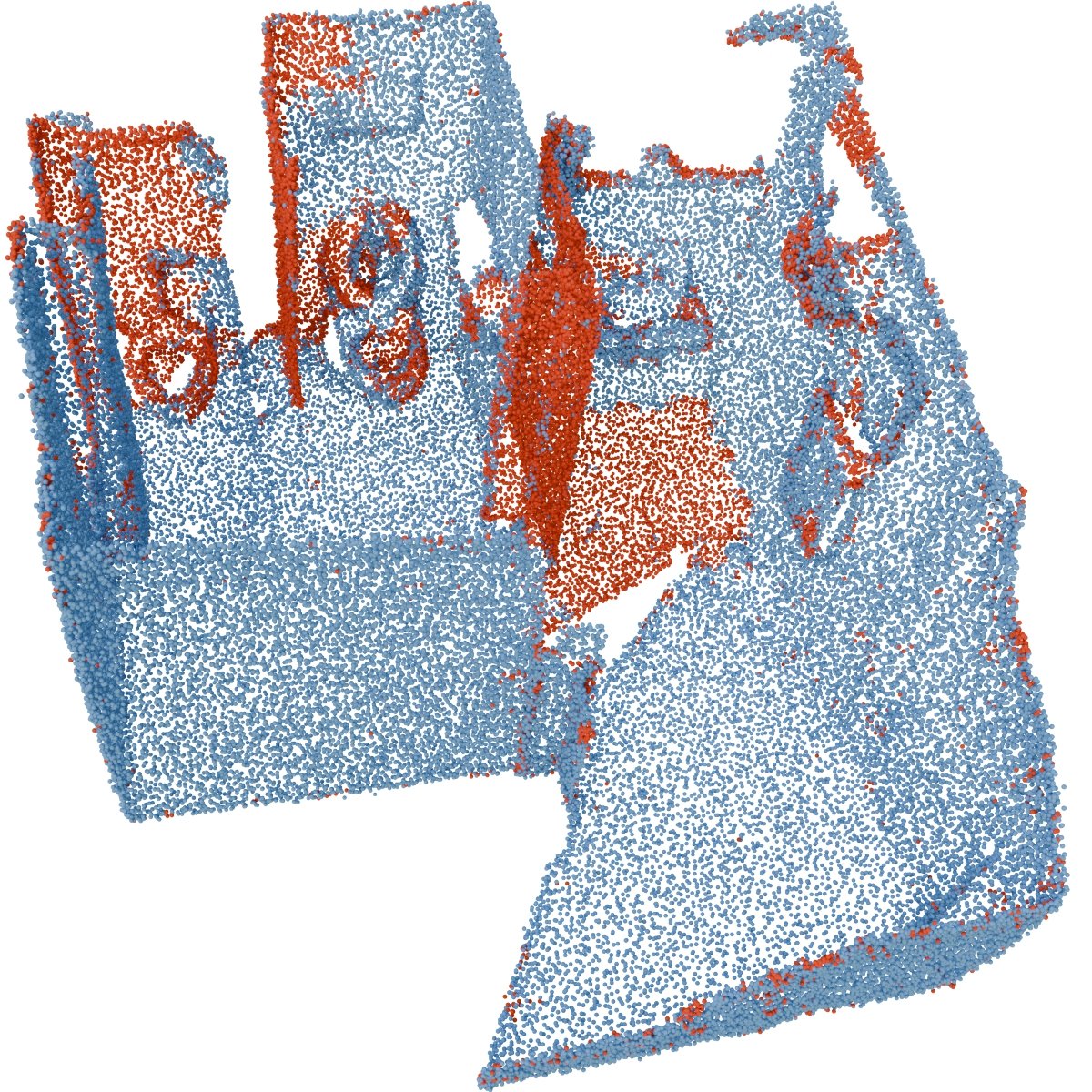}};
    \draw[black,thin]  (0.84,1.15) rectangle (0.89,1.25);
    \node[anchor=west,inner sep=0] (near) at (far.east) {\fcolorbox{black}{white}{\includegraphics[width=0.045\linewidth]{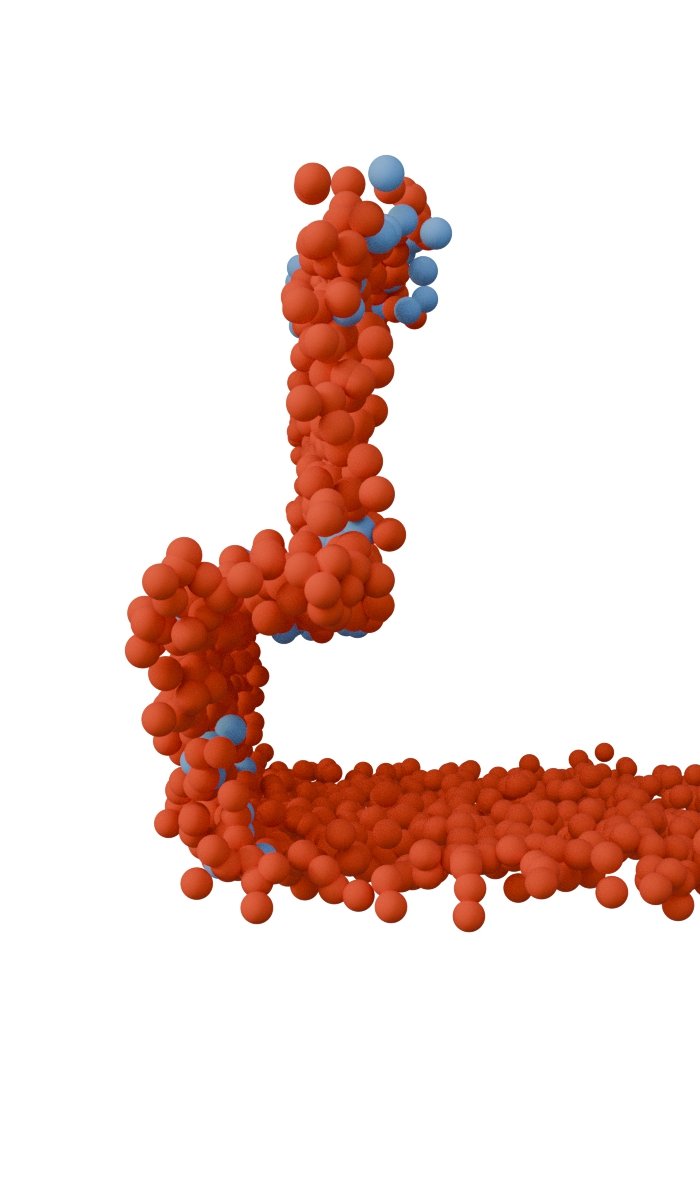}}};
\end{tikzpicture} &
\begin{tikzpicture}[baseline=(current bounding box.center)]
    \node[anchor=south west,inner sep=0] (far) at (0,0) {\includegraphics[width=0.09\linewidth]{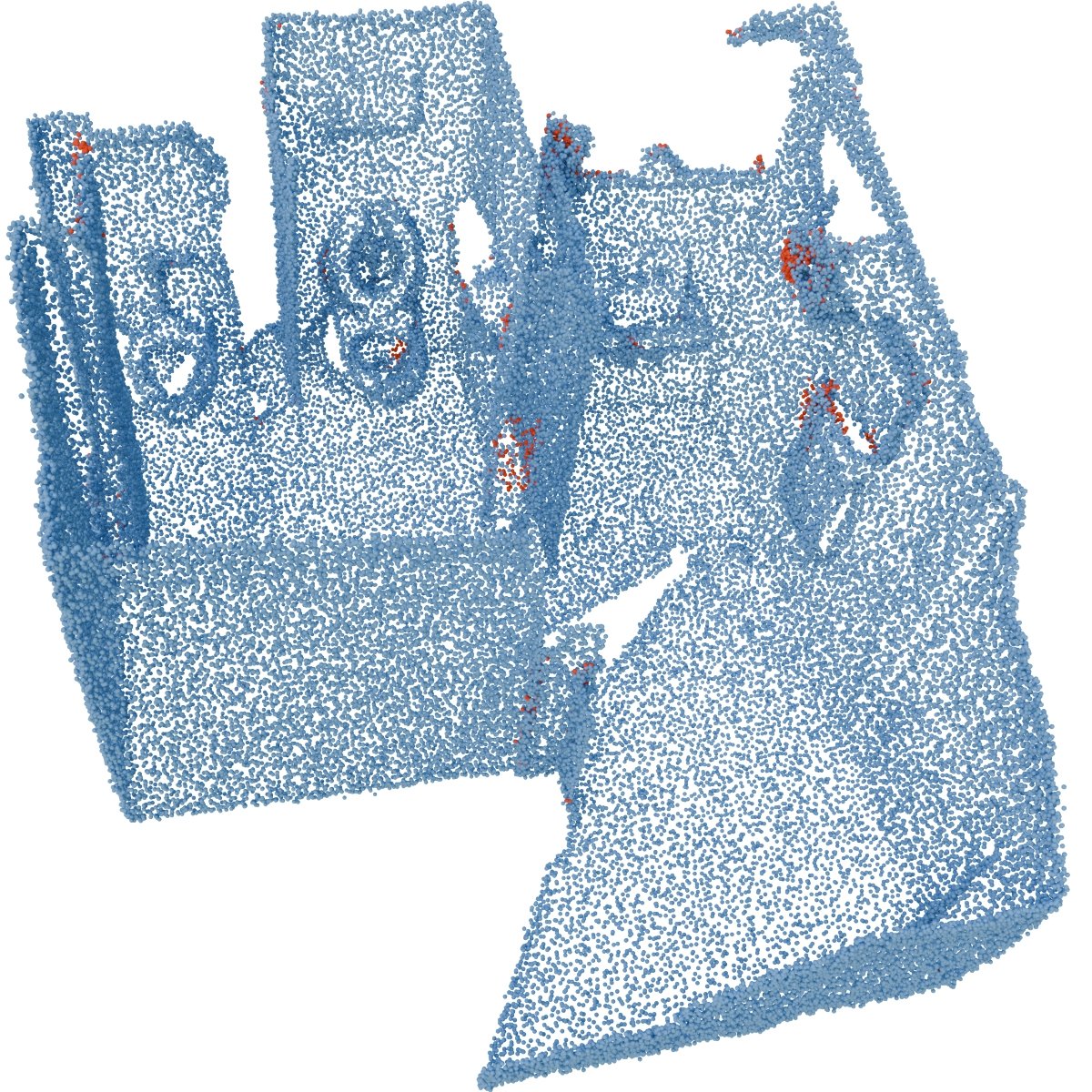}};
    \draw[black,thin]  (0.84,1.15) rectangle (0.89,1.25);
    \node[anchor=west,inner sep=0] (near) at (far.east) {\fcolorbox{black}{white}{\includegraphics[width=0.045\linewidth]{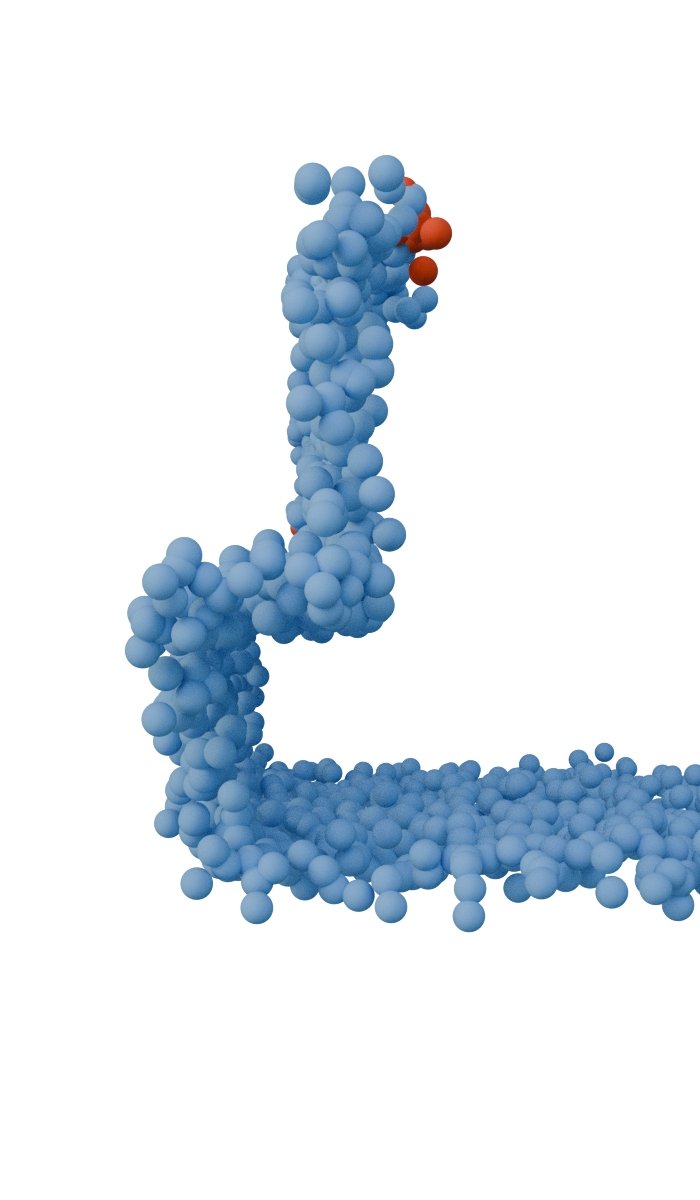}}};
\end{tikzpicture} \\
\end{tabular}
    \caption{Normal orientation results for a noisy point cloud with 86,094 points. We also provide a close-up view for the rectangular region. Blue points indicate correct orientations, while red points denote incorrect orientations. From top to bottom: the original ScanNet v2, ScanNetV2-Noise1 and ScanNetV2-Noise2.}
    \label{fig:comparison_scannetv2}
\end{figure*}

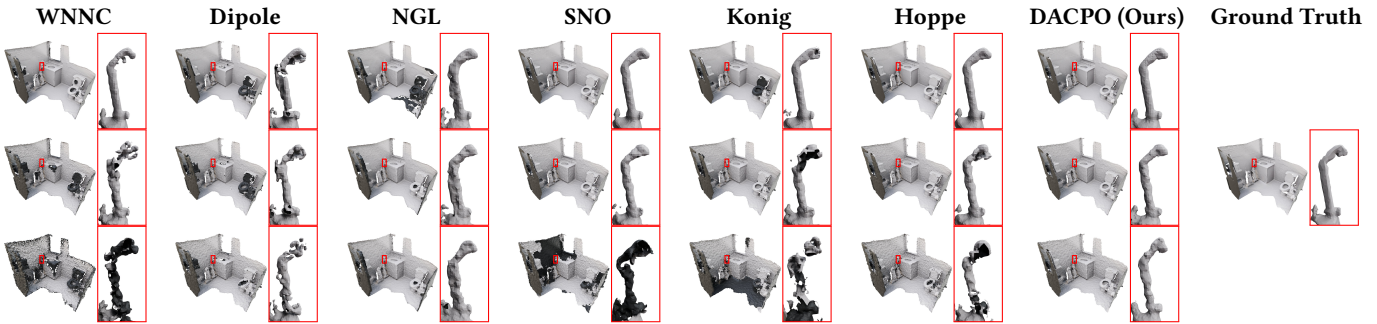
\begin{figure*}[htbp]
    \centering
    \begin{tabular}{cccccccc}
    \textbf{WNNC} & \textbf{Dipole} & \textbf{NGL} & \textbf{SNO} & \textbf{Konig} & \textbf{Hoppe} & \textbf{DACPO (Ours)} & \textbf{Ground Truth} \\

    \begin{tikzpicture}[baseline=(current bounding box.center)]
        \node[anchor=south west,inner sep=0] (far) at (0,0) {\includegraphics[width=0.070\linewidth]{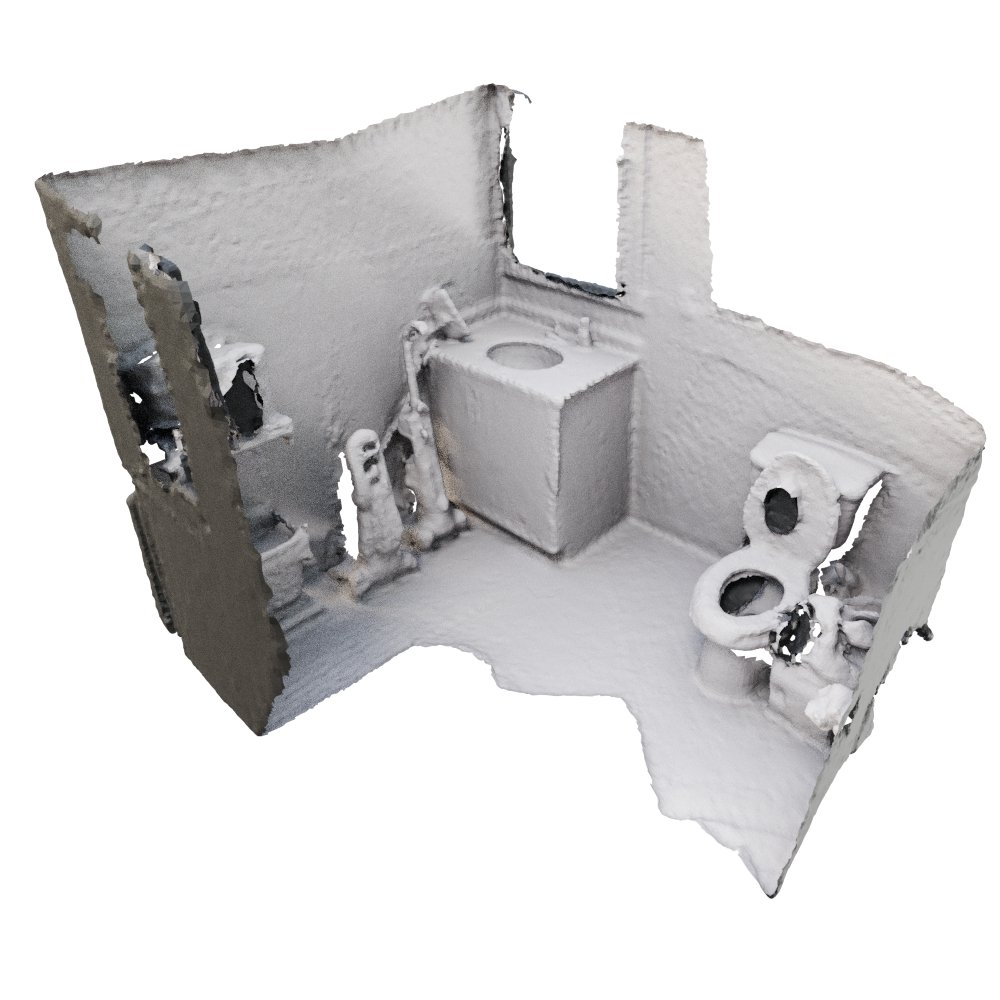}};
        \draw[red,thin]  (0.515,0.78) rectangle (0.565,0.88);
        \node[anchor=west,inner sep=0] (near) at (far.east) {\fcolorbox{red}{white}{\includegraphics[width=0.035\linewidth]{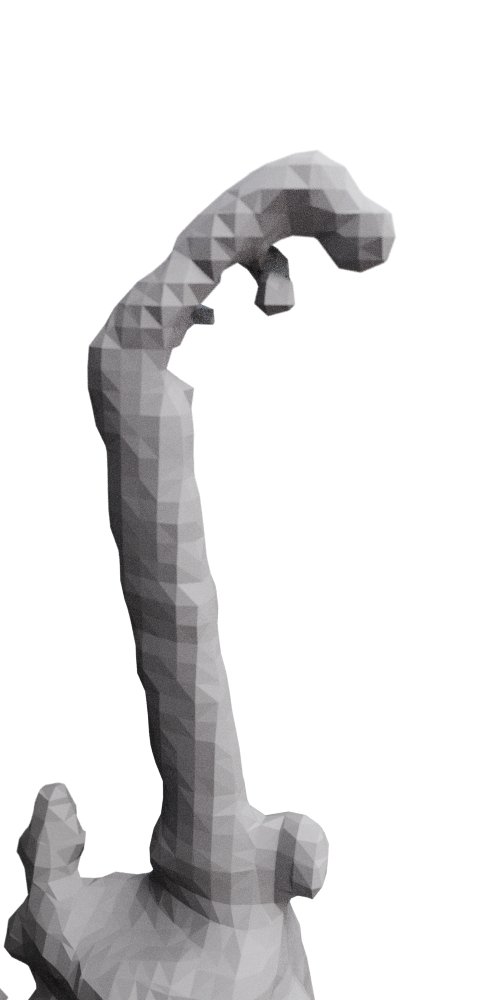}}};
    \end{tikzpicture} &
    \begin{tikzpicture}[baseline=(current bounding box.center)]
        \node[anchor=south west,inner sep=0] (far) at (0,0) {\includegraphics[width=0.070\linewidth]{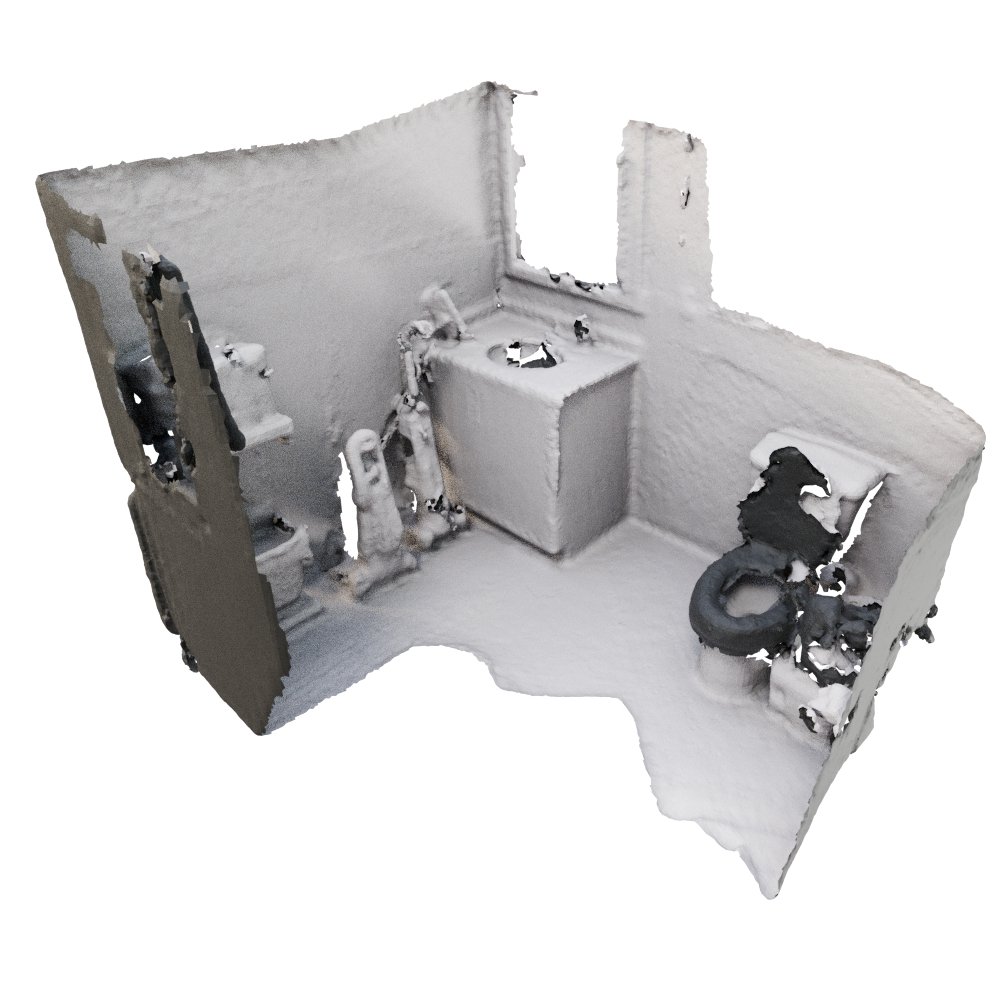}};
        \draw[red,thin]  (0.515,0.78) rectangle (0.565,0.88);
        \node[anchor=west,inner sep=0] (near) at (far.east) {\fcolorbox{red}{white}{\includegraphics[width=0.035\linewidth]{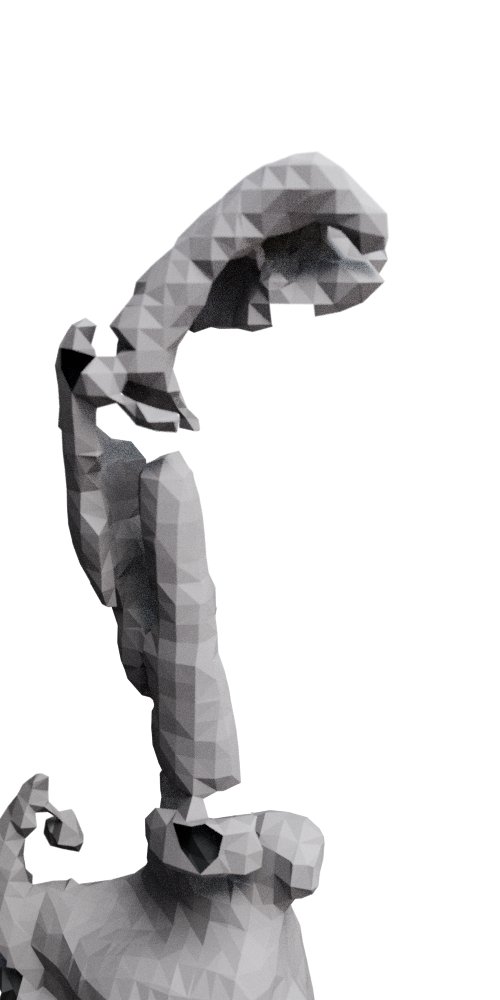}}};
    \end{tikzpicture} &
    \begin{tikzpicture}[baseline=(current bounding box.center)]
        \node[anchor=south west,inner sep=0] (far) at (0,0) {\includegraphics[width=0.070\linewidth]{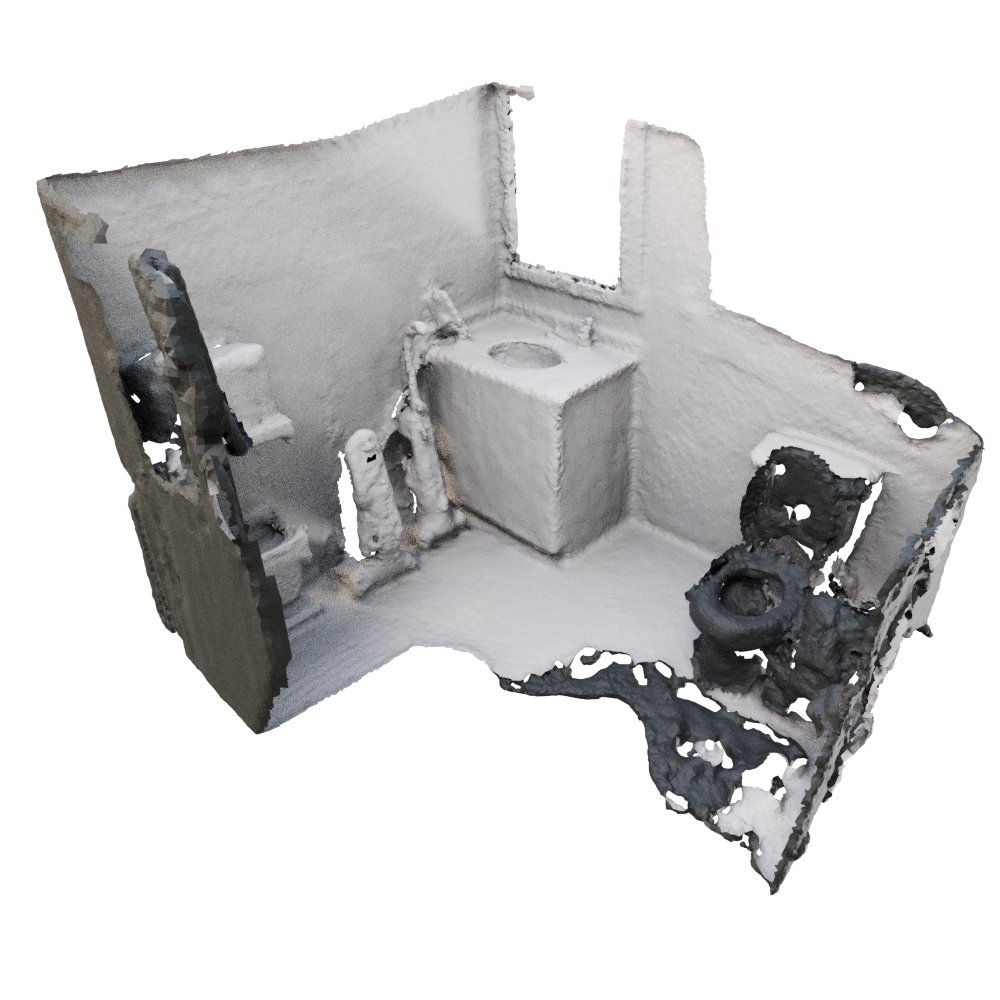}};
        \draw[red,thin]  (0.515,0.78) rectangle (0.565,0.88);
        \node[anchor=west,inner sep=0] (near) at (far.east) {\fcolorbox{red}{white}{\includegraphics[width=0.035\linewidth]{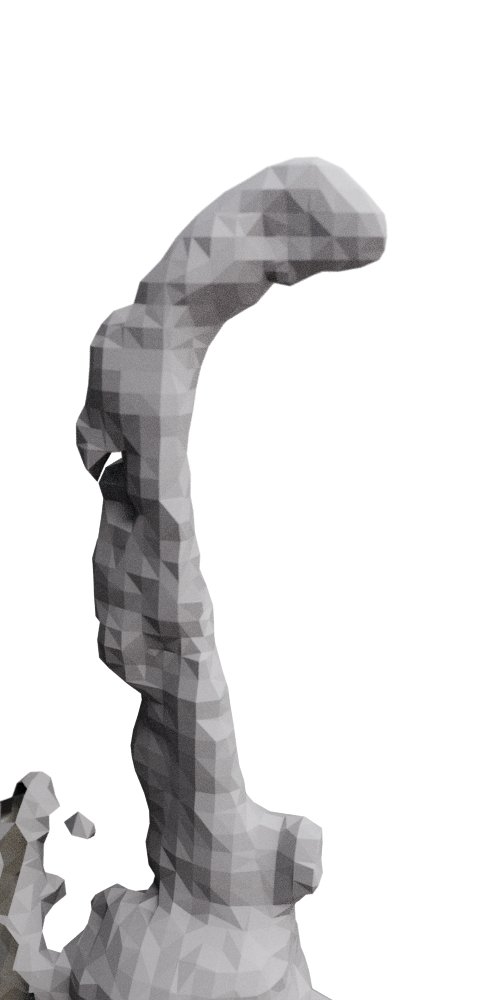}}};
    \end{tikzpicture} &
    \begin{tikzpicture}[baseline=(current bounding box.center)]
        \node[anchor=south west,inner sep=0] (far) at (0,0) {\includegraphics[width=0.070\linewidth]{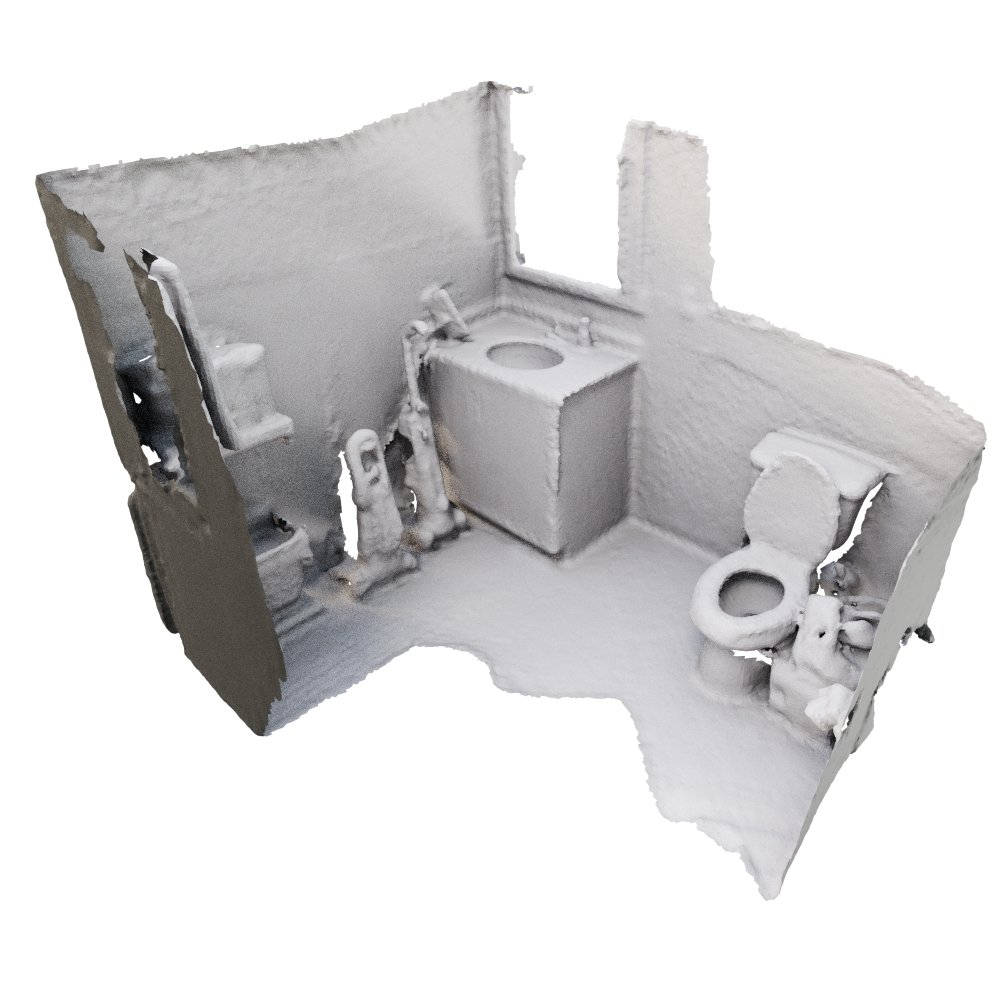}};
        \draw[red,thin]  (0.515,0.78) rectangle (0.565,0.88);
        \node[anchor=west,inner sep=0] (near) at (far.east) {\fcolorbox{red}{white}{\includegraphics[width=0.035\linewidth]{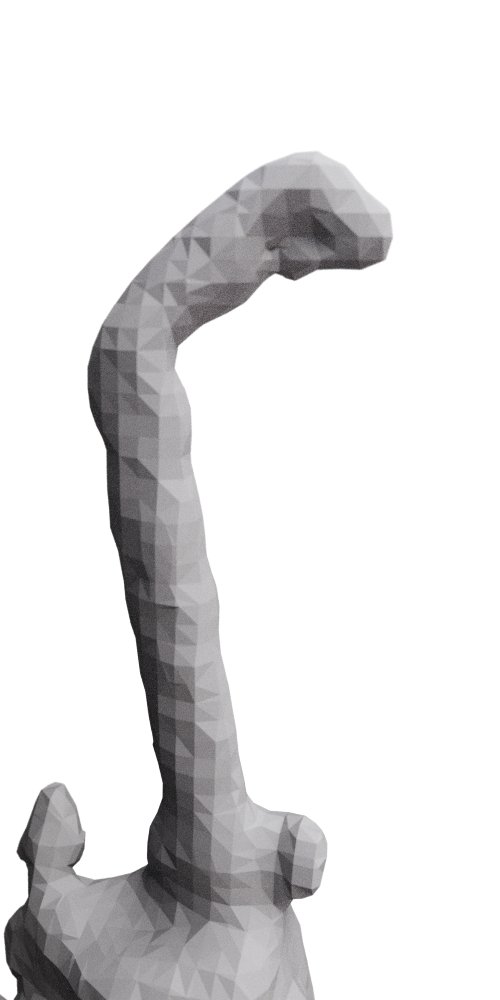}}};
    \end{tikzpicture} &
    \begin{tikzpicture}[baseline=(current bounding box.center)]
        \node[anchor=south west,inner sep=0] (far) at (0,0) {\includegraphics[width=0.070\linewidth]{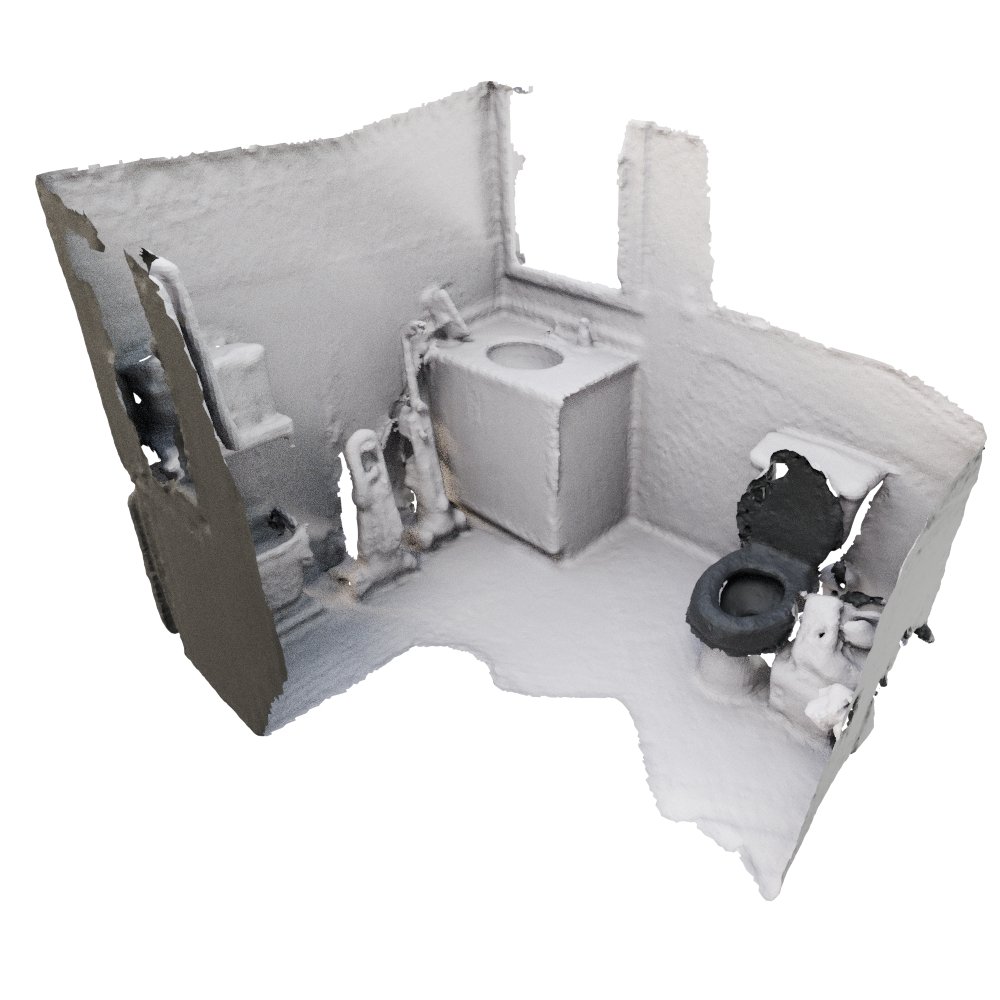}};
        \draw[red,thin]  (0.515,0.78) rectangle (0.565,0.88);
        \node[anchor=west,inner sep=0] (near) at (far.east) {\fcolorbox{red}{white}{\includegraphics[width=0.035\linewidth]{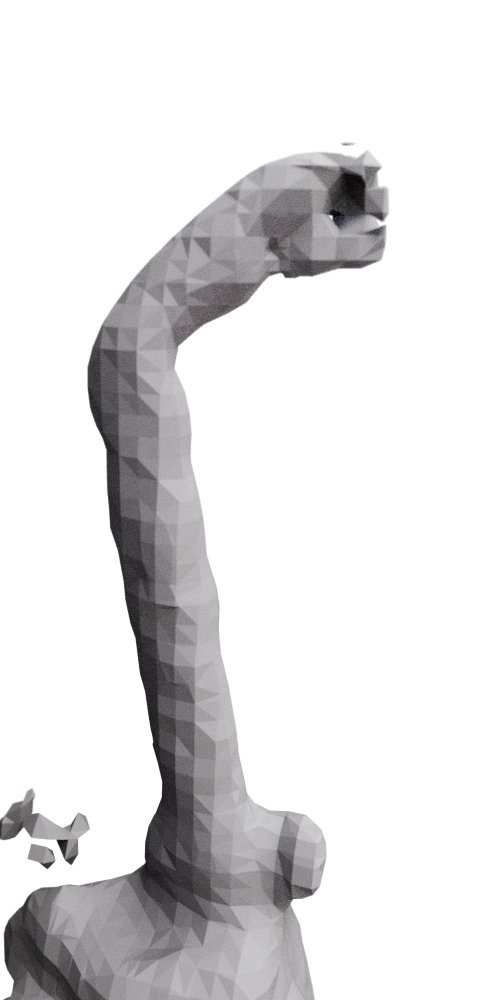}}};
    \end{tikzpicture} &
    \begin{tikzpicture}[baseline=(current bounding box.center)]
        \node[anchor=south west,inner sep=0] (far) at (0,0) {\includegraphics[width=0.070\linewidth]{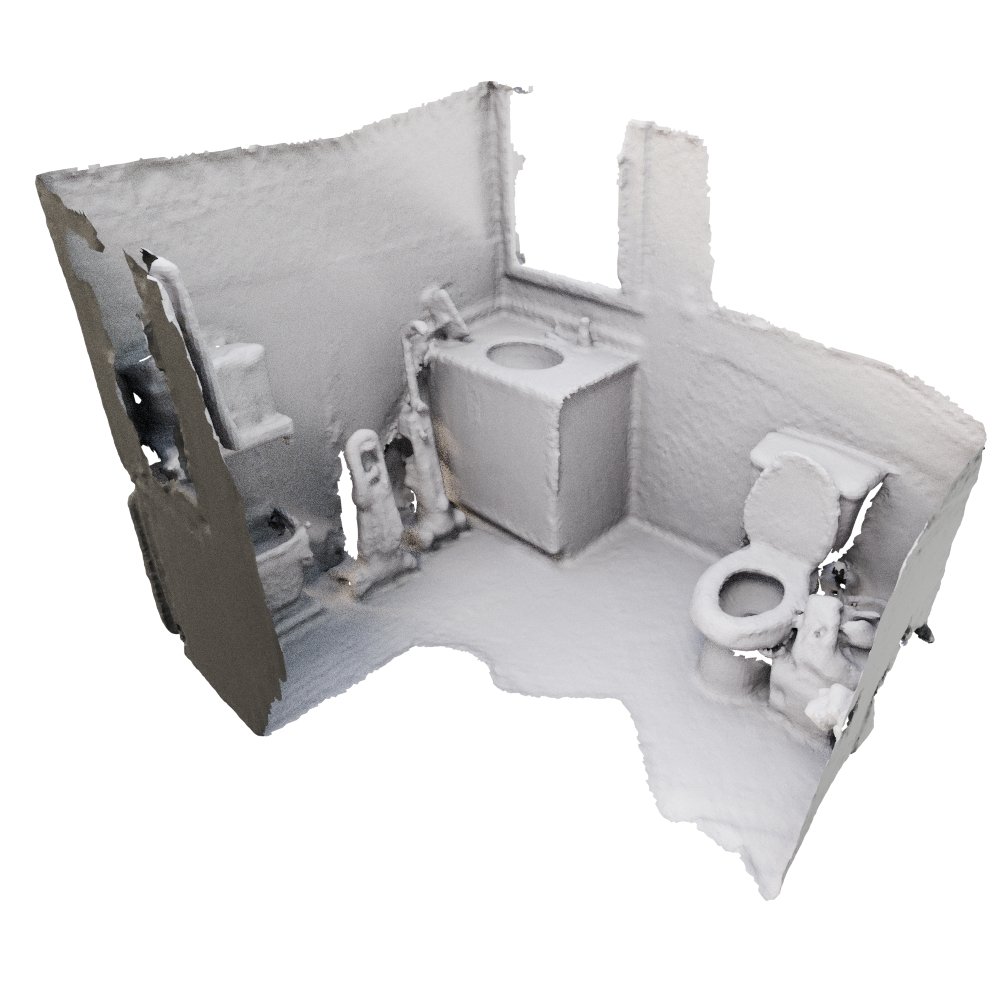}};
        \draw[red,thin]  (0.515,0.78) rectangle (0.565,0.88);
        \node[anchor=west,inner sep=0] (near) at (far.east) {\fcolorbox{red}{white}{\includegraphics[width=0.035\linewidth]{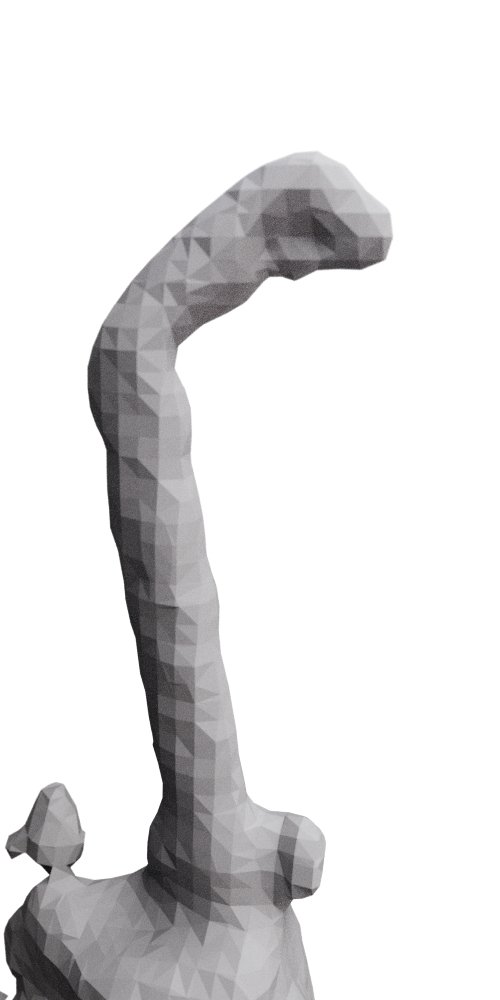}}};
    \end{tikzpicture} &
    \begin{tikzpicture}[baseline=(current bounding box.center)]
        \node[anchor=south west,inner sep=0] (far) at (0,0) {\includegraphics[width=0.070\linewidth]{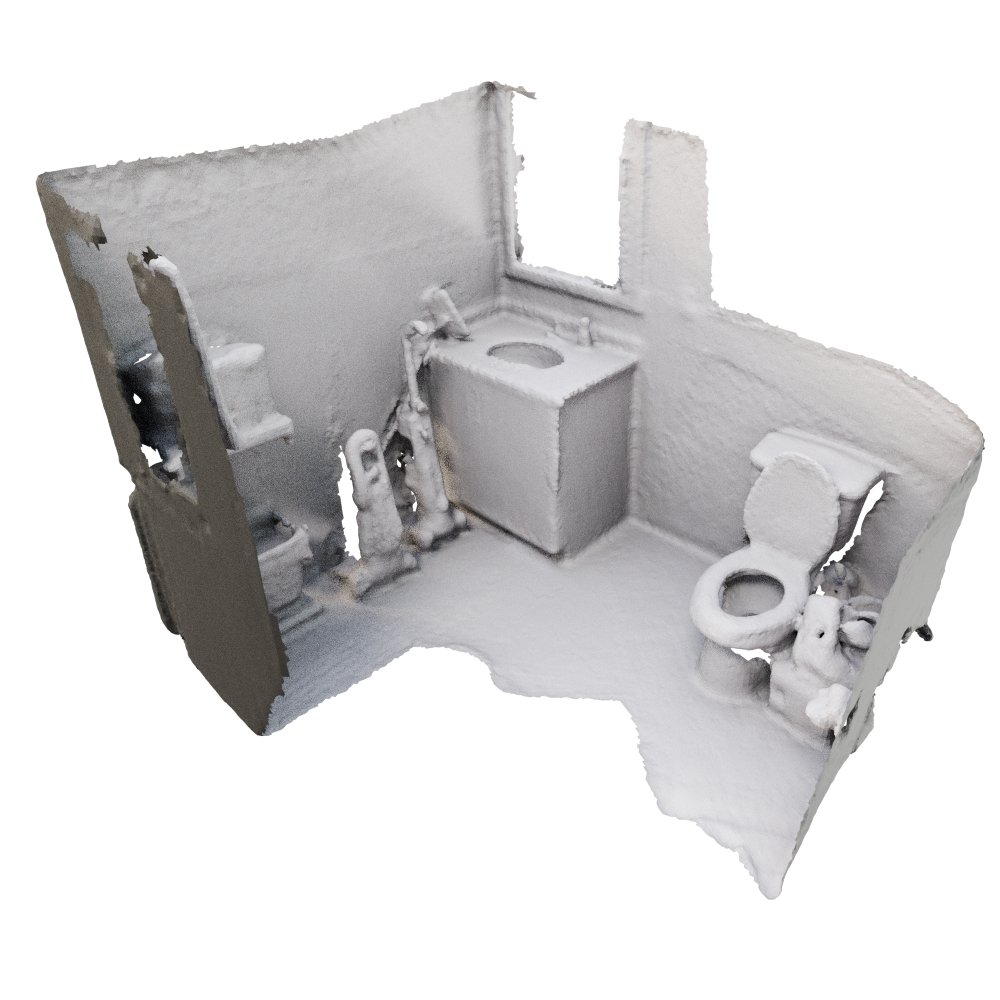}};
        \draw[red,thin]  (0.515,0.78) rectangle (0.565,0.88);
        \node[anchor=west,inner sep=0] (near) at (far.east) {\fcolorbox{red}{white}{\includegraphics[width=0.035\linewidth]{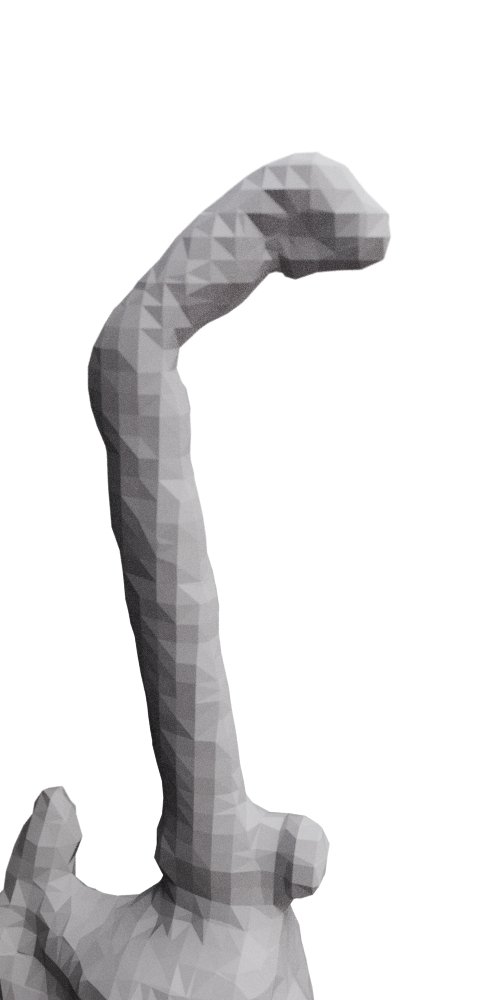}}};
    \end{tikzpicture} & \\[2ex]

    \begin{tikzpicture}[baseline=(current bounding box.center)]
        \node[anchor=south west,inner sep=0] (far) at (0,0) {\includegraphics[width=0.070\linewidth]{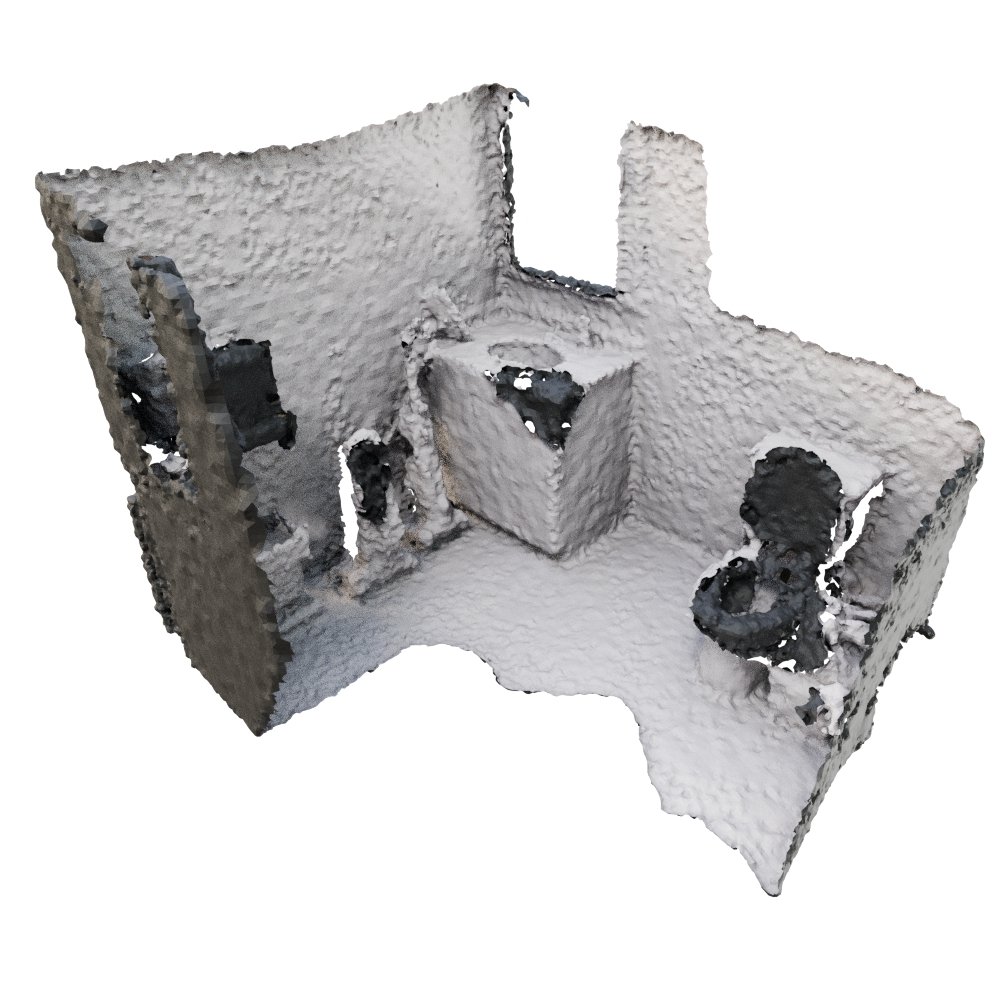}};
        \draw[red,thin]  (0.515,0.78) rectangle (0.565,0.88);
        \node[anchor=west,inner sep=0] (near) at (far.east) {\fcolorbox{red}{white}{\includegraphics[width=0.035\linewidth]{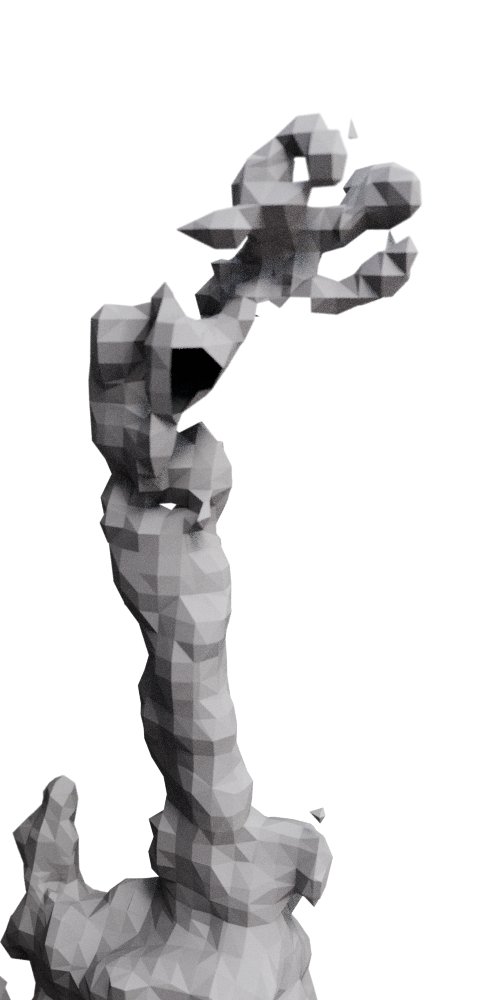}}};
    \end{tikzpicture} &
    \begin{tikzpicture}[baseline=(current bounding box.center)]
        \node[anchor=south west,inner sep=0] (far) at (0,0) {\includegraphics[width=0.070\linewidth]{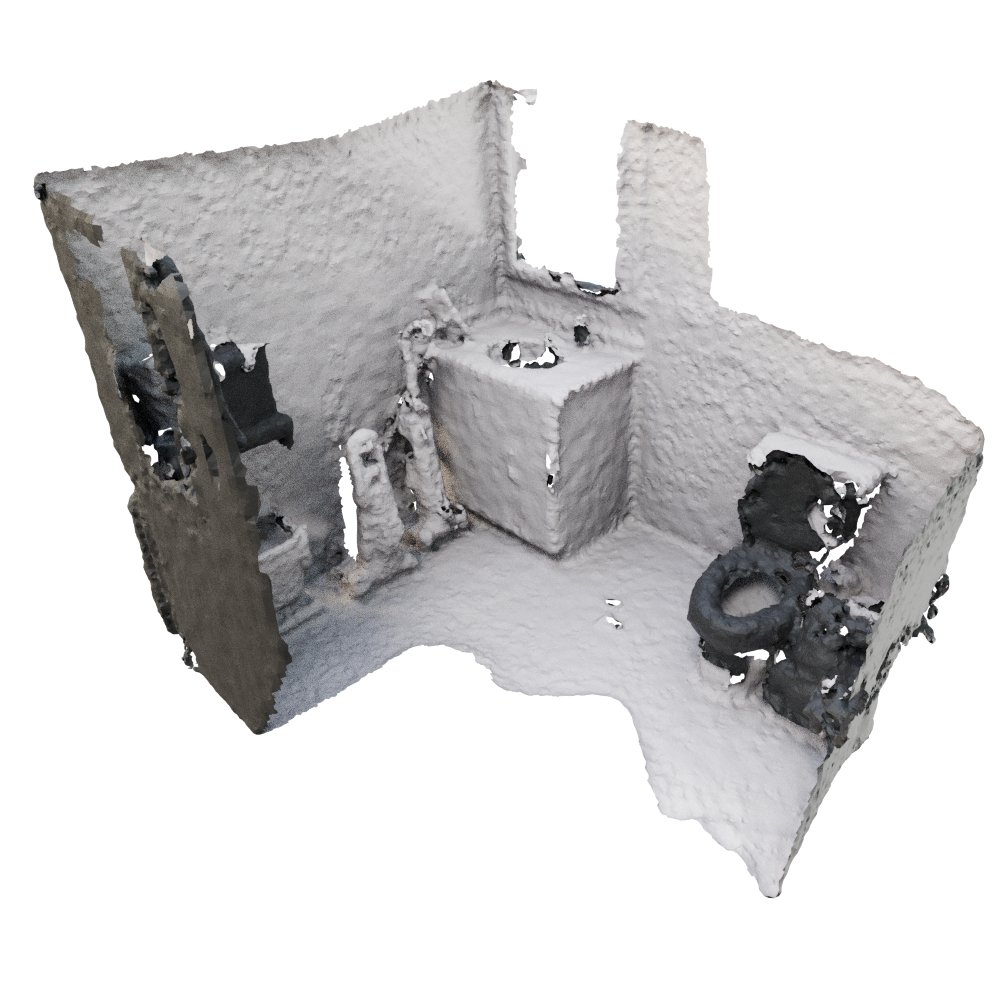}};
        \draw[red,thin]  (0.515,0.78) rectangle (0.565,0.88);
        \node[anchor=west,inner sep=0] (near) at (far.east) {\fcolorbox{red}{white}{\includegraphics[width=0.035\linewidth]{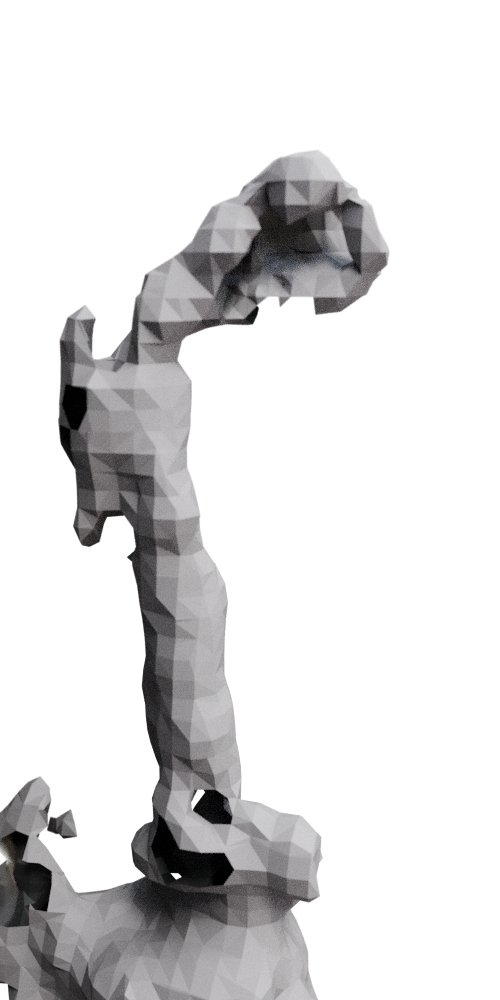}}};
    \end{tikzpicture} &
    \begin{tikzpicture}[baseline=(current bounding box.center)]
        \node[anchor=south west,inner sep=0] (far) at (0,0) {\includegraphics[width=0.070\linewidth]{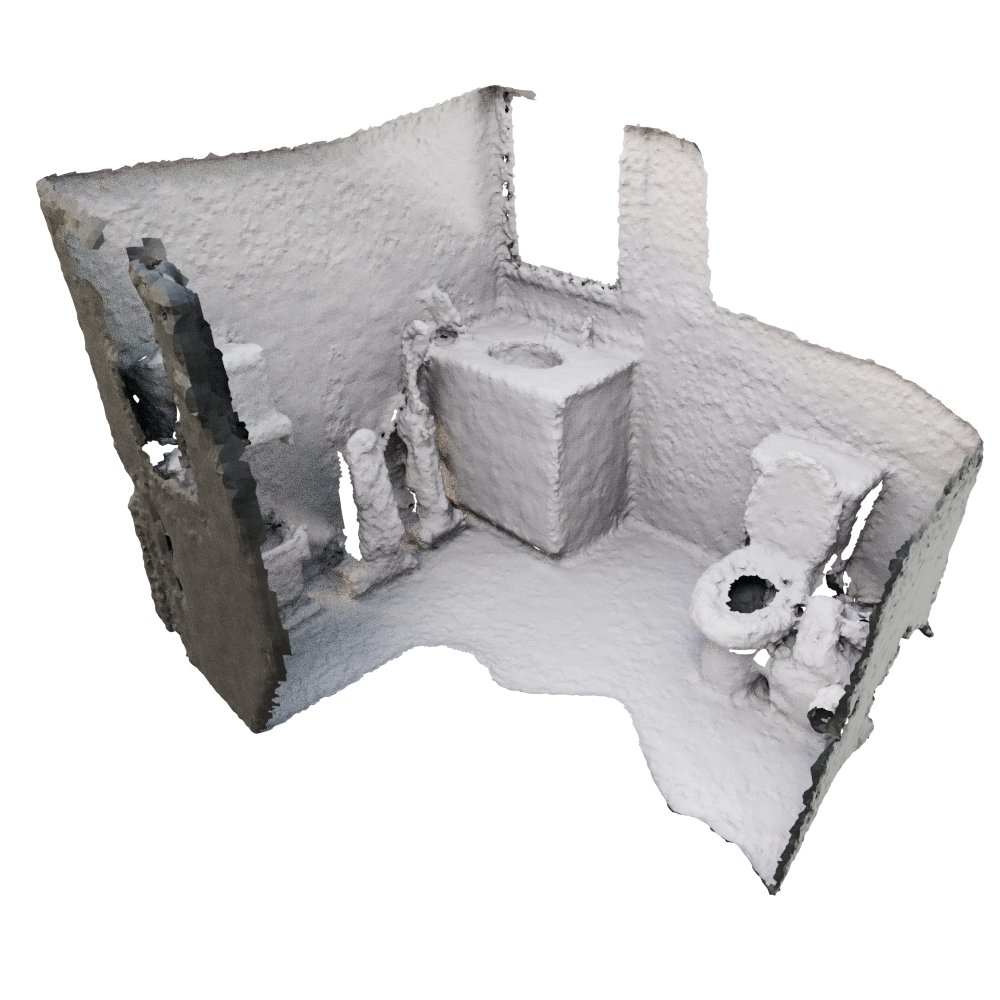}};
        \draw[red,thin]  (0.515,0.78) rectangle (0.565,0.88);
        \node[anchor=west,inner sep=0] (near) at (far.east) {\fcolorbox{red}{white}{\includegraphics[width=0.035\linewidth]{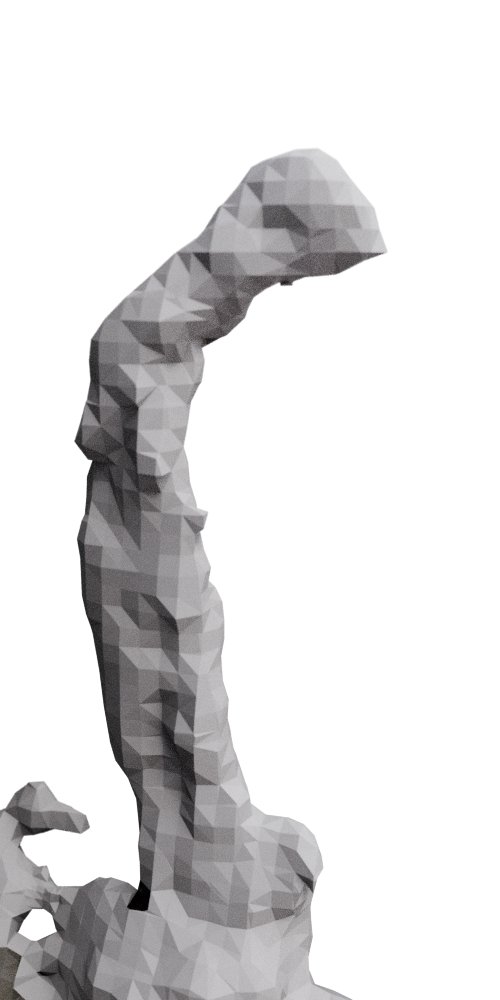}}};
    \end{tikzpicture} &
    \begin{tikzpicture}[baseline=(current bounding box.center)]
        \node[anchor=south west,inner sep=0] (far) at (0,0) {\includegraphics[width=0.070\linewidth]{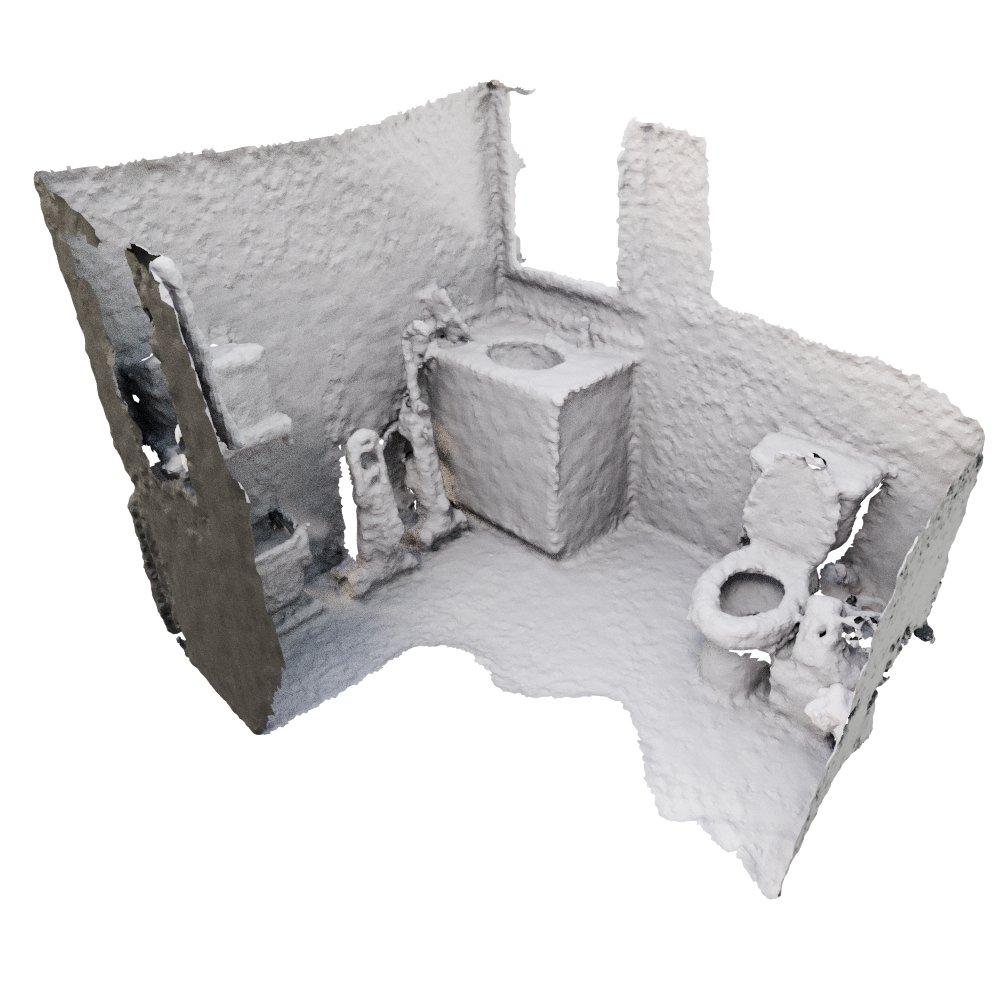}};
        \draw[red,thin]  (0.515,0.78) rectangle (0.565,0.88);
        \node[anchor=west,inner sep=0] (near) at (far.east) {\fcolorbox{red}{white}{\includegraphics[width=0.035\linewidth]{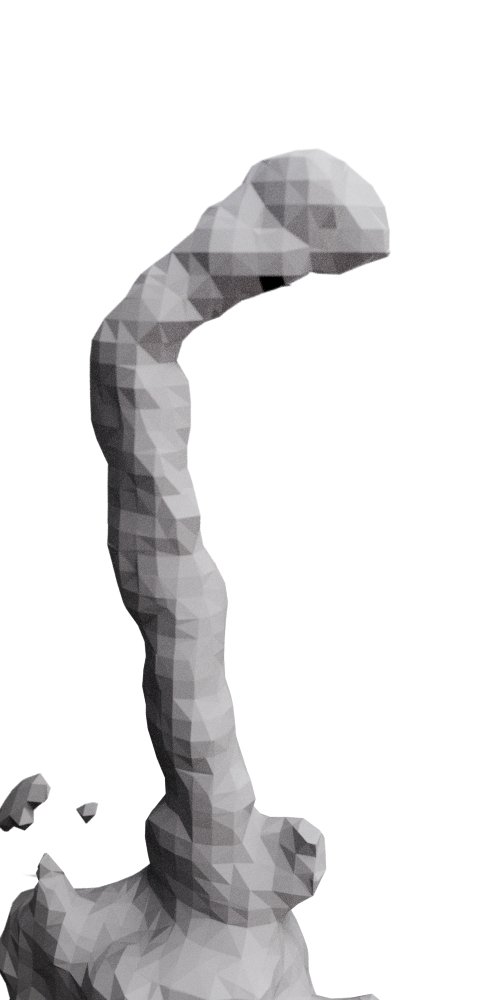}}};
    \end{tikzpicture} &
    \begin{tikzpicture}[baseline=(current bounding box.center)]
        \node[anchor=south west,inner sep=0] (far) at (0,0) {\includegraphics[width=0.070\linewidth]{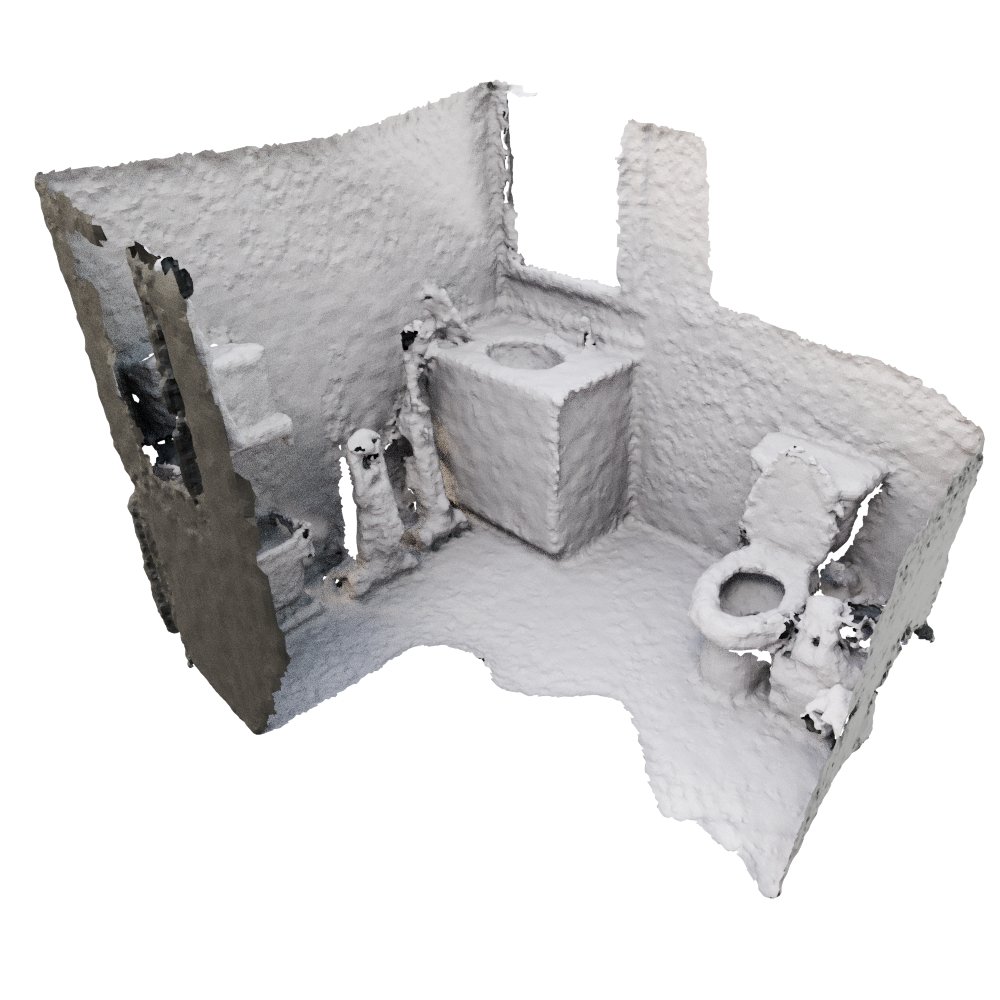}};
        \draw[red,thin]  (0.515,0.78) rectangle (0.565,0.88);
        \node[anchor=west,inner sep=0] (near) at (far.east) {\fcolorbox{red}{white}{\includegraphics[width=0.035\linewidth]{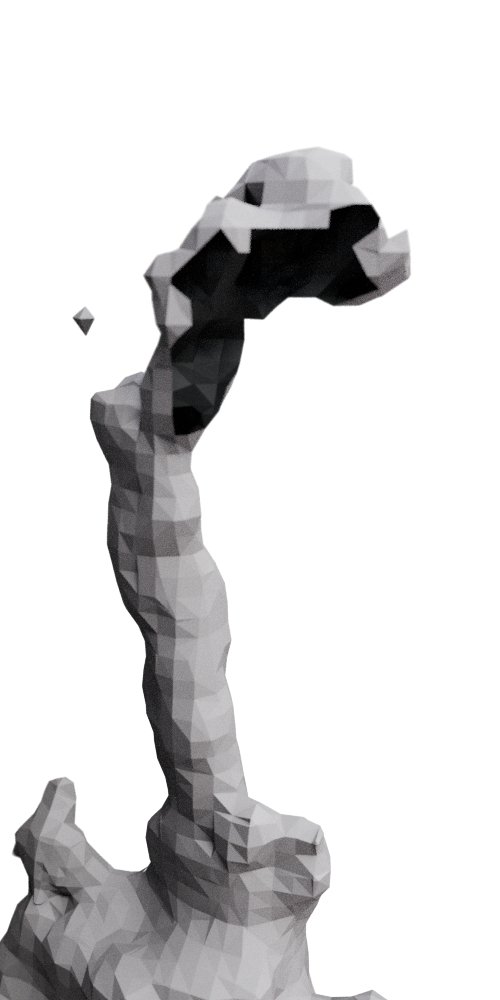}}};
    \end{tikzpicture} &
    \begin{tikzpicture}[baseline=(current bounding box.center)]
        \node[anchor=south west,inner sep=0] (far) at (0,0) {\includegraphics[width=0.070\linewidth]{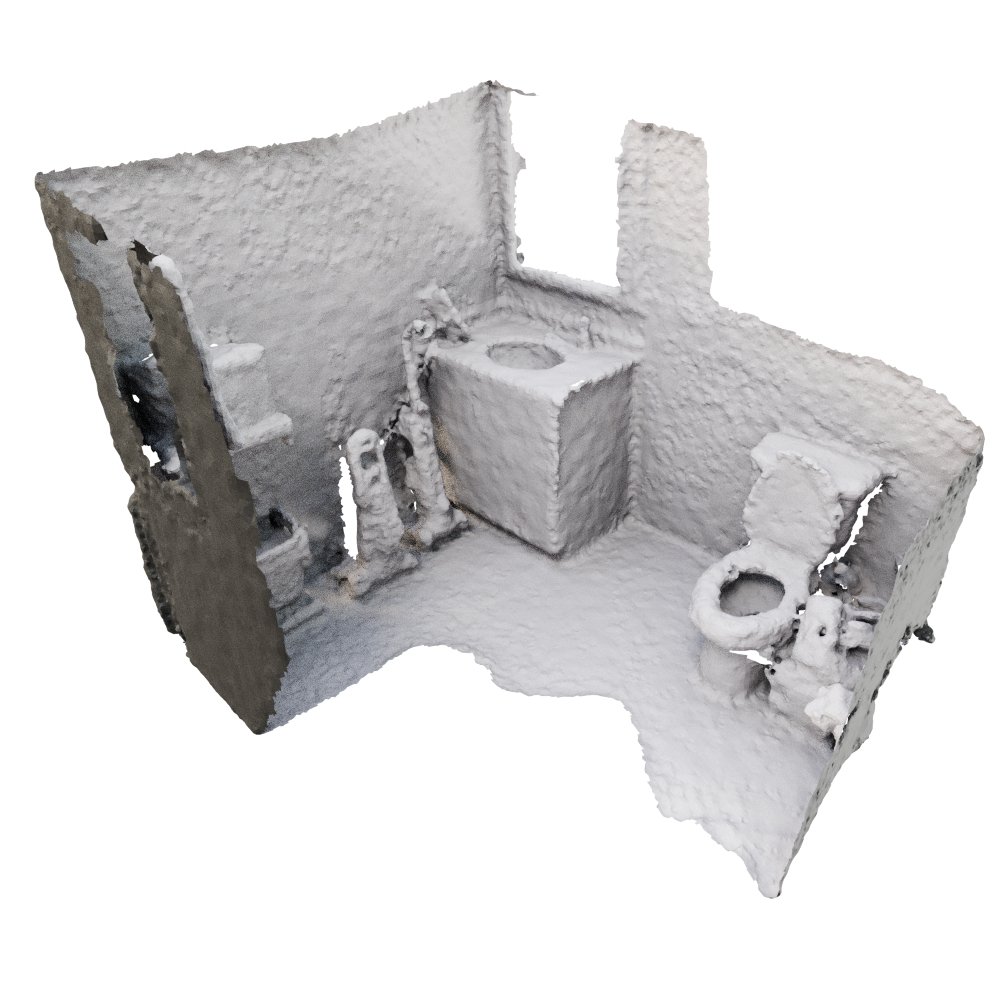}};
        \draw[red,thin]  (0.515,0.78) rectangle (0.565,0.88);
        \node[anchor=west,inner sep=0] (near) at (far.east) {\fcolorbox{red}{white}{\includegraphics[width=0.035\linewidth]{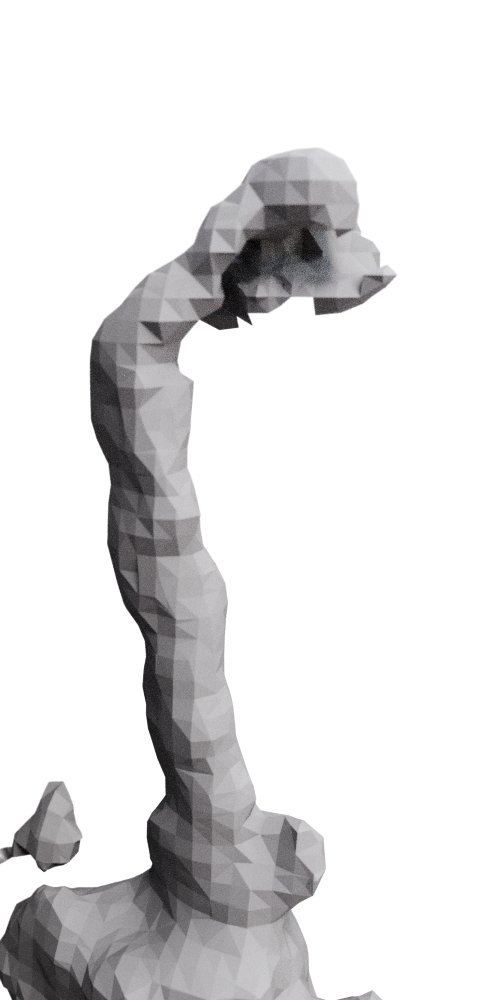}}};
    \end{tikzpicture} &
    \begin{tikzpicture}[baseline=(current bounding box.center)]
        \node[anchor=south west,inner sep=0] (far) at (0,0) {\includegraphics[width=0.070\linewidth]{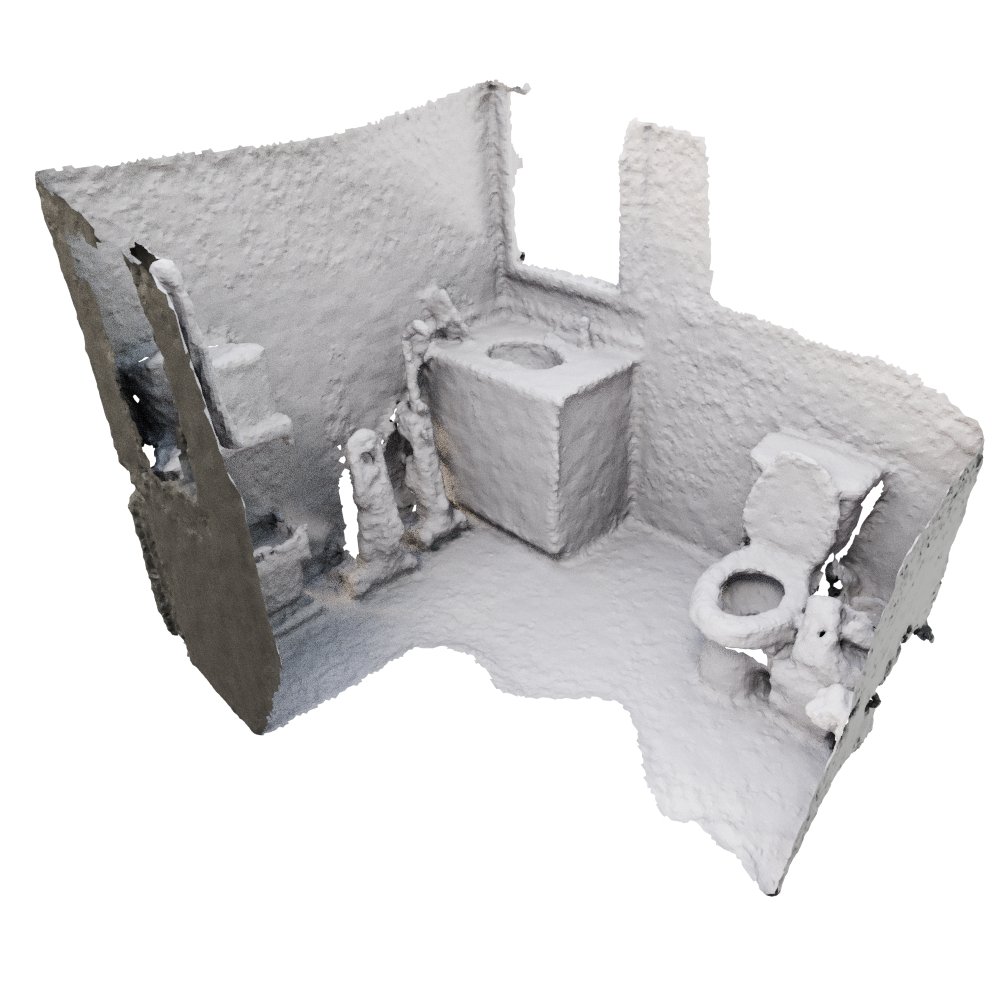}};
        \draw[red,thin]  (0.515,0.78) rectangle (0.565,0.88);
        \node[anchor=west,inner sep=0] (near) at (far.east) {\fcolorbox{red}{white}{\includegraphics[width=0.035\linewidth]{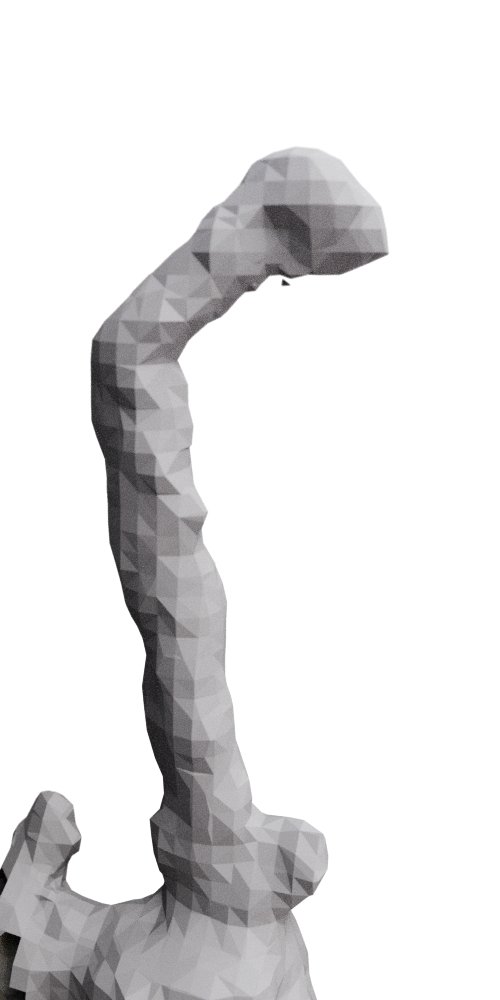}}};
    \end{tikzpicture} &
    \begin{tikzpicture}[baseline=(current bounding box.center)]
        \node[anchor=south west,inner sep=0] (far) at (0,0) {\includegraphics[width=0.070\linewidth]{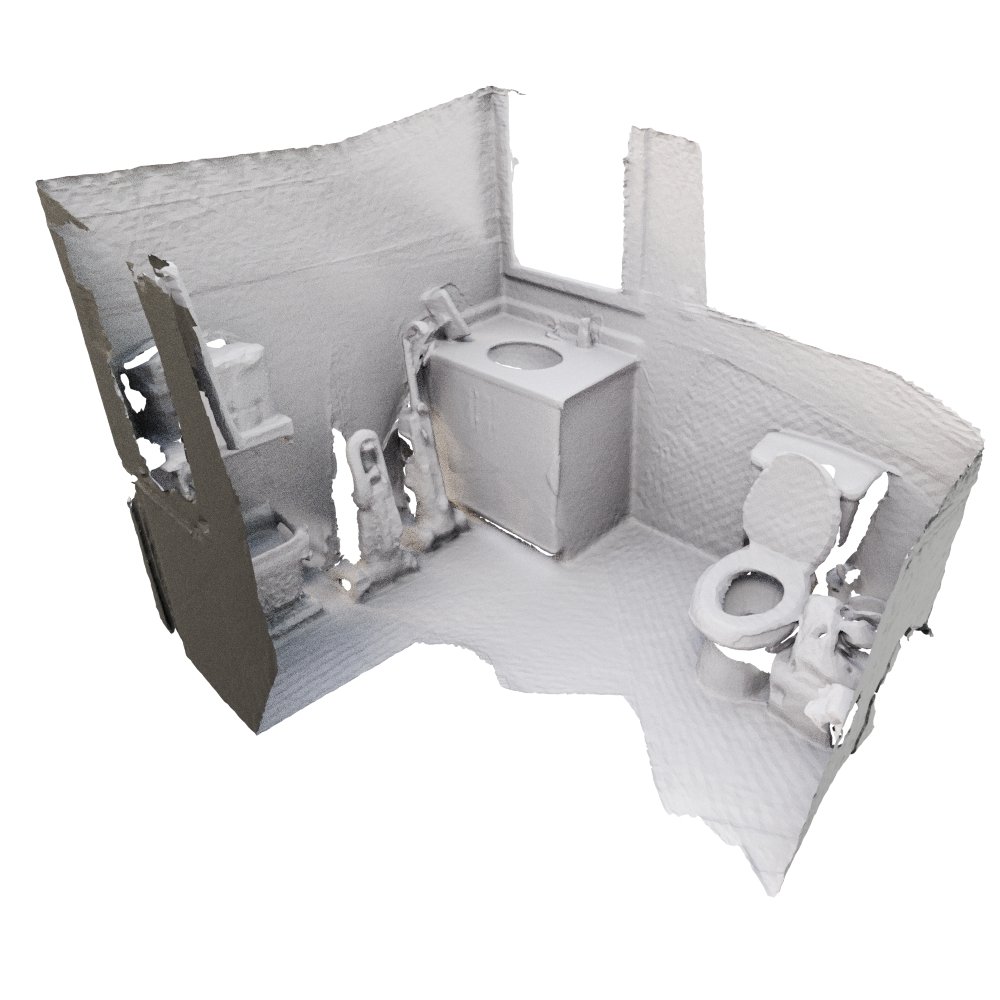}};
        \draw[red,thin]  (0.515,0.78) rectangle (0.565,0.88);
        \node[anchor=west,inner sep=0] (near) at (far.east) {\fcolorbox{red}{white}{\includegraphics[width=0.035\linewidth]{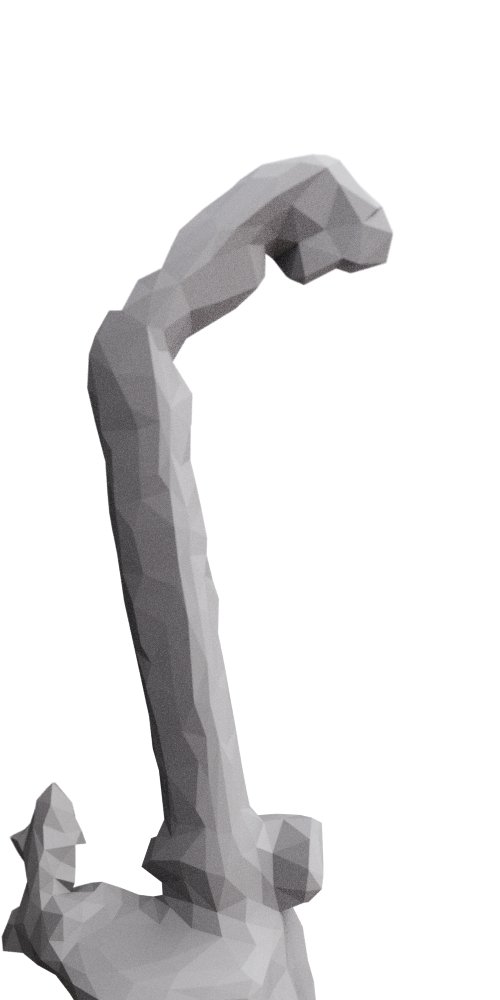}}};
    \end{tikzpicture} \\[2ex]

    \begin{tikzpicture}[baseline=(current bounding box.center)]
        \node[anchor=south west,inner sep=0] (far) at (0,0) {\includegraphics[width=0.070\linewidth]{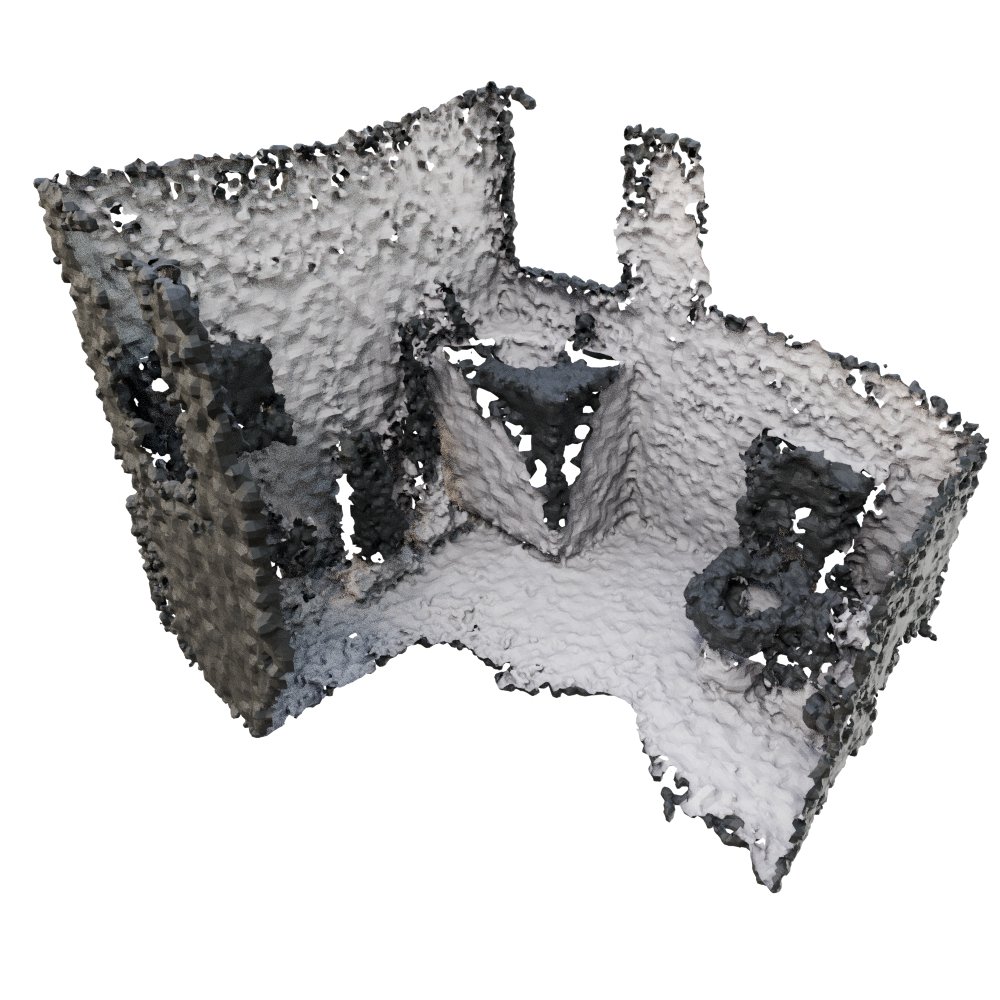}};
        \draw[red,thin]  (0.515,0.78) rectangle (0.565,0.88);
        \node[anchor=west,inner sep=0] (near) at (far.east) {\fcolorbox{red}{white}{\includegraphics[width=0.035\linewidth]{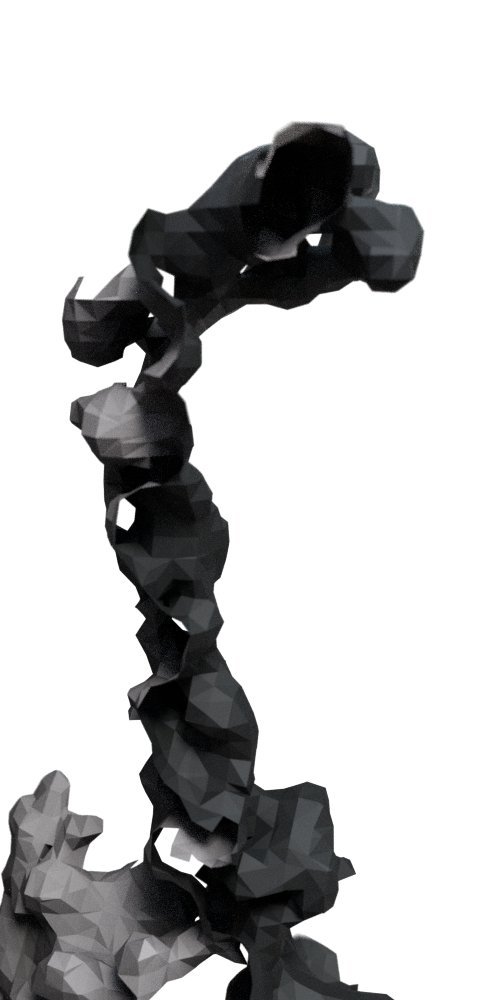}}};
    \end{tikzpicture} &
    \begin{tikzpicture}[baseline=(current bounding box.center)]
        \node[anchor=south west,inner sep=0] (far) at (0,0) {\includegraphics[width=0.070\linewidth]{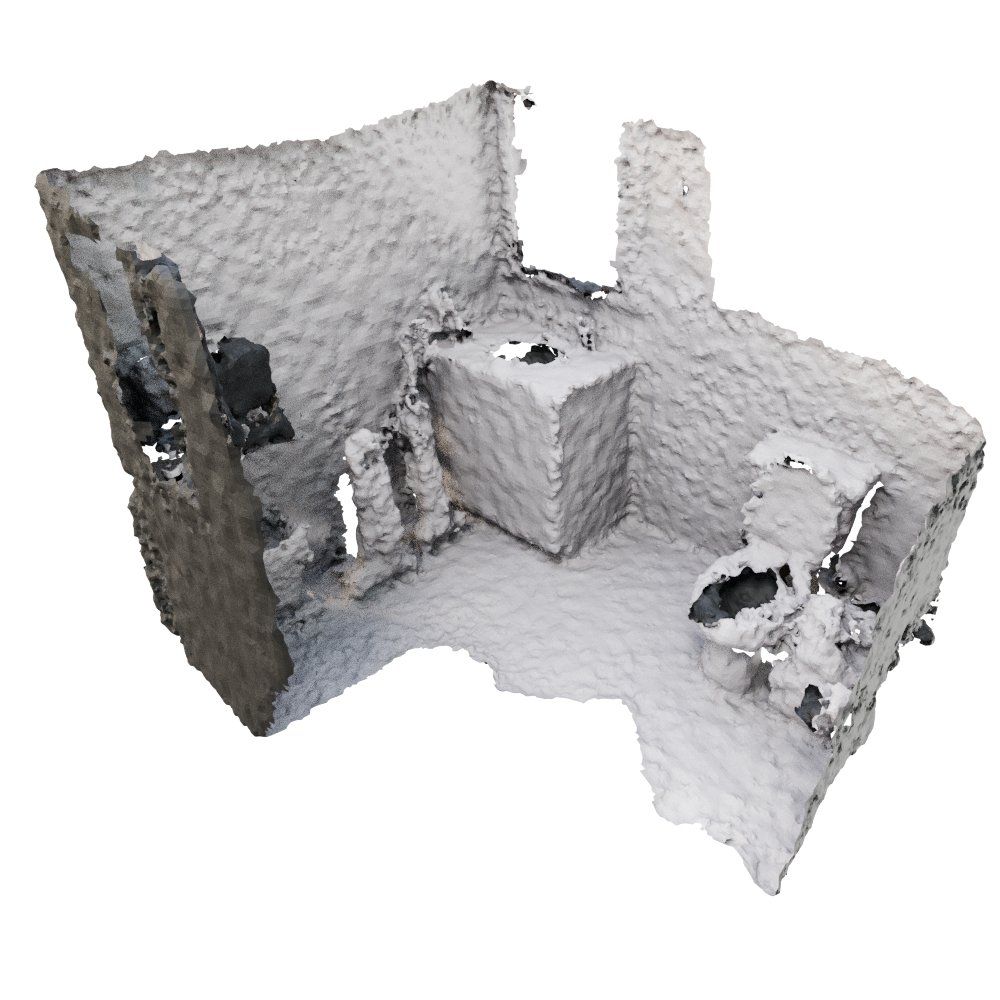}};
        \draw[red,thin]  (0.515,0.78) rectangle (0.565,0.88);
        \node[anchor=west,inner sep=0] (near) at (far.east) {\fcolorbox{red}{white}{\includegraphics[width=0.035\linewidth]{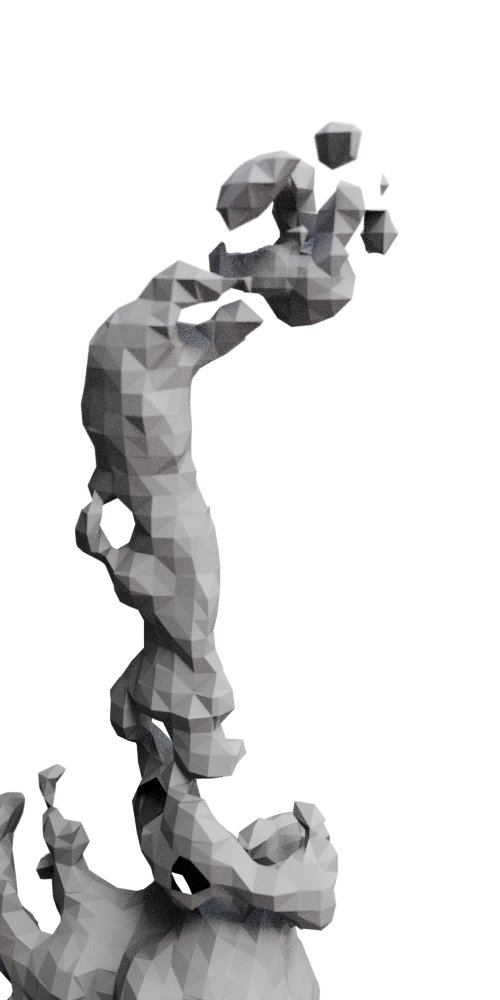}}};
    \end{tikzpicture} &
    \begin{tikzpicture}[baseline=(current bounding box.center)]
        \node[anchor=south west,inner sep=0] (far) at (0,0) {\includegraphics[width=0.070\linewidth]{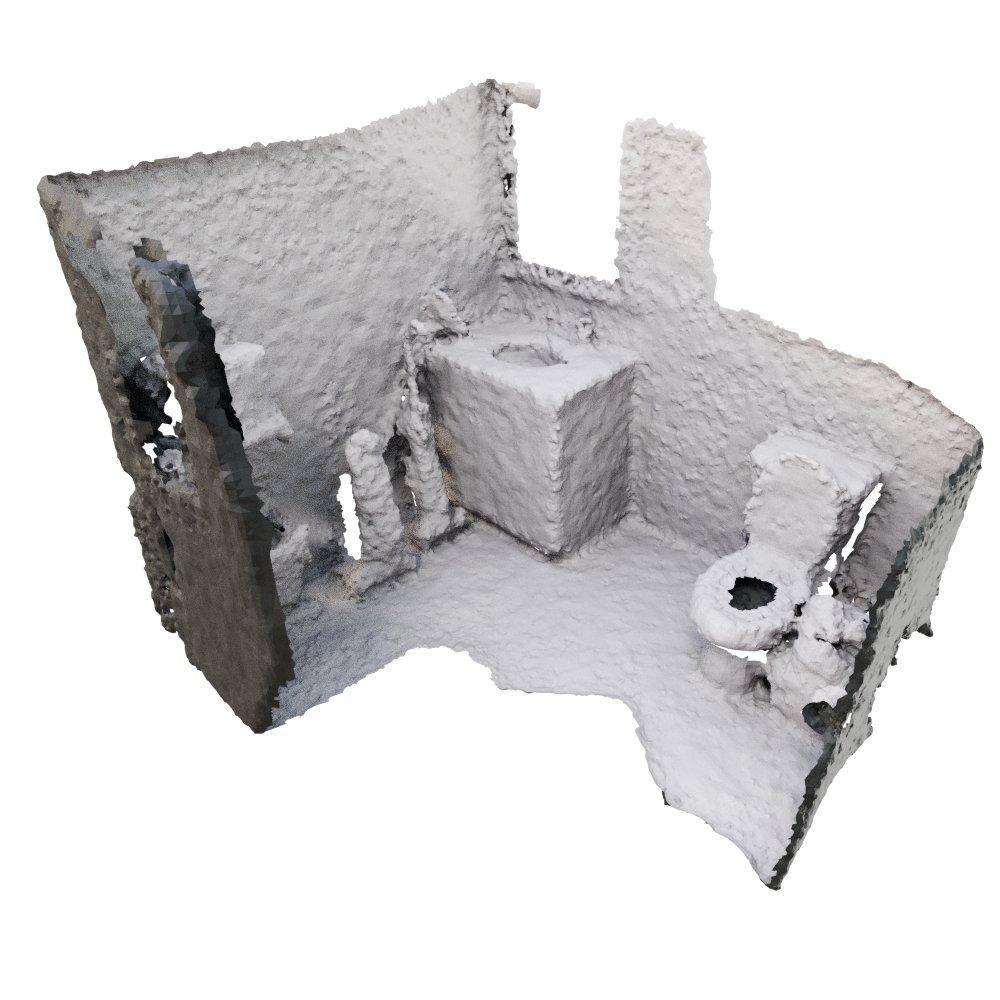}};
        \draw[red,thin]  (0.515,0.78) rectangle (0.565,0.88);
        \node[anchor=west,inner sep=0] (near) at (far.east) {\fcolorbox{red}{white}{\includegraphics[width=0.035\linewidth]{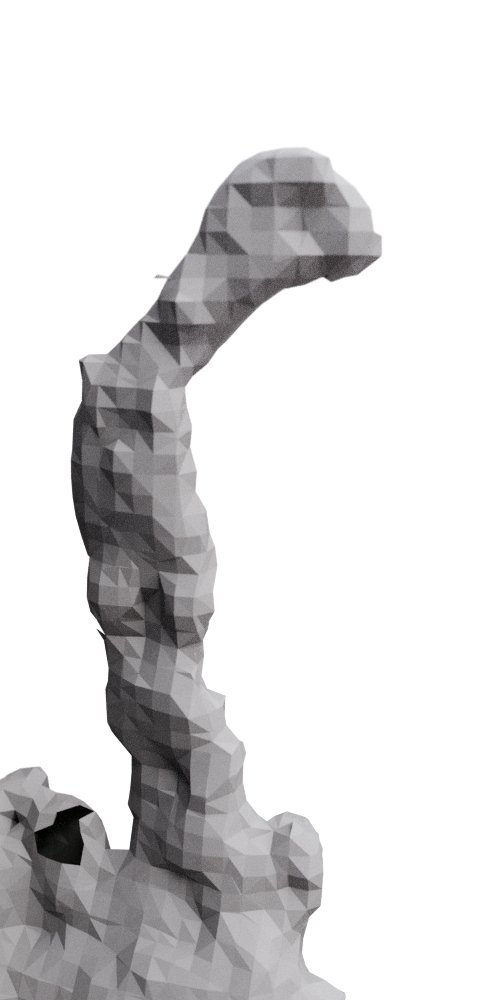}}};
    \end{tikzpicture} &
    \begin{tikzpicture}[baseline=(current bounding box.center)]
        \node[anchor=south west,inner sep=0] (far) at (0,0) {\includegraphics[width=0.070\linewidth]{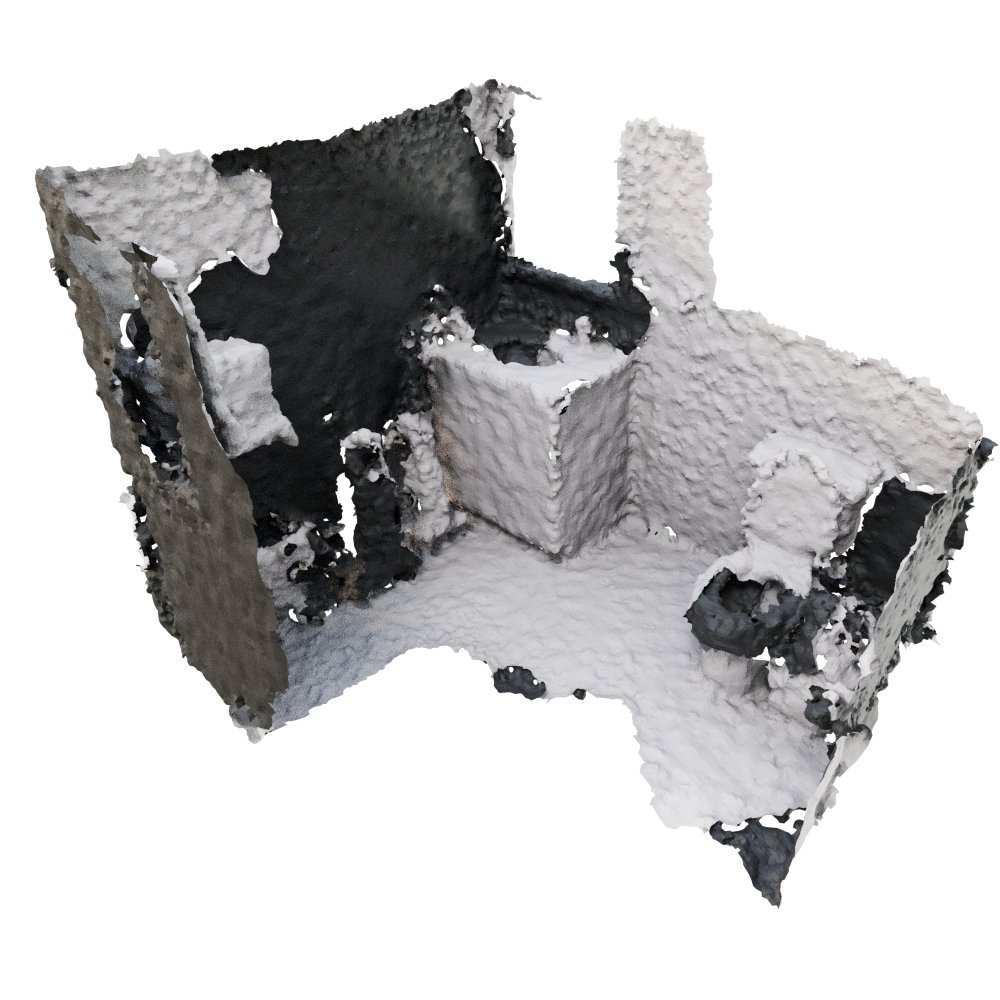}};
        \draw[red,thin]  (0.515,0.78) rectangle (0.565,0.88);
        \node[anchor=west,inner sep=0] (near) at (far.east) {\fcolorbox{red}{white}{\includegraphics[width=0.035\linewidth]{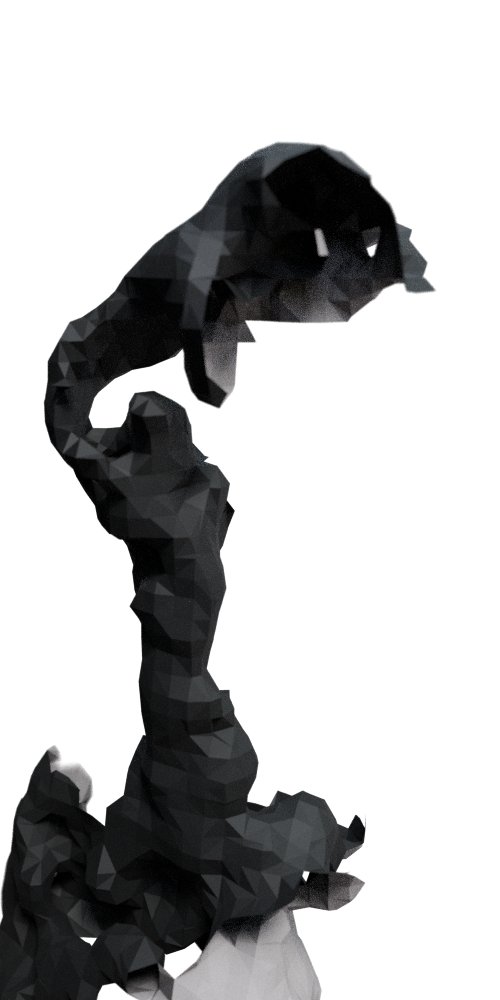}}};
    \end{tikzpicture} &
    \begin{tikzpicture}[baseline=(current bounding box.center)]
        \node[anchor=south west,inner sep=0] (far) at (0,0) {\includegraphics[width=0.070\linewidth]{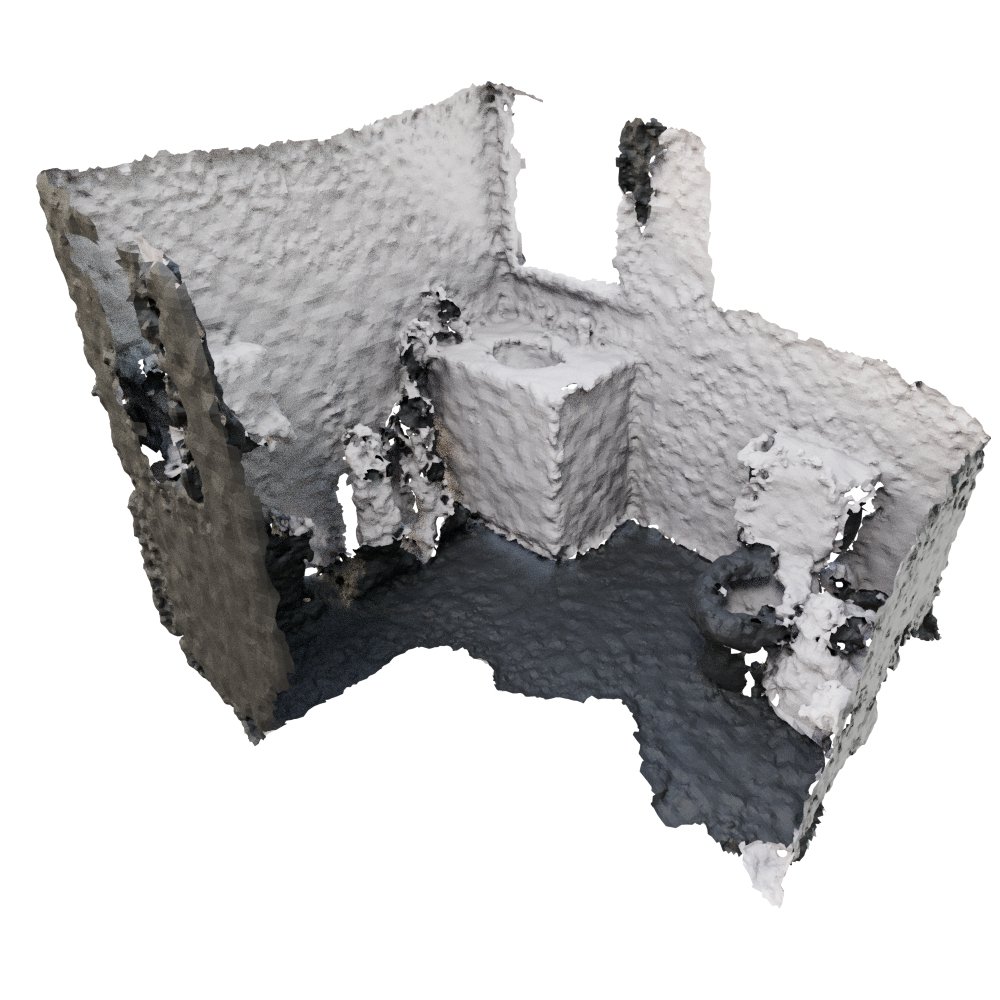}};
        \draw[red,thin]  (0.515,0.78) rectangle (0.565,0.88);
        \node[anchor=west,inner sep=0] (near) at (far.east) {\fcolorbox{red}{white}{\includegraphics[width=0.035\linewidth]{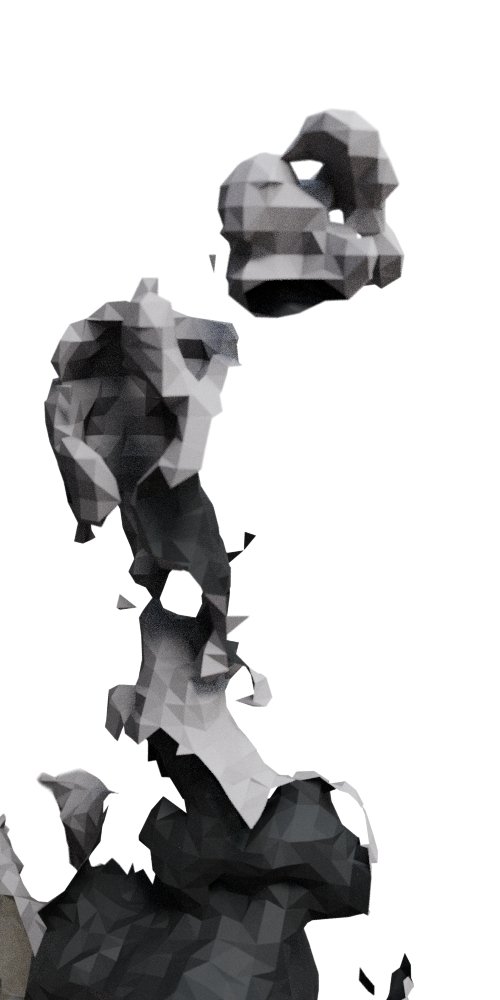}}};
    \end{tikzpicture} &
    \begin{tikzpicture}[baseline=(current bounding box.center)]
        \node[anchor=south west,inner sep=0] (far) at (0,0) {\includegraphics[width=0.070\linewidth]{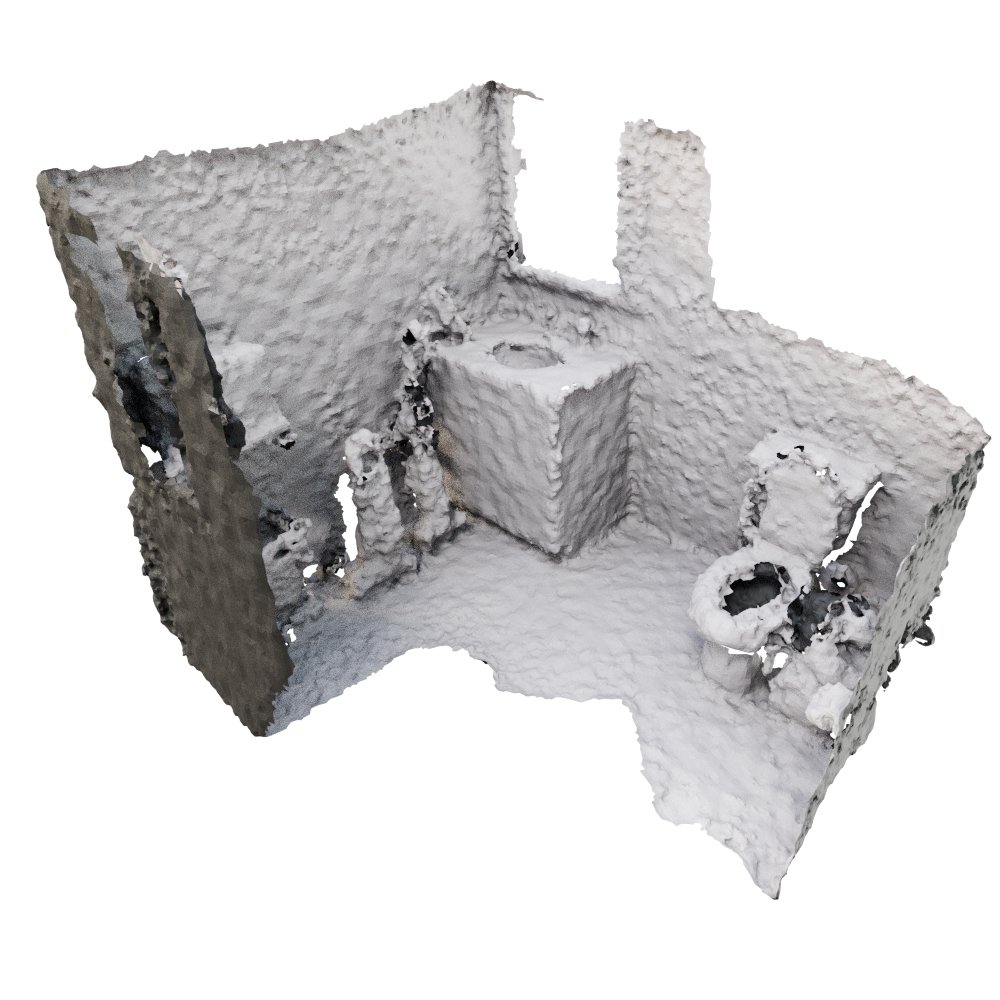}};
        \draw[red,thin]  (0.515,0.78) rectangle (0.565,0.88);
        \node[anchor=west,inner sep=0] (near) at (far.east) {\fcolorbox{red}{white}{\includegraphics[width=0.035\linewidth]{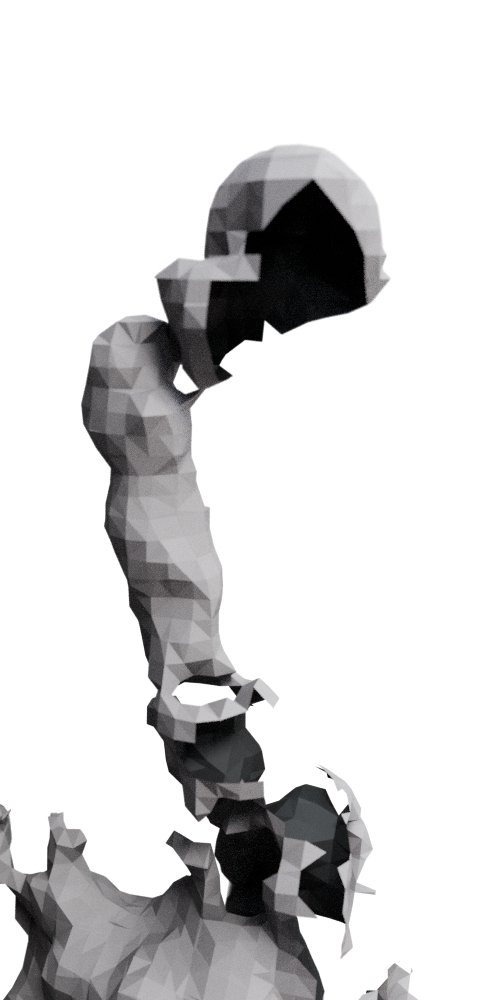}}};
    \end{tikzpicture} &
    \begin{tikzpicture}[baseline=(current bounding box.center)]
        \node[anchor=south west,inner sep=0] (far) at (0,0) {\includegraphics[width=0.070\linewidth]{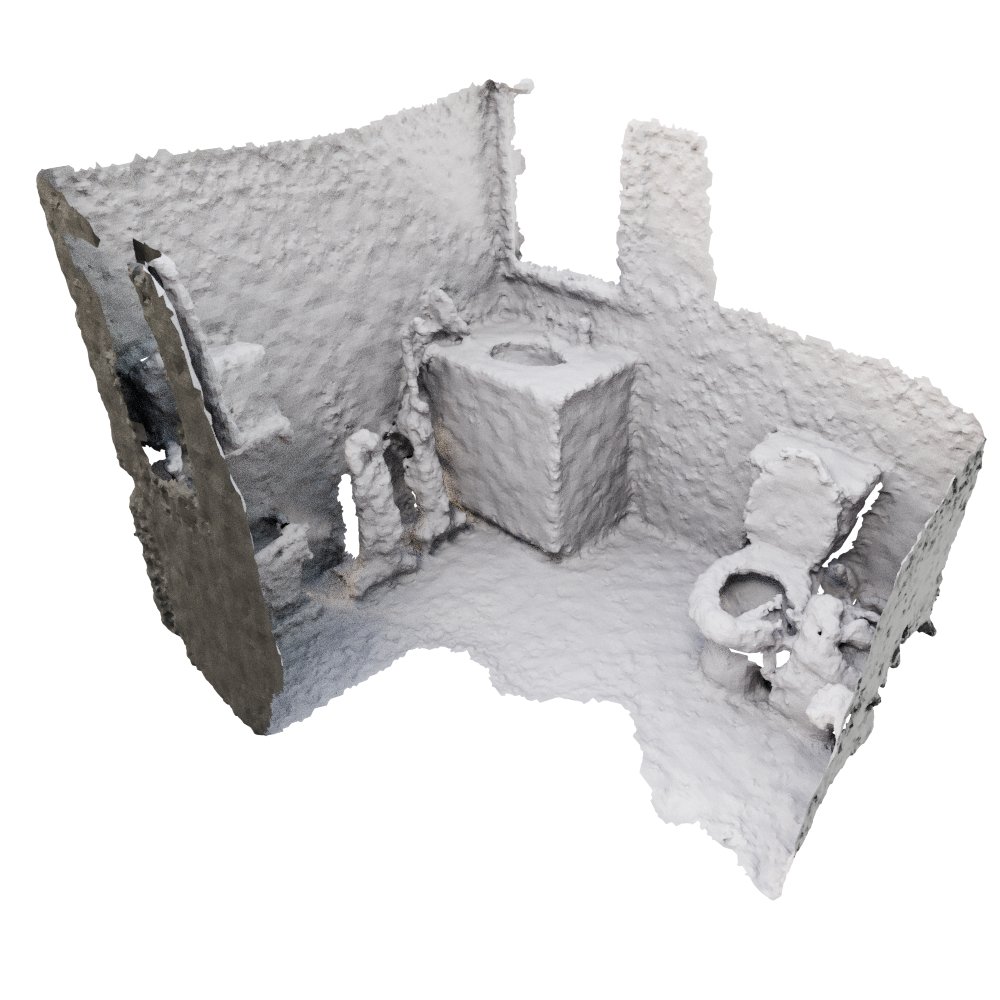}};
        \draw[red,thin]  (0.515,0.78) rectangle (0.565,0.88);
        \node[anchor=west,inner sep=0] (near) at (far.east) {\fcolorbox{red}{white}{\includegraphics[width=0.035\linewidth]{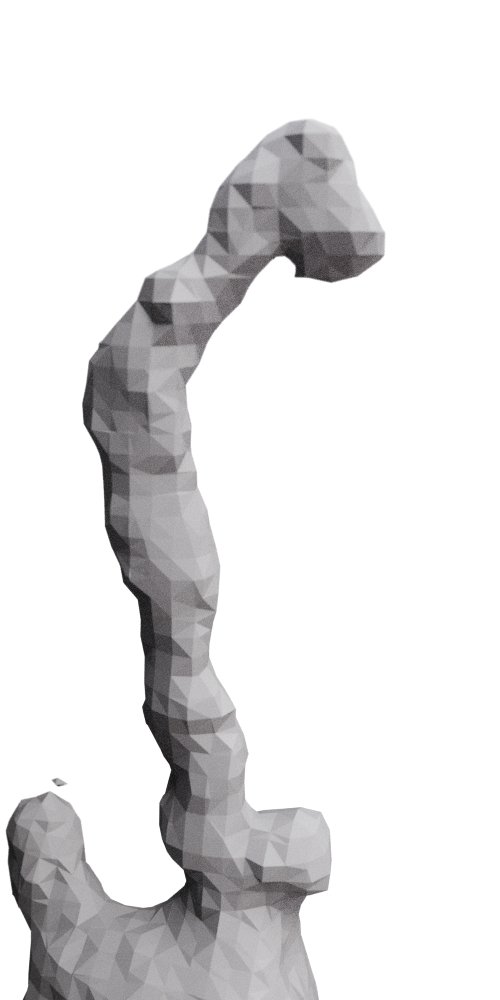}}};
    \end{tikzpicture} &\\[2ex]
    \end{tabular}
    
    \caption{Surface reconstruction results from a noisy point cloud with 39,981 points. We also provide a close-up view for the red rectangular region. Row 1: Original ScanNet v2~\cite{Dai2017ScanNet} for comparison; Row 2: ScanNet v2 with Gaussian noise (standard deviation 0.004); Row 3: ScanNet v2 with Gaussian noise (standard deviation 0.008). }
    \label{fig:comparison_scannetv2_mesh}
\end{figure*}

\begin{figure*}[!htbp]
    \centering
    \begin{tabular}{cccccc}
        & \textbf{Hoppe} & \textbf{Konig} & \textbf{NGL} & \textbf{SNO} & \textbf{DACPO (Ours)} \\[2ex]
        \raisebox{.3in}{\rotatebox{90}{\textbf{Original}}} & \includegraphics[width=0.15\textwidth]{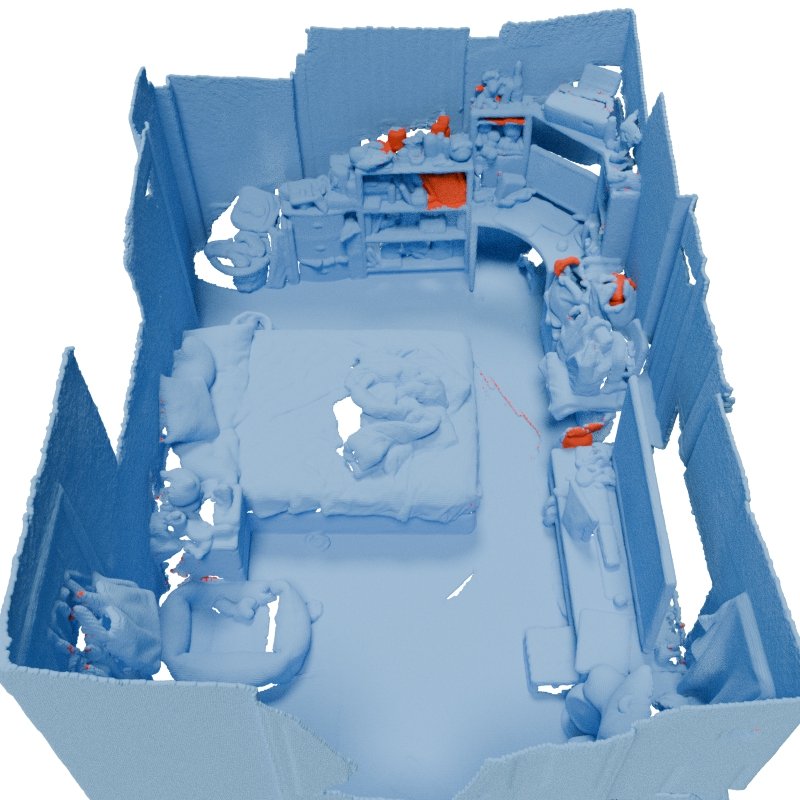} & \includegraphics[width=0.15\textwidth]{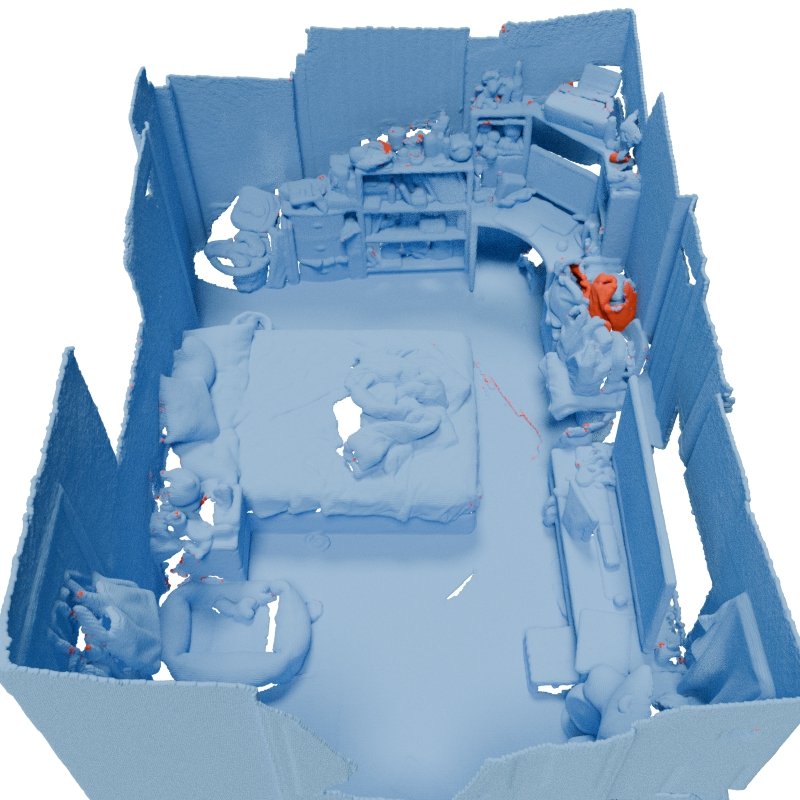} & \includegraphics[width=0.15\textwidth]{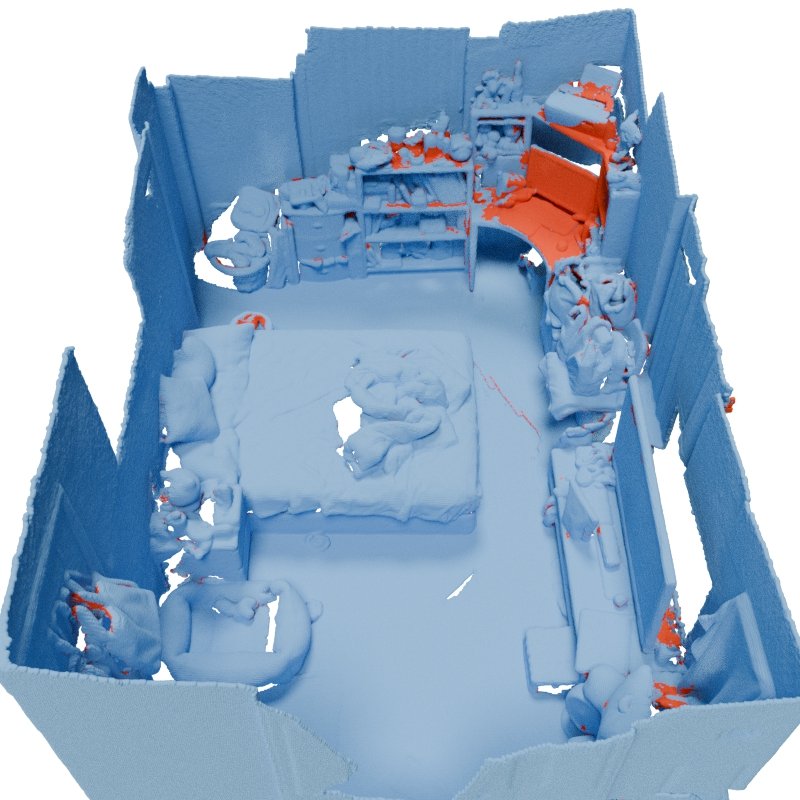} & \includegraphics[width=0.15\textwidth]{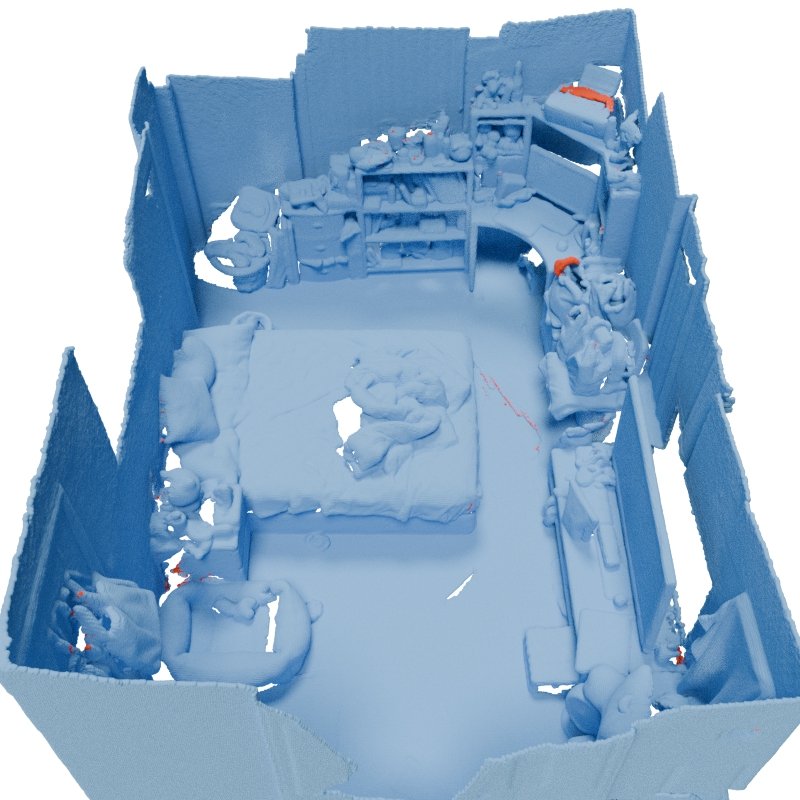} & \includegraphics[width=0.15\textwidth]{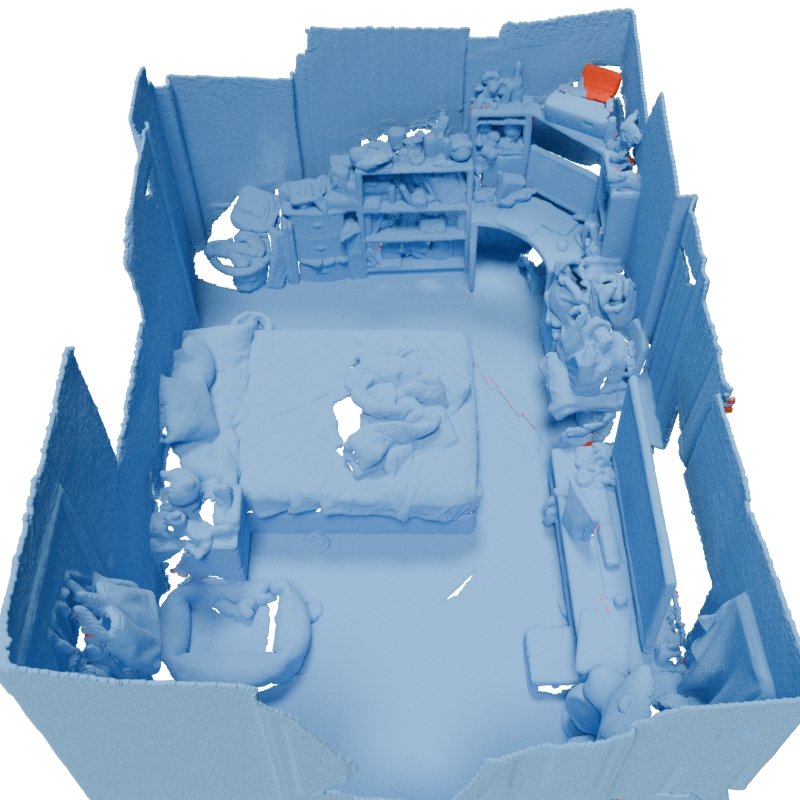} \\[2ex]
        \raisebox{.4in}{\rotatebox{90}{\textbf{100K}}} & \includegraphics[width=0.15\textwidth]{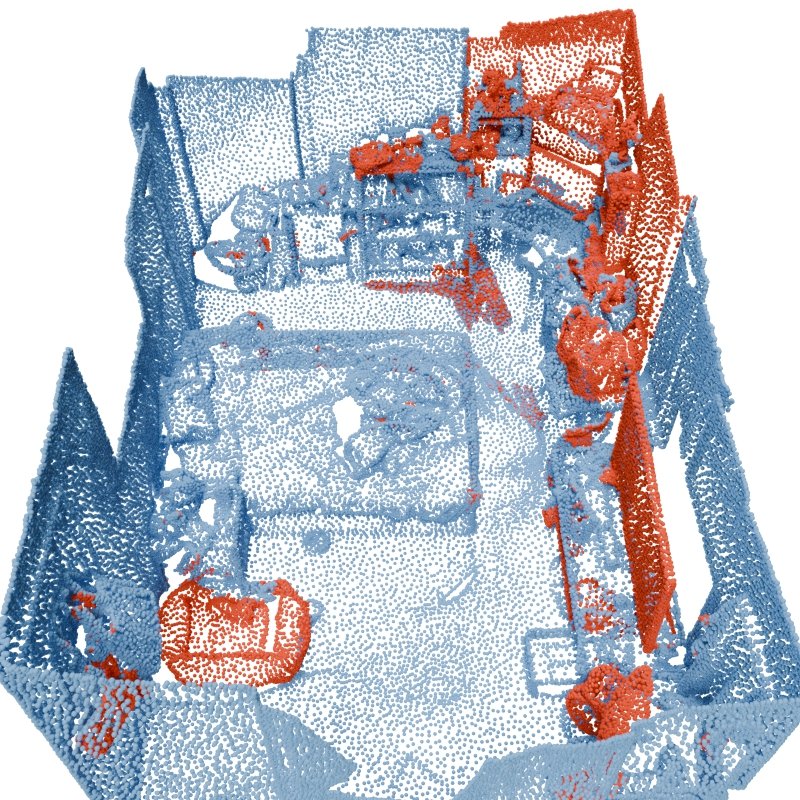} & \includegraphics[width=0.15\textwidth]{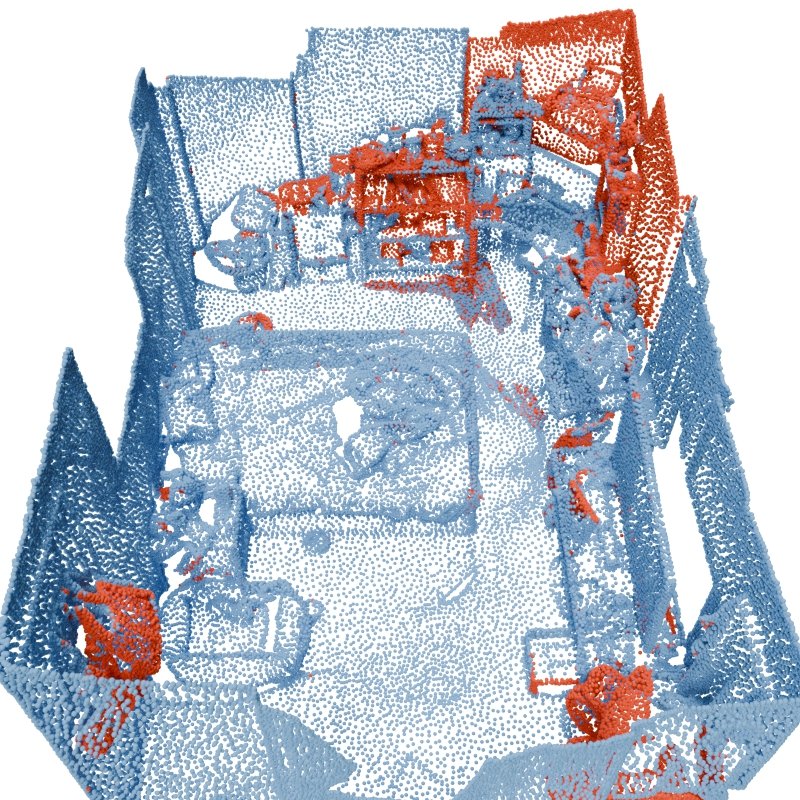} & \includegraphics[width=0.15\textwidth]{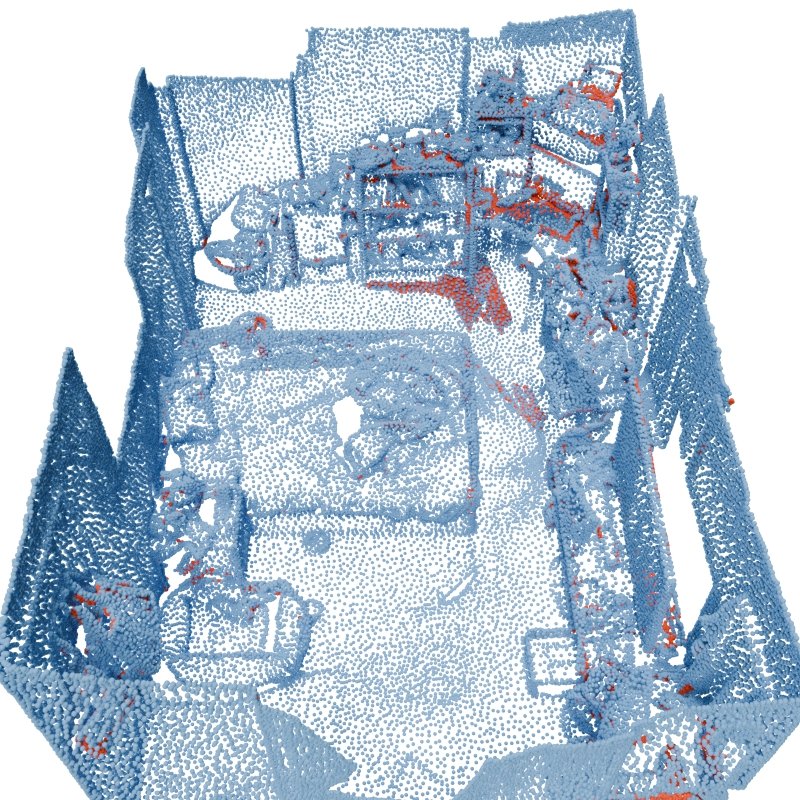} & \includegraphics[width=0.15\textwidth]{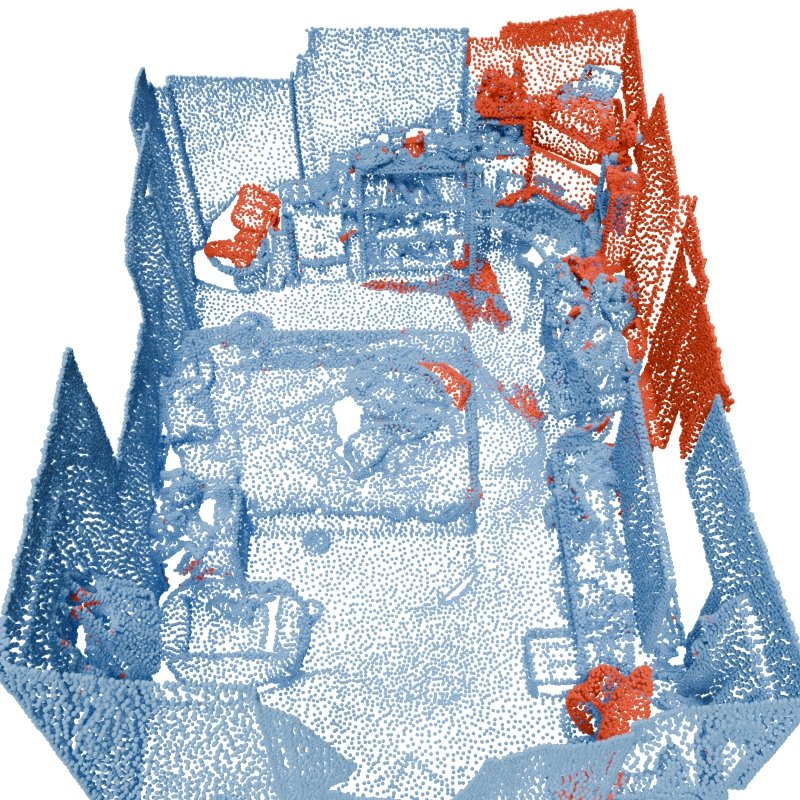} & \includegraphics[width=0.15\textwidth]{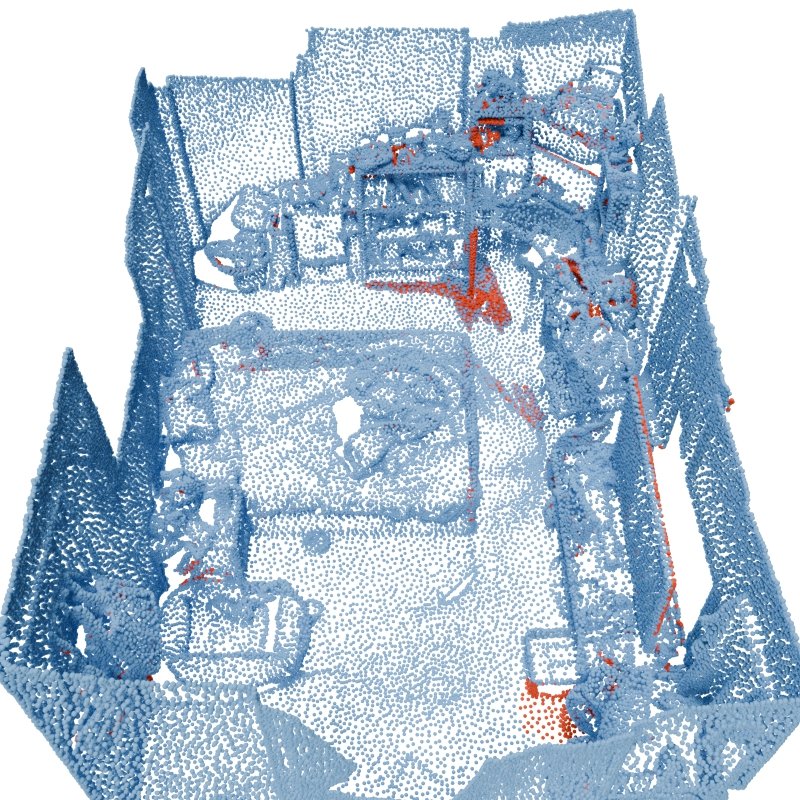} \\[2ex]
        \raisebox{.44in}{\rotatebox{90}{\textbf{10K}}} & \includegraphics[width=0.15\textwidth]{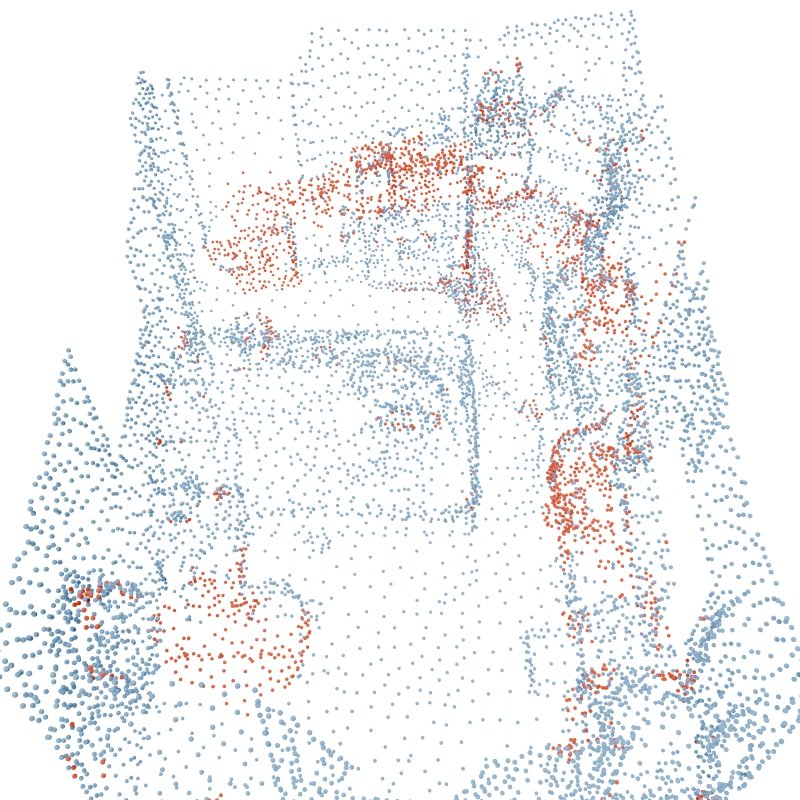} & \includegraphics[width=0.15\textwidth]{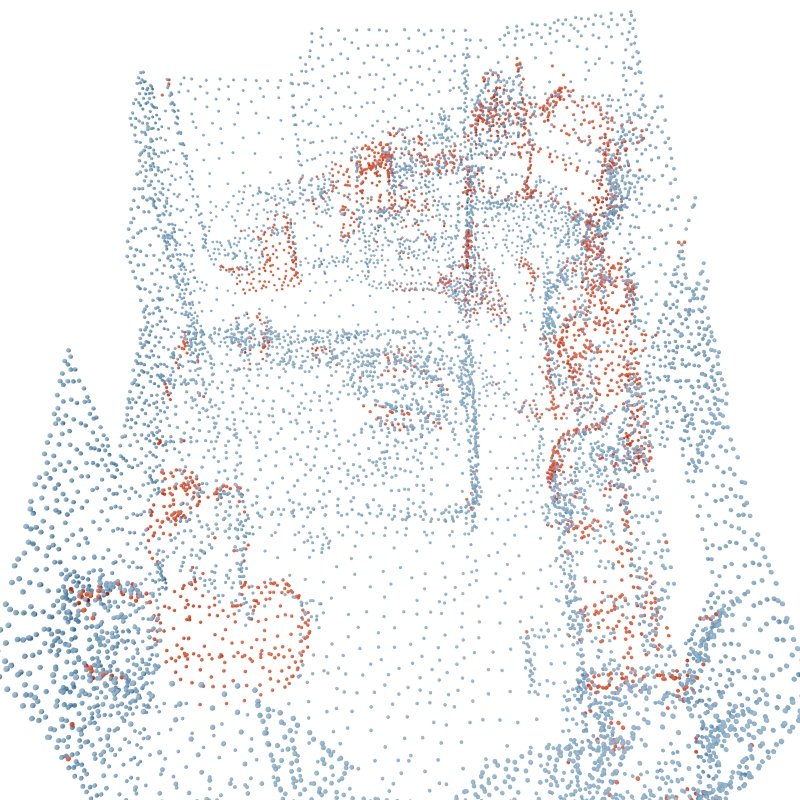} & \includegraphics[width=0.15\textwidth]{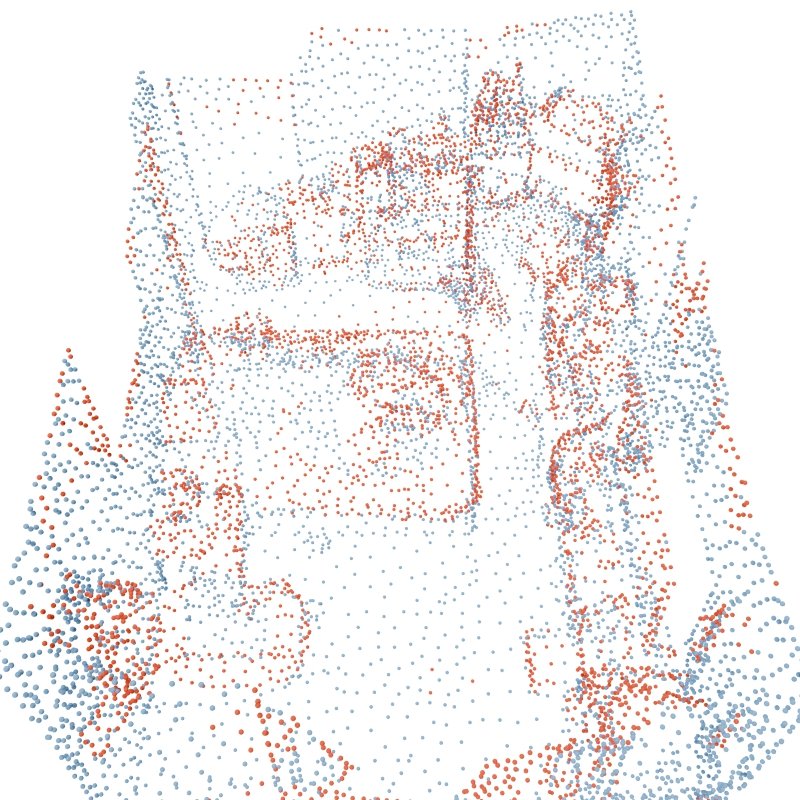} & \includegraphics[width=0.15\textwidth]{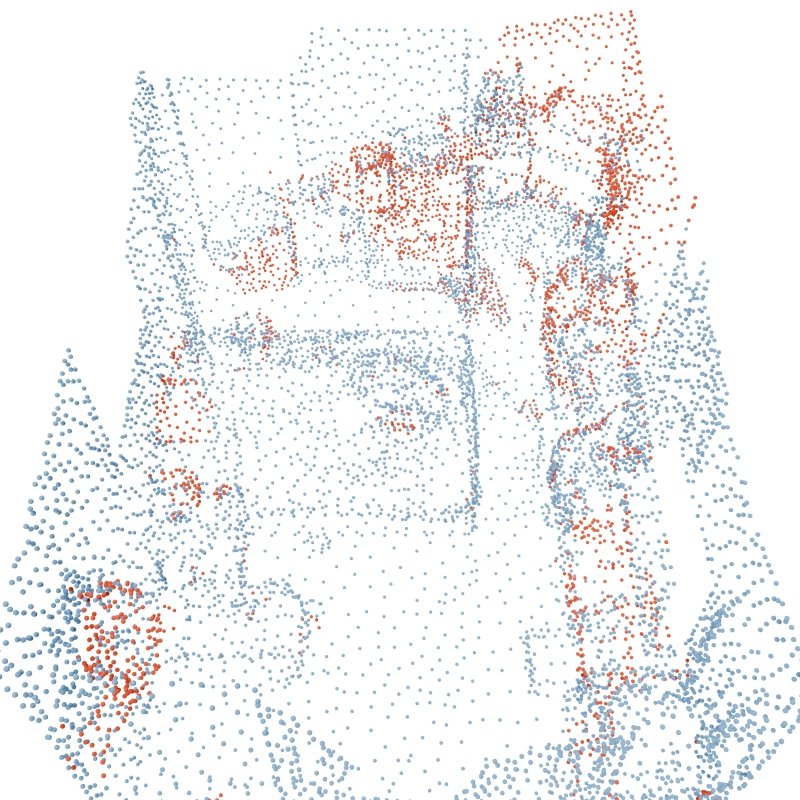} & \includegraphics[width=0.15\textwidth]{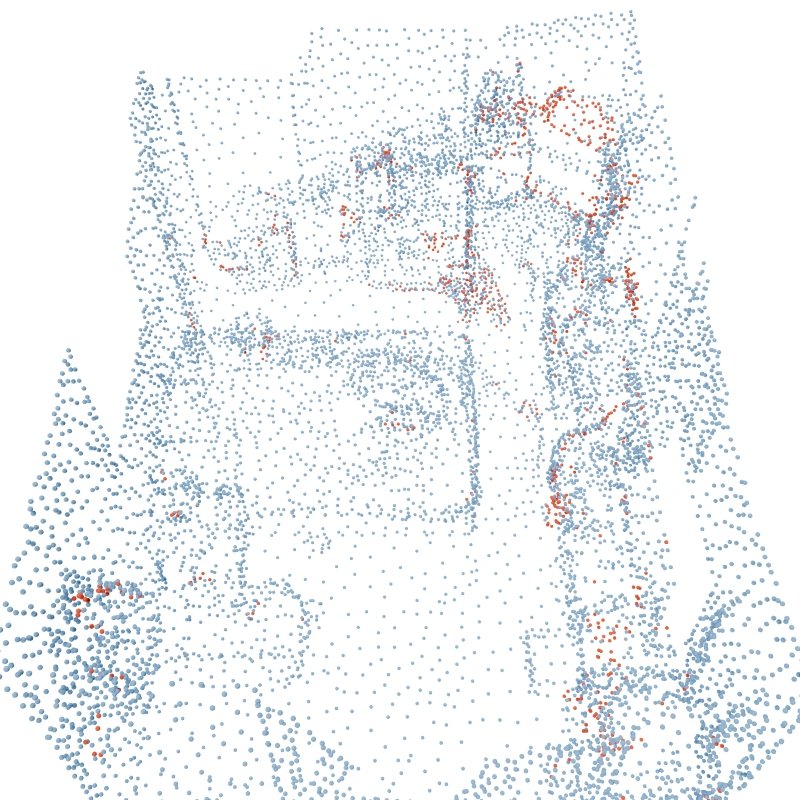}
    \end{tabular}
    \caption{Results on sparse data. Blue points represent correctly oriented normals, while red points denote incorrectly oriented normals. Row 1: Original SceneNN~\cite{Hua2018} (1,017,269 points) for comparison; Row 2: Downsampled SceneNN (100K points); Row 3: Downsampled SceneNN (10K points).}
    \label{fig:comparison_scenenn}
\end{figure*}

\begin{figure}[!htbp]
    \centering
    \begin{tabular}{ccc}
\textbf{Ground Truth} & \textbf{WNNC} & \textbf{DACPO (Ours)} \\
\multicolumn{3}{c}{~} \\
\includegraphics[width=0.3\linewidth]{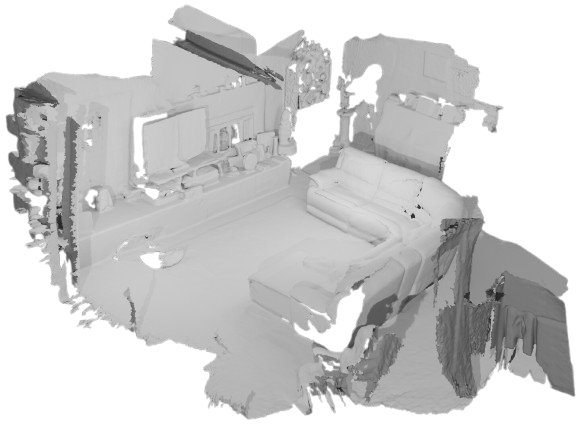} & \includegraphics[width=0.3\linewidth]{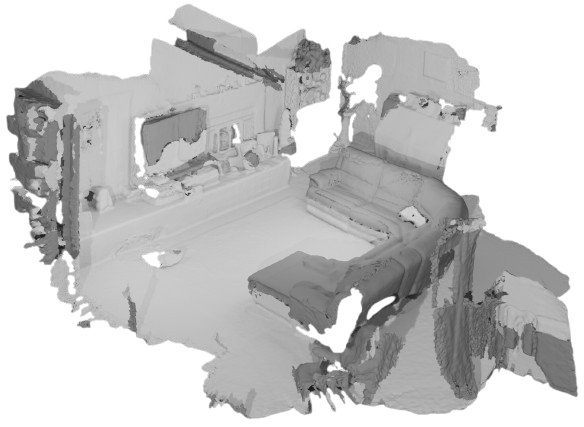} & \includegraphics[width=0.3\linewidth]{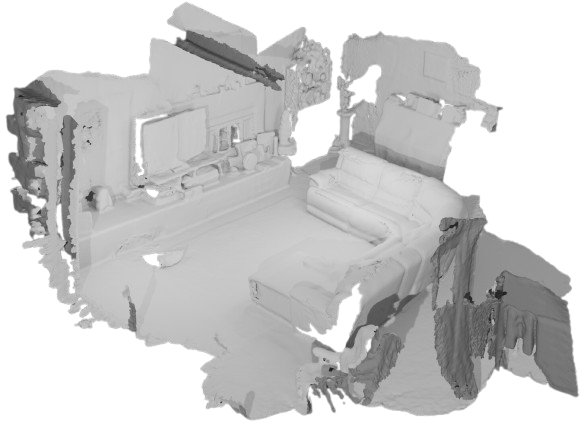} \\ 
\includegraphics[width=0.3\linewidth]{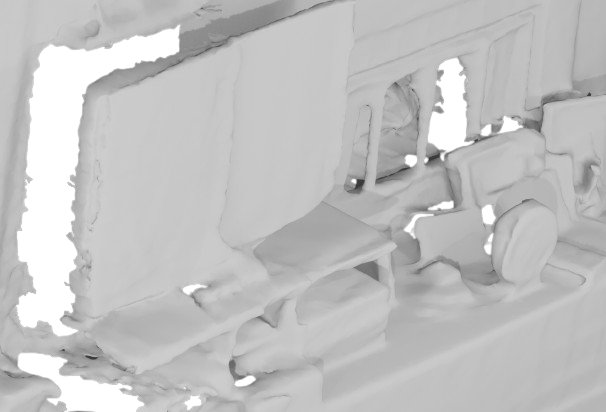} & \includegraphics[width=0.3\linewidth]{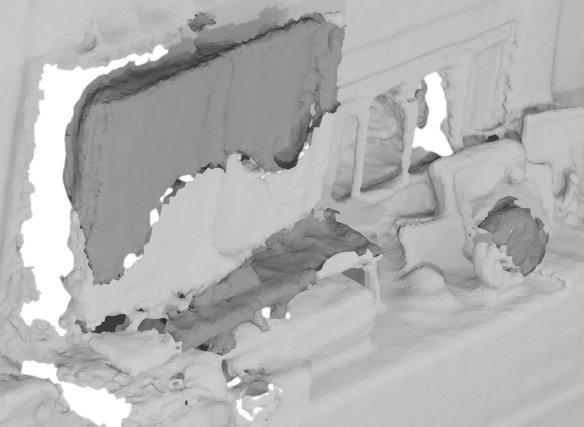} & \includegraphics[width=0.3\linewidth]{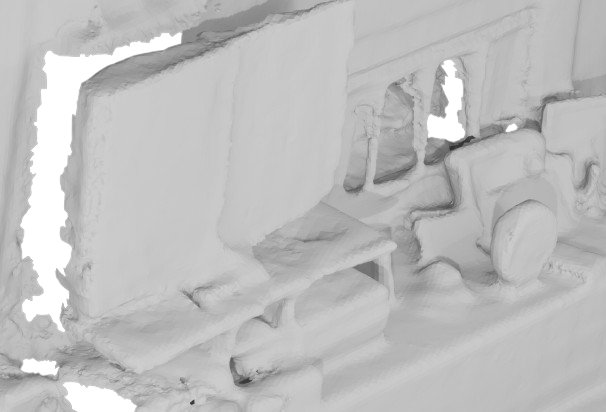} \\ 
\end{tabular}
    \caption{Comparison with WNNC~\cite{Lin2024WNNC}, a GWN-based method. The flipped normals result in gaps in the reconstructed mesh.}
    \label{fig:comparison_WNNC}
\end{figure}

\paragraph{Original Data}
 We report the ratios of incorrectly oriented normals are provided in the ``Original'' columns in Table~\ref{tab:normal_ratio}. The normal orientations are visualized in Figures~\ref{fig:comparison_original}, \ref{fig:comparison_scannetv2} (row 1) and \ref{fig:comparison_scenenn} (row 1). The reconstructed surfaces are shown in Figure~\ref{fig:comparison_scannetv2_mesh} (row 1) and the Chamfer distances are provided in Table~\ref{tab:chamfer}. Incorrectly flipped normals often create gaps in the reconstructed surfaces, as shown in Figure~\ref{fig:comparison_WNNC}. However, these errors do not significantly affect the Chamfer distance, resulting in only minor differences across the methods.

\paragraph{Noisy Data}
To evaluate the noise resilience of our method and the baseline methods, we added Gaussian noise to the ScanNet v2 dataset. While the point count and sizes of ScanNet v2 scenes vary significantly, the average edge length of the triangular faces in the ground truth meshes remains relatively consistent (approximately $0.022$), indicating similar point densities across the dataset. For testing on noisy models, we introduced noise that is proportional to the average edge length rather than the model size. Specifically, we applied Gaussian noise with standard deviations of $0.004$ and $0.008$, corresponding to average displacements of $0.0032$ and $0.0064$, or approximately $14.5\%$ and $29\%$ of the average edge length, respectively. We refer to the two variants as ScanNetV2-Noise1 and ScanNetV2-Noise2. The experimental results are provided in the ``Noise1'' and ``Noise2'' columns of Table~\ref{tab:normal_ratio}. Figures~\ref{fig:comparison_scannetv2} and~\ref{fig:comparison_scannetv2_mesh} provide visualizations of normal orientation and surface reconstruction for two representative models. In each figure, the first row illustrates the results of the original points for comparison, while the second and third rows visualize the results for ScanNetV2-Noise1 and ScanNetV2-Noise2, respectively. The Chamfer distances are reported in Table~\ref{tab:chamfer}.

\setlength{\fboxsep}{0pt}

\begin{figure*}[htb]
    \centering
\begin{tabular}{ccccccc}
\textbf{Dipole} & \textbf{Hoppe} & \textbf{Konig} & \textbf{NGL} & \textbf{SNO} & \textbf{WNNC} & \textbf{DACPO (Ours)} \\
\begin{tikzpicture}[baseline=(current bounding box.center),x=0.11\textwidth,y=0.07333333333333333\textwidth]
    \node[anchor=south west,inner sep=0] {\includegraphics[width=0.11\textwidth,height=0.07333333333333333\textwidth]{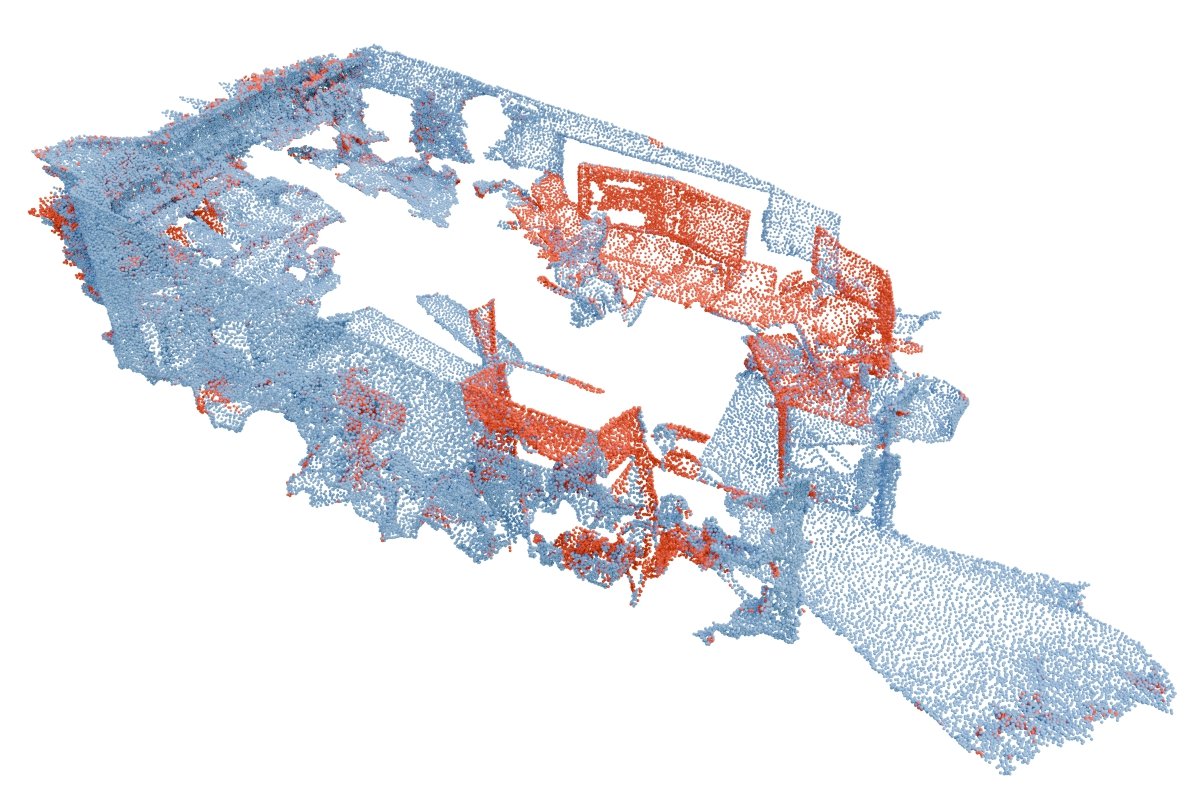}};
    \draw[black] (0.428,0.60) rectangle (0.628,0.80);
\end{tikzpicture} &
\begin{tikzpicture}[baseline=(current bounding box.center),x=0.11\textwidth,y=0.07333333333333333\textwidth]
    \node[anchor=south west,inner sep=0] {\includegraphics[width=0.11\textwidth,height=0.07333333333333333\textwidth]{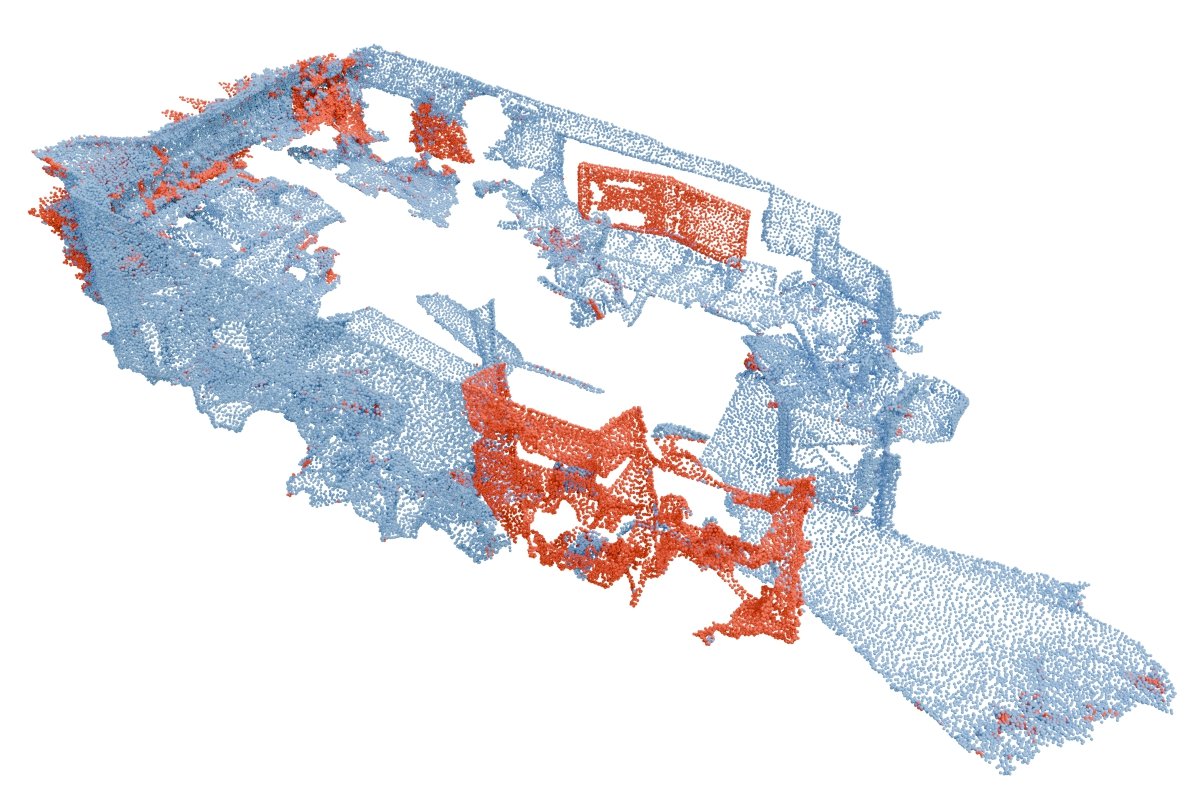}};
    \draw[black] (0.428,0.60) rectangle (0.628,0.80);
\end{tikzpicture} &
\begin{tikzpicture}[baseline=(current bounding box.center),x=0.11\textwidth,y=0.07333333333333333\textwidth]
    \node[anchor=south west,inner sep=0] {\includegraphics[width=0.11\textwidth,height=0.07333333333333333\textwidth]{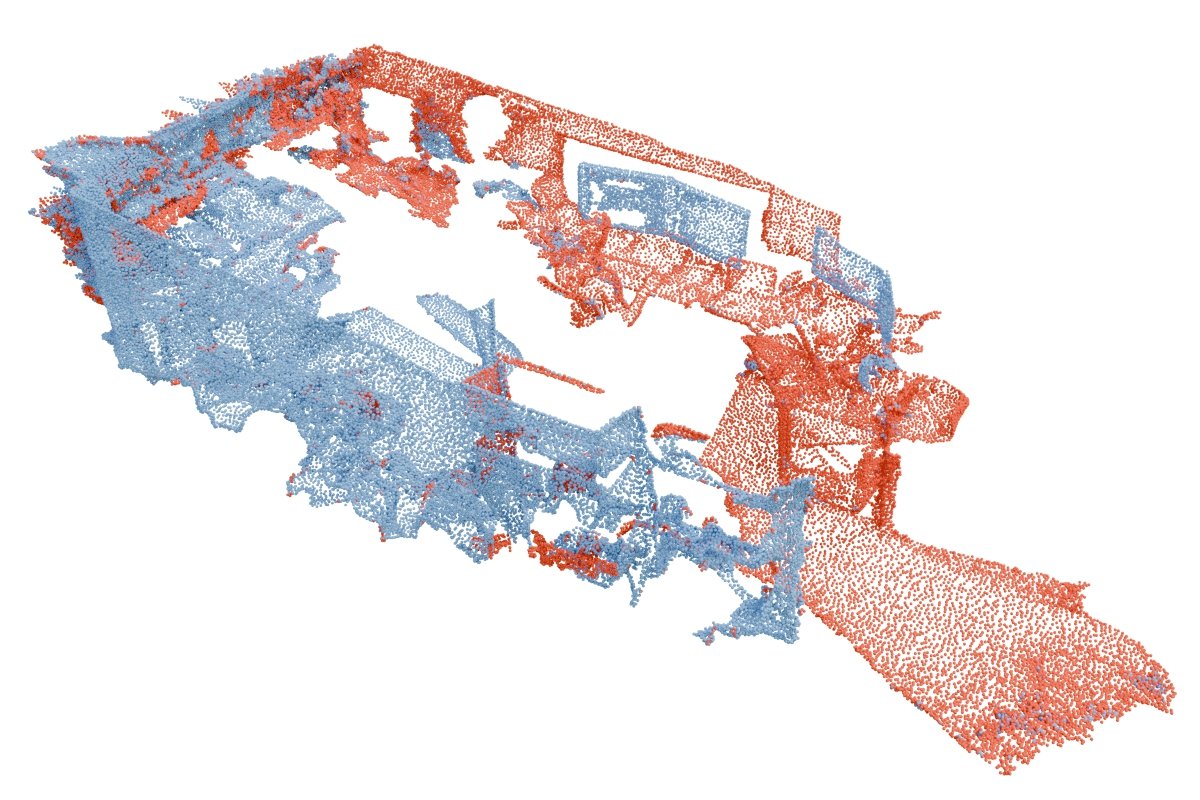}};
    \draw[black] (0.428,0.60) rectangle (0.628,0.80);
\end{tikzpicture} &
\begin{tikzpicture}[baseline=(current bounding box.center),x=0.11\textwidth,y=0.07333333333333333\textwidth]
    \node[anchor=south west,inner sep=0] {\includegraphics[width=0.11\textwidth,height=0.07333333333333333\textwidth]{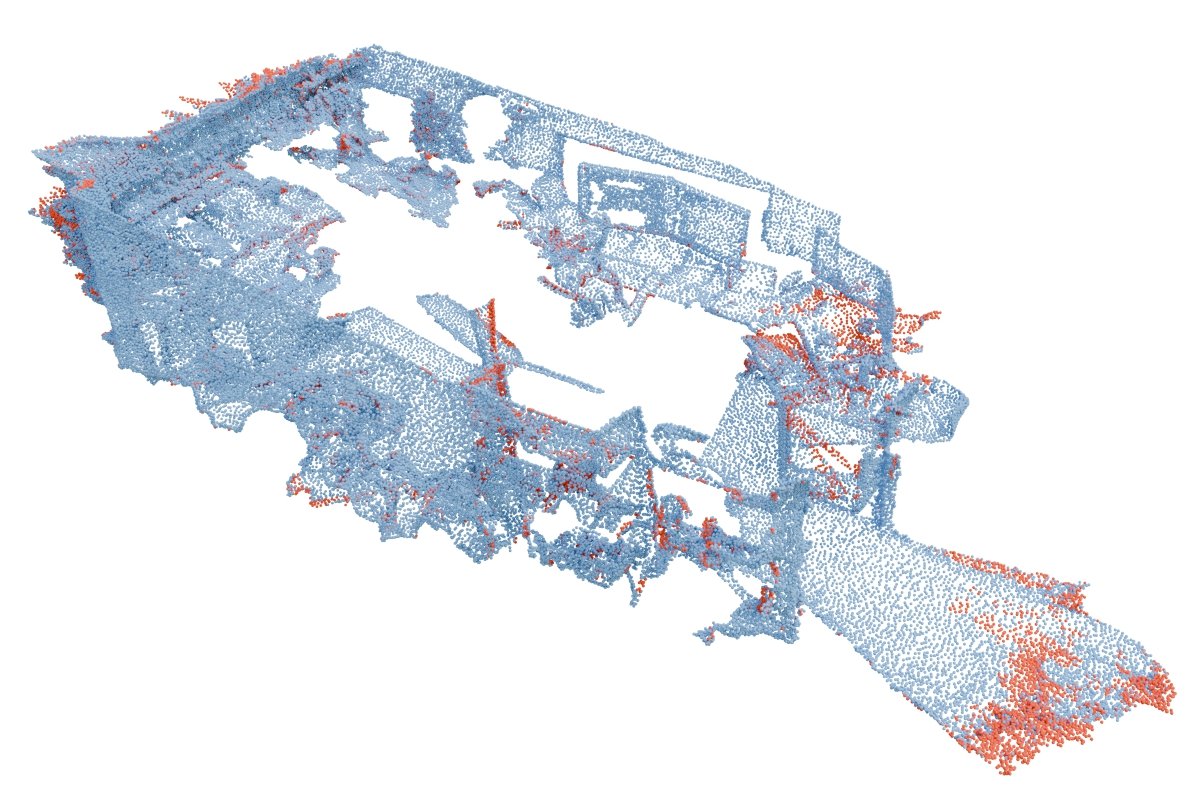}};
    \draw[black] (0.428,0.60) rectangle (0.628,0.80);
\end{tikzpicture} &
\begin{tikzpicture}[baseline=(current bounding box.center),x=0.11\textwidth,y=0.07333333333333333\textwidth]
    \node[anchor=south west,inner sep=0] {\includegraphics[width=0.11\textwidth,height=0.07333333333333333\textwidth]{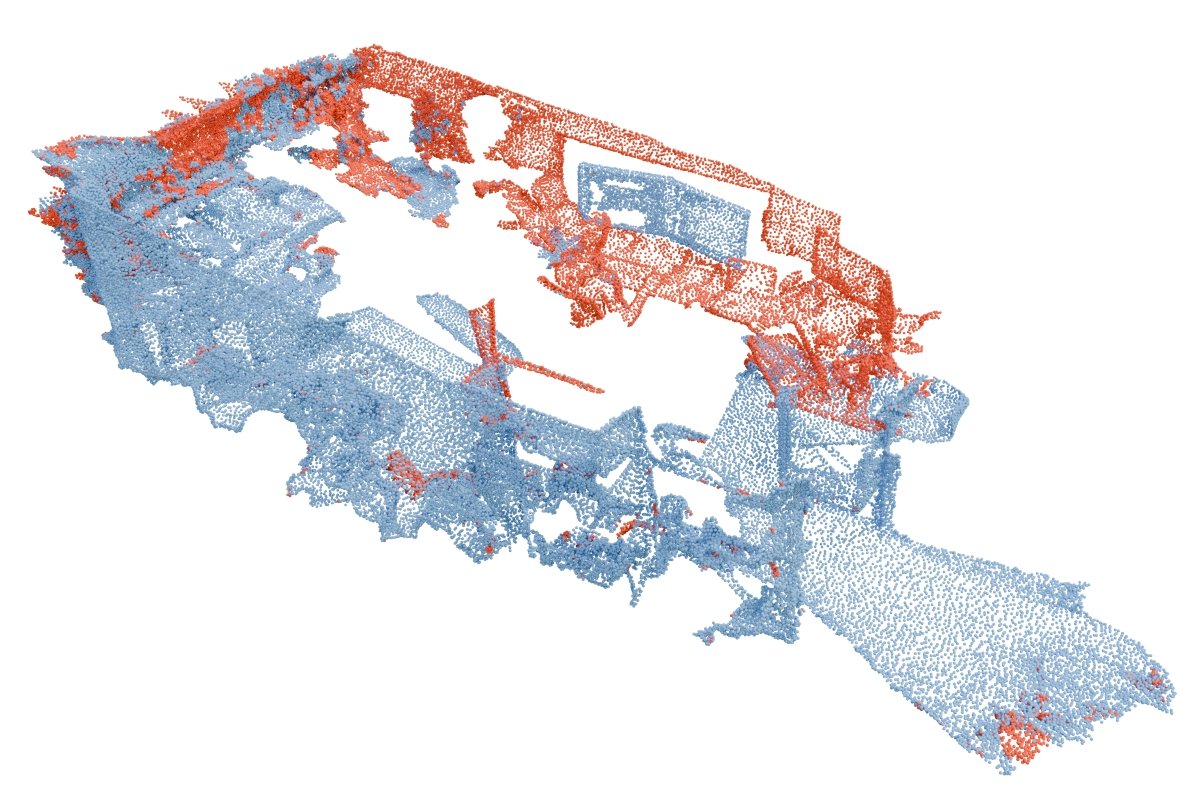}};
    \draw[black] (0.428,0.60) rectangle (0.628,0.80);
\end{tikzpicture} &
\begin{tikzpicture}[baseline=(current bounding box.center),x=0.11\textwidth,y=0.07333333333333333\textwidth]
    \node[anchor=south west,inner sep=0] {\includegraphics[width=0.11\textwidth,height=0.07333333333333333\textwidth]{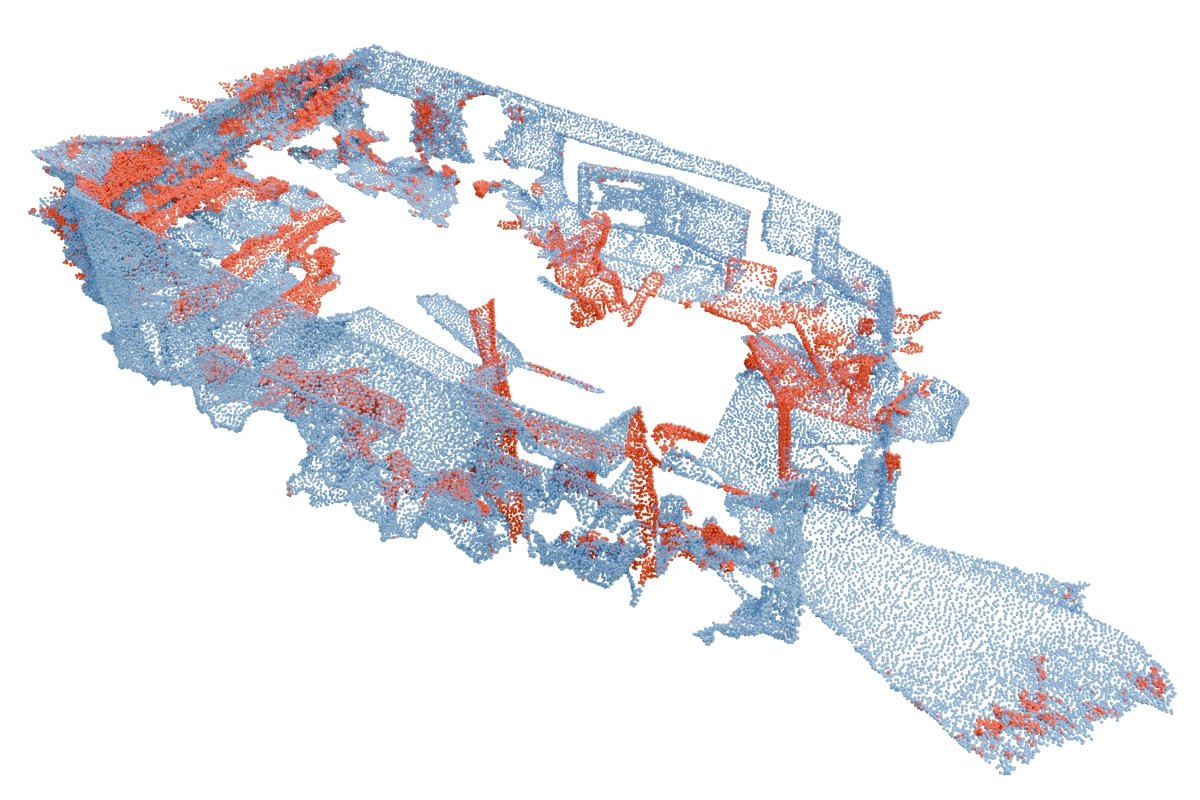}};
    \draw[black] (0.428,0.60) rectangle (0.628,0.80);
\end{tikzpicture} &
\begin{tikzpicture}[baseline=(current bounding box.center),x=0.11\textwidth,y=0.07333333333333333\textwidth]
    \node[anchor=south west,inner sep=0] {\includegraphics[width=0.11\textwidth,height=0.07333333333333333\textwidth]{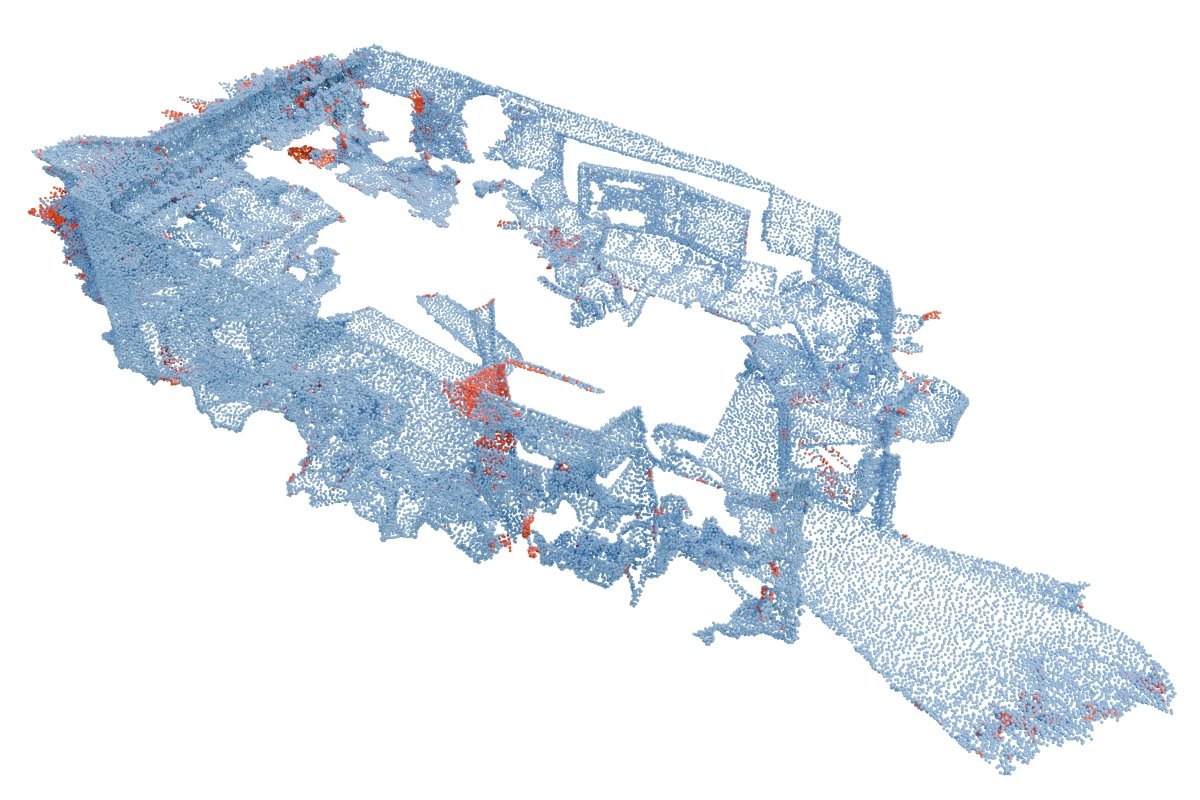}};
    \draw[black] (0.428,0.60) rectangle (0.628,0.80);
\end{tikzpicture} \\ 

\fcolorbox{black}{white}{\includegraphics[width=0.11\textwidth, height=0.07333333333333333\textwidth]{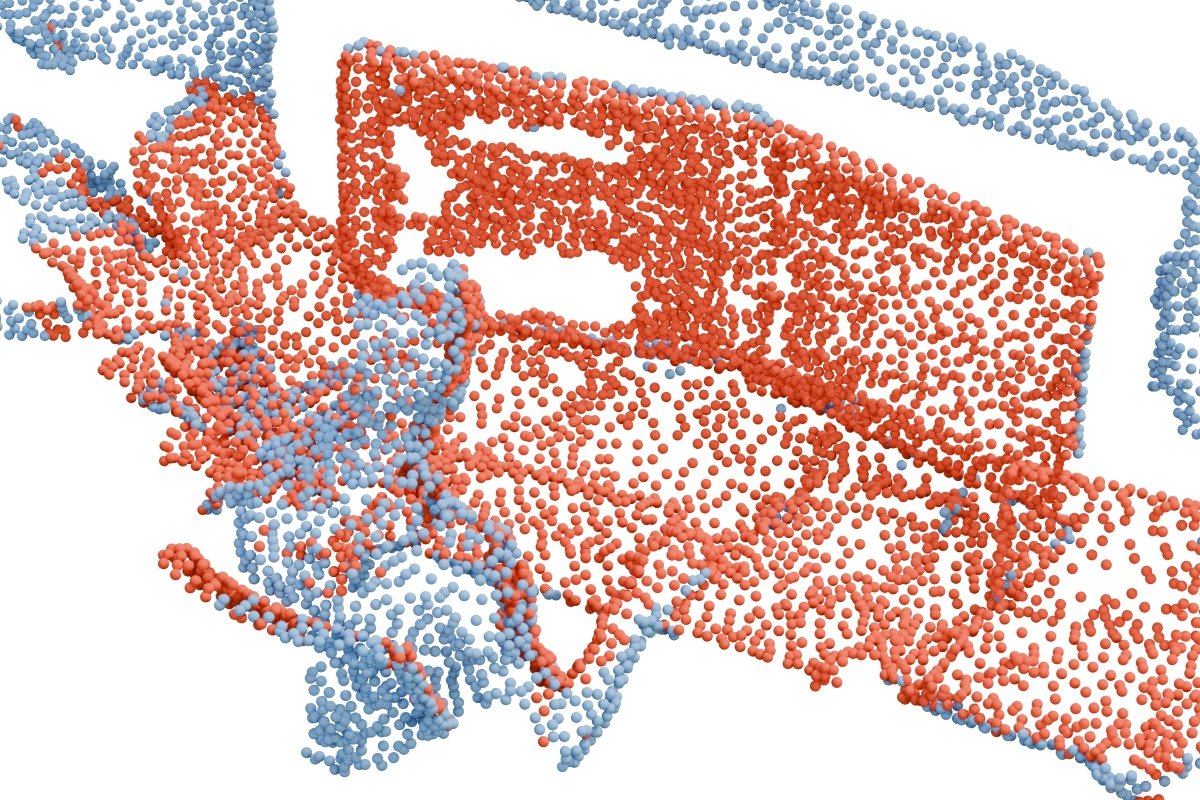}} & \fcolorbox{black}{white}{\includegraphics[width=0.11\textwidth, height=0.07333333333333333\textwidth]{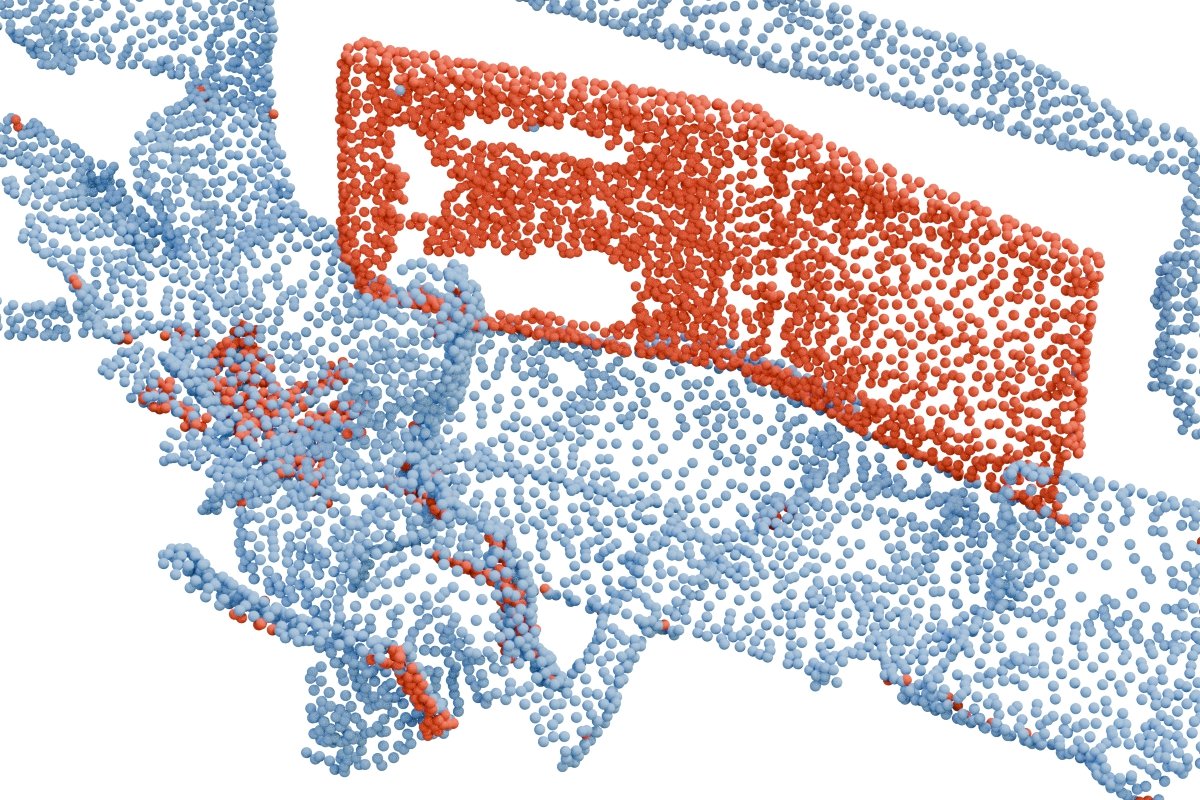}} & \fcolorbox{black}{white}{\includegraphics[width=0.11\textwidth, height=0.07333333333333333\textwidth]{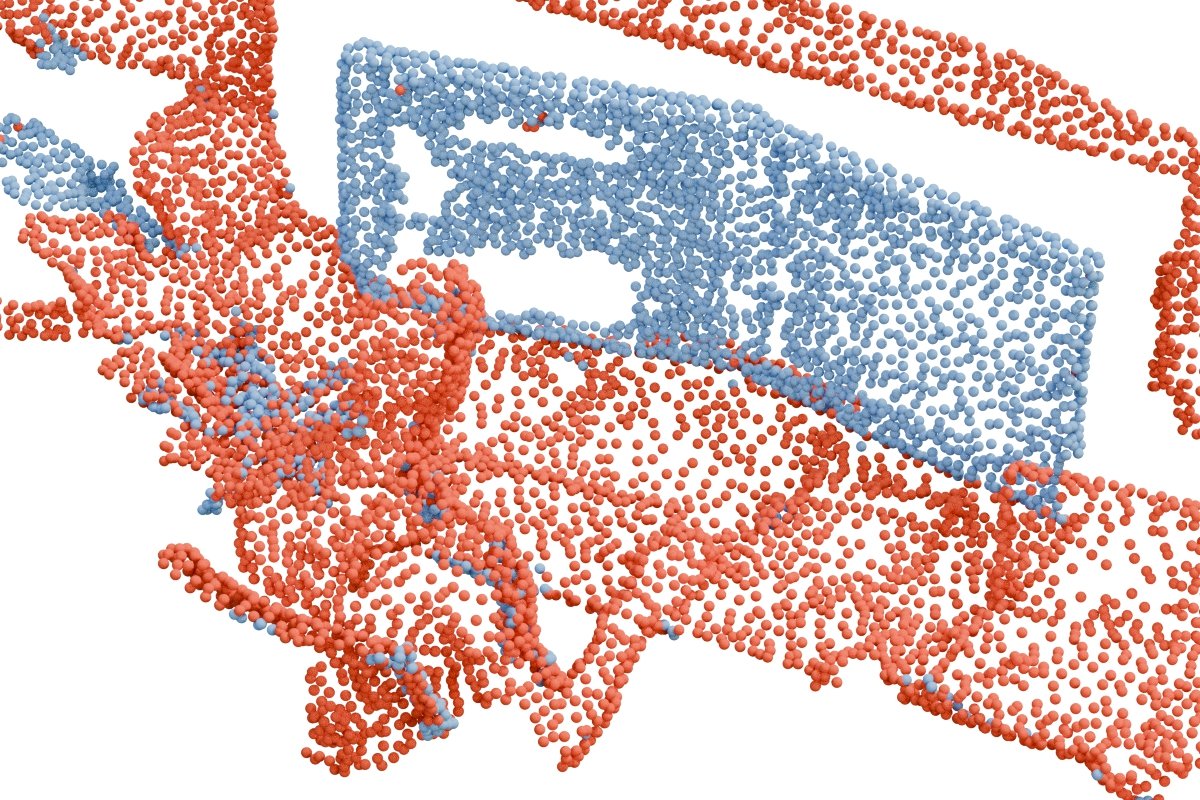}} & \fcolorbox{black}{white}{\includegraphics[width=0.11\textwidth, height=0.07333333333333333\textwidth]{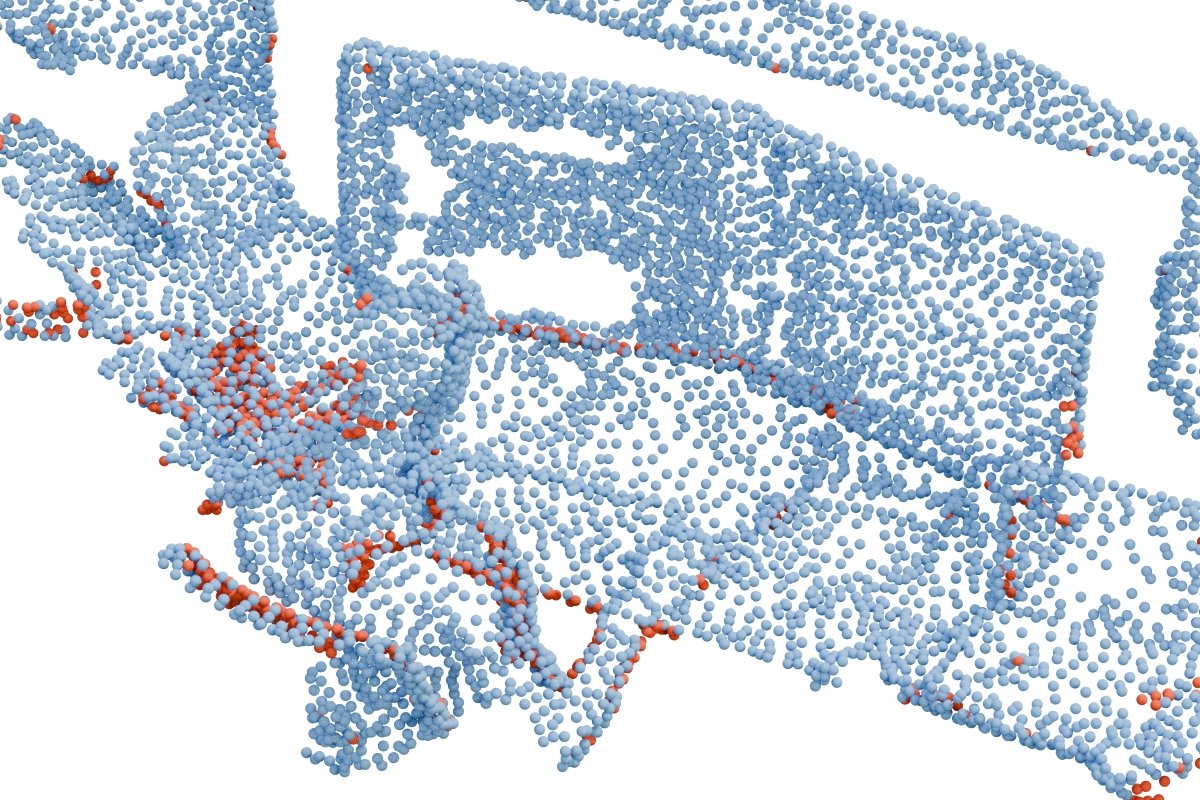}} & \fcolorbox{black}{white}{\includegraphics[width=0.11\textwidth, height=0.07333333333333333\textwidth]{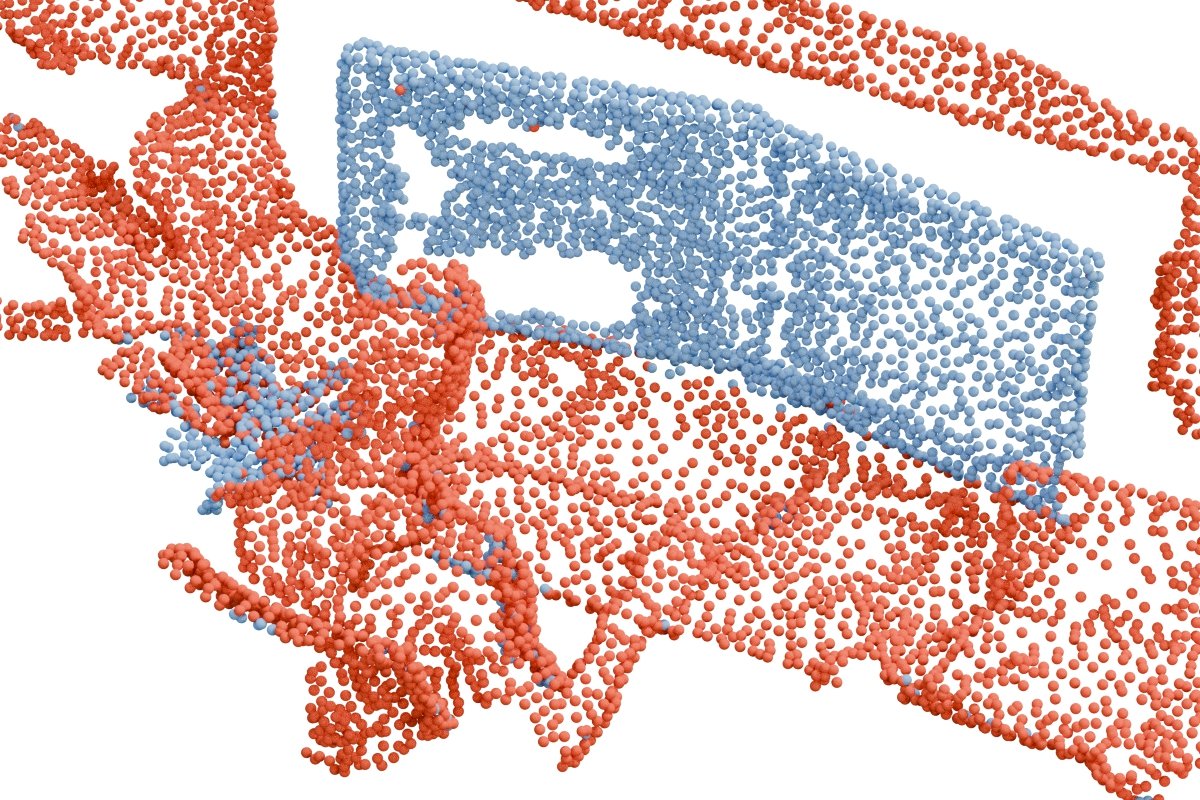}} & \fcolorbox{black}{white}{\includegraphics[width=0.11\textwidth, height=0.07333333333333333\textwidth]{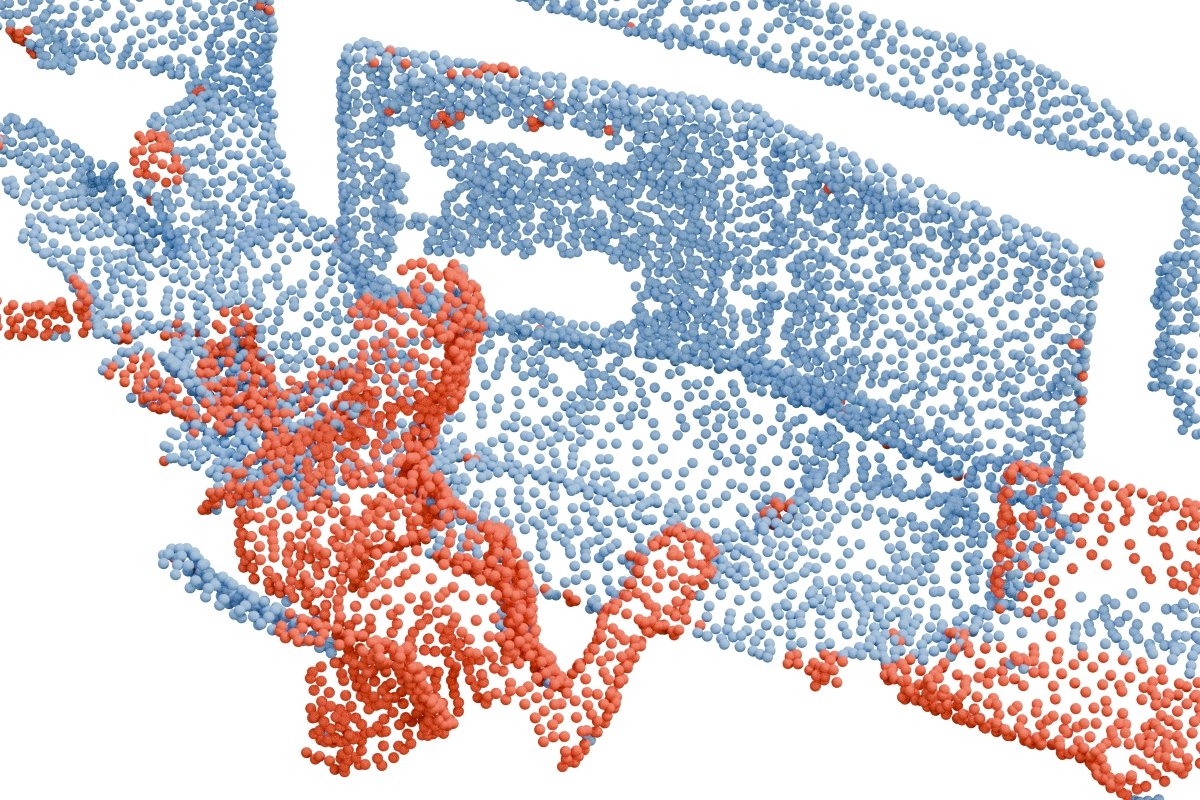}} & \fcolorbox{black}{white}{\includegraphics[width=0.11\textwidth, height=0.07333333333333333\textwidth]{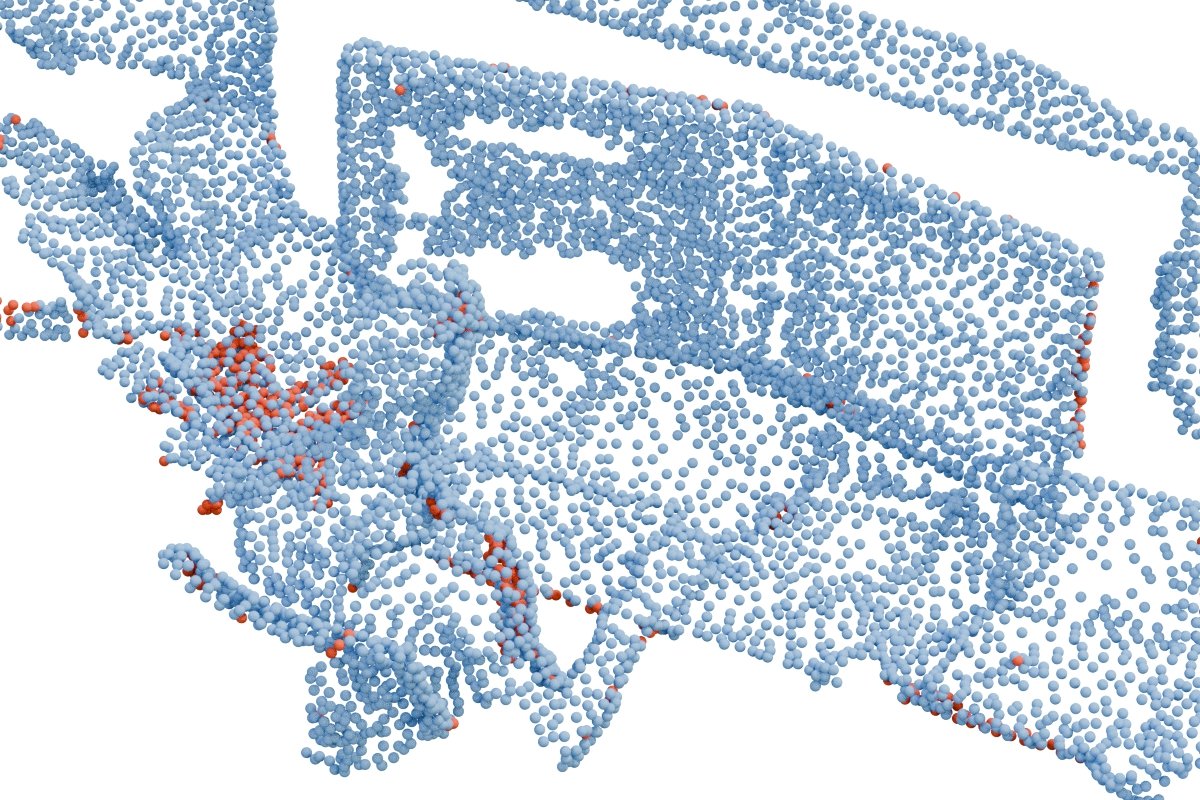}} \\ 

\begin{tikzpicture}[baseline=(current bounding box.center),x=0.11\textwidth,y=0.07333333333333333\textwidth]
    \node[anchor=south west,inner sep=0] {\includegraphics[width=0.11\textwidth,height=0.07333333333333333\textwidth]{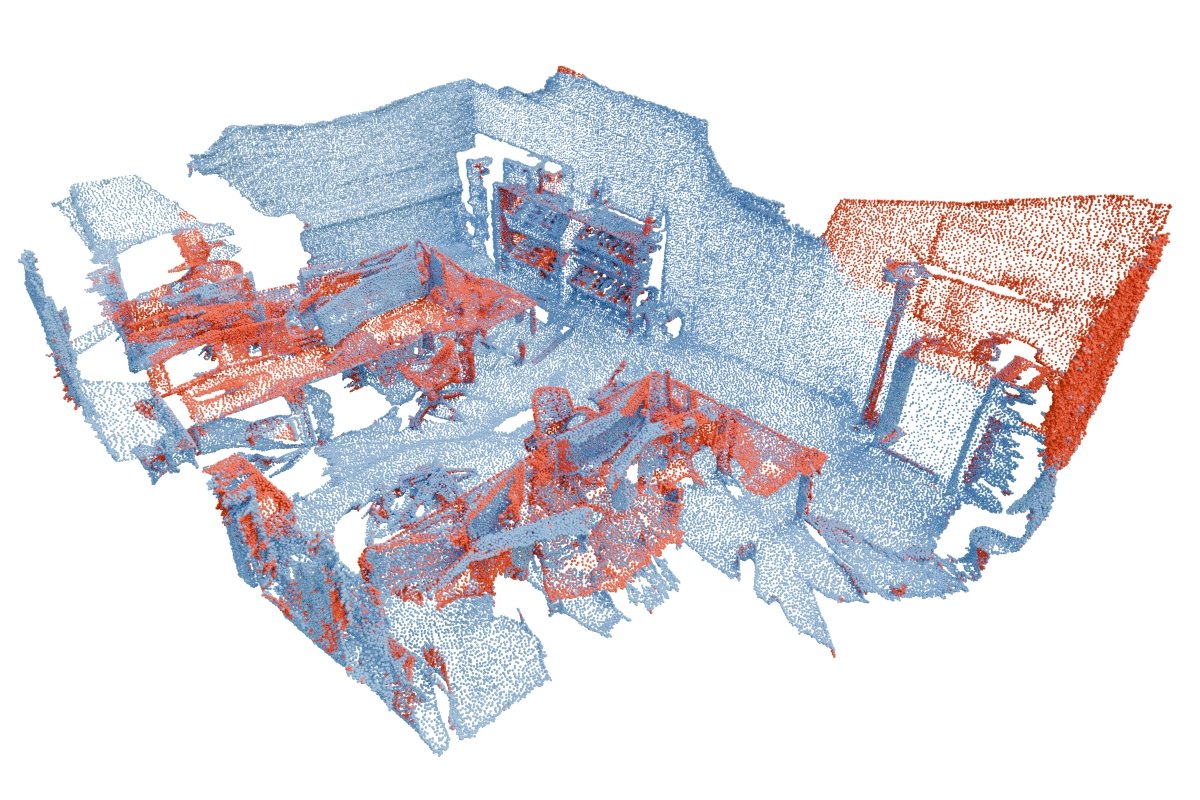}};
    \draw[black] (0.25,0.53) rectangle (0.40,0.68);
\end{tikzpicture} &
\begin{tikzpicture}[baseline=(current bounding box.center),x=0.11\textwidth,y=0.07333333333333333\textwidth]
    \node[anchor=south west,inner sep=0] {\includegraphics[width=0.11\textwidth,height=0.07333333333333333\textwidth]{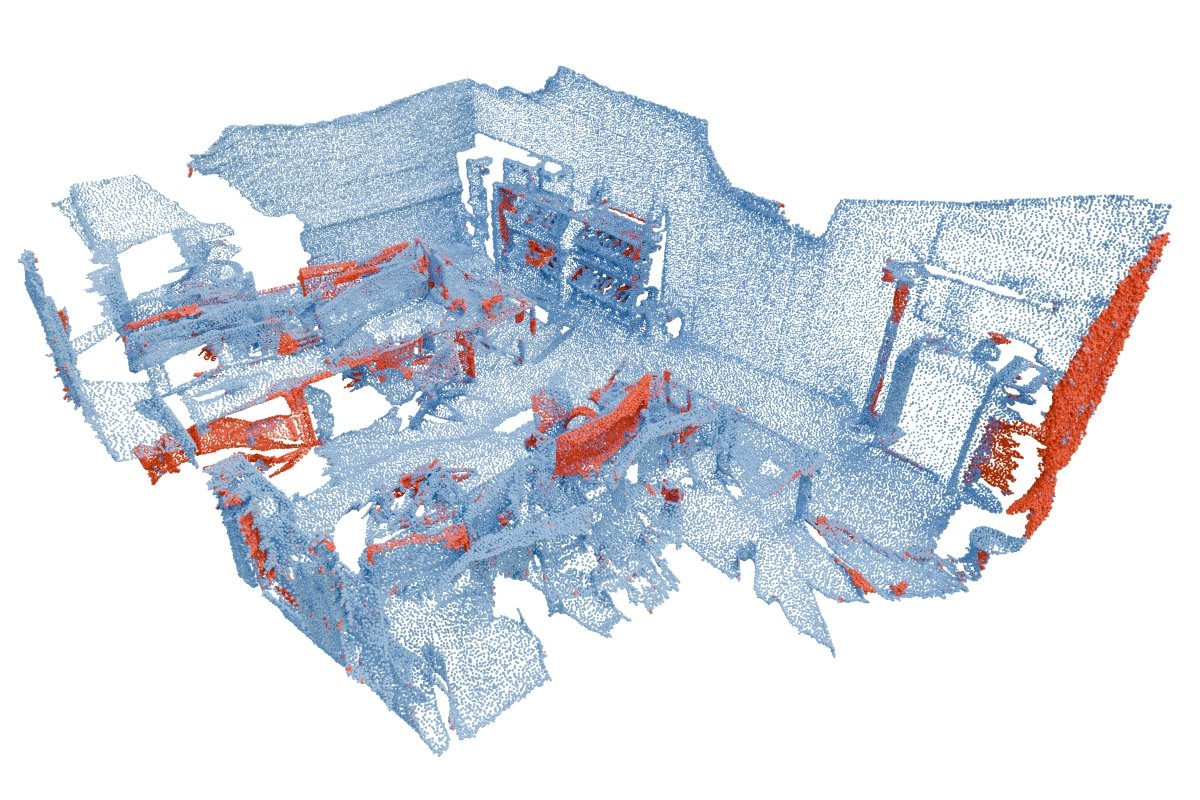}};
    \draw[black] (0.25,0.53) rectangle (0.40,0.68);
\end{tikzpicture} &
\begin{tikzpicture}[baseline=(current bounding box.center),x=0.11\textwidth,y=0.07333333333333333\textwidth]
    \node[anchor=south west,inner sep=0] {\includegraphics[width=0.11\textwidth,height=0.07333333333333333\textwidth]{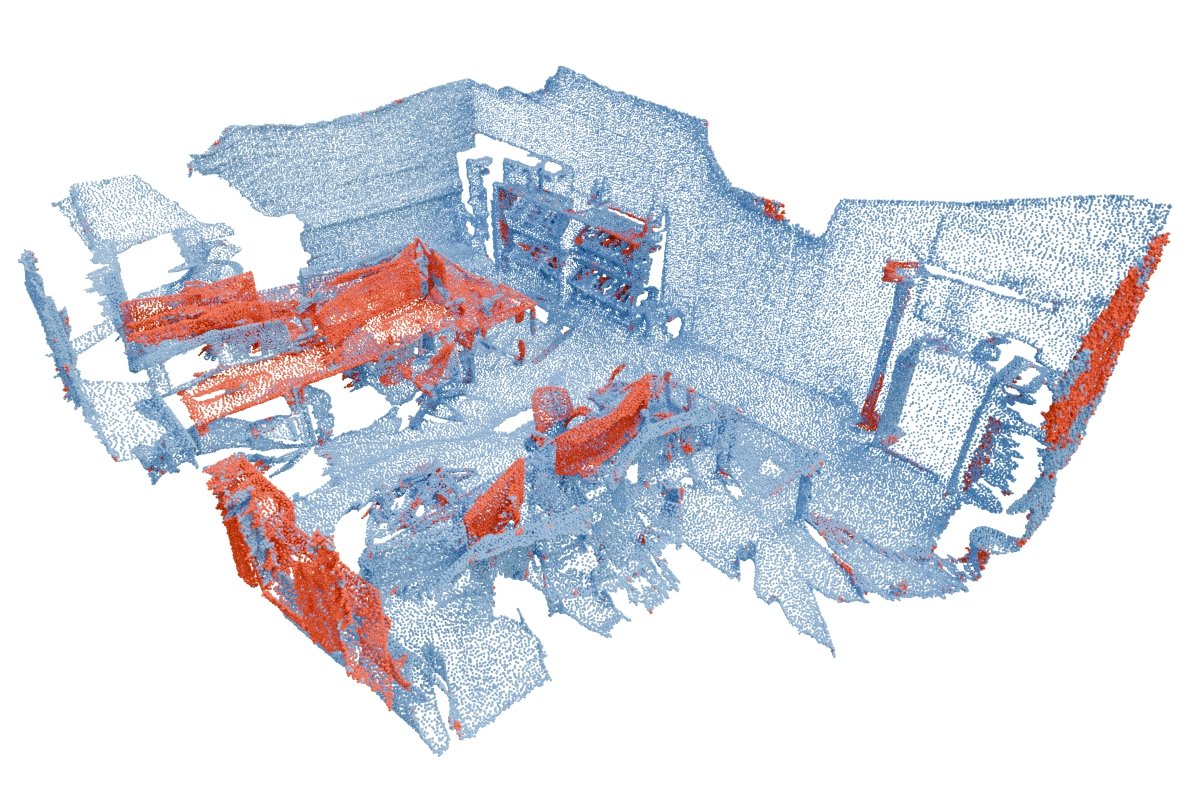}};
    \draw[black] (0.25,0.53) rectangle (0.40,0.68);
\end{tikzpicture} &
\begin{tikzpicture}[baseline=(current bounding box.center),x=0.11\textwidth,y=0.07333333333333333\textwidth]
    \node[anchor=south west,inner sep=0] {\includegraphics[width=0.11\textwidth,height=0.07333333333333333\textwidth]{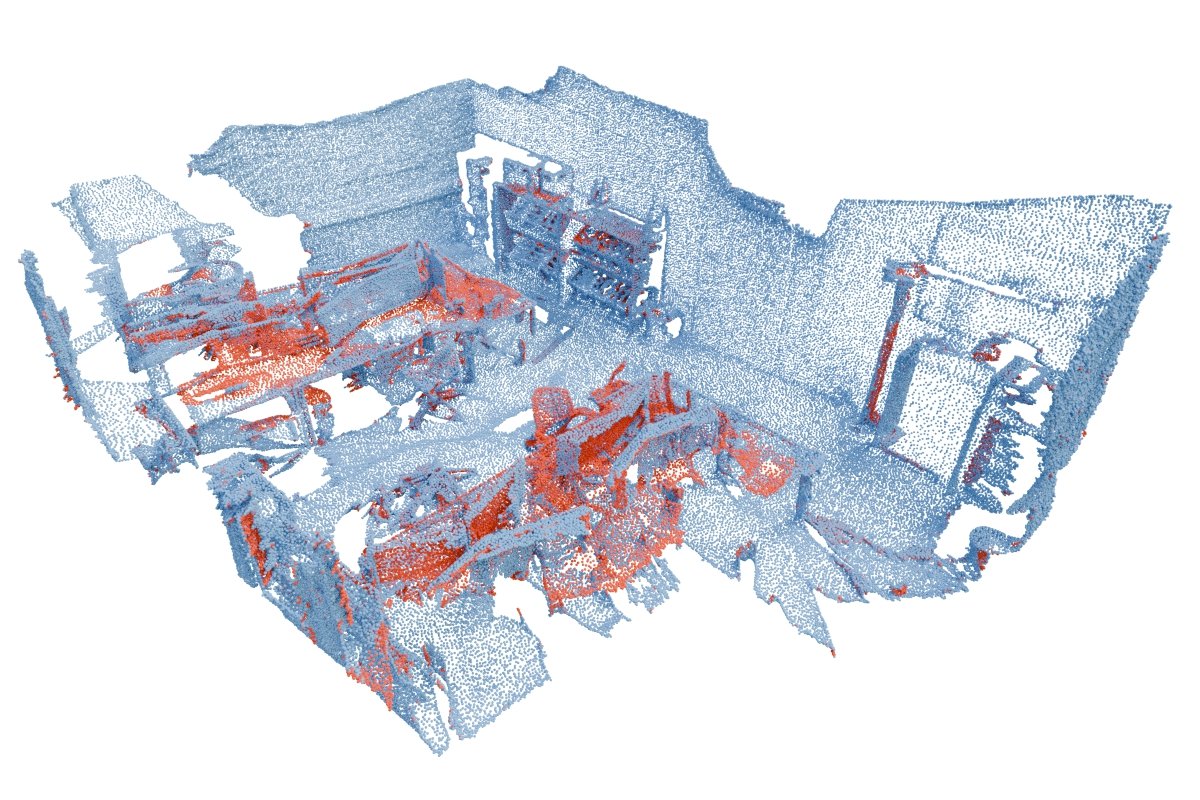}};
    \draw[black] (0.25,0.53) rectangle (0.40,0.68); 
\end{tikzpicture} &
\begin{tikzpicture}[baseline=(current bounding box.center),x=0.11\textwidth,y=0.07333333333333333\textwidth]
    \node[anchor=south west,inner sep=0] {\includegraphics[width=0.11\textwidth,height=0.07333333333333333\textwidth]{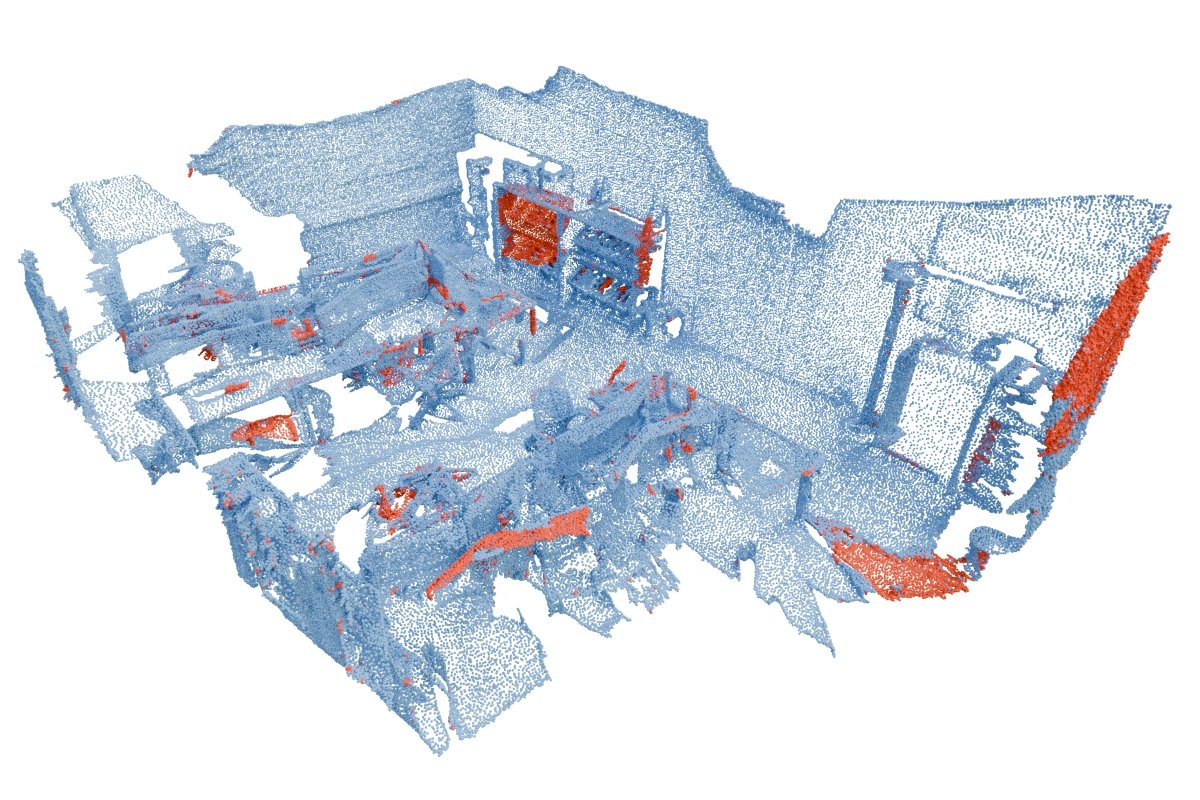}};
    \draw[black] (0.25,0.53) rectangle (0.40,0.68);
\end{tikzpicture} &
\begin{tikzpicture}[baseline=(current bounding box.center),x=0.11\textwidth,y=0.07333333333333333\textwidth]
    \node[anchor=south west,inner sep=0] {\includegraphics[width=0.11\textwidth,height=0.07333333333333333\textwidth]{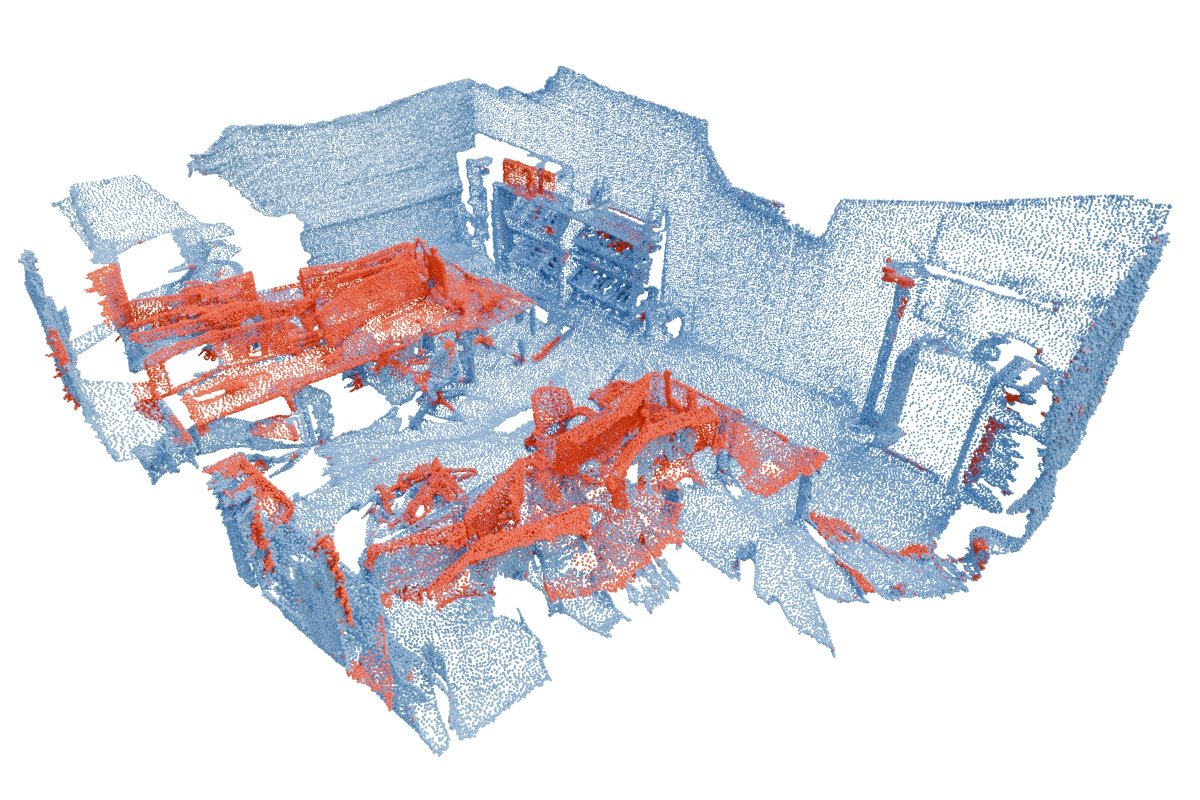}};
    \draw[black] (0.25,0.53) rectangle (0.40,0.68);
\end{tikzpicture} &
\begin{tikzpicture}[baseline=(current bounding box.center),x=0.11\textwidth,y=0.07333333333333333\textwidth]
    \node[anchor=south west,inner sep=0] {\includegraphics[width=0.11\textwidth,height=0.07333333333333333\textwidth]{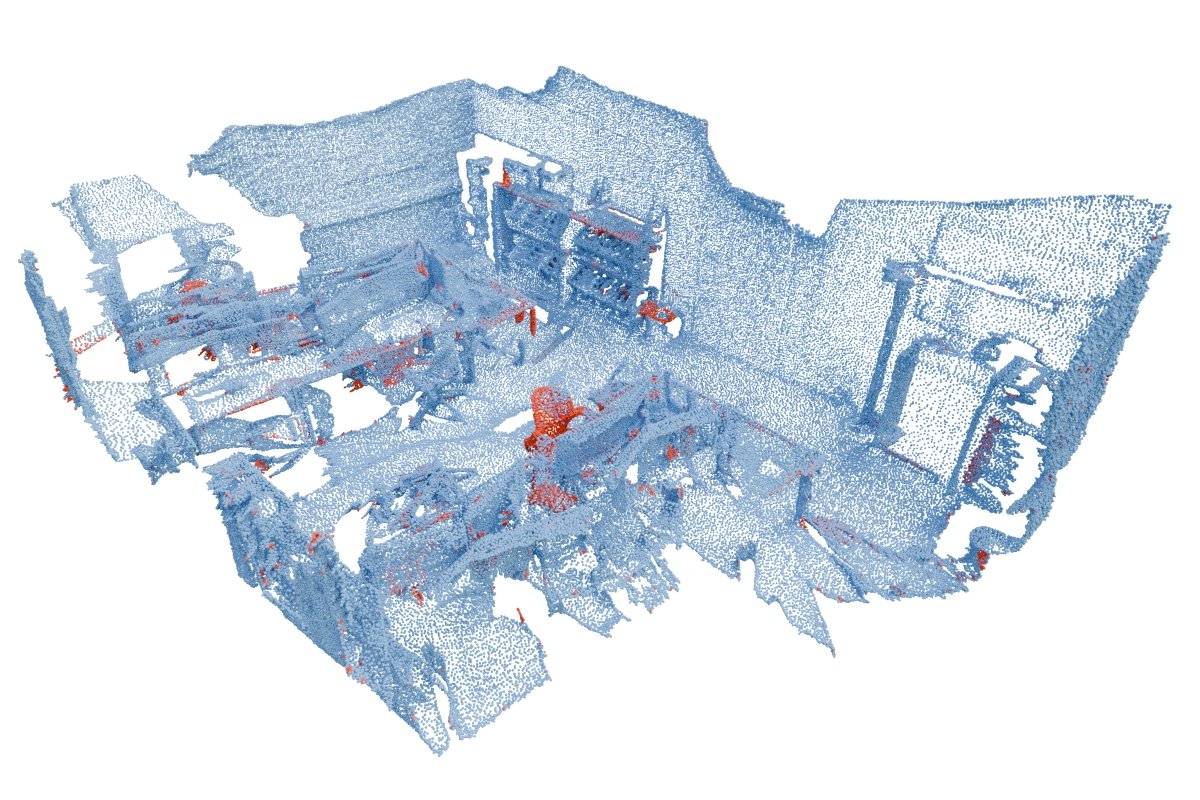}};
    \draw[black] (0.25,0.53) rectangle (0.40,0.68);
\end{tikzpicture} \\ 

\fcolorbox{black}{white}{\includegraphics[width=0.11\textwidth, height=0.07333333333333333\textwidth]{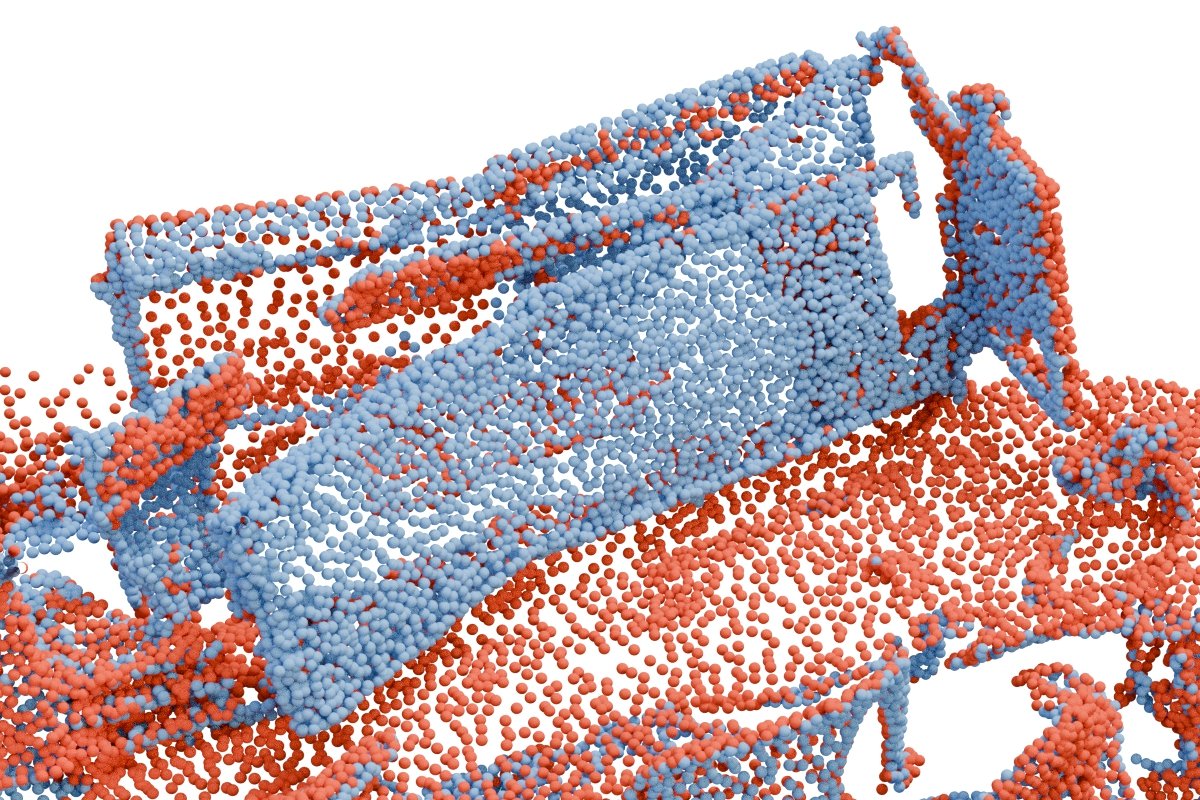}} & \fcolorbox{black}{white}{\includegraphics[width=0.11\textwidth, height=0.07333333333333333\textwidth]{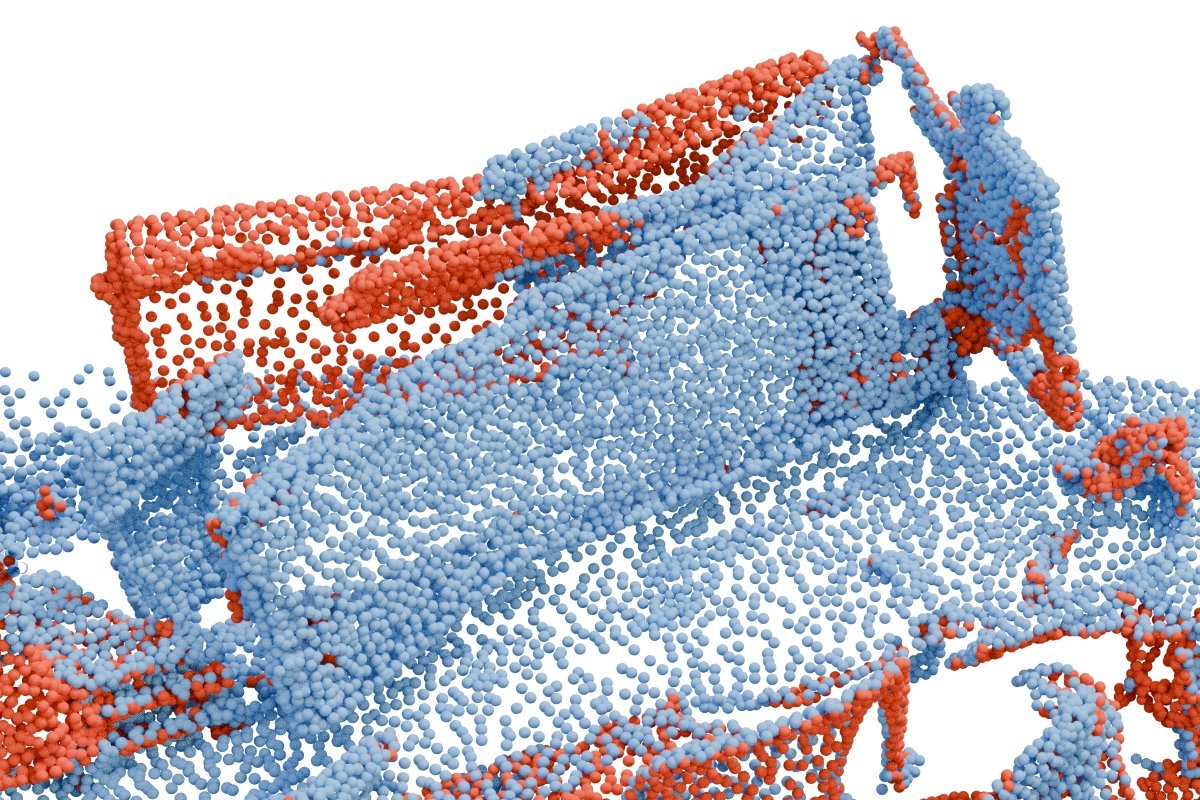}} & \fcolorbox{black}{white}{\includegraphics[width=0.11\textwidth, height=0.07333333333333333\textwidth]{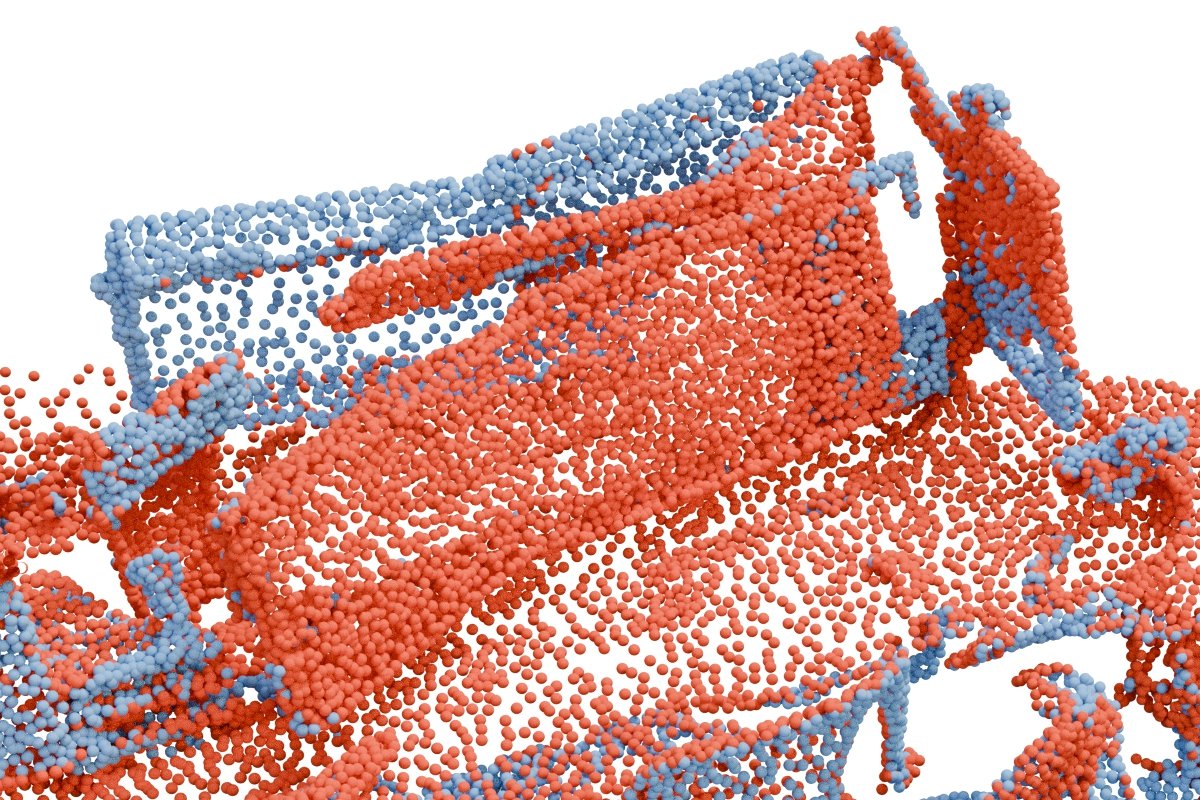}} & \fcolorbox{black}{white}{\includegraphics[width=0.11\textwidth, height=0.07333333333333333\textwidth]{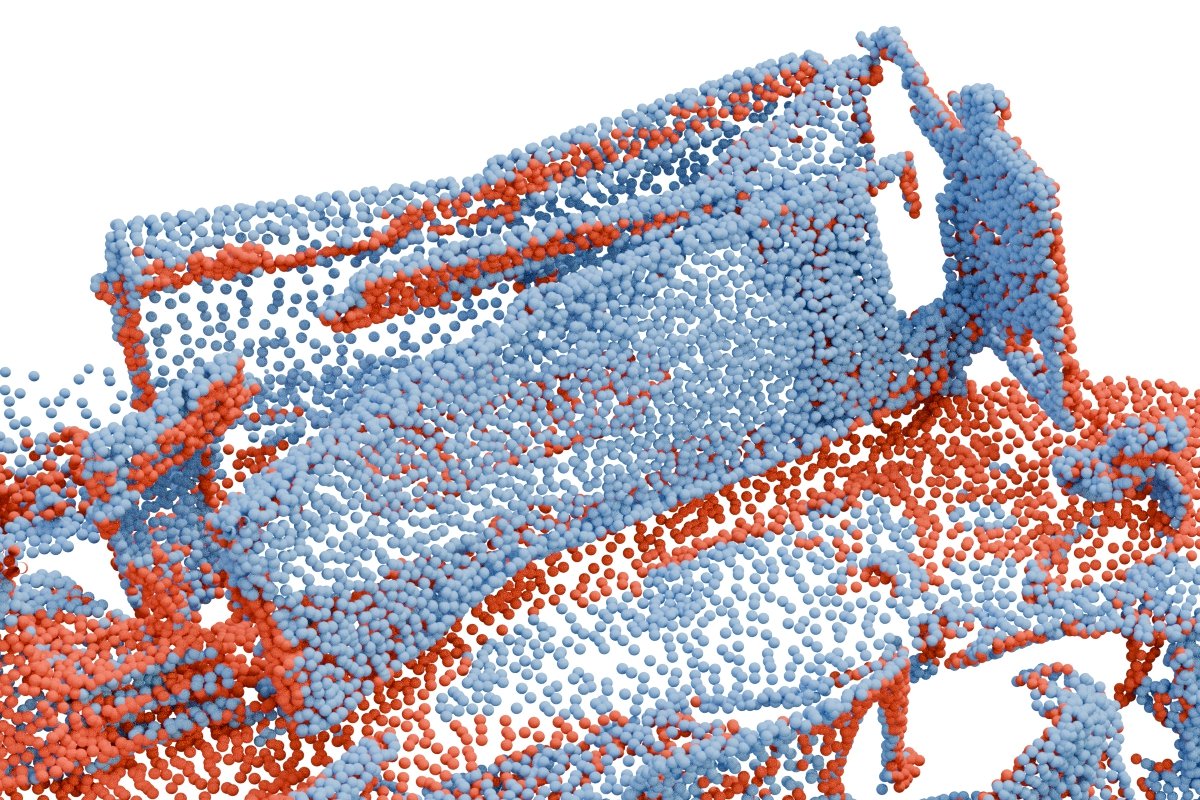}} & \fcolorbox{black}{white}{\includegraphics[width=0.11\textwidth, height=0.07333333333333333\textwidth]{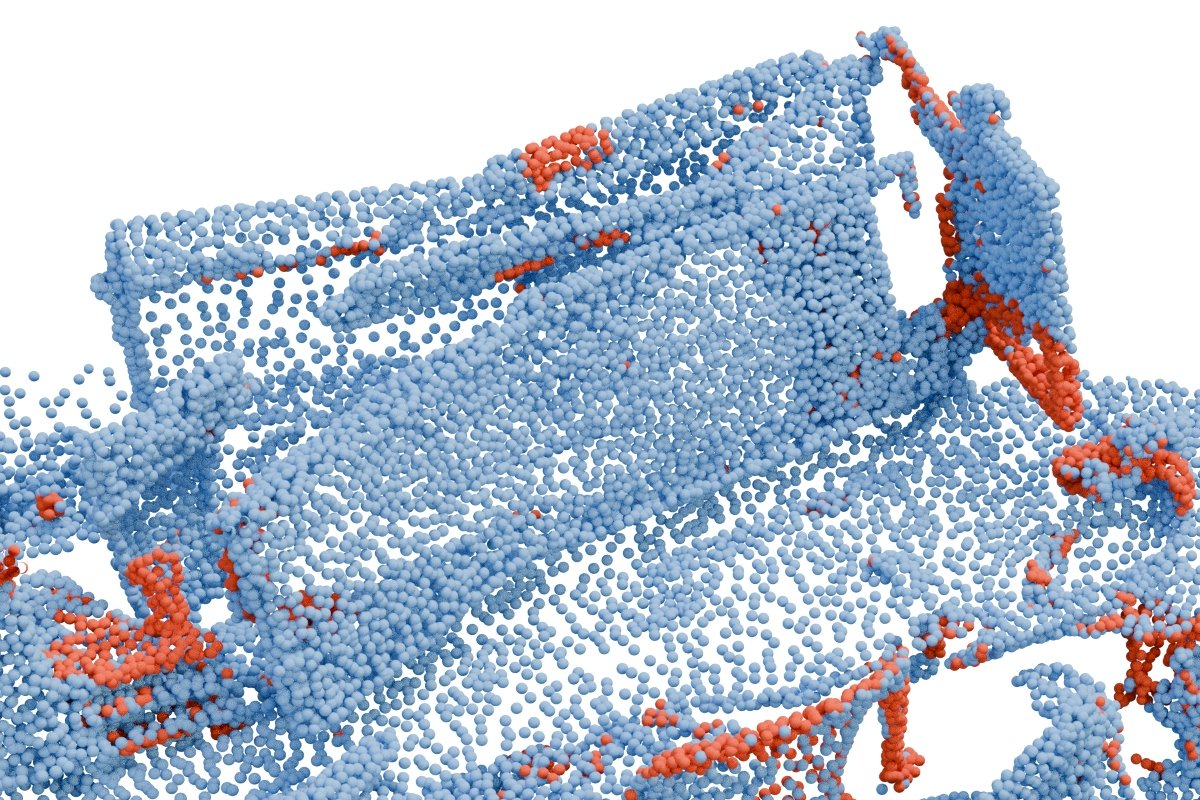}} & \fcolorbox{black}{white}{\includegraphics[width=0.11\textwidth, height=0.07333333333333333\textwidth]{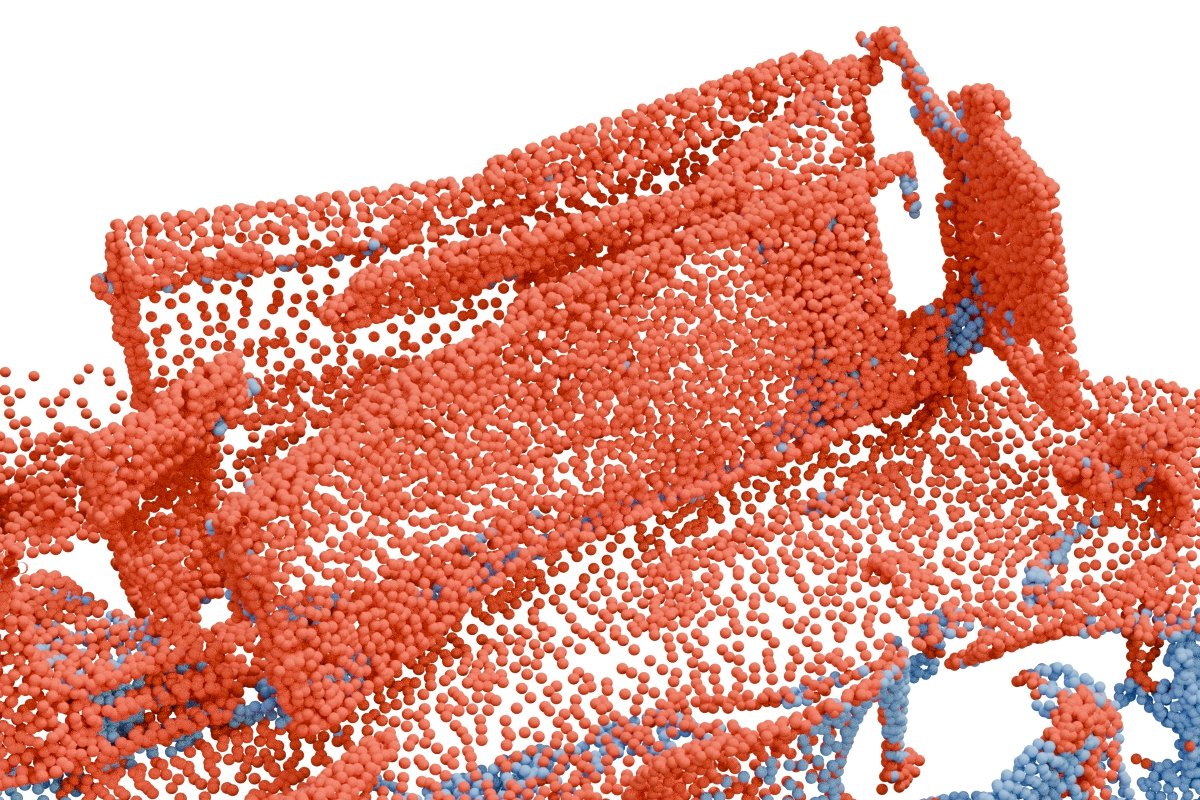}} & \fcolorbox{black}{white}{\includegraphics[width=0.11\textwidth, height=0.07333333333333333\textwidth]{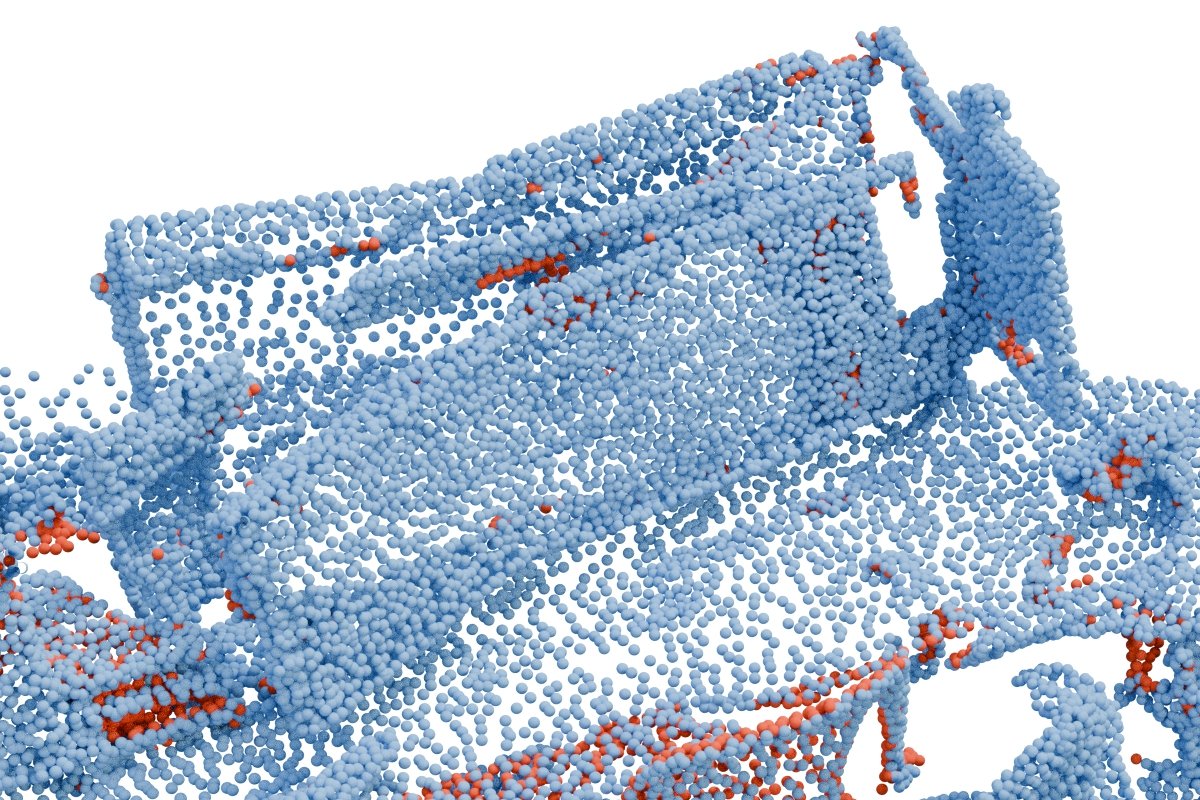}} \\ 

\begin{tikzpicture}[baseline=(current bounding box.center),x=0.11\textwidth,y=0.07333333333333333\textwidth]
    \node[anchor=south west,inner sep=0] {\includegraphics[width=0.11\textwidth,height=0.07333333333333333\textwidth]{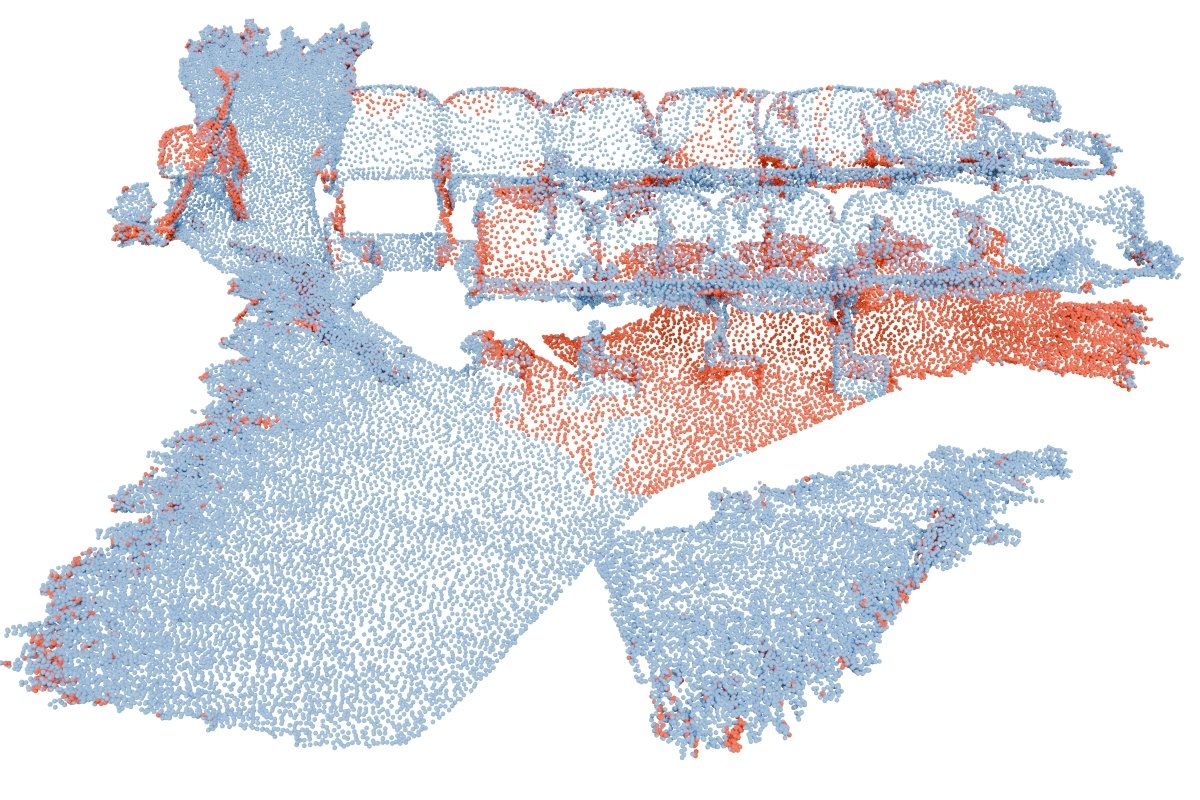}};
    \draw[black] (0.525,0.51) rectangle (0.815,0.80);
\end{tikzpicture} &
\begin{tikzpicture}[baseline=(current bounding box.center),x=0.11\textwidth,y=0.07333333333333333\textwidth]
    \node[anchor=south west,inner sep=0] {\includegraphics[width=0.11\textwidth,height=0.07333333333333333\textwidth]{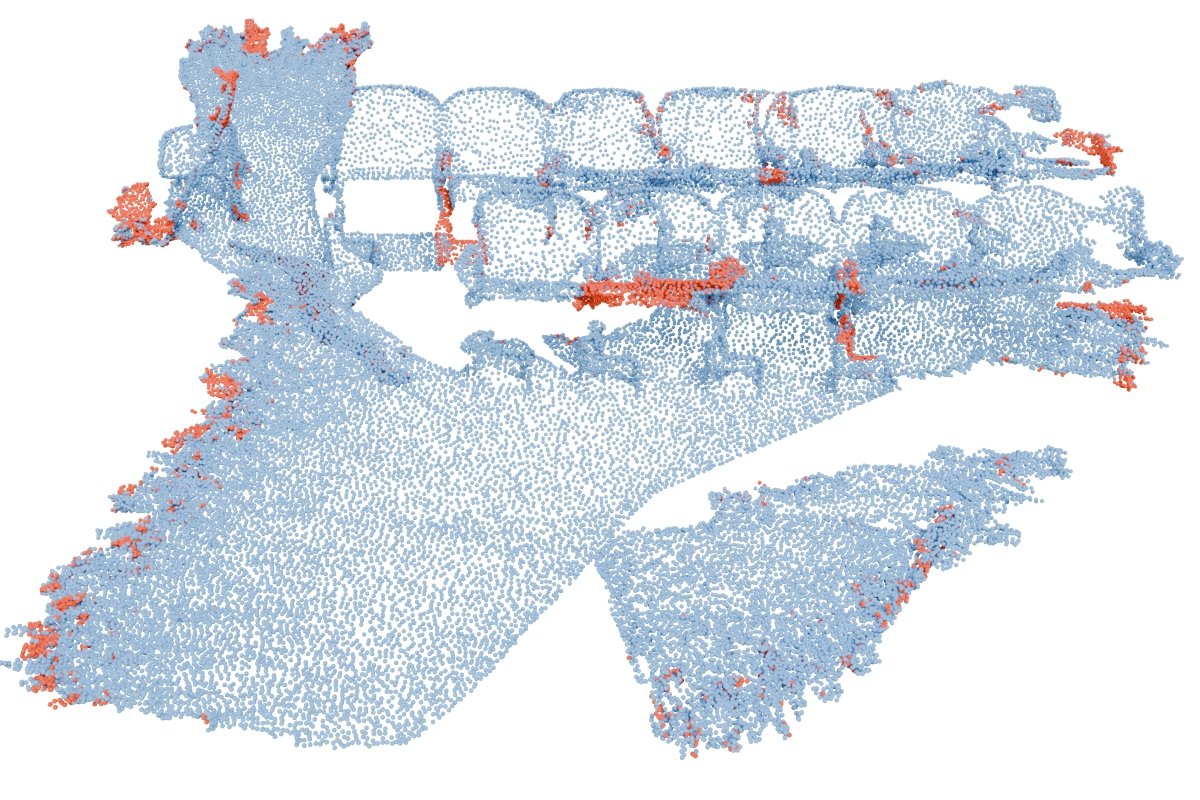}};
    \draw[black] (0.525,0.51) rectangle (0.815,0.80);
\end{tikzpicture} &
\begin{tikzpicture}[baseline=(current bounding box.center),x=0.11\textwidth,y=0.07333333333333333\textwidth]
    \node[anchor=south west,inner sep=0] {\includegraphics[width=0.11\textwidth,height=0.07333333333333333\textwidth]{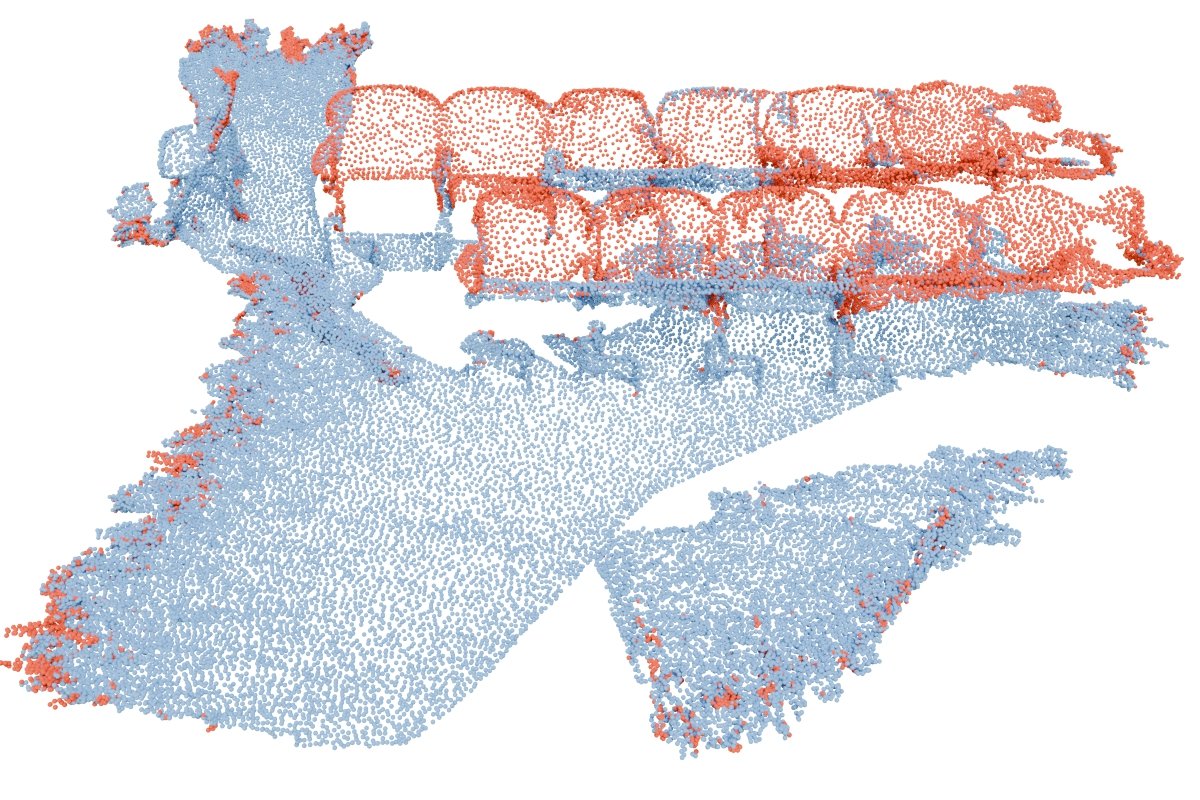}};
    \draw[black] (0.525,0.51) rectangle (0.815,0.80);
\end{tikzpicture} &
\begin{tikzpicture}[baseline=(current bounding box.center),x=0.11\textwidth,y=0.07333333333333333\textwidth]
    \node[anchor=south west,inner sep=0] {\includegraphics[width=0.11\textwidth,height=0.07333333333333333\textwidth]{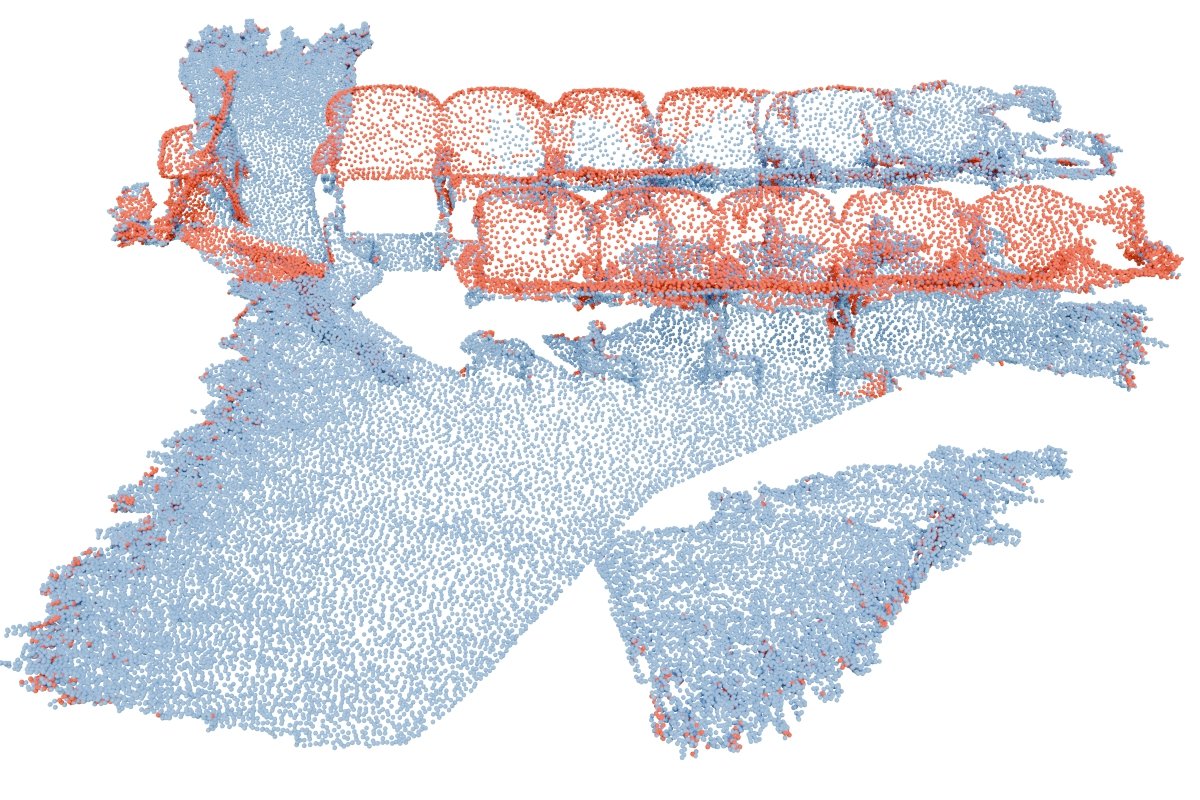}};
    \draw[black] (0.525,0.51) rectangle (0.815,0.80);
\end{tikzpicture} &
\begin{tikzpicture}[baseline=(current bounding box.center),x=0.11\textwidth,y=0.07333333333333333\textwidth]
    \node[anchor=south west,inner sep=0] {\includegraphics[width=0.11\textwidth,height=0.07333333333333333\textwidth]{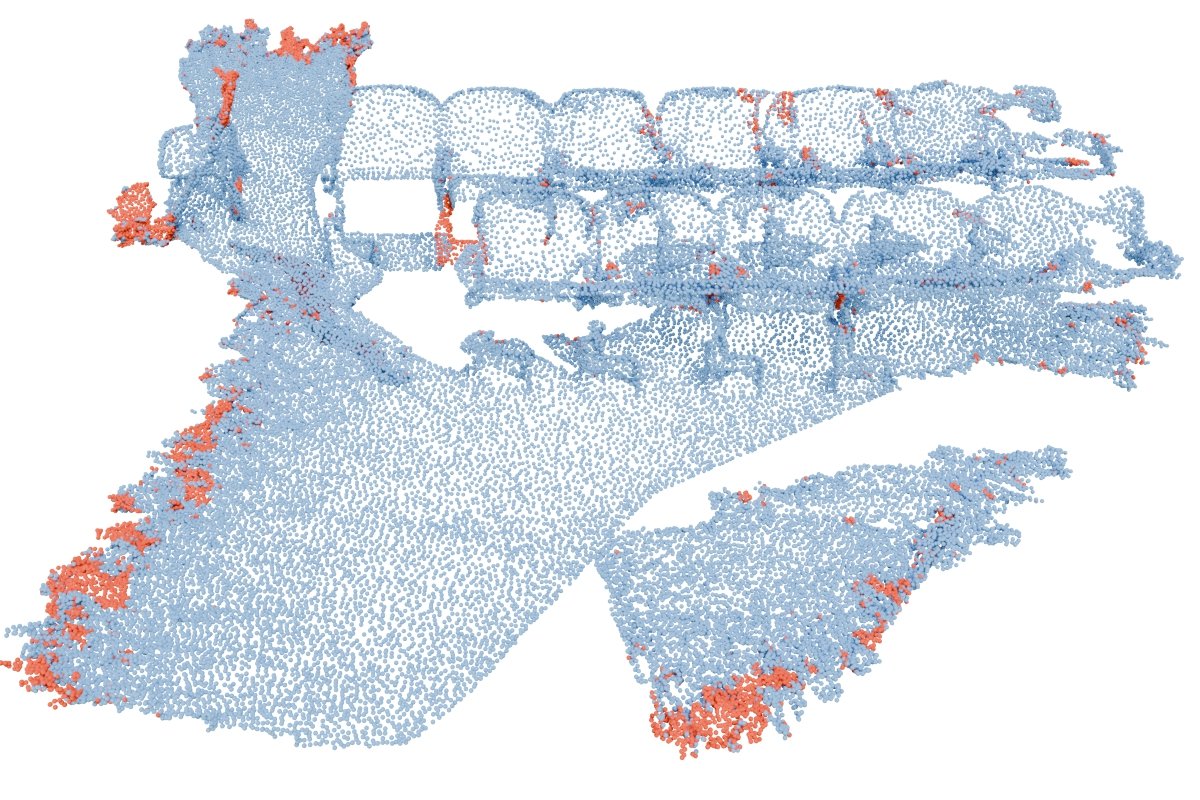}};
    \draw[black] (0.525,0.51) rectangle (0.815,0.80);
\end{tikzpicture} &
\begin{tikzpicture}[baseline=(current bounding box.center),x=0.11\textwidth,y=0.07333333333333333\textwidth]
    \node[anchor=south west,inner sep=0] {\includegraphics[width=0.11\textwidth,height=0.07333333333333333\textwidth]{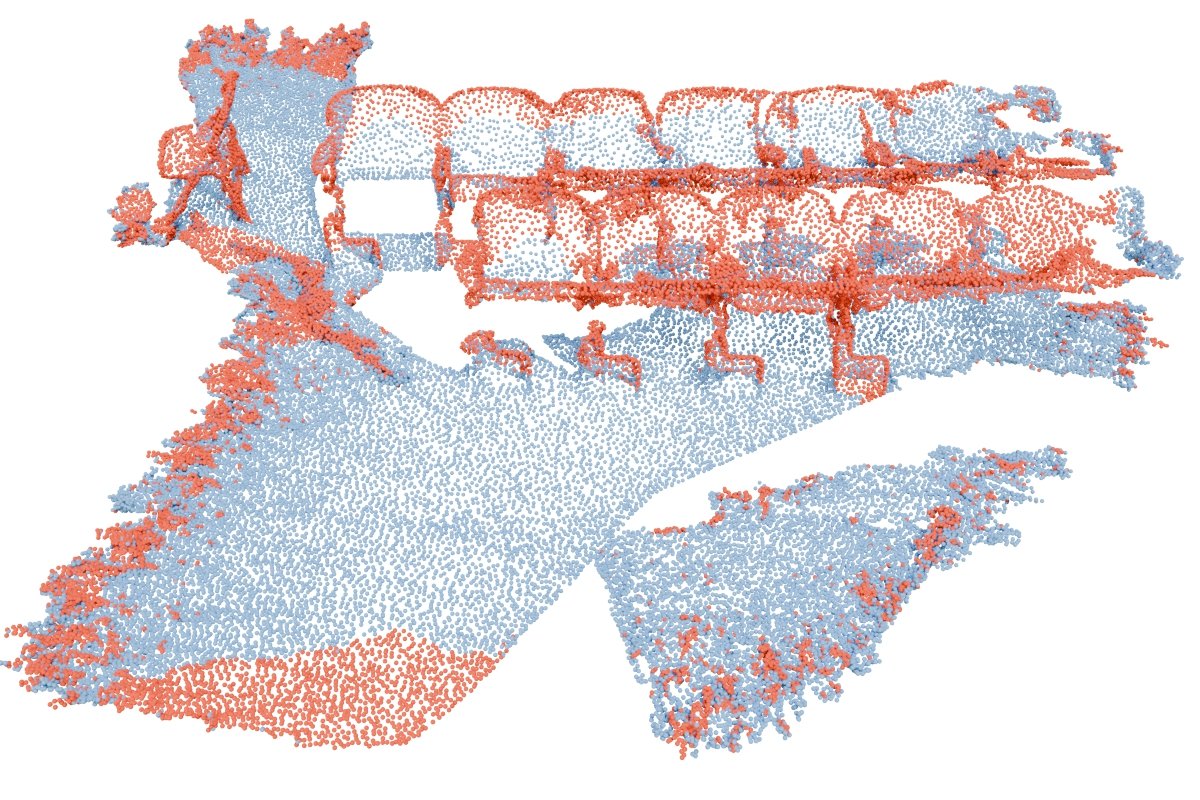}};
    \draw[black] (0.525,0.51) rectangle (0.815,0.80);
\end{tikzpicture} &
\begin{tikzpicture}[baseline=(current bounding box.center),x=0.11\textwidth,y=0.07333333333333333\textwidth]
    \node[anchor=south west,inner sep=0] {\includegraphics[width=0.11\textwidth,height=0.07333333333333333\textwidth]{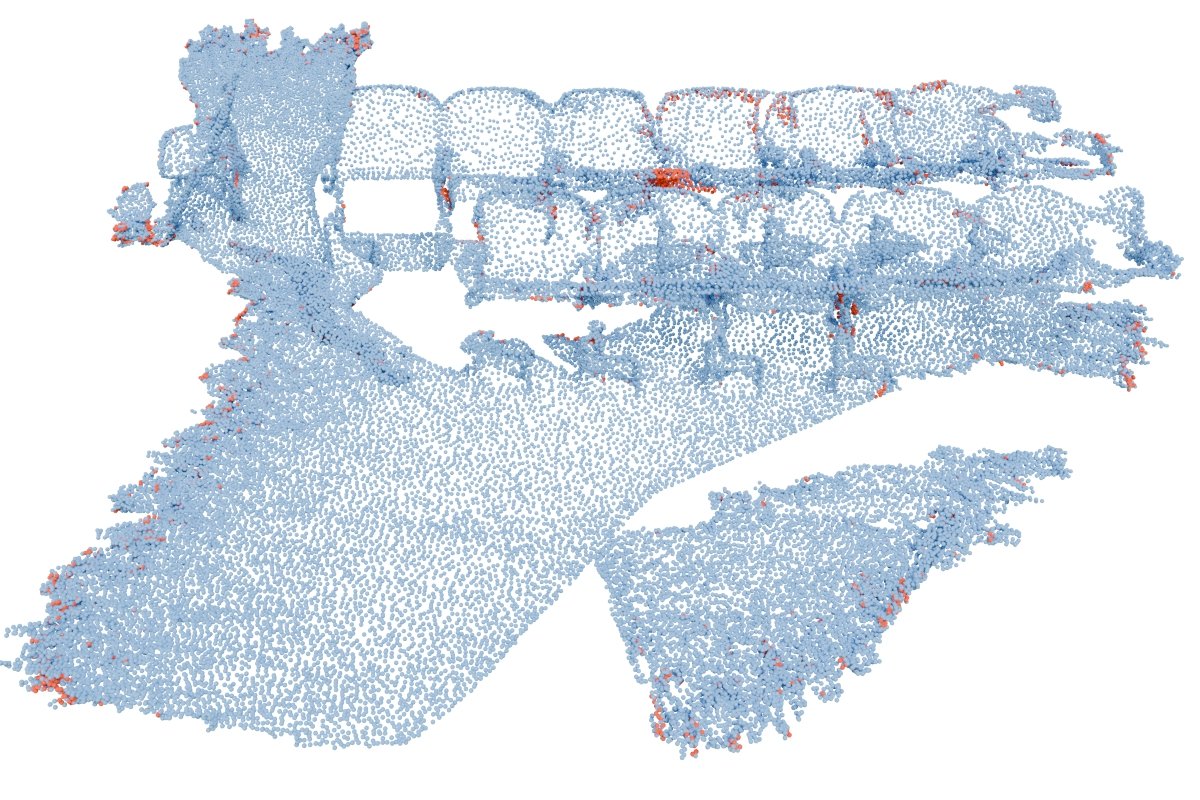}};
    \draw[black] (0.525,0.51) rectangle (0.815,0.80);
\end{tikzpicture} \\ 

\fcolorbox{black}{white}{\includegraphics[width=0.11\textwidth, height=0.07333333333333333\textwidth]{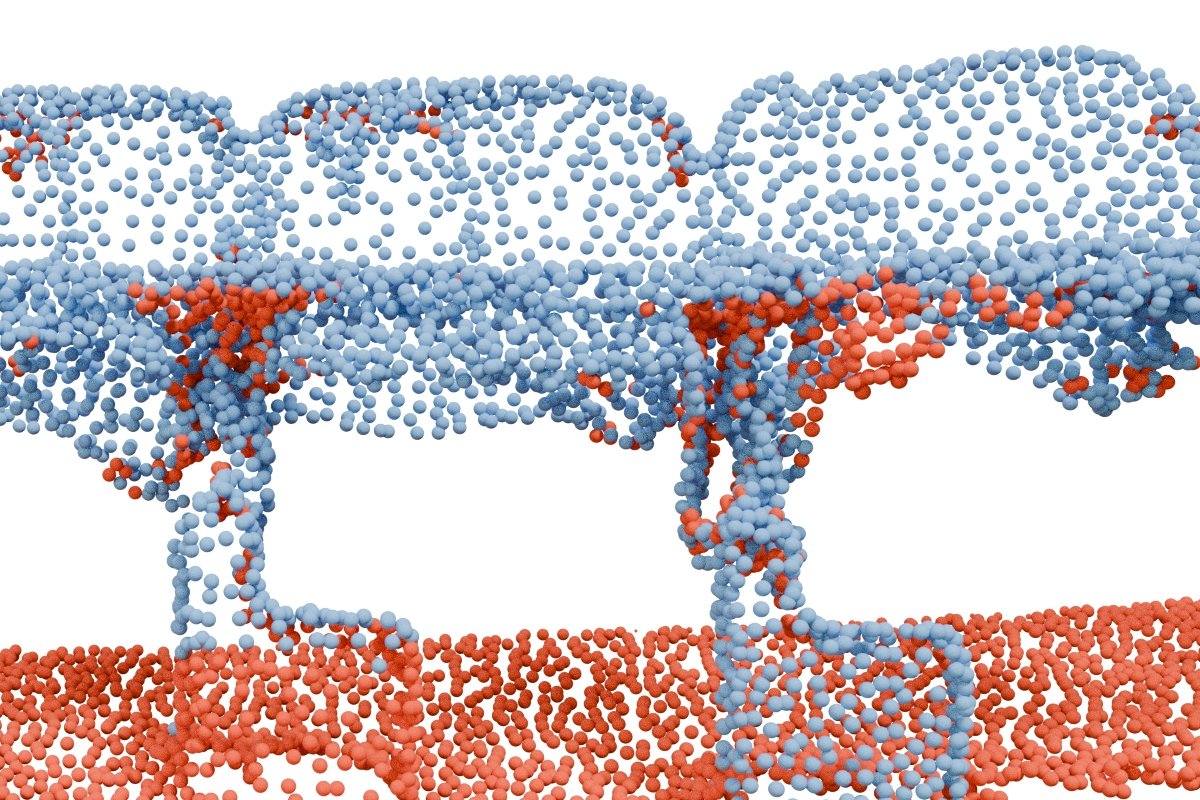}} & \fcolorbox{black}{white}{\includegraphics[width=0.11\textwidth, height=0.07333333333333333\textwidth]{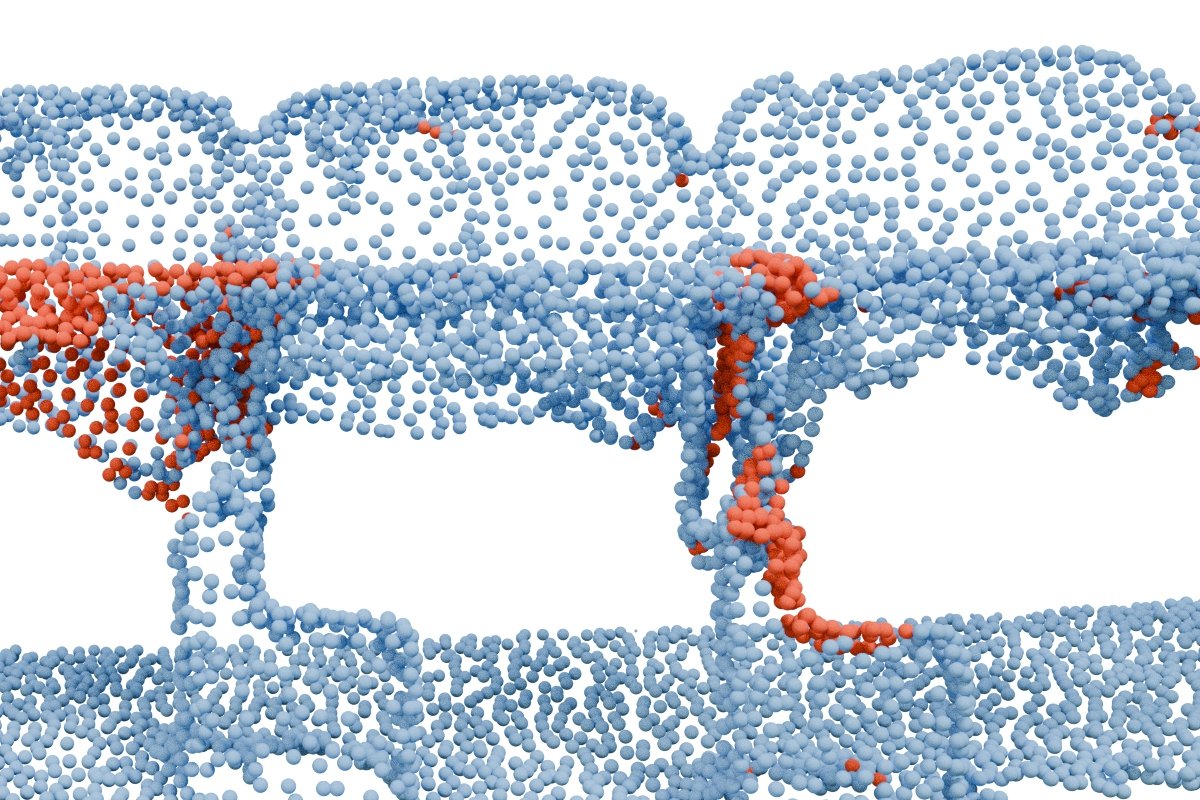}} & \fcolorbox{black}{white}{\includegraphics[width=0.11\textwidth, height=0.07333333333333333\textwidth]{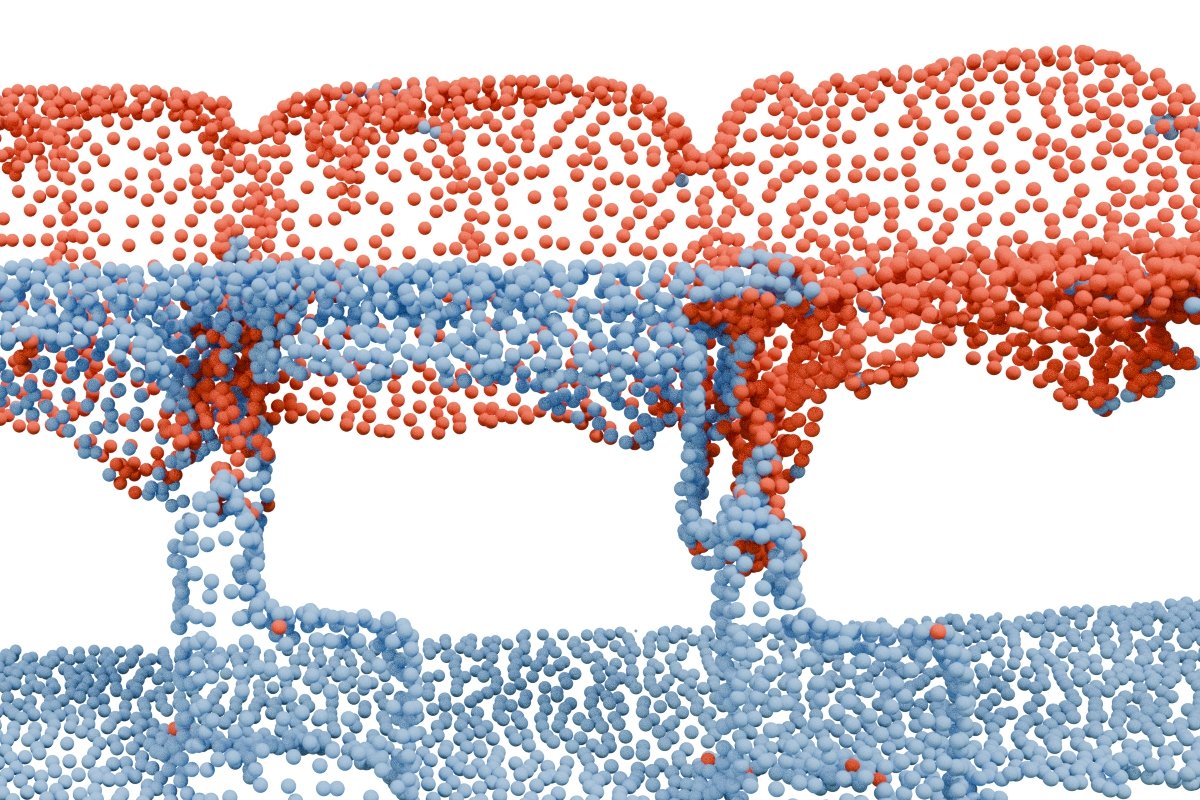}} & \fcolorbox{black}{white}{\includegraphics[width=0.11\textwidth, height=0.07333333333333333\textwidth]{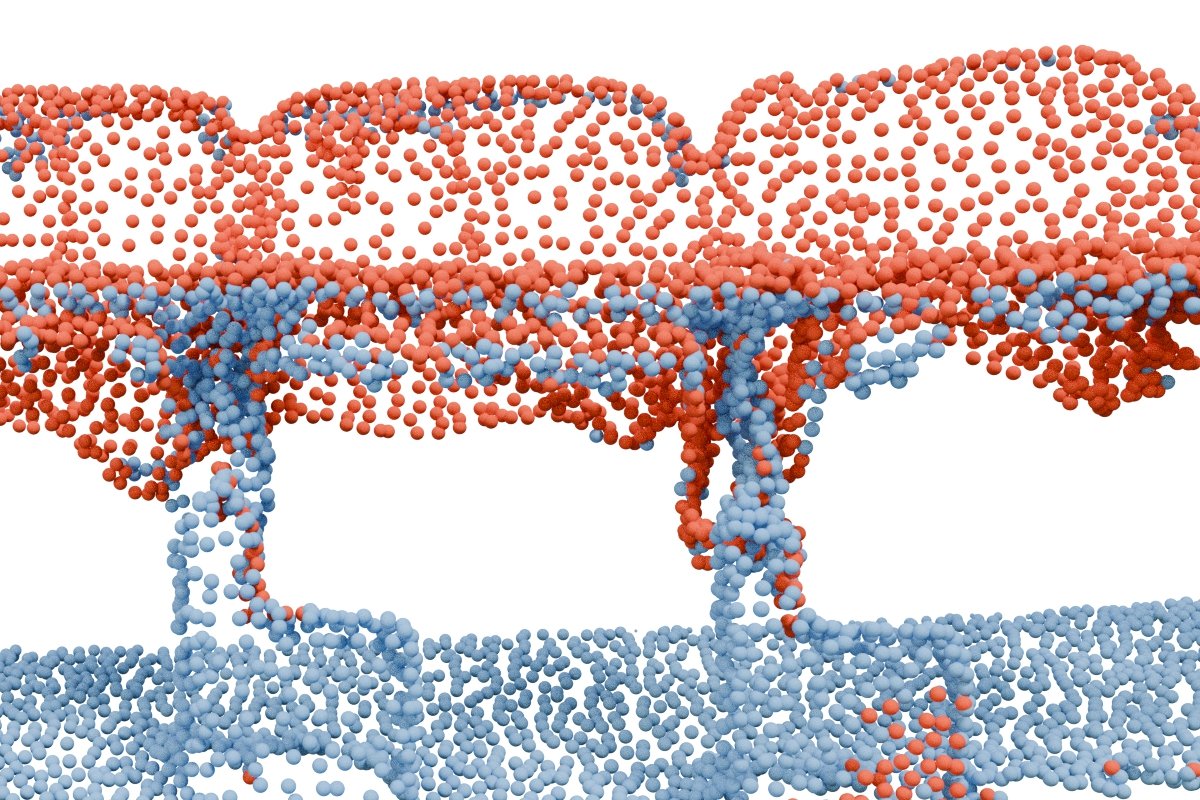}} & \fcolorbox{black}{white}{\includegraphics[width=0.11\textwidth, height=0.07333333333333333\textwidth]{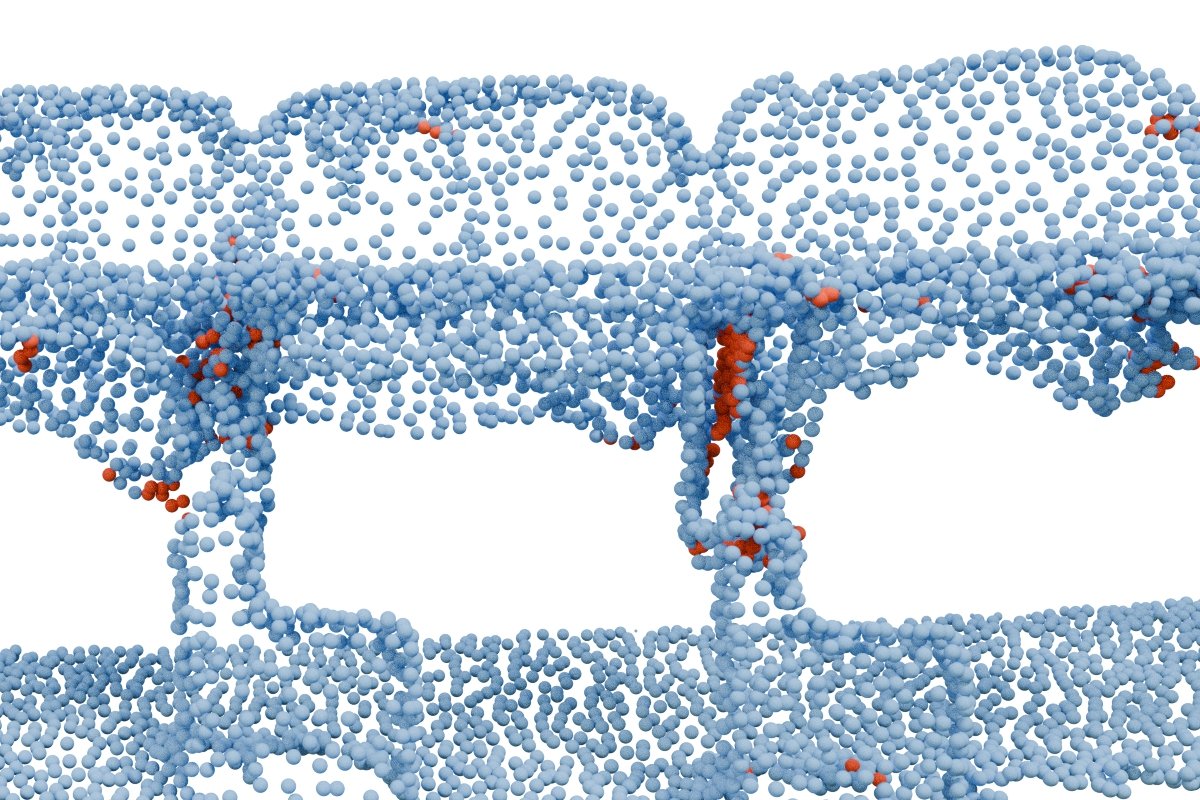}} & \fcolorbox{black}{white}{\includegraphics[width=0.11\textwidth, height=0.07333333333333333\textwidth]{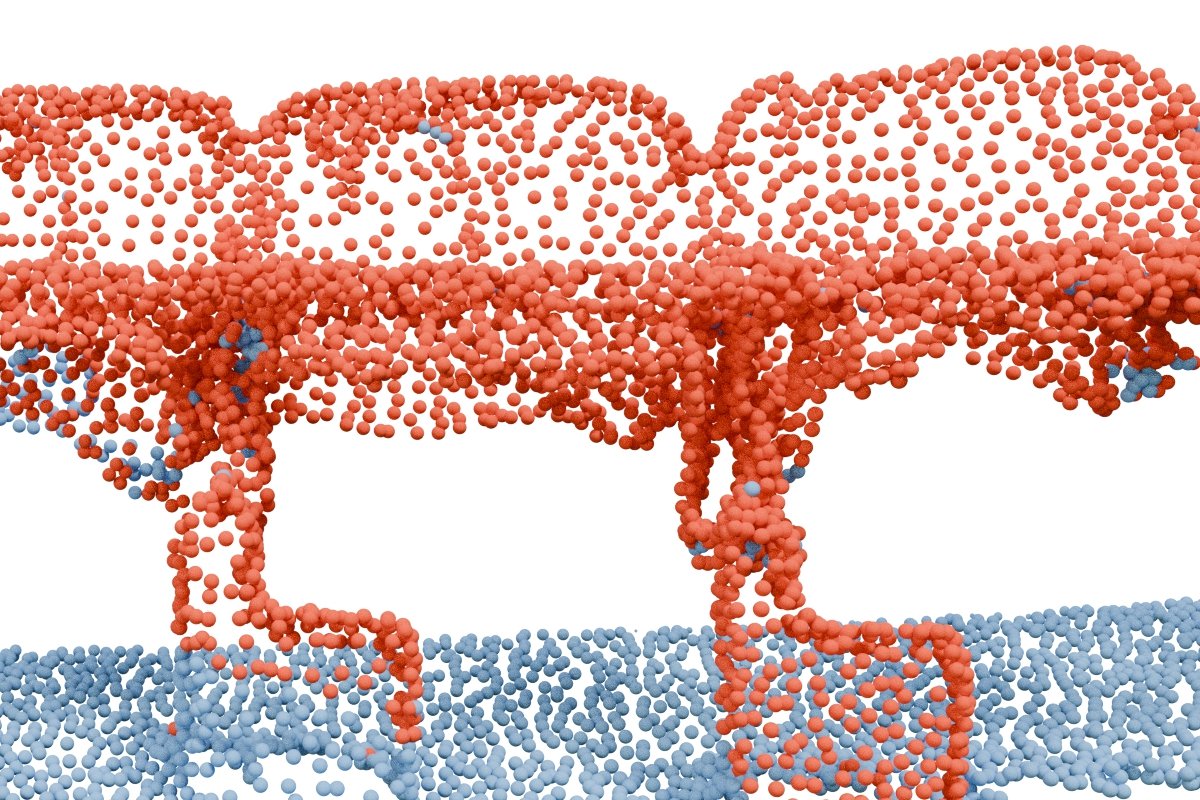}} & \fcolorbox{black}{white}{\includegraphics[width=0.11\textwidth, height=0.07333333333333333\textwidth]{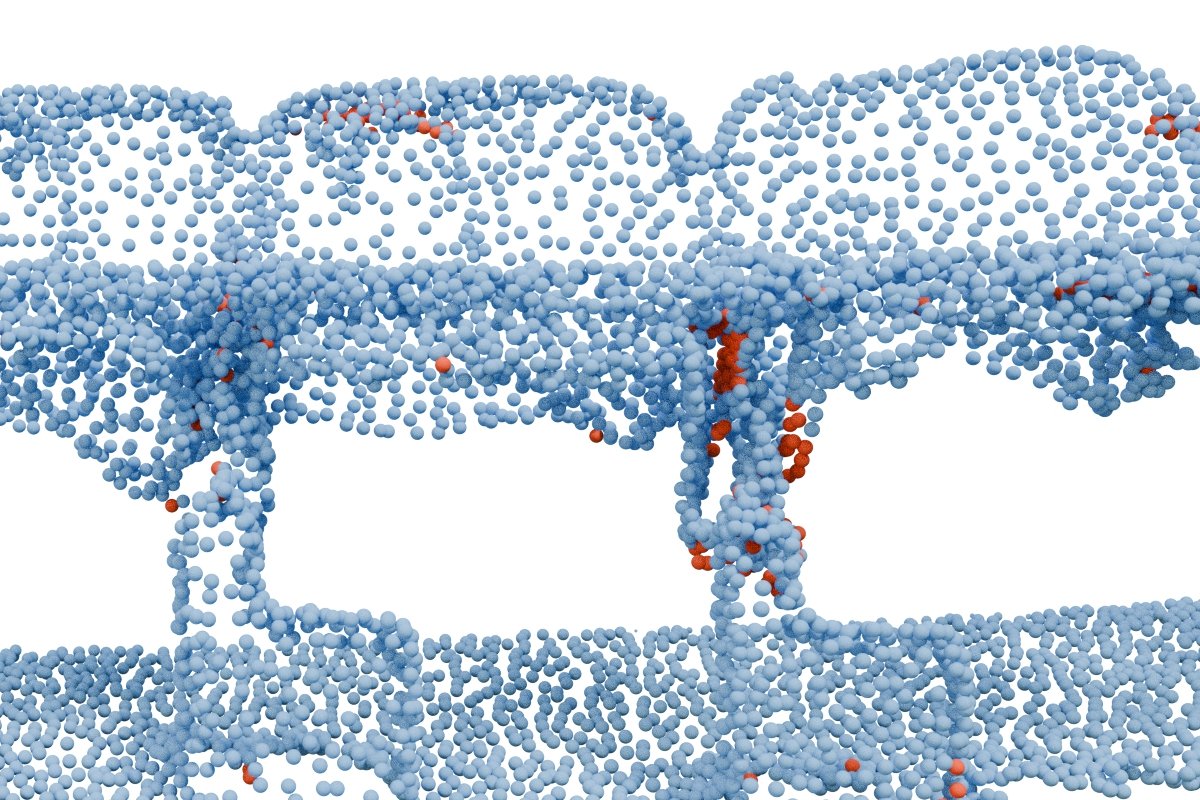}} \\ 
\end{tabular}
    \caption{  
    Scenes with weakly connected regions are challenging to orient. Blue points represent correctly oriented normals, while red points indicate incorrectly oriented normals. Each of the three models includes a full view of the scene and a corresponding close-up view highlighting areas with weak connections. From top to bottom, each scene contains 108K, 274K, 59K points, respectively.}
    \label{fig:comparison_weak_connection}
\end{figure*}

\paragraph{Sparse Data}
Sparsity introduces significant challenges in orienting scene-level point clouds, especially for propagation-based methods. {In the downsampled SceneNN-100K and SceneNN-10K datasets, the 100K version represents sparse data, while the 10K version is extremely sparse. We evaluate both our method and the baseline methods on these downsampled point clouds, with results provided in Table~\ref{tab:normal_ratio}. Figure~\ref{fig:comparison_scenenn} shows the orientation results for the original point clouds (row 1) and the downsampled sparse point clouds (rows 2 and 3).}

\paragraph{Weak Connections}
Scene data often contains numerous weak connections, which are narrow regions that provide limited useful information for orientation. As a result, they often introduce ambiguity in the orientation process. When normals are incorrectly oriented in these regions, it can lead to the misorientation of large adjacent areas. This issue is especially problematic for propagation-based methods. Figure~\ref{fig:comparison_weak_connection} illustrates examples where weak connections cause baseline methods to flip large areas incorrectly. Thanks to our global optimization, our method demonstrates enhanced robustness against weak connections.

\subsection{Ablation Studies}
\label{subsec:ablation}
To evaluate the necessity of the divide-and-conquer approach, we applied iPSR with the Neumann boundary condition directly to the scene point clouds from ScanNet v2~\cite{Dai2017ScanNet} {(107 scenes)} without block segmentation. Due to the complexity of the scenes, iPSR fails to properly orient points. The ratio of incorrectly oriented normals increases significantly, from $5.270\%$ to $29\%$. {Given this high failure rate, we decided not to test iPSR on more diverse datasets}.

Next, we assess the effects of orientation initialization (Section~\ref{subsec:initialization}) and iPSR (Section~\ref{subsec:ipsr}) separately. Table~\ref{tab:ablation} presents the ratio of incorrectly oriented normals when either iPSR or  initialization is omitted. The best results are achieved only when both iPSR and initialization are applied together.

\begin{table}[!htbp]
    \centering
    \caption{Ablation studies. We evaluate our method by skipping the orientation initialization and iPSR steps, and report the ratio of incorrectly oriented normals. ``Original'' refers to the original, noise-free point clouds from the the ScanNet v2~\cite{Dai2017ScanNet} dataset, while ``Noise1'' and ``Noise2'' correspond to point clouds with Gaussian noises applied, with standard deviations of $0.004$ and $0.008$, respectively. {The studies demonstrate that both the initialization and iPSR steps are crucial for accurate per-block orientation.} }
    \label{tab:ablation}
    \label{tab:ablation}
    \begin{tabular}{cccc}
        \toprule
        & w/o iPSR & w/o initialization  &  w/ both\\
        \midrule
        Original & 8.16\% & 9.07\% & 5.27\%\\
        Noise1 & 16.63\% & 10.22\% & 7.69\% \\
        Noise2 & 30.00\% & 12.40\% &  9.80\%\\
        \bottomrule
    \end{tabular}
\end{table}

\subsection{Runtime Performance}
\label{subsec:time}
{Thanks to the divide-and-conquer strategy, our method scales well to large models. For a typical input point cloud with 500K points, the total processing time is about 900 seconds, with approximately 800 seconds spent on iPSR for per-block orientation. In contrast, the global 0-1 integer-constrained optimization is highly efficient, typically requiring only a few seconds. This is because the number of blocks $N$ is relatively small, usually in the hundreds, which allows the Gurobi solver to find the optimal solution quickly.}

\subsection{Discussions}
\label{subsec:discussions}

Point orientation can sometimes be inferred from scanning devices, such as when the position and viewing direction of a laser scanner are available. However, many point clouds lack this information. For instance, point clouds generated by 3D generative method~\cite{Nichol2022PointE} typically do not include normal vectors. Moreover, orientation information can be disrupted by downstream operations; for example, editing or deforming point clouds may invalidate previously estimated normals. Scanning thin objects presents another challenge, as it often leads to ambiguous orientations. These cases highlight the need for reliable point cloud orientation methods.

Non-watertight manifold point clouds are typically generated due to incomplete scans or broken objects. For instance, ground scans often omit the underside, leading to non-watertight surfaces. Thin objects may also result in non-watertight point clouds due to scanner limitations. In fact, many scanned datasets contain a large number of non-watertight models, some of which do not provide normal vectors (e.g.,~\cite{Tan2020Toronto}). Since ScanNet v2 and SceneNN provide ground truth normals for quantitative evaluation, we use them in our experiments.

\section{Conclusion, Limitations and Future Works}
\label{sec:conclusions}

We propose a novel divide-and-conquer method for surface orientation in large-scale, scene-level point clouds. Existing global methods struggle with non-watertight models, while  propagation-based methods are susceptible to noise and sparsity in the point clouds. Our approach addresses these challenges by combining a divide-and-conquer strategy with a global optimization formulation. We begin by segmenting the input point cloud into smaller, manageable blocks, each initialized using a propagation-based method. We then apply an iPSR variant with the Neumann boundary condition, to refine the orientation within each block. To ensure consistent orientations across all blocks, we introduce the concept of visible consistent regions, which evaluates the consistency between adjacent blocks based on visibility. This leads to the formulation of an integer-constrained 0-1 optimization problem on the adjacency graph. By limiting the number of blocks to a few hundred, our method remains  computationally efficient and scalable for handling large-scale models.

\paragraph{Limitations} Despite showing good experimental results on benchmark datasets, our method still has several limitations. First, our method struggles with extremely sparse point clouds, such as wireframed data. After segmentation, wireframe points often fail to preserve geometry within a block, making per-block orientation highly challenging. Second, when producing the final reconstructed surface, our method requires a postprocessing step. Specifically, sPSR with the Neumann boundary condition generates an extended mesh from the boundary, and redundant faces must be deleted. This can result in zigzag artifacts along the boundary. Third, although our method can handle certain thin structures, it may fail to orient points sampled from two extremely close surfaces. This limitation is shared by all other  methods. Fourth, our method assumes the input point cloud represents a connected surface, therefore, extremely weak connections are likely to result in incorrect orientations. Last but not least, our method uses a randomized greedy method for initializing point orientations within each block, followed by an adapted iPSR to further refine orientation accuracy. However, this approach does not guarantee that point orientations within each block are always correct. The subsequent global 0-1 optimization only determines the flipping states of the blocks and does not correct orientation errors within individual blocks. Therefore, if a point orientation within a block is incorrect, the error persists throughout the global optimization stage.

\paragraph{Future Works} DACPO is inherently parallelizable due to its divide-and-conquer nature. Our current implementation runs on a single CPU using multithreading. In future work, we plan to deploy it on CPU clusters to improve runtime performance. Additionally, we aim to replace the current per-block orientation method with a more robust, learning-based approach to further enhance orientation accuracy.


\begin{acks}
This research was supported in part by the National Key R\&D Program of China (2023YFB3002901), in part by the Basic Research Project of ISCAS (ISCAS-JCMS-202303), in part by the Major Research Project of ISCAS (ISCAS-ZD-202401, ISCAS-JCZD-202402), and in part by the Ministry of Education, Singapore, under its Academic Research Fund Grant (RT19/22).
\end{acks}


\bibliographystyle{ACM-Reference-Format}
\bibliography{reference}

\appendix
\section{Orientation of VCR}
\label{sec:app_vcr}
Given a particular viewpoint, we observe that the orientations of a visible connected region are view-aligned. Below, we explain why this is the case.

\begin{figure}[!htbp]
    \centering
    \includegraphics[width=\linewidth]{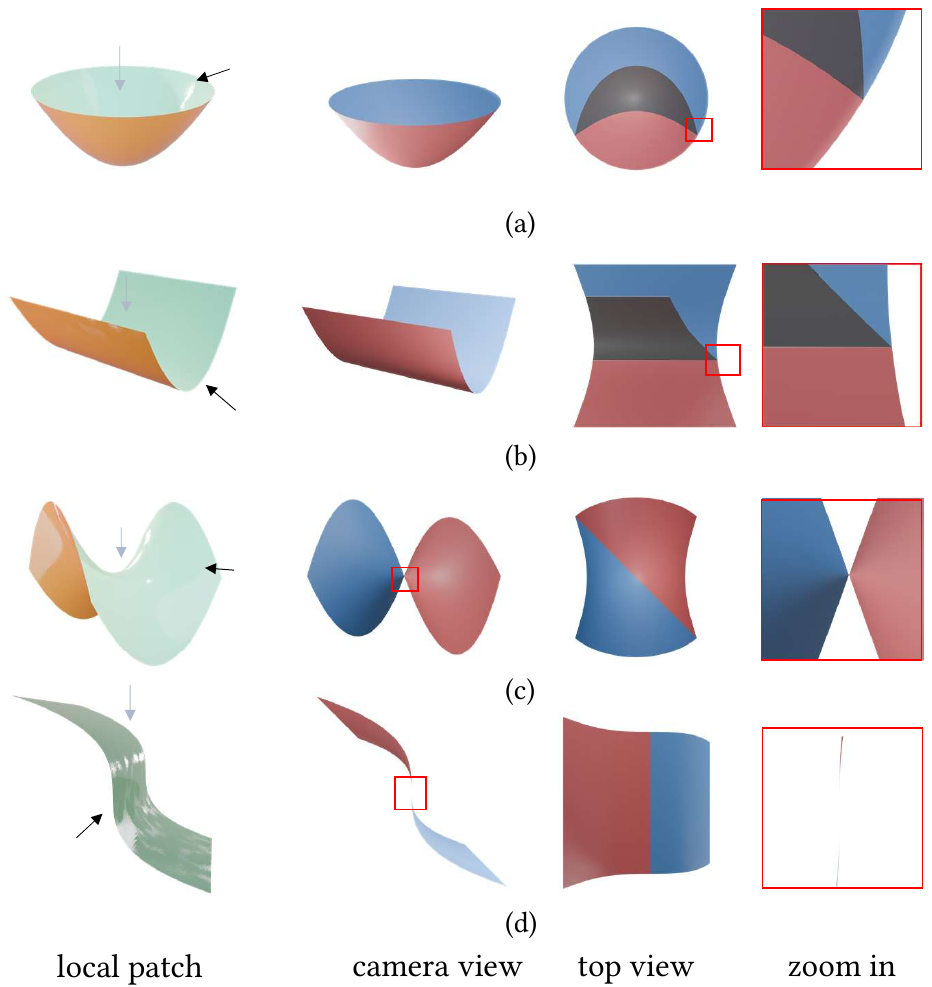}
    \caption{Visible regions of quadratic surfaces. The front and back sides of each surface are colored in green and yellow, respectively. The black and gray arrows indicate the camera view and top view directions, respectively. From the camera viewpoint, we paint the visible front side in blue, the visible back side in red, and the invisible regions in black. The top view illustrates the surface from another perspective. (a) Elliptic paraboloid, (b) Parabolic cylinder, and (c) Hyperbolic paraboloid.  In (a) and (b), the visible regions are disconnected (an isolated intersection point is treated as disconnected), while in (c), the visible regions may appear connected on the surface but remain disconnected in their projections on the viewing plane. In the degenerate case (d), the S-shaped cylindrical surface with all points along the central line are flat umbilics (i.e., $\kappa_1=\kappa_2=0$), the visible regions are connected on the surface when viewed from the central line but still disconnected in their projections.}
    \label{fig:quadratic_expansion}
\end{figure}

Let $S$ be a smooth manifold surface and $\mathbf{p}\in S$ an interior point. Consider the local neighborhood $U\subset S$ around $\mathbf{p}$. We can construct a local coordinate system at $\bf p$, with the $z$-axis aligned with the normal. According to differential geometry, we can approximate $U$ using a quadratic function:
\[
z(x,y)=\frac{1}{2}(\kappa_1 x^2 + \kappa_2 y^2),
\]
where $\kappa_1$ and $\kappa_2$ are the principal curvatures at $\mathbf{p}$. 
The surface is classified as:
\begin{itemize}
    \item Elliptic paraboloid if $\kappa_1\kappa_2>0$,
    \item Hyperbolic paraboloid if $\kappa_1\kappa_2<0$,
    \item Parabolic cylinder if one of $\kappa_1$ or $\kappa_2$ is 0.
\end{itemize}
Since local patches can be approximated by quadratic functions, we examine VCR orientation by checking four different cases as illustrated in 
Figure~\ref{fig:quadratic_expansion}.
In (a) and (b), it is clear that, on elliptic paraboloids and parabolic cylinders, it is impossible to view both the front and back sides simultaneously from any viewpoint. 

In the case of a hyperbolic paraboloid (c), a geometrically connected visible region could contain both front and back sides, but the projection remains disconnected, resulting in two VCRs. In each of these VCRs, the orientations are view-aligned.

In the degenerate case where $\kappa_1=0$ and $\kappa_2=0$, point $\bf p$ is a flat umbilic. A plane is the only surface where every point is a flat umbilic. While flat umbilics are rare in general surfaces, certain surfaces can have all points along a line as flat umbilics. Let $C$ be a planar curve where a point $\mathbf{p}\in C$ has zero curvature. When we sweep $C$ along a ray orthogonal to its plane to generate a cylindrical surface, the trajectory of the point $\mathbf{p}$ on the surface traces out a set of flat umbilics. Figure~\ref{fig:quadratic_expansion}(d) illustrates an S-shaped cylindrical surface where every point along the central line is a flat umbilic. When observing the surface along this ray, the visible front and back sides are connected on the surface. However, the projection of this connection remains disconnected, leading to two VCRs. In each of these VCRs, the orientations are view-aligned.

\end{document}